%% file: icml2024.tex
\documentclass{article}

\usepackage{microtype}
\usepackage{graphicx}
\usepackage{booktabs} 

\usepackage{url}
\usepackage{subcaption}
\usepackage{wrapfig}

\usepackage{multirow}
\usepackage[normalem]{ulem}
\useunder{\uline}{\ul}{}

\usepackage{bm}

\usepackage{xspace}
\def \linf {$\ell_{\infty}$\xspace}
\def \lt {$\ell_2$\xspace}
\def \nonlp {non-$\ell_p$\xspace}
\def \lp {$\ell_p$\xspace}
\def \rsquare {$R^2$\xspace}
\def \pnorm {$p$-norm\xspace}
\def \oodd {$\text{OOD}_d$\xspace}
\def \oodt {$\text{OOD}_t$\xspace}

\usepackage{xcolor}

\usepackage{hyperref}

\usepackage[accepted]{icml2024}

\usepackage{amsmath}
\usepackage{amssymb}
\usepackage{mathtools}
\usepackage{amsthm}

\usepackage[capitalize,noabbrev]{cleveref}

\theoremstyle{plain}

\theoremstyle{definition}

\theoremstyle{remark}

\usepackage[textsize=tiny]{todonotes}

\icmltitlerunning{OODRobustBench: a benchmark and large-scale analysis of adversarial robustness under distribution shift}

\begin{document}

\twocolumn[
\icmltitle{OODRobustBench: a Benchmark and Large-Scale Analysis of\\ Adversarial Robustness under Distribution Shift}




\begin{icmlauthorlist}
\icmlauthor{Lin Li}{kcl}
\icmlauthor{Yifei Wang}{mit}
\icmlauthor{Chawin Sitawarin}{ucb}
\icmlauthor{Michael Spratling}{kcl,uol}
\end{icmlauthorlist}

\icmlaffiliation{kcl}{Department of Informatics, King's College London, UK}
\icmlaffiliation{mit}{MIT CSAIL, USA}
\icmlaffiliation{ucb}{UC Berkeley, USA}
\icmlaffiliation{uol}{University of Luxembourg, Luxembourg}

\icmlcorrespondingauthor{Lin Li}{lin.3.li@kcl.ac.uk}

\icmlkeywords{adversarial robustness, attacks, AI safety, OOD generalization}

\vskip 0.3in
]



\printAffiliationsAndNotice{} 

\begin{abstract}
Existing works have made great progress in improving adversarial robustness, but typically test their method only on data from the same distribution as the training data, i.e. in-distribution (ID) testing. 
As a result, it is unclear how such robustness generalizes under input distribution shifts, i.e. out-of-distribution (OOD) testing. 
To address this issue we propose a benchmark named OODRobustBench to comprehensively assess OOD adversarial robustness using 23 dataset-wise shifts (i.e. naturalistic shifts in input distribution) and 6 threat-wise shifts (i.e., unforeseen adversarial threat models). 
OODRobustBench is used to assess 706 robust models using 60.7K adversarial evaluations. This large-scale analysis shows that: 
1) adversarial robustness suffers from a severe OOD generalization issue;
2) ID robustness correlates strongly with OOD robustness in a positive linear way. 
The latter enables the prediction of OOD robustness from ID robustness. 
We then predict and verify that existing methods are unlikely to achieve high OOD robustness. 
Novel methods are therefore required to achieve OOD robustness beyond our prediction.
To facilitate the development of these methods, we investigate a wide range of techniques and identify several promising directions. 
Code and models are available at: \url{https://github.com/OODRobustBench/OODRobustBench}.

\end{abstract}

\vspace{-5mm}
\section{Introduction}
Adversarial attack poses a serious threat to real-world machine learning models, and various approaches have been developed to defend against such attacks. Previous work \citep{athalye_obfuscated_2018} has shown that adversarial evaluation is critical to the study of adversarial robustness since an unreliable evaluation can often give a false sense of robustness. However, we believe that even state-of-the-art evaluation benchmarks ~\citep[like RobustBench][]{croce_robustbench_2021} suffer from a severe limitation: they only consider ID generalization where test data comes from the same distribution as the training data.
%
Since distribution shifts are inevitable in the real world, it is crucial to assess how adversarial robustness is affected when the test distribution differs from the training one.

\begin{figure*}[!ht]
    \centering
    \begin{subfigure}{\linewidth}
        \includegraphics[width=\linewidth]{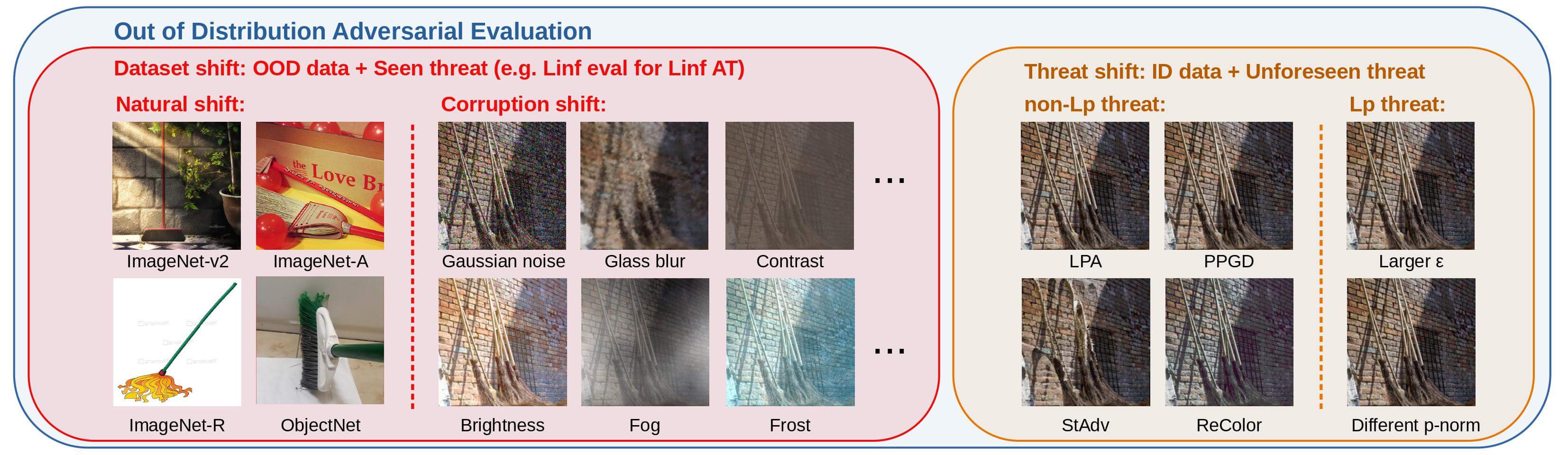}
    \end{subfigure}
    
    \begin{subfigure}{.245\linewidth}
        \includegraphics[width=\linewidth]{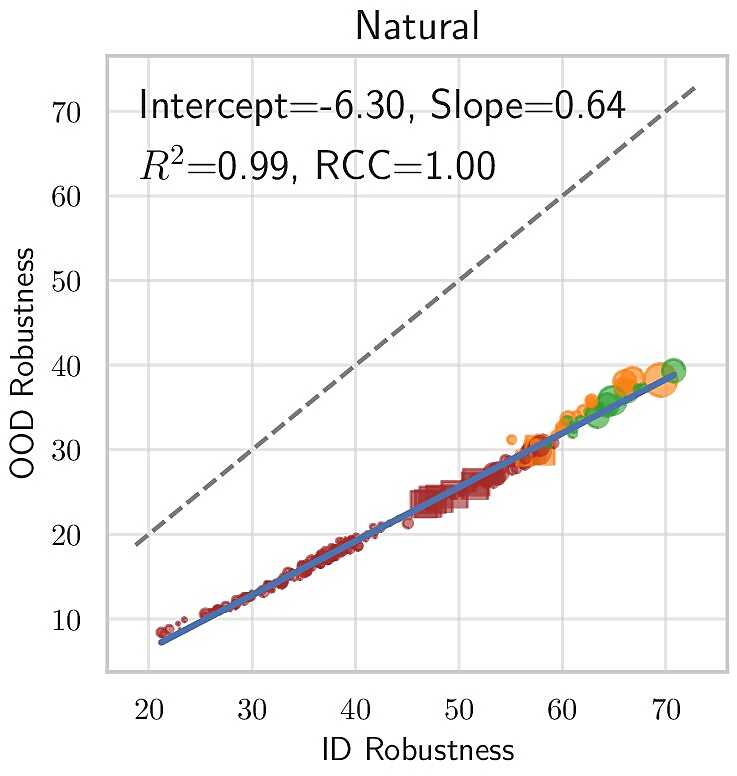}
    \end{subfigure}
    \begin{subfigure}{.245\linewidth}
        \includegraphics[width=\linewidth]{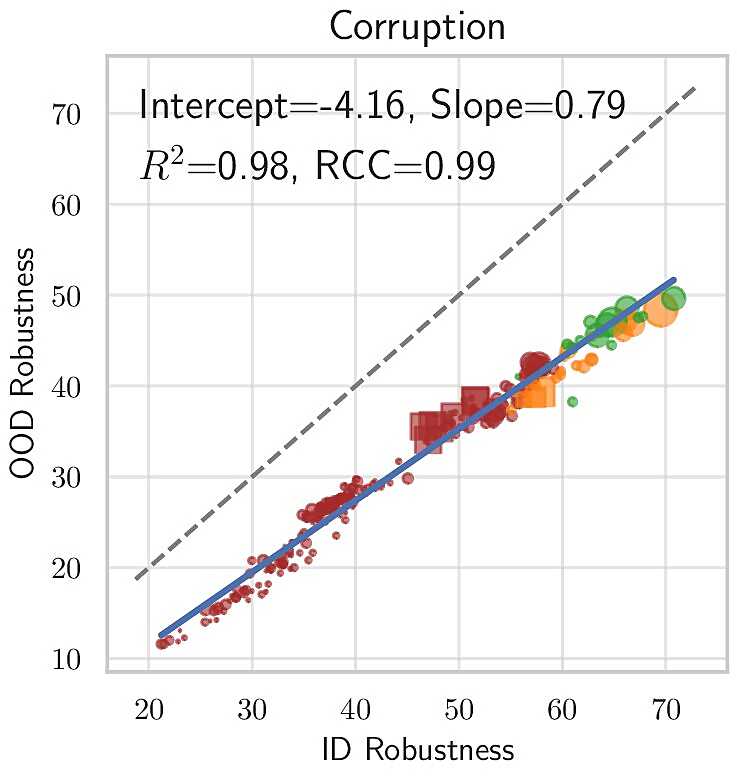}
    \end{subfigure}
    \begin{subfigure}{.245\linewidth}
        \includegraphics[width=\linewidth]{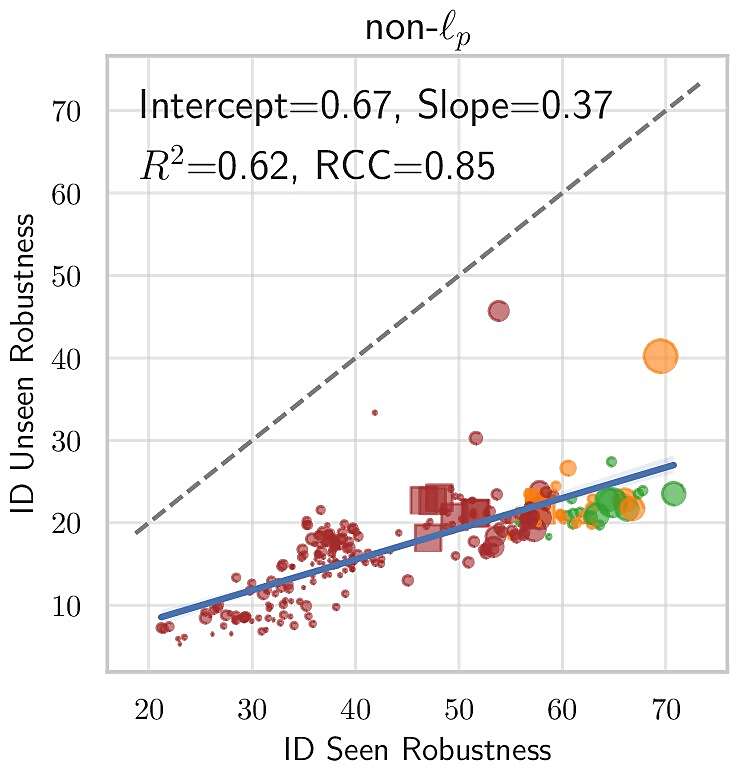}
    \end{subfigure}
    \begin{subfigure}{.245\linewidth}
        \includegraphics[width=\linewidth]{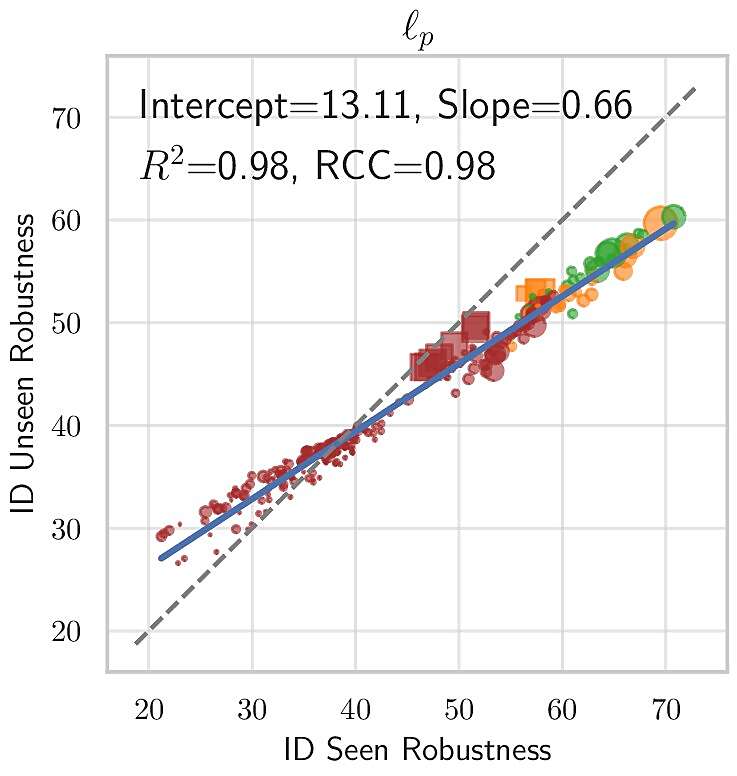}
    \end{subfigure}

    \begin{subfigure}{\linewidth}
        \includegraphics[width=\linewidth, trim=0 25 0 25, clip]{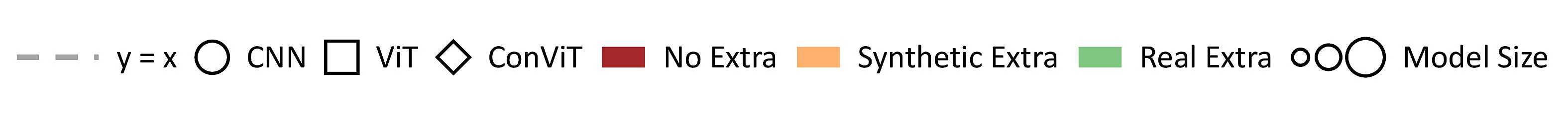}
    \end{subfigure}
    
    \caption{\textbf{The construction of OODRobustBench} (top) and \textbf{the correlation between ID and OOD robustness under 4 types of distribution shift for CIFAR10 \linf} (bottom). Each marker represents a model and is annotated by its training set-up. The solid blue line is the fitted linear correlation. The dashed gray line ($y=x$) represents perfect generalization where OOD robustness equals ID robustness. Deviation from the dashed line indicates robustness degradation under the respective distribution shift.}
    \label{fig: benchmark construction}
\end{figure*}

Although OOD generalization has been extensively studied for clean accuracy~\citep{hendrycks_benchmarking_2019,taori_measuring_2020,miller_accuracy_2021,baek_agreementontheline_2022,zhao_oodcv_2022,yang_openood_2022}, there is little known about the OOD generalization of adversarial robustness.
To fill this void, this paper presents for the first time, a comprehensive benchmark, \textbf{OODRobustBench}, for assessing out-of-distribution adversarial robustness. Furthermore, it reports results of a large-scale analysis of existing robust models performed using the new benchmark to answer the following questions:
\begin{enumerate}
    \vspace{-1mm}
    \item How resilient are current adversarially robust models to distribution shift?
    \vspace{-1mm}
    \item Can we predict OOD robustness from ID robustness?
    \vspace{-1mm}
    \item How can we achieve OOD robustness?
\end{enumerate}
\vspace{-1mm}
OODRobustBench is analogous and complementary to RobustBench which is used for assessing in-distribution adversarial robustness. It includes two categories of distribution shift: dataset shift and threat shift (see \cref{fig: benchmark construction}).
Dataset shift denotes test data that has different characteristics from the training data due to varying conditions under which the samples are collected: for images, these include but are not limited to corruptions, background, and viewpoint.
OODRobustBench contains 23 such dataset shifts and assesses adversarial robustness to such data using the attack seen by the model during training. Threat shift denotes a variation between training and test adversarial threat models. In other words, threat shift assesses a model's robustness to unseen adversarial attacks applied to ID test data. OODRobustBench employs six different types of threat shifts. Adversarial robustness is evaluated for each type of shift to comprehensively assess OOD robustness.

With OODRobustBench, we analyze the OOD generalization behavior of 706 well-trained robust models (a total of 60.7K adversarial evaluations). This model zoo covers a diversity of architectures, robust training methods, data augmentation techniques and training set-ups to ensure the conclusions drawn from this assessment are general and comprehensive. This large-scale analysis reveals that:
\begin{itemize}
    \vspace{-1mm}
    \item \textbf{Adversarial robustness suffers from a severe OOD generalization issue}. Robustness degrades on average by 18\%/31\%/24\% under distribution shifts for CIFAR10 \linf, CIFAR10 \lt and ImageNet \linf respectively. 
    \vspace{-1mm}
    
    \item \textbf{ID and OOD accuracy/robustness have a strong linear correlation under many shifts} (visualized in \cref{fig: benchmark construction}). This enables the prediction of OOD performance from ID performance.
    %
    %
    \vspace{-1mm}
\end{itemize}
The findings above are rigorously identified by a large-scale, systematic, analysis for the first time. 
Furthermore, our analysis also offer several novel insights into the OOD generalization behavior of adversarial robustness:
\begin{itemize}
    \vspace{-1mm}
    \item \textbf{The higher the ID robustness of the model, the more robustness degrades under distribution shift.} 
    This suggests that while great progress has been made on improving ID robustness, we only gain \emph{diminishing returns} under distribution shift.
    \vspace{-1mm}
    \item \textbf{An abnormal catastrophic drop in robustness under noise shifts is observed in some methods.} 
    For instance, under Gaussian noise shift, HAT \citep{rade_reducing_2022} suffers from a severe drop of robustness by 46\% whereas the average drop is 9\%.
    \vspace{-1mm}
    \item \textbf{Adversarial training boosts the correlation between ID and OOD performance under corruption shifts}, 
    and thus, improves the fidelity of using ID performance for model selection and OOD performance prediction.
    \vspace{-1mm}
    \item \textbf{\lp robustness correlates poorly with \nonlp robustness}.
    This suggests that \nonlp robustness cannot be predicted from \lp robustness, and enhancing \lp robustness does not necessarily result in improved \nonlp robustness. 
    
    \vspace{-1mm}
\end{itemize}

Last, we investigate how to achieve OOD adversarial robustness.
First, based on the discovered linear trend, we predict the best available OOD performance for the existing \lp-based robustness methodology and find that \textbf{existing methods are unlikely to achieve high OOD adversarial robustness} (e.g. the predicted upper bound of OOD robustness under the dataset shifts is only 43\% on ImageNet \linf). 
Next, we examine a wide range of techniques for achieving OOD adversarial robustness beyond the above prediction.
Most of these techniques, including training with extra data, data augmentation, advanced model architectures, scaling-up models and unsupervised representation learning, have limited or no benefit. However, we do identify several adversarial training methods \citep{dai_formulating_2022,pang_boosting_2020,ding_mma_2020,bai_improving_2023} that have the potential to exceed the prediction and produce higher OOD adversarial robustness.

Overall, this work reveals that most existing robust models including the state-of-the-art ones are vulnerable to distribution shifts and demonstrates that the existing approaches to improve ID robustness may be insufficient to achieve high OOD robustness. 
To ensure safe deployment in the wild, we advocate for the assessment of OOD robustness in future models and for the development of new approaches that can cope with distribution shifts better and achieve OOD robustness beyond our prediction.

%
%

\section{Related Works}
\textbf{Robustness under dataset shift}. Early work \citep{sehwag_analyzing_2019} studied the generalization of robustness to novel classes that are unseen during training.
On the other hand, our setup only considers the input distribution shift and not the unforeseen classes.
Recently, \citet{sun_spectral_2022} studied the OOD generalization of certified robustness under corruption shifts for a few state-of-the-art methods. 
In contrast, we focus on empirical robustness instead of certified robustness. \citet{alhamoud_generalizability_2023} is the most relevant work. 
They studied the generalization of robustness from multiple source domains to an unseen domain. 
Different from them, the models we examine are trained on only one source domain, which is the most common set-up in the existing works of adversarial training \citep{croce_robustbench_2021}. 
Moreover, we also cover much more diverse distribution shifts, models and training methods than \citet{sun_spectral_2022} and \citet{alhamoud_generalizability_2023} so that the conclusion drawn in this work is more general and comprehensive. 

\textbf{Robustness against unforeseen adversarial threat models}.
It was observed that naive adversarial training \citep{madry_towards_2018} with only one single \lp threat model generalizes poorly to unforeseen \lp threat models, e.g., higher perturbation bound \citep{stutz_confidence-calibrated_2020}, different \pnorm \citep{tramer_adversarial_2019, maini_adversarial_2020,croce2022adversarial}, or \nonlp threat models including color transformation ReColor~\citep{laidlaw_functional_2019}, spatial transformation StAdv~\citep{xiao_spatially_2018}, LPIPS-bounded attacks PPGD and LPA~\citep{laidlaw_perceptual_2021} and many others~\citep{kaufmann_testing_2023}.
We complement the existing works by conducting a large-scale analysis on the unforeseen robustness of \lp robust models trained by varied methods and training set-ups. 
We are thus able to provide new insights into the generalization of robustness to unforeseen threat models and identify effective yet previously unknown approaches to enhance unforeseen robustness.

More related works are discussed in \cref{appendix: additional related works}. 

\section{OOD Adversarial Robustness Benchmark}
This section first proposes the benchmark OODRobustBench and then presents the benchmark results for state-of-the-art robust models.

\input{tables/benchmark-cifar10-linf}

\subsection{OODRobustBench}
OODRobustBench is designed to simulate the possible data distribution shifts that might occur in the wild and evaluate adversarial robustness in the face of them.
It focuses on two types of distribution shifts: dataset shift and threat shift.
\emph{Dataset shift}, \oodd, denotes the distributional difference between training and test raw datasets.
\emph{Threat shift}, \oodt, denotes the difference between training and evaluation \emph{threat models}, a special type of distribution shift.
The original test set drawn from the same distribution as the training set is considered ID.
The variant dataset with the same classes yet where the distribution of the inputs differs is considered OOD.
%

\textbf{Dataset shift}.
To represent diverse data distribution in the wild, OODRobustBench includes multiple types of dataset shifts from two sources: \emph{natural} and \emph{corruption}.
For natural shifts, we adopt four different variant datasets per source dataset:
CIFAR10.1 \citep{recht_cifar-10_2018}, CIFAR10.2 \citep{lu_harder_2020}, CINIC \citep{darlow_cinic-10_2018}, and CIFAR10-R \citep{hendrycks_many_2021} for CIFAR10, and ImageNet-v2 \citep{recht_imagenet_2019}, ImageNet-A \citep{hendrycks_natural_2021}, ImageNet-R \citep{hendrycks_many_2021}, and ObjectNet \citep{barbu_objectnet_2019} for ImageNet.
For corruption shifts, we adopt, from the corruption benchmarks~\citep{hendrycks_benchmarking_2019}, 15 types of common corruption in four categories: Noise (gaussian, impulse, shot), Blur (motion, defocus, glass, zoom), Weather (fog, snow, frost) and Digital (brightness, contrast, elastic, pixelate, JPEG).
Each corruption has five levels of severity.
Overall, the dataset-shift testbed consists of 79 ($4+15\times5$) subsets.
\cref{appendix: benchmark datasets} gives the details of the above datasets and data processing.

Accuracy and robustness are evaluated on the ID and OOD dataset.
To compute the overall performance of \oodd, we first average the result of natural and corruption shifts: 
\begin{align}
    &R_{c}(f) = \mathbb{E}_{i \in \{\text{corruptions}\}, j \in \{\text{severity}\}} R_{i, j}(f) \\
    &R_n(f) = \mathbb{E}_{i \in \{\text{naturals}\}} R_i(f)
\end{align}
where $R(\cdot)$ returns accuracy or adversarial robustness and $f$ denotes the model to be assessed. Next, we average the above two results to get the overall performance of the dataset shift as
\begin{equation}
    R_{ood}(f) = (R_c(f) + R_n(f)) / 2 \label{equ: overall eval metric}
\end{equation}

To evaluate a model, OODRobustBench performs 80 (79 for \oodd and 1 for ID) runs of adversarial evaluation.
This makes computationally expensive attacks like AutoAttack \citep{croce_reliable_2020} impractical to use.
To balance efficiency and effectiveness, we use MM5 \citep{gao_fast_2022} for robustness evaluation. 
MM5 is approximately 32$\times$ faster than AutoAttack \citep{gao_fast_2022} while achieving similar results, as verified in \cref{appendix: adversarial evaluation} alongside the results of evaluations using alternative attacks.
The perturbation bound $\epsilon$ is 8/255 for CIFAR10 \linf, 0.5 for CIFAR10 \lt and 4/255 for ImageNet \linf.

\textbf{Threat shift}.
OODRobustBench adopts six unforeseen attacks as in \citet{laidlaw_perceptual_2021,dai_formulating_2022} to simulate threat shifts.
They are categorized into two groups, \lp and \nonlp, according to whether they are bounded by the \lp norm or not.
The \lp shift group includes MM attacks with the same \pnorm but larger $\epsilon$ and with different \pnorm.
The \nonlp shift group includes the imperceptible, PPGD and LPA, and perceptible, ReColor and StAdv, attacks.
The overall robustness under threat shift, \oodt, is simply the mean of these six unforeseen attacks.
These attacks are selected because they cover a wide range of different scenarios of threat shift and each of them is representative of its corresponding category (100+ cites). 
We are aware of alternative \nonlp attacks \citep{kaufmann_testing_2023} but do not include them due to the constraint of computational resource. 

We follow the same setting as \citet{laidlaw_perceptual_2021,dai_formulating_2022} to configure the above attacks since this has been well tested to be effective. 
The \lp attacks use $\epsilon=12/255$ and $\epsilon=0.5$ for \linf and \lt threats on CIFAR10 \linf, $\epsilon=8/255$ and $\epsilon=1$ for \linf and \lt threats on CIFAR10 \lt and on ImageNet \linf.
The perturbation bound is 0.5 for PPGD, 0.5 for LPA, 0.05 for StAdv and 0.06 for ReColor. The number of iterations is 40 for PPGD and LPA regardless of dataset, is 100 for StAdv and ReColor on CIFAR10 and 200 on ImageNet.

\textbf{Criteria for robust models}
are described in \cref{appendix: model criteria} and are the same as RobustBench \cite{croce_robustbench_2021}.

\subsection{OOD Performance and Ranking}
\label{ssec:ranking}
\begin{figure*}
    \centering
    \begin{subfigure}{\linewidth}
        \includegraphics[width=\linewidth, trim=4 7 3 4,clip]{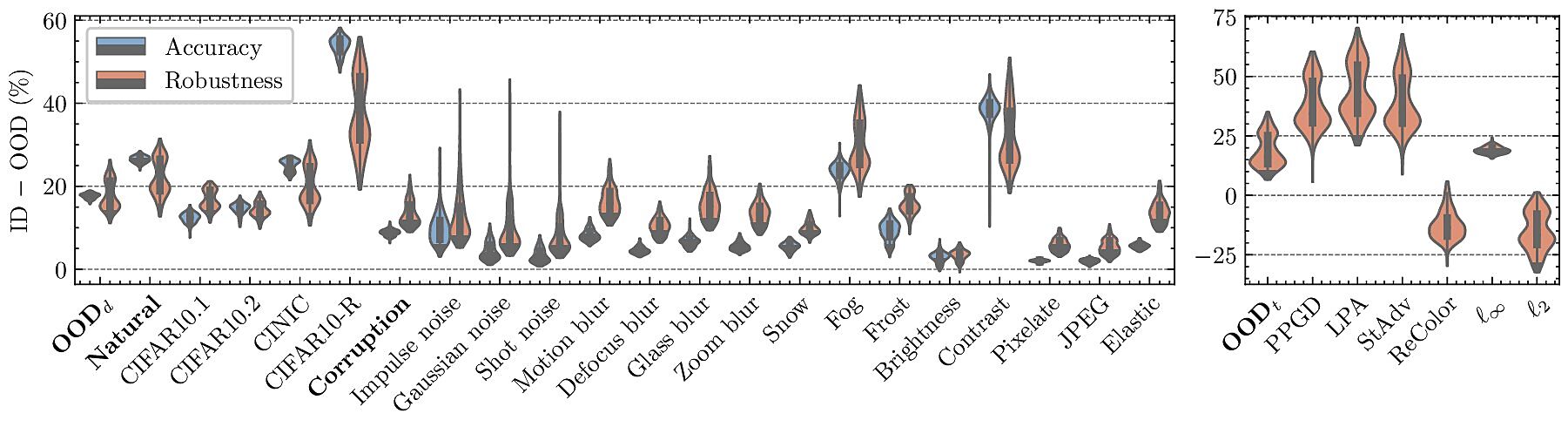}
    \end{subfigure}    
    \caption{\textbf{Degradation of accuracy and robustness under various distribution shifts for CIFAR10 \linf}.}
    \label{fig: degradation violin CIFAR10 linf}
\end{figure*}

The benchmark results for CIFAR10 \linf, \lt and ImageNet \linf are in \cref{tab: benchmark result,tab: benchmark result cifar10 l2,tab: benchmark result imagenet linf} respectively.

\textbf{Robustness degrades significantly under distribution shift}.
For models trained to be robust for CIFAR10 \linf (\cref{fig: degradation violin CIFAR10 linf}), CIFAR10 \lt (\cref{fig: degradation violin cifar10 l2}) and ImageNet \linf (\cref{fig: degradation violin imagenet}), the average drop in robustness (ID adversarial accuracy - OOD adversarial accuracy) is 18\%/20\%/27\% under dataset shift and 18\%/42\%/22\% under threat shift.
Robustness degradation is much severe for a subset of shifts: whereas the average robustness degradation of \oodd is 18\% on CIFAR10 \linf, some shifts like CIFAR10-R, fog and contrast degrade by 38\%, 30\% and 32\%, respectively.

\textbf{The higher the ID robustness of the model, the more robustness degrades under the shifts}.
For example, the top method in \cref{tab: benchmark result} degrades by 30\% of robustness, while the bottom method degrades by only 18\%.
This suggests that while the great progress has been made on improving ID robustness, we only gain diminishing returns under the distribution shifts.
%
Besides, in \cref{fig: degradation violin CIFAR10 linf}, the distribution of robustness degradation for most shifts spreads over a wide range, suggesting a large variation across individual models.

\textbf{Robustness degradation under noise shifts can be abnormally catastrophic} (the outliers under noise shifts in \cref{fig: degradation violin CIFAR10 linf}).
This issue is most severe on \cite{rade_reducing_2022} whose robustness falls by 43\%/46\%/38\% under impulse/Gaussian/shot noise, whereas the average drop is 12\%/9\%/8\% (discussed in \cref{appendix: catastrophic degradation}).
A similar yet milder drop is also observed on \citet{debenedetti_light_2023} and models trained with some advanced data augmentations like AutoAugment \citep{cubuk_autoaugment_2019}.

\textbf{Higher ID robustness generally implies higher OOD robustness but not always} (see the last two columns of \cref{tab: benchmark result,tab: benchmark result cifar10 l2,tab: benchmark result imagenet linf}).
For example, in \cref{tab: benchmark result}, the ranking of \citet{rade_reducing_2022} drops from 22 to 57 due to catastrophic degradation, while the ranking of \citet{pang_boosting_2020} jumps from 70 to 3 due to its superior robustness under threat shift (analyzed in \cref{sec: robust intervention training}).



\section{Linear Trend and OOD Prediction}\label{sec: ID OOD correlation}
It was previously observed that OOD accuracy is strongly correlated with ID accuracy under many dataset shifts for Standardly-Trained (ST) models \citep{miller_accuracy_2021}. 
This property is important since it enables the model selection and OOD performance prediction through ID performance.
Nevertheless, it is unclear if such correlation still holds for adversarial robustness.
This is particularly intriguing because accuracy and robustness usually go in opposite directions: i.e. there is a trade-off between accuracy and robustness \citep{tsipras_robustness_2019}. 
Furthermore, the threat shifts as a scenario of OOD are unique to adversarial evaluation and were, thus, never explored in the previous studies of accuracy trends.
Surprisingly, we find that ID and OOD robustness also have a linear correlation under many distribution shifts. 
It is even more surprising that the correlation for AT models is much stronger than that for ST models.


%
The following result is based on a large-scale analysis including over 60K OOD evaluations of 706 models.
187 of these models were retrieved from RobustBench or other published works so as to include current state-of-the-art methods, and the remaining models were trained by ourselves.
These models are mainly trained in three set-ups: CIFAR10 \linf, CIFAR10 \lt and ImageNet $\ell_{\infty}$.
They cover a wide range of model architectures, model sizes, data augmentation methods, training and regularization techniques. More detail is given in \cref{appendix: model zoo}.

\begin{figure*}
    \centering
    \begin{subfigure}{\linewidth}
        \includegraphics[width=\linewidth]{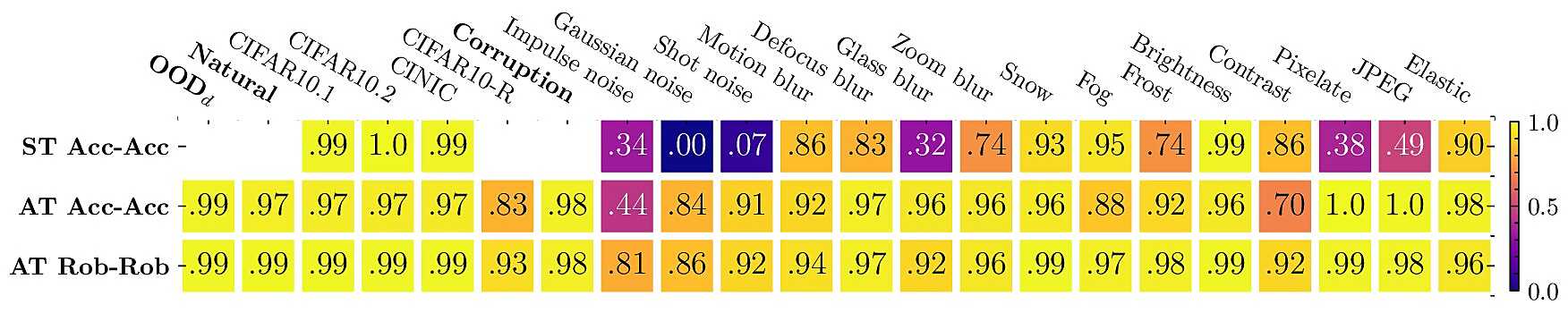}
    \end{subfigure}

    \caption{\textbf{\rsquare of regression between ID and OOD performance for Standardly-Trained (ST) and Adversarially-Trained (AT) models under various dataset shifts for CIFAR10 \linf}. Higher \rsquare implies stronger linear correlation. The results for ST models were copied from \cite{miller_accuracy_2021}. Some results of ST are missing (blank cells) because they were not reported in \cite{miller_accuracy_2021}.}
    \label{fig: dataset shfit correlation r2}
\end{figure*}

\subsection{Linear Trend under Dataset Shift} \label{sec: dataset shifts correlation dataset shfit}
This section studies how ID and OOD accuracy/robustness correlate under dataset shifts.
We fit a linear regression on four pairs of metrics (Acc-Acc, Rob-Rob, Acc-Rob, and Rob-Acc) for each dataset shift and each training setup (CIFAR10 \linf, CIFAR10 \lt and ImageNet \linf).
Taking Acc-Rob as an example, a linear model is fitted with ID accuracy as the observed variable $\bm{x}$ and OOD adversarial robustness as the target variable $\bm{y}$.
The result for each shift is given in \cref{appendix: correlation per dataset shift}.
Below are the major findings. 
 
\textbf{ID accuracy (resp. robustness) strongly correlates with OOD accuracy (resp. robustness) in a linear relationship} for most dataset shifts.
In \cref{fig: dataset shfit correlation r2,fig: dataset shfit correlation r2 cifar10 l2,fig: dataset shfit correlation r2 imagenet linf}, the regression of Acc-Acc and Rob-Rob for most shifts achieve very high $R^2$ ($> 0.9$), i.e., their relationship can be well explained by a linear model.
This suggests for these shifts \uline{ID performance is a good indication of OOD performance}, and more importantly, \uline{OOD performance can be reliably predicted by ID performance using the fitted linear model}.

\textbf{Nevertheless, under some shifts, ID and OOD performance are only weakly correlated.}
Natural shifts like CIFAR10-R and ImageNet-A and corruption shifts like noise,
fog and contrast are observed to have relatively low \rsquare across varied training set-ups in \cref{fig: dataset shfit correlation r2,fig: dataset shfit correlation r2 cifar10 l2,fig: dataset shfit correlation r2 imagenet linf}.
It can be seen from \cref{fig: inferior dataset shifts cifar linf,fig: inferior dataset shifts cifar l2} that the correlation for these shifts becomes even weaker, and the gap of \rsquare between them and the others expands, as more inferior (relatively worse accuracy and/or robustness) models are excluded from the regression.
This suggests that the models violating the linear trend are mostly high-performance. 
\cref{appendix: inferior models correlation} discusses how inferior models are identified and how they influence the correlation.

\textbf{AT models exhibit a stronger linear correlation between ID and OOD accuracy} under most corruption shifts on CIFAR10 in \cref{fig: dataset shfit correlation r2,fig: dataset shfit correlation r2 cifar10 l2}.
The improvement is dramatic for particular shifts.
For example, \rsquare surges from nearly 0 (no linear correlation) for ST models to around 0.8 (evident linear correlation) for AT models with Gaussian and shot noise data shifts.
These results are contrary to the previous finding on ST models~\citep{miller_accuracy_2021}. However, note that we measure linear correlation for the raw data, whereas \cite{miller_accuracy_2021} applies a nonlinear transformation to the data to promote linearity.
Overall, \uline{adversarial training boosts linear correlation for corruption shifts, and hence, improves the faithfulness of using ID performance for model selection and OOD performance prediction}.

We attribute this to AT improving accuracy on the corrupted data~\citep{kireev_effectiveness_2022}.
Intuitively, ST models have less correlated corruption accuracy because corruption significantly impairs accuracy and such effect varies a lot among models.
Compared to ST, AT effectively mitigates the effect of corruption on accuracy, and hence, reduces the divergence of corruption accuracy so that corruption accuracy is more correlated to ID accuracy. 

Last, we observe \textbf{no evident correlation when ID and OOD metrics misalign, i.e., Acc-Rob and Rob-Acc for CIFAR10}, but weak correlation for ImageNet \linf as shown in \cref{fig: dataset shfit correlation r2 CAAC}. 
This is due to the varied trade-off between accuracy and robustness of different models (discussed in details in
\cref{appendix: why no correlation metrics misalign})

\subsection{Linear Trend under Threat Shift}

\begin{figure}
    \centering
    \includegraphics[width=.7\linewidth]{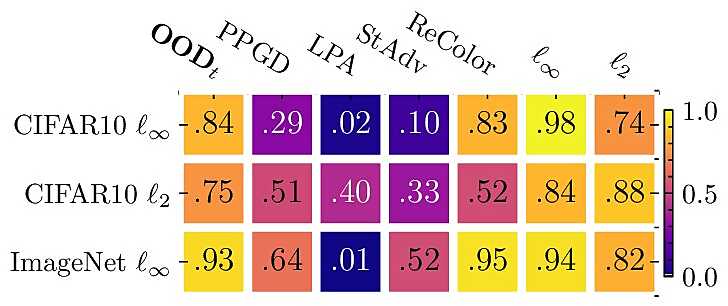}
    \caption{\textbf{\rsquare of regression between ID seen robustness and OOD unforeseen robustness, i.e., threat shift}.}
    \label{fig: seen unforeseen correlation}
\end{figure}

This section studies the relationship between seen and unforeseen robustness. 
Both seen and unforeseen robustness are computed using only ID data yet with different attacks.
%
Linear regression is then conducted between seen robustness ($\bm{x}$) and unforeseen robustness ($\bm{y}$). 
The result of regression for each threat shift is given in \cref{appendix: correlation per threat shift}. The sensitivity of the regression results to the composition of the model zoo is discussed in \cref{appendix: inferior models correlation}.

\textbf{\lp robustness correlates poorly with \nonlp robustness}.
\rsquare of the regression between ID \lp robustness and PPGD, LPA and StAdv robustness is low in \cref{fig: seen unforeseen correlation}.
Particularly, \rsquare is close to 0 for \linf-LPA and \linf-StAdv on CIFAR10 \linf suggesting no correlation at all.
As shown in \cref{fig: threat shift plots cifar10 linf,fig: threat shift plots cifar10 l2,fig: threat shift imagenet}, the increase in ID \lp robustness leads to only slight or even no improvement on unforeseen robustness esp. for LPA and StAdv.
Interestingly, despite poor correlation with PPGD, LPA and StAdv, ID \lp robustness is well correlated with ReColor unforeseen robustness.

\textbf{\lp robustness correlates strongly with \lp robustness of different $\epsilon$ and $p$-norm}. 
\rsquare of their regression is higher than 0.7 across all assessed set-ups in \cref{fig: seen unforeseen correlation} suggesting a consistently strong linear correlation. The correlation between different $\epsilon$ of the same \pnorm is stronger than the correlation between different $p$-norm. 

\subsection{Unsupervised OOD Robustness Prediction}
\begin{figure}
    \centering

    \begin{subfigure}{.49\linewidth}
        \includegraphics[width=\linewidth]{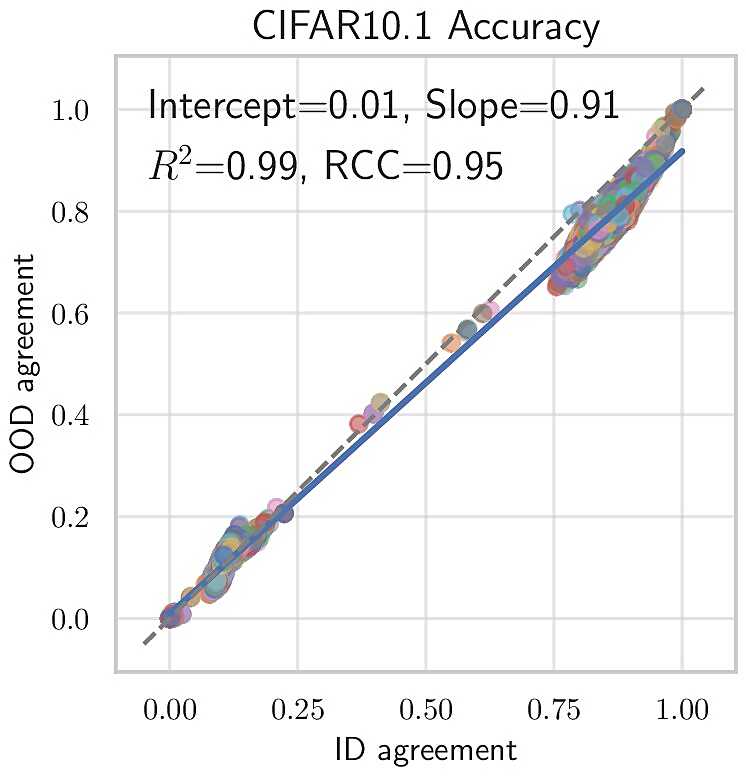}
    \end{subfigure}
    \begin{subfigure}{.49\linewidth}
        \includegraphics[width=\linewidth]{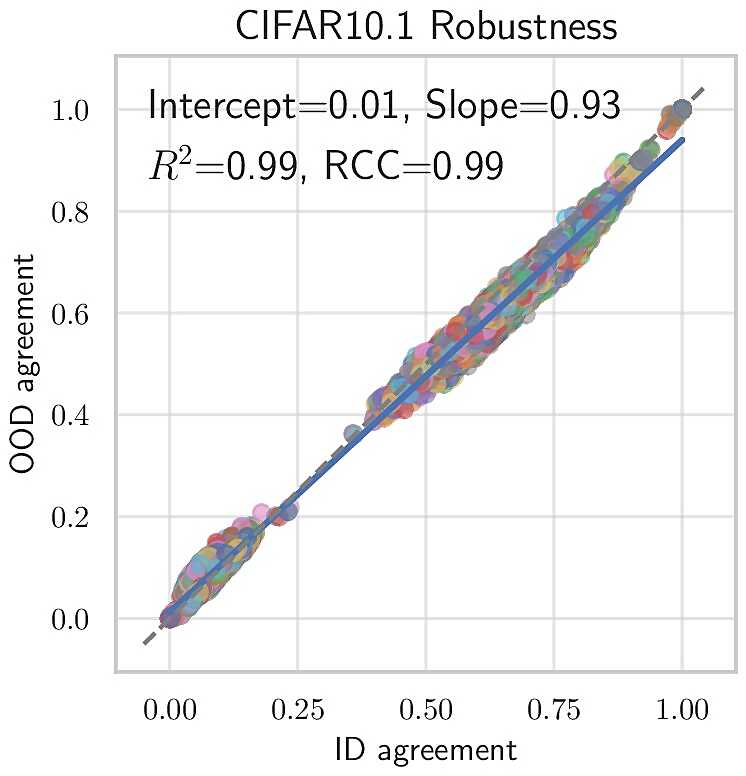}
    \end{subfigure}

    \caption{\textbf{Correlation between ID and OOD prediction agreement on adversarial examples for CIFAR10 \linf AT models}. Each point represents the prediction agreement of two models.}
    \label{fig: agreement line}
\end{figure}

The linear trends discovered above enable the prediction of OOD performance only if labeled OOD data is available. 
There is a line of works \citep{baek_agreementontheline_2022,deng_are_2021,garg_leveraging_2022} showing that OOD accuracy can be predicted with only unlabeled OOD data. 
We study here if OOD adversarial robustness can be predicted, similarly, in an unsupervised manner. 
We run the experiments with CIFAR-10 \linf models for CIFAR-10.1 (\cref{fig: agreement line}) and Impulse noise (\cref{fig: agreement line impulse noise}) shifts and find that a linear trend is also observed in the agreement between the predictions of any pair of two robust models: \rsquare is 0.99 for CIFAR-10.1 shift and 0.95 for Impulse noise shift. This suggests that the unsupervised method \citep{baek_agreementontheline_2022} is also effective in predicting OOD adversarial robustness.

\section{Incompetence in OOD Generalization}
\label{sec: predict upper limit}

Based on the precise linear trend observed above for existing robust training methods, we can predict the OOD performance of a model trained by such a method from its ID performance using the fitted linear model. 
Furthermore, we can extrapolate from current trends to predict the maximum OOD robustness that can be expected from a hypothetical future model that achieves perfect robustness on ID data (assuming the linear trend continues). 
\begin{equation}
    \text{slope} \times 100 + \text{intercept}.
\end{equation}
This estimates the best OOD performance one can expect by fully exploiting existing robust training techniques. Note that a wide range of models and techniques (\cref{appendix: model zoo}) are covered by our correlation analysis so their, as well as their variants', OOD performance should be (approximately) bounded by the predicted upper limit. The accuracy of the prediction depends on the \rsquare of the correlation.

\begin{figure}
    \centering
    \includegraphics[width=\linewidth]{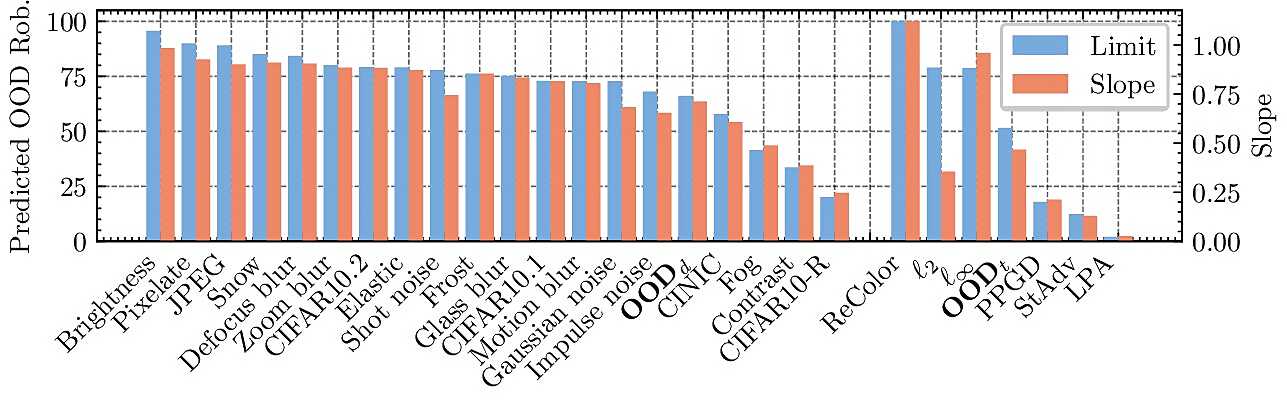}
    \caption{\textbf{The estimated upper limit of OOD robustness, and the slope, of OOD robustness from ID robustness under various distribution shifts for CIFAR10 \linf}. The estimated upper limit is capped by 100\% as robustness can not surpass 100\%.}
    \label{fig: predicted upper OOD limit and conversion rate cifar10 linf}
\end{figure}

We find that \textbf{continuously improving ID \lp robustness following existing practice is unlikely to achieve high OOD adversarial robustness}.
The upper limit of OOD robustness under dataset shift, \oodd, is 66\%/71\%/43\% for CIFAR10 \linf (\cref{fig: predicted upper OOD limit and conversion rate cifar10 linf}), CIFAR10 \lt (\cref{fig: predicted upper limit and slope}) and ImageNet \linf (\cref{fig: predicted upper limit and slope}) respectively, and under threat shift \oodt is 52\%/35\%/52\% correspondingly.
Hence, if current trends continue, the resulting models are likely to be very unreliable in real-world applications.
The vulnerability of models is most evident for ImageNet \linf under dataset shift and for CIFAR10 \lt under threat shift.
The expected upper limit of OOD robustness also varies greatly among individual shifts ranging from nearly 0 to 100\%.

One of the accounts for this issue is that \uline{the existing methods have poor conversion rate to OOD robustness from ID robustness} as shown by the slope of the linear trend in \cref{fig: predicted upper OOD limit and conversion rate cifar10 linf,fig: predicted upper limit and slope}. Taking an example of fog shift on ImageNet, the slope is roughly 0.1 so improving 10\% ID robustness can only lead to 1\% improvement on fog robustness. Besides, the upper limit and conversion rate of robustness are observed to be much lower than those of accuracy in \cref{fig: predicted upper limit and slope}, suggesting the OOD generalization issue is more severe for robustness. 
Overall, this issue calls for developing novel methods that can improve OOD robustness beyond our prediction.

\section{Improving OOD Adversarial Robustness}
\begin{figure*}
    \centering
    \begin{subfigure}{.24\linewidth}
        \includegraphics[width=\linewidth]{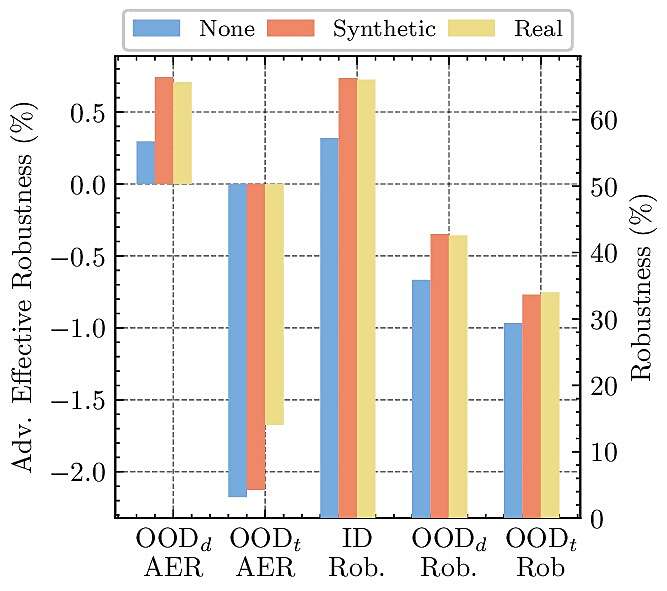}
        \caption{Extra data}
        \label{fig: intervention extra data}
    \end{subfigure}
    \begin{subfigure}{.245\linewidth}
        \includegraphics[width=\linewidth]{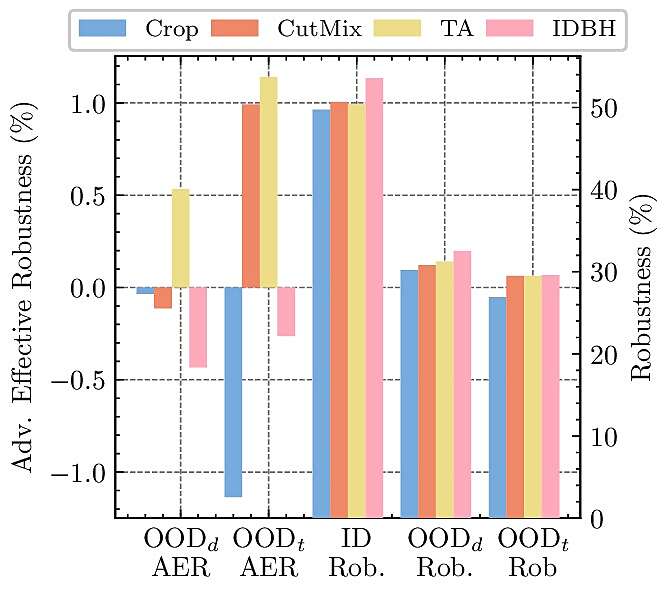}
        \caption{Data aug.}
        \label{fig: intervention data aug}
    \end{subfigure}
    \begin{subfigure}{.24\linewidth}
        \includegraphics[width=\linewidth]{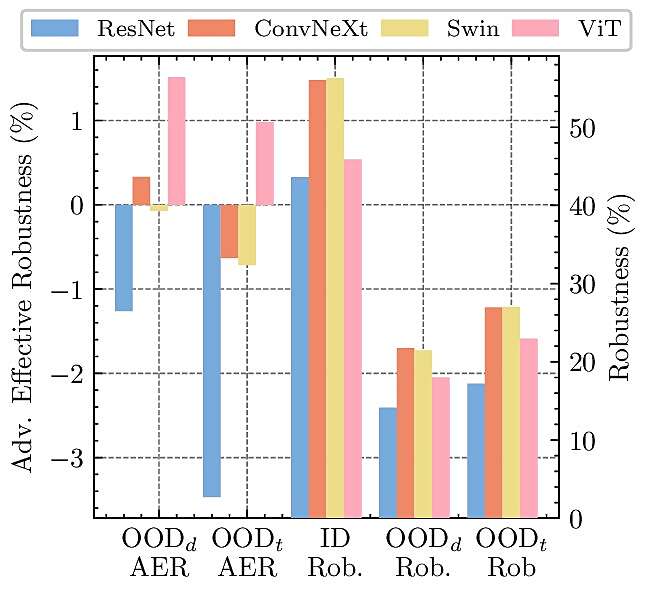}
        \caption{Model architecture}
        \label{fig: intervention model architecture}
    \end{subfigure}
    \begin{subfigure}{.245\linewidth}
        \includegraphics[width=\linewidth]{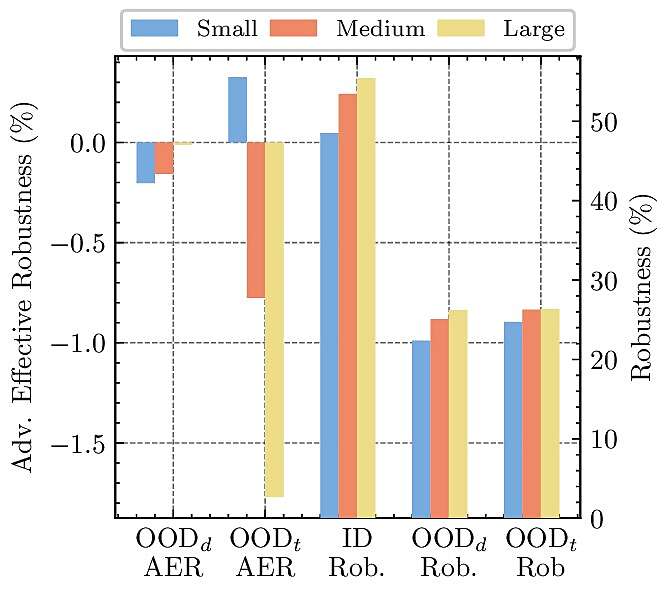}
        \caption{Model size}
        \label{fig: intervention model size}
    \end{subfigure}
    
    \begin{subfigure}{.245\linewidth}
        \includegraphics[width=\linewidth]{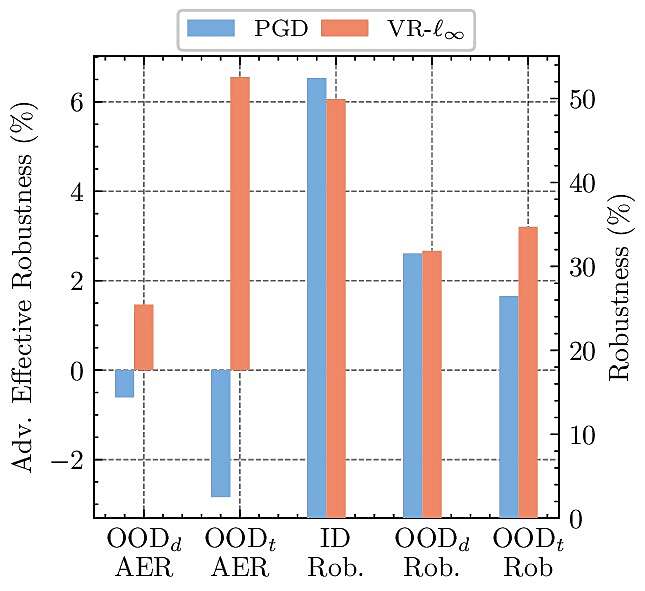}
        \caption{VR}
        \label{fig: intervention vr}
    \end{subfigure}
    \begin{subfigure}{.245\linewidth}
        \includegraphics[width=\linewidth]{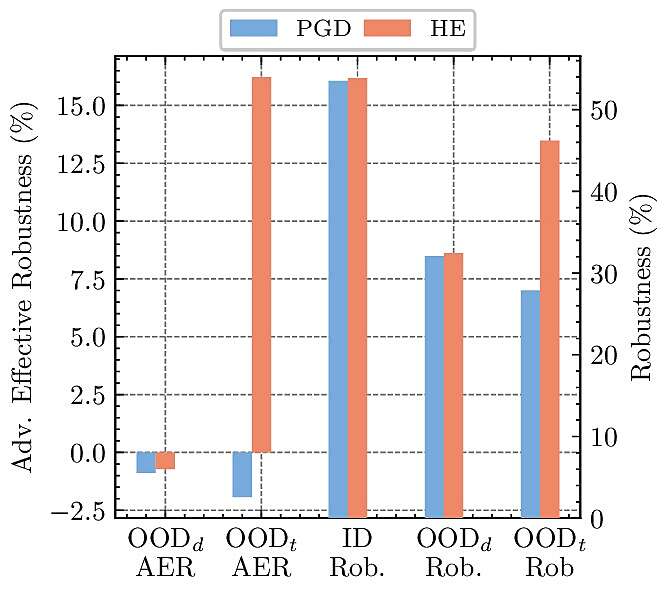}
        \caption{HE}
        \label{fig: intervention he}
    \end{subfigure}
    \begin{subfigure}{.245\linewidth}
        \includegraphics[width=\linewidth]{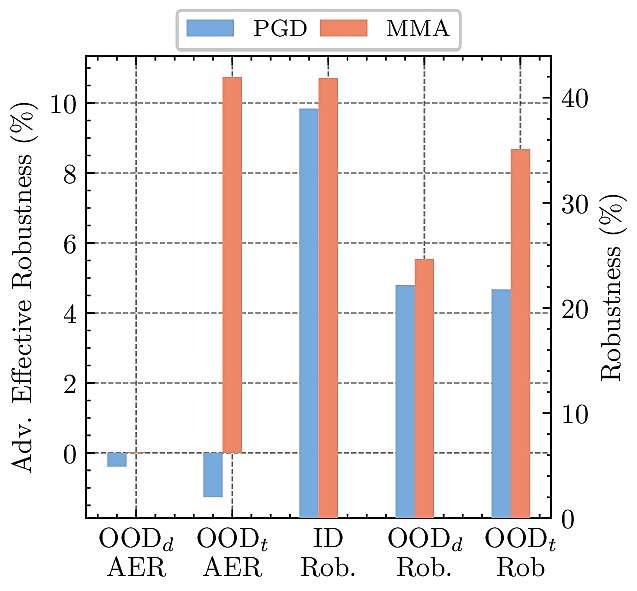}
        \caption{MMA}
        \label{fig: intervention mma}
    \end{subfigure}
    \begin{subfigure}{.245\linewidth}
        \includegraphics[width=\linewidth]{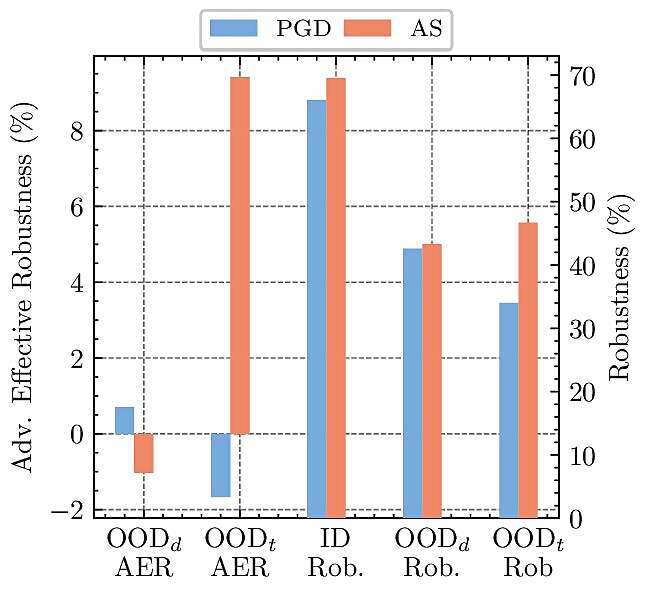}
        \caption{AS}
        \label{fig: intervention as}
    \end{subfigure}
    \caption{\textbf{The robustness (Rob.) and adversarial effective robustness (AER) of various robust techniques}.}
    \label{fig: effective robust intervention main text}
\end{figure*}

To inspire the design of methods that have OOD robustness exceeding the above prediction, this section investigates methods that have the potential to be effective for boosting the OOD generalization of robustness. The effectiveness is quantified by two metrics: OOD performance and effective performance. Effective performance measures the extra resilience of a model under distribution shift when compared to a group of models by adapting the metric of ``Effective Robustness" \cite{taori_measuring_2020}: 
\begin{equation}
    R'(f) = R_{ood}(f) - \beta (R_{id}(f))
    \label{equ: effective accuracy and robustness}
\end{equation}
where $\beta(\cdot)$ is a linear mapping from ID to OOD metric fitted on a group of models. We name this metric effective robustness (adversarial effective robustness) when $R_{id}$ and $R_{ood}$ are accuracy (robustness). A positive adversarial effective robustness means that $f$ achieves adversarial robustness above what the linear trend predicts based on its ID performance, i.e., $f$ is advantageous over the fitted models on OOD generalization. 
Note that higher adversarial effective robustness is not equivalent to higher OOD robustness since the model may have a lower ID robustness. 
The specific set-ups and detailed results of the following experiments are described in \cref{appendix: effective robust intervention}.

\begin{table*}[!htp]
\centering
\caption{\textbf{The performance of non-single-\lp defense and OOD generalization methods under distribution shift on CIFAR10 \linf}. 
The AER of StAdv AT, PLAT and CARD-Deck is invalid (``-'') because of their (nearly) 0 ID/ODD robustness.}
\label{tab: nonlp defense}
\resizebox{\textwidth}{!}{%
\begin{tabular}{@{}lllccccccccc@{}}
\toprule
\multicolumn{1}{c}{\multirow{2}{*}{Defense}} & \multicolumn{1}{c}{\multirow{2}{*}{Method}} & \multicolumn{1}{c}{\multirow{2}{*}{Model}} & \multicolumn{2}{c}{ID} & \multicolumn{4}{c}{\oodd} & \multicolumn{3}{c}{\oodt} \\ \cmidrule(l){4-5}\cmidrule(l){6-9}\cmidrule(l){10-12} 
\multicolumn{1}{c}{} & \multicolumn{1}{c}{} & \multicolumn{1}{c}{} & Acc. & Rob. & Acc. & Rob. & ER & AER & \nonlp & \lp & Avg. \\ \midrule
Supervised \lp defense & PGD \lp AT & ResNet18 & 83.4 & \textbf{49.4} & 64.6 & {\ul 30.2} & -1.1 & 0.2 & 19.1 & \textbf{45.5} & 27.9 \\ \midrule
Self-supervised \lp defense & ACL & ResNet18 & 82.3 & \textbf{49.4} & 64.2 & 30.1 & -0.6 & 0.1 & 20.2 & {\ul 45.1} & 28.5 \\ \midrule
\multirow{4}{*}{\nonlp attack defense} & ReColor AT & ResNet50 & 93.4 & 8.5 & 78.0 & 3.4 & 2.8 & \textbf{2.6} & 16.4 & 18.1 & 17.0 \\
 & StAdv AT & ResNet50 & 86.2 & 0.1 & 66.8 & 0.0 & -1.6 & - & 9.5 & 3.4 & 7.4 \\
 & PAT-Alexnet & ResNet50 & 71.6 & 28.6 & 57.1 & 16.8 & 2.3 & 1.6 & {\ul 48.4} & 33.1 & {\ul 43.3} \\
 & PAT-Self & ResNet50 & 82.4 & 30.4 & 66.3 & 16.8 & 1.4 & 0.3 & 32.1 & 36.6 & 33.6 \\ \midrule
\multirow{2}{*}{Unforeseen attack defense} & VR & ResNet18 & 72.9 & {\ul 48.9} & 56.4 & \textbf{31.4} & 0.5 & {\ul 1.7} & 24.8 & 43.3 & 31.0 \\
 & PAT+VR & ResNet50 & 72.5 & 29.4 & 56.8 & 17.4 & 1.3 & {\ul 1.7} & \textbf{55.0} & 33.6 & \textbf{47.8} \\ \midrule
\multirow{2}{*}{Composite attack defense} & GAT-f & ResNet50 & 82.3 & 38.7 & 66.2 & 22.2 & 1.4 & -0.2 & 14.8 & 17.0 & 15.5 \\
 & GAT-fs & ResNet50 & 82.1 & 41.9 & 65.7 & 24.8 & 1.2 & 0.1 & 17.2 & 18.3 & 17.5 \\ \midrule
\multirow{3}{*}{Multi-attack defense} & MAAT-Average & ResNet50 & 86.8 & 39.9 & 70.7 & 22.4 & 1.7 & -0.8 & 25.0 & 41.7 & 30.5 \\
 & MAAT-Max & ResNet50 & 84.0 & 25.6 & 68.1 & 13.1 & 1.8 & 0.0 & 30.2 & 29.8 & 30.0 \\
 & MAAT-Random & ResNet50 & 85.2 & 22.1 & 67.7 & 10.5 & 0.2 & 0.0 & 12.7 & 31.0 & 18.8 \\ \midrule
 \multirow{2}{*}{OOD generalization method} & PLAT  & ResNet18               & {\ul 94.7}          & 0.1           & {\ul 80.3}          & 0.0           &  {\ul 4.0}          & -             & 0.0           & 0.0   & 0.0          \\
& CARD-Deck  &     WRN-18-2      & \textbf{96.5} & 1.0           & \textbf{83.5} & 0.5           & \textbf{5.4} & -             & 0.0           & 0.0    & 0.0         \\ \midrule
Multi-attack + OOD defense & MSD+REx & ResNet18 & 78.0 & 43.3 & 59.9 & 26.2 & -0.7 & 0.5 & 19.0 & 40.4 & 26.1 \\ \bottomrule
\end{tabular}%
}
\end{table*}

\subsection{Data}
\textbf{Training with extra data}
boosts both robustness and adversarial effective robustness compared to training schemes without extra data (see \cref{fig: intervention extra data}). 
There is no clear advantage to training with extra real data \citep{carmon_unlabeled_2019} rather than synthetic data \citep{gowal_improving_2021} except for the adversarial effective robustness under threat shift which is improved more by real data. 

\textbf{Advanced data augmentation}
improves robustness under both types of shifts and adversarial effective robustness under threat shift over the baseline augmentation RandomCrop (see \cref{fig: intervention data aug}). Nevertheless, advanced data augmentation methods other than TA \citep{muller_trivialaugment_2021} degrade adversarial effective robustness under dataset shift. 

\subsection{Model}
\textbf{Advanced model architecture}
greatly boosts robustness and adversarial effective robustness under both types of shift over the baseline ResNet \citep{he_deep_2016} (\cref{fig: intervention model architecture}). 
Among all tested architectures, ViT \citep{dosovitskiy_image_2021} achieves the highest adversarial effective robustness.

\textbf{Scaling model up}
improves robustness under both types of shift and adversarial effective robustness under dataset shift, but dramatically impairs adversarial effective robustness under threat shift (\cref{fig: intervention model size}). The latter is because increasing model size greatly improves ID robustness but not OOD robustness so that the real OOD robustness is much below the OOD robustness predicted by linear correlation.

\subsection{Adversarial Training} 
\label{sec: robust intervention training}

\textbf{VR} \citep{dai_formulating_2022}, the state-of-the-art defense against unforeseen attacks, greatly boosts adversarial effective robustness under threat shifts in spite of inferior ID robustness. 
Surprisingly, VR also clearly boosts adversarial effective robustness under dataset shift even though not designed for dealing with these shifts. 

Training methods \textbf{HS} \citep{pang_boosting_2020}, \textbf{MMA} \citep{ding_mma_2020} and \textbf{AS} \citep{bai_improving_2023} achieve an AER of 16.22\%, 10.74\% and 9.41\%, respectively, under threat shift, which are much higher than the models trained with PGD. Importantly, in contrast to VR, these methods also improve ID robustness resulting in a further boost on OOD robustness. This makes them a potentially promising defense against multi-attack \citep{dai_multirobustbench_2023}.

\subsection{OOD Generalization Methods}
Two leading methods, CARD-Deck \citep{diffenderfer_winning_2021} (ranked 1st) and PLAT \citep{kireev_effectiveness_2022}, from the common corruptions leaderboard of RobustBench are evaluated using our benchmark in \cref{tab: nonlp defense}. 
Despite the expected remarkable OOD clean generalization under \oodd shifts, they offer little or no adversarial robustness regardless of ID or OOD setting.
It suggests that OOD generalization methods alone do not help OOD adversarial robustness unless combined with adversarial training.

To test the effect of combining adversarial training with OOD generalization method, we evaluate a recent attempt in this direction, MSD+REx \citep{ibrahim_towards_2023}. 
This approach trains using the multi-attack defense MSD with various attacks and applies REx \citep{krueger2021out} by treating different attacks as separate domains. 
Surprisingly, as shown in \cref{tab: nonlp defense}, this purpose-built solution impairs OOD adversarial robustness under both dataset and threat shifts and offers no evident improvement in AER when compared to supervised $\ell_p$ adversarial training.

While these findings suggest that this specific implementation might be ineffective, the combination of AT and OOD methods remains a promising direction. 
Future work should focus on a more careful design for this integration.
One potential strategy could be to treat different groups of data, rather than attacks, as different domains like \citet{huang2023robust}.

\subsection{Unsupervised Representation Learning}
Unsupervised learning has been observed to train models that generalize to distribution shifts better than supervised learning \citep{shi2023robust,shen_towards_2021}. 
However, it is unclear whether or not unsupervised learning will benefit OOD adversarial robustness.
To test this we evaluated a model trained by Adversarial Contrastive Learning (ACL) \citep{jiang_robust_2020} which combines self-supervised contrastive learning with adversarial training. 
The effective robustness under dataset shift is 0.1\% (\cref{tab: nonlp defense}), suggesting only marginal benefit in improving OOD robustness.

\subsection{Non-single-\lp Defenses}
This section evaluates the OOD generalization capability of various defenses beyond the supervised single-attack \lp defense. 
We compared several specific methods, including AT with the color-based attack ReColor \citep{laidlaw_perceptual_2021}, the spatial attack StAdv \citep{laidlaw_perceptual_2021}, and the LPIPS-bound attack PAT-Alexnet/Self \citep{laidlaw_perceptual_2021}. 
Additionally, we examined composite attacks defense (e.g., color plus \lp) GAT-f/fs \citep{hsiung_towards_2023} and multiple attacks adversarial training (MAAT) \citep{tramer_adversarial_2019,maini_adversarial_2020} involving \lt, \linf, StAdv, and ReColor. 
MAAT has three variants: "Average" optimizes the average loss across all attacks, "Max" optimizes the maximum loss across all attacks, and "Random" selects a random attack at each training iteration.

Unfortunately, none of these defenses achieve high \oodd ER and AER in \cref{tab: nonlp defense}, indicating that they are not significantly better than the supervised single-attack \lp AT at handling OOD dataset distribution shifts. 
This reinforces our conclusion that achieving OOD adversarial robustness is challenging with existing methods and underscores the need to develop new approaches that effectively address distribution shifts.

While MAAT and PAT show substantial improvements over \lp AT in terms of robustness against \nonlp attacks, this is partly because some of the \nonlp attacks used were already encountered during their training, making them no longer unforeseen. 
This highlights the difficulty in benchmarking unforeseen robustness across different types of defenses.

\subsection{Summary}

The evaluated techniques, except for some AT methods (\cref{sec: robust intervention training}), achieve relatively limited or even no adversarial effective robustness. 
This suggests that applying them is unlikely to significantly change the linear trend in \cref{sec: ID OOD correlation} and thus the predicted upper limit of OOD robustness (\cref{sec: predict upper limit}).
In contrast, the methods identified in \cref{sec: robust intervention training} show the promise in achieving OOD performance beyond our prediction.
Another promising direction is to combine OOD generalization methods with adversarial training.

\section{Conclusions}
This work proposes a new benchmark to assess OOD adversarial robustness, provides many insights into the generalization of existing robust models under distribution shift and identifies several robust interventions beneficial to OOD generalization. We have analyzed the OOD robustness of hundreds of diverse models to ensure that we obtain generally applicable insights.
As we focus on general trends, our analysis does not provide a detailed investigation into individual methods or explain the observed outliers such as the catastrophic robustness degradation. However, OODRobustBench provides a tool for performing such more detailed investigations in the future. It also provides a means of measuring progress towards models that are more robust in real-world conditions and will, hopefully, spur the future development of such models. 

\section*{Acknowledgment}
The authors acknowledge the use of the research computing facility at King’s College London, King's Computational Research, Engineering and Technology Environment (CREATE). 
Lin Li was funded by the King's - China Scholarship Council (K-CSC).
Yifei Wang was supported by Office of Naval Research under grant N00014-20-1-2023 (MURI ML-SCOPE), NSF AI Institute TILOS (NSF CCF-2112665), and NSF award 2134108.
Chawin was supported in part by funds provided by the National Science Foundation (under grant 2229876), the KACST-UCB Center for Secure Computing, the Department of Homeland Security, IBM, the Noyce Foundation, Google, Open Philanthropy, and the Center for AI Safety Compute Cluster.
Any opinions, findings, and conclusions or recommendations expressed in this material are those of the author(s) and do not necessarily reflect the views of the sponsors.

\section*{Impact Statement}
This paper presents work whose goal is to advance the field of adversarial machine learning. 
There are many potential societal consequences of our work, none of which we feel are negative and must be specifically highlighted here.

\bibliography{references,more_ref}
\bibliographystyle{icml2024}

\newpage
\appendix
\onecolumn

\section{Additional Related Works}
\label{appendix: additional related works}

Except for a few exceptions \citep{geirhos_shortcut_2020,Sun_etal21,Rusak_etal20,gilmer_adversarial_2019}, previous work on generalization to input distribution shifts has not considered adversarial robustness. 
Hence, work on robustness to OOD data and adversarial attacks has generally happened in parallel, as exemplified by RobustBench~\citep{croce_robustbench_2021} which provides independent benchmarks for assessing performance on corrupt data and adversarial threats.

A line of works \citep{tramer_adversarial_2019, maini_adversarial_2020} defends against a union of \lp threat models by training with multiple \lp threat models jointly, which makes these threat models no longer unforeseen. PAT \citep{laidlaw_perceptual_2021} replaces \lp bound with LPIPS \citep{zhang2018unreasonable} in adversarial training and achieves high robustness against several unforeseen attacks. 
Alternatively, \cite{dai_formulating_2022} proposes variation regularization in addition to \lp adversarial training and improves unforeseen robustness.

\textbf{Robustness benchmarks.}
There is a line of works on benchmarking adversarial robustness, including \citet{dong_benchmarking_2020}, RobustBench \citep{croce_robustbench_2021}, RobustART \citep{tang_robustart_2022}, MultiRobustBench \citep{dai_multirobustbench_2023}, UA \citep{kaufmann_testing_2023}, and \citet{liu_comprehensive_2023}. 
RobustBench focuses on ID adversarial evaluation. \citet{dong_benchmarking_2020} evaluates adversarial robustness under \lp threat shifts in addition to ID adversarial evaluation. 
RobustART, compared to \citet{dong_benchmarking_2020}, also supports the evaluation of OOD clean accuracy under dataset shifts. 
MultiRobustBench and UA both extend the evaluation set to include more unforeseen attacks, assessing adversarial robustness under threat shifts. \citet{liu_comprehensive_2023} is similar to RobustART but supports a larger number of OOD datasets. 
Compared to these benchmarks, OODRobustBench is unique in supporting the evaluation of OOD adversarial robustness under dataset shifts while also incorporating the functionalities of the other benchmarks.

\subsection{Comparison with Related Works}

\begin{quote}
    Is the linear trend of robustness really expected given the linear trend of accuracy?
\end{quote}

No. There is a well-known trade-off between accuracy and robustness in the ID setting \citep{tsipras_robustness_2019}. We further confirm this fact for the models we evaluate in \cref{fig: dataset shfit correlation r2 CAAC} in the appendix. This means that accuracy and robustness usually go in opposite directions making the linear trend we discover in both particularly interesting. Furthermore, the threat shifts as a scenario of OOD are unique to adversarial evaluation and were thus never explored in the previous studies of accuracy trends.

\begin{quote}
    How does the linear trends observed by us differ from the previously discovered ones?
\end{quote}

Robust models exhibit a stronger linear correlation between ID and OOD accuracy for most corruption shifts (\cref{fig: dataset shfit correlation r2}). Particularly, the boost on linearity is dramatic for shifts including Impulse, Shot and Gaussian noises, Glass blur, Pixelate, and JPEG. For instance, R2 surges from 0 (no linear correlation) for non-robust models to 0.84 (evident linear correlation) for robust models with Gaussian noise data shifts. This suggests that, for robust models, predicting OOD accuracy from ID accuracy is more faithful and applicable to more shifts. 

The linear trend of robustness is even stronger than that of accuracy for dataset shifts (\cref{fig: dataset shfit correlation r2}) but with a lower slope (\cref{sec: predict upper limit}). The latter leads to a predicted upper limit of OOD robustness that is way lower than that of OOD accuracy suggesting that the OOD generalization of robustness is much more challenging.

\begin{quote}
    How does our analysis differ from the similar analysis in the prior works?
\end{quote}

The scale of these previous works is rather small. For instance, RobustBench observes linear correlation only for three shifts on CIFAR-10 based on 39 models with either ResNet or WideResNet architectures. In such a narrow setting, it is actually neither surprising to see a linear trend nor reliable for predicting OOD performance. By contrast, our conclusion is derived from much more shifts on CIFAR-10 and ImageNet based on 706 models. Importantly, our model zoo covers a diverse set of architectures, robust training methods, data augmentation techniques, and training set-ups. This makes our conclusion more generalizable and the observed (almost perfect) linear trend much more significant.

Similarly, the existing works only test a few models under threat shifts. Those methods are usually just the baseline AT method plus different architectures or the relevant defenses, e.g., jointly trained with multiple threats. It is unclear how the state-of-the-art robust models perform under threat shifts. By conducting a large-scale analysis, we find that those SOTA models generalize poorly to other threats while also discovering several methods that have relatively inferior ID performance but superior OOD robustness under threat shift. Our analysis therefore facilitates future works in this direction by identifying what techniques are ineffective and what are promising.

\begin{quote}
    How does you benchmark differ from RobustBench?
\end{quote}

Our benchmark focuses on OOD adversarial robustness while RobustBench focuses on ID adversarial robustness. Specifically, our benchmark contrasts RobustBench in the datasets and the attacks. We use CIFAR-10.1, CIFAR-10.2, CINIC, and CIFAR-10-R (ImageNet-V2, ImagetNet-A, ImageNet-R, ObjectNet) to simulate input data distribution shift for the source datasets CIFAR-10 (ImageNet), while RobustBench only uses the latter source datasets. We use PPGD, LPA, ReColor, StAdv, Linf-12/255, L2-0.5 (PPGD, LPA, ReColor, StAdv, Linf-8/255, L2-1) to simulate threat shift for the training threats Linf-8/255 (L2-0.5), while RobustBench only evaluates the same threats as the training ones.

\section{Benchmark Set-up}

\subsection{Datasets} \label{appendix: benchmark datasets}
This section introduces the OOD datasets of natural shifts.
For ImageNet, we have:
\begin{itemize}
    \item \textbf{ImageNet-V2} is a reproduction of ImageNet using a completely new set of images. It has the same 1000 classes as ImageNet and each class has 10 images so 10K images in total. 

    \item \textbf{ImageNet-A} is an adversarially-selected reproduction of ImageNet. The images in this dataset were selected to be those most misclassified by an ensemble of ResNet-50s. It has 200 ImageNet classes and 7.5K images. 

    \item \textbf{ImageNet-R} contains various artistic renditions of objects from ImageNet, so there is a domain shift. It has 30K images and 200 ImageNet classes.

    \item \textbf{ObjectNet} is a large real-world dataset for object recognition. It is constructed with controls to randomize background, object rotation and viewpoint. It has 313 classes but only 104 classes compatible with ImageNet classes so we only use this subset. The selected subset includes 17.2K images. 
\end{itemize}

For CIFAR10, we have:
\begin{itemize}
    \item \textbf{CIFAR10.1} is a reproduction of CIFAR10  using a completely new set of images. It has 2K images sampled from the same source as CIFAR10, i.e., 80M TinyImages \citep{torralba_80_2008}. It has the same number of classes as CIFAR10.
    
    \item \textbf{CIFAR10.2} is another reproduction of CIFAR10. It has 12K (10k for training and 2k for test) images sampled from the same source as CIFAR10, i.e., 80M TinyImages. It has the same number of classes as CIFAR10. We only use the test set of CIFAR10.2.
    
    \item \textbf{CINIC} is a downscaled subset of ImageNet with the same image resolution and classes as CIFAR10. Its test set has 90K images in total, of which 20K images are from CIFAR10 and 70K images are from ImageNet. We use only the ImageNet part. 

    \item \textbf{CIFAR10-R} is a new dataset created by us. The images in CIFAR10-R and CIFAR10 have different styles so there is a domain shift. We follow the same procedure as CINIC to downscale the images from ImageNet-R to the same resolution as CIFAR10 and select images from the classes of ImageNet corresponding to CIFAR10 classes. We follow the same class mapping between ImageNet and CIFAR10 as CINIC. Note that ImageNet-R does not have images of the ImageNet classes corresponding to CIFAR10 classes of "airplane" and "horse", so there are only 8 classes in CIFAR10-R.
\end{itemize}

In practice, we evaluate models using a random sample of 5K images from each of the ImageNet variant datasets, and 10K images from each of the CIFAR10 variant datasets, if those datasets contain more images than that number. This is done to accelerate the evaluation and follows the practice used in RobustBench \citep{croce_robustbench_2021}.  

The OODRobustBench framework is designed for easy integration of alternative datasets representing input data distribution shifts. 
We plan to maintain and update our code to continually incorporate new OOD datasets such as ImageNet-V \citep{dong_viewfool_2022}, Stylized-ImageNet \citep{geirhos_imagenet-trained_2019}, and ImageNet-Sketch \citep{wang_learning_2019-1}. 
For the latest developments, we invite readers to visit our GitHub repository: \url{https://github.com/OODRobustBench/OODRobustBench}.

\subsection{Verification of the Effectiveness of the MM5 Attack} 
\label{appendix: adversarial evaluation}

\subsubsection{Comparison of MM5 against AutoAttack}
\begin{figure}[t]
    \centering
    \includegraphics[width=.3\textwidth]{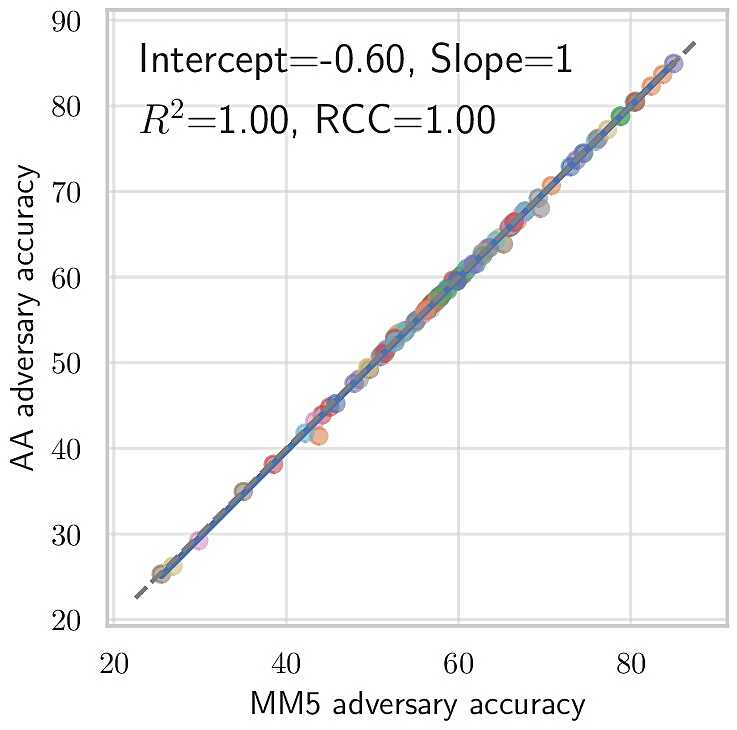}
    \vspace{-5mm}
    \caption{\textbf{Comparison of MM5 adversarial accuracy against AutoAttack adversarial accuracy}.
    Each data point represents a model.}
    \label{fig: mm5 aa}
    \vspace{-3mm}
\end{figure}

To verify the effectiveness of MM5, we compare its result with the result of AutoAttack on the ID dataset across all publicly available models from RobustBench for CIFAR10 \linf, CIFAR10 \lt and ImageNet \linf. 
As shown in \cref{fig: mm5 aa}, almost all models\footnote{Two models, \citet{ding_mma_2020} and \citet{xu_exploring_2023}, are observed to have a slightly higher adversarial accuracy compared to the corresponding AutoAttack results. We use MM+ \citep{gao_fast_2022} attack to evaluate these two models for a more reliable evaluation and the result of MM+ is close to AutoAttack.} are approximately on the line of $y=x$ (gray dashed line) suggesting that their MM5 adversarial accuracy is very close to AA adversarial accuracy. 
Specifically, the mean gap between MM5 and AA adversarial accuracy is 0.16 and the standard deviation is 0.32. 

\subsubsection{Comparison of MM5 against Diverse Attacks}
To test the effectiveness of MM5 against various attacks, we selected 10 models from RobustBench and evaluated their robustness against PGD100, CW100, and AutoAttack. As shown in \cref{tab: more attacks}, MM5 achieved an adversarial accuracy comparable to that of the strongest attack, AutoAttack, in both in-distribution (ID) and out-of-distribution (OOD) settings. This suggests that the MM5 attack is a reliable method for adversarial evaluation, even under distribution shifts.

\begin{table}[t]
\centering
\caption{\textbf{The accuracy and adversarial robustness evaluated by various attacks of models on CIFAR10 and ImageNet for \linf threat model}. These models are sourced from RobustBench respective leaderboards and identified by their RobustBench identifiers. The results for corruption shifts are reported for severity level 3 .}
\label{tab: more attacks}
\resizebox{\textwidth}{!}{%
\begin{tabular}{@{}llcccccccccc@{}}
\toprule
\multicolumn{1}{c}{\multirow{2}{*}{Dataset}} & \multicolumn{1}{c}{\multirow{2}{*}{Model}} & \multicolumn{5}{c}{ID} & \multicolumn{5}{c}{\oodd} \\ \cmidrule(l){3-7} \cmidrule(l){8-12} 
\multicolumn{1}{c}{} & \multicolumn{1}{c}{} & Acc. & PGD & CW & AA & MM5 & Acc. & PGD & CW & AA & MM5 \\ \midrule
\multirow{5}{*}{CIFAR10} & Wong2020Fast & 83.3 & 46.6 & 46.3 & 43.2 & 43.3 & 65.0 & 27.6 & 27.3 & 24.8 & 25.0 \\
 & Engstrom2019Robustness & 87.0 & 52.3 & 52.6 & 49.3 & 49.7 & 70.6 & 31.7 & 31.8 & 29.1 & 29.3 \\
 & Huang2020Self & 83.5 & 56.2 & 54.0 & 52.9 & 53.0 & 65.0 & 34.4 & 32.9 & 31.7 & 31.8 \\
 & Sehwag2021Proxy\_R18 & 84.6 & 58.7 & 57.2 & 55.6 & 55.7 & 67.0 & 37.8 & 36.5 & 34.6 & 34.8 \\
 & Wang2020Improving & \textbf{87.5} & \textbf{62.6} & \textbf{58.7} & \textbf{56.3} & \textbf{56.8} & \textbf{70.5} & \textbf{40.3} & \textbf{37.1} & \textbf{34.7} & \textbf{35.1} \\ \midrule
\multirow{5}{*}{ImageNet} & Salman2020Do\_R18 & 52.9 & 29.8 & 27.3 & 25.3 & 25.5 & 21.4 & 9.5 & 8.5 & 7.5 & 7.7 \\
 & Wong2020Fast & 55.6 & 30.0 & 28.9 & 26.3 & 26.8 & 21.6 & 8.7 & 8.5 & 7.2 & 7.5 \\
 & Engstrom2019Robustness & 62.5 & 32.9 & 32.6 & 29.2 & 29.8 & 27.2 & 10.1 & 10.2 & 8.4 & 8.7 \\
 & Salman2020Do\_R50 & 64.1 & 39.0 & 37.6 & 34.7 & 35.0 & 27.6 & 12.5 & 12.0 & 10.5 & 10.7 \\
 & Singh2023Revisiting\_ViT-S-ConvStem & \textbf{72.6} & \textbf{51.4} & \textbf{50.6} & \textbf{48.1} & \textbf{48.5} & \textbf{39.8} & \textbf{19.8} & \textbf{19.4} & \textbf{17.5} & \textbf{17.7} \\ \bottomrule
\end{tabular}%
}
\end{table}


\section{Model Zoo}

\subsection{Criteria for Robust Models}
\label{appendix: model criteria}
We follow the same criteria as the popular benchmarks (RobustBench \citep{croce_robustbench_2021}, MultiRobustBench \citep{dai_multirobustbench_2023}, etc), which only include robust models that (1) have in general non-zero gradients w.r.t. the inputs, (2) have a fully deterministic forward pass (i.e. no randomness) and (3) do not have an optimization loop. These criteria include most AT models, while excluding most preprocessing methods because they rely on randomness like \citet{guo_countering_2018} or inner optimization loop like \citet{samangouei_defense-gan_2018} which leads to false security, i.e., high robustness to the non-adaptive attack but vulnerable to the adaptive attack. 

Meanwhile, we acknowledge that evaluating dynamic preprocessing-based defenses is still an active area of research. It is tricky \citep{croce2022evaluating}, and there has not been a consensus on how to evaluate them. So now, we exclude them for a more reliable evaluation. We will keep maintaining this benchmark, and we would be happy to include them in the future if the community has reached a consensus on that (e.g., if these models are merged into RobustBench).

\subsection{Model Zoo}
\label{appendix: model zoo}
Our model zoo consists of 706 models, of which: 
\begin{itemize}
    \item 396 models are trained on CIFAR10 by \linf 8/255
    \item 239 models are trained on CIFAR10 by \lt 0.5
    \item 56 models are trained on ImageNet by \linf 4/255
    \item 10 models are trained on CIFAR10 for \nonlp adversarial robustness
    \item 5 models are trained on CIFAR10 for common corruption robustness
\end{itemize}

Among the above models, 66 models of CIFAR10 \linf, 19 models of CIFAR10 \lt and 18 models of ImageNet \linf are retrieved from RobustBench. 84 models are retrieved from the published works including \cite{li_aroid_2023,li_data_2023,li_understanding_2023,li_improved_2023,liu_comprehensive_2023,singh_revisiting_2023,dai_formulating_2022,hsiung_towards_2023,mao2022easyrobust}. The remaining models are trained by ourselves.

We locally train additional models with varying architectures and training parameters to complement the public models from RobustBench on CIFAR-10.
We consider 20 model architectures: DenseNet-121~\citep{huang_densely_2017}, GoogLeNet~\citep{szegedy_going_2015}, Inception-V3~\citep{szegedy_rethinking_2016}, VGG-11/13/16/19~\citep{simonyan_very_2015}, ResNet-34/50/101/152~\citep{he_deep_2016}, EfficientNet-B0~\citep{tan_efficientnet_2019}, MobileNet-V2~\citep{sandler_2018_mobilenetv2}, DLA~\citep{yu_2018_dla}, ResNeXt-29 (2x64d/4x64d/32x4d/8x64d)~\citep{xie_aggregated_2017}, SeNet-18~\citep{hu_squeeze-and-excitation_2018}, and ConvMixer~\citep{trockman_2023_patches}.
For each architecture, we vary the training procedure to obtain 15 models across four adversarial training methods: PGD~\citep{madry_towards_2018}, TRADES~\citep{zhang_theoretically_2019}, PGD-SCORE, and TRADES-SCORE~\citep{pang_robustness_2022}.

We train all models under both $\ell_\infty$ and $\ell_2$ threat models with the following steps:
\begin{enumerate}
    \item We use PGD adversarial training to train eight models with batch size $\in \{128, 512\}$, a learning rate $\in \{0.1, 0.05\}$, and weight decay $\in \{10^{-4}, 10^{-5}\}$. We also save the overall best hyperparameter choice. For the $\ell_2$ threat model, we fix the learning rate to 0.1 since we observe that with $\ell_\infty$, 0.1 is strictly better than 0.05.
    \item Using the best hyperparameter choice, we train one model with PGD-SCORE, three with TRADES, and three with TRADES-SCORE. For TRADES and TRADES-SCORE, we take their $\beta$ parameter from ${0.1, 0.3, 1.0}$.
\end{enumerate}

After training, we observe that some locally trained models exhibit inferior accuracy and/or robustness that is abnormally lower than others. The influence of inferior models on the correlation analysis is discussed in \cref{appendix: inferior models correlation}. Finally, we filter out all models with an overall performance (accuracy + robustness) below 110. This threshold is determined to exclude only those evidently inferior models so that the size of model zoo (557 after filtering) is still large enough to ensure the generality and comprehensiveness of the conclusions drawn on it.


\clearpage
\section{Additional Results} 
\label{appendix: additional resutl}

\subsection{OOD Performance and Ranking}

\input{tables/benchmark-cifar10-l2}
\input{tables/benchmark-imagenet}

\subsection{Performance Degradation Distribution}

\begin{figure}[!h]
    \centering
    
    \begin{subfigure}{\linewidth}
        \includegraphics[width=\linewidth, trim=4 7 3 4,clip]{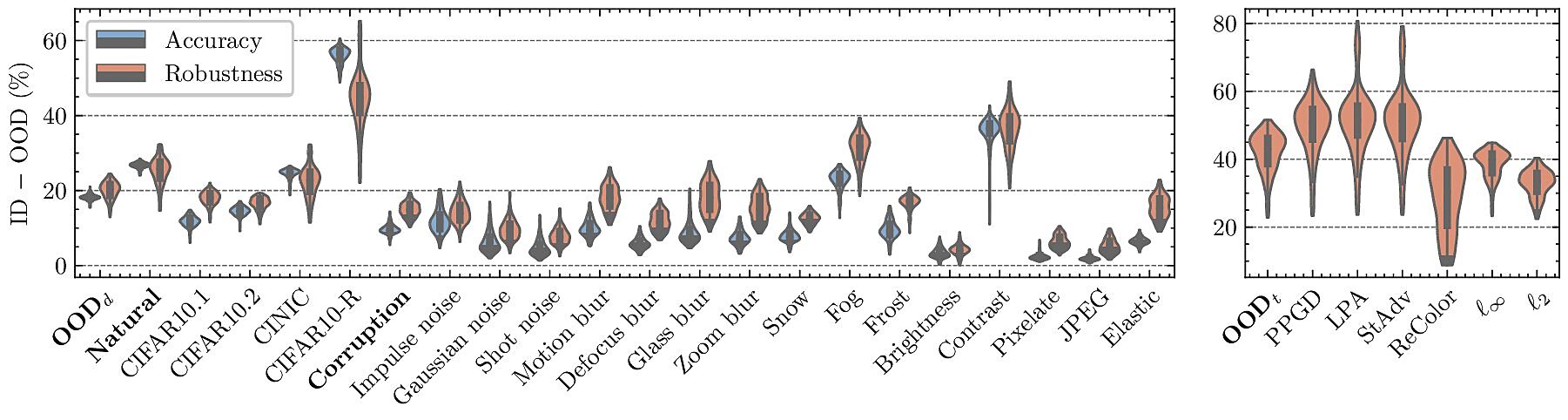}
    \end{subfigure}
    \vspace{-7mm}
    \caption{\textbf{Degradation of accuracy and robustness under various distribution shifts for CIFAR10 \lt}.}
    \label{fig: degradation violin cifar10 l2}
\end{figure}

\begin{figure}[!h]
    \centering

    \begin{subfigure}{\linewidth}
        \includegraphics[width=\linewidth, trim=4 7 3 4,clip]{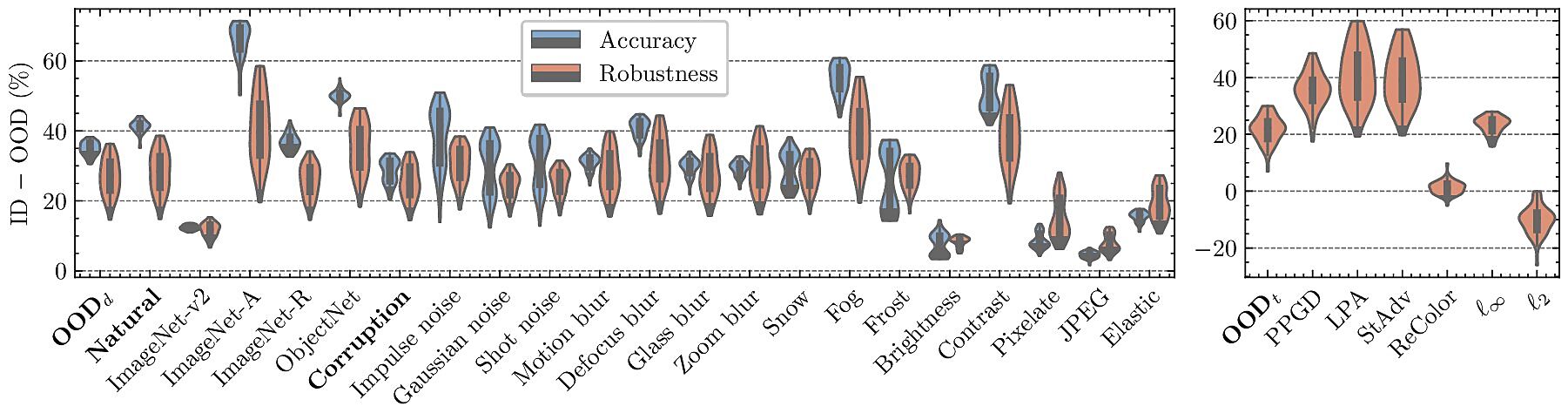}
    \end{subfigure}  
    \vspace{-7mm}
    \caption{\textbf{Degradation of accuracy and robustness under various distribution shifts for ImageNet \linf}.}
    \label{fig: degradation violin imagenet}
\end{figure}

\subsection{Correlation Between ID and OOD Performance under Dataset Shifts} \label{appendix: dataset shifts correlation}



\begin{figure}[!h]
    \centering
    \begin{subfigure}{\linewidth}
        \includegraphics[width=\linewidth]{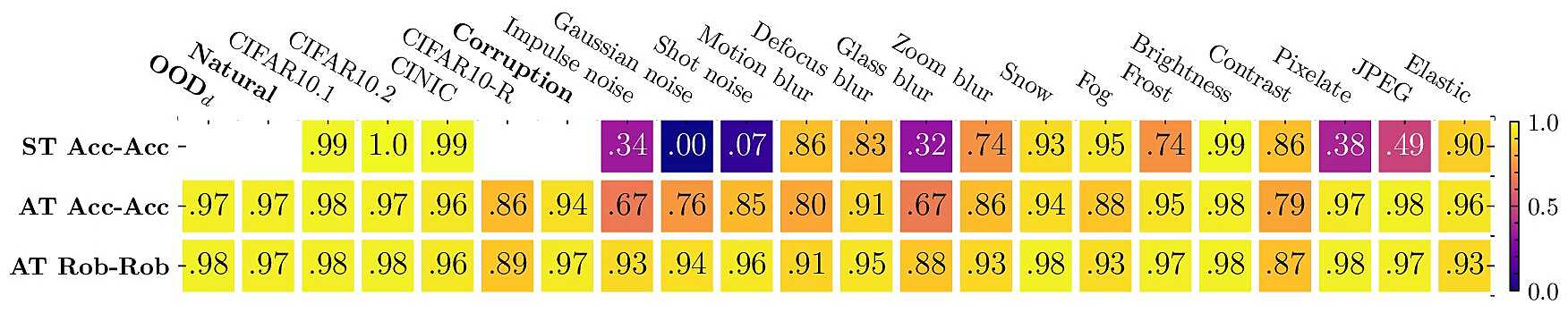}
    \end{subfigure}
    \vspace{-7mm}
    \caption{\textbf{\rsquare of regression between ID and OOD performance for Standardly-Trained (ST) and Adversarially-Trained (AT) models under dataset shifts for CIFAR10 \lt}. Higher \rsquare implies stronger linear correlation. The result of ST is copied from \cite{miller_accuracy_2021}.}
    \label{fig: dataset shfit correlation r2 cifar10 l2}
    \vspace{-2mm}
\end{figure}

\begin{figure}[!h]
    \centering
    \begin{subfigure}{\linewidth}
        \includegraphics[width=\linewidth]{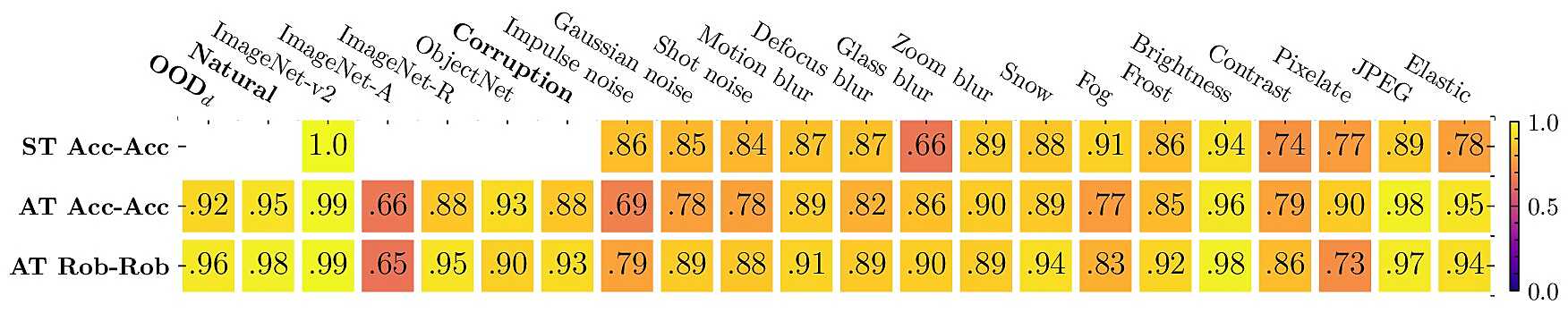}
    \end{subfigure}
    \vspace{-7mm}
    \caption{\textbf{\rsquare of regression between ID and OOD performance for Standardly-Trained (ST) and Adversarially-Trained (AT) models under dataset shifts for ImageNet \linf}. Higher \rsquare implies stronger linear correlation. The result of ST is copied from \cite{miller_accuracy_2021}.}
    \label{fig: dataset shfit correlation r2 imagenet linf}
    \vspace{-2mm}
\end{figure}

\begin{figure}[!h]
    \centering
    
    \begin{subfigure}{\linewidth}
        \includegraphics[width=\linewidth]{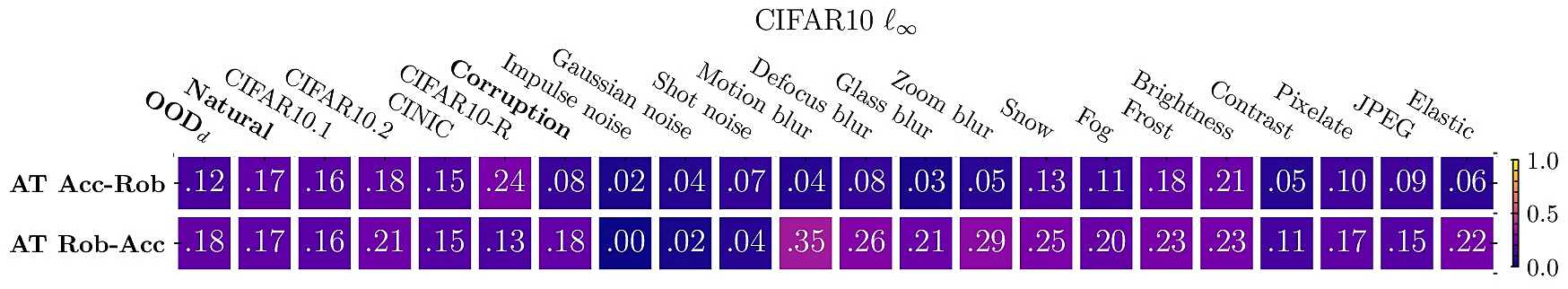}
    \end{subfigure}

    \begin{subfigure}{\linewidth}
        \includegraphics[width=\linewidth]{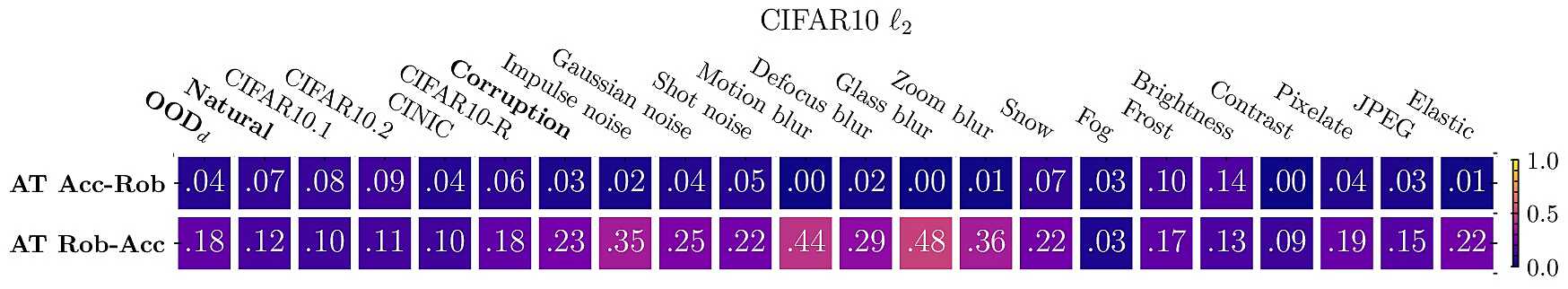}
    \end{subfigure}

    \begin{subfigure}{\linewidth}
        \includegraphics[width=\linewidth]{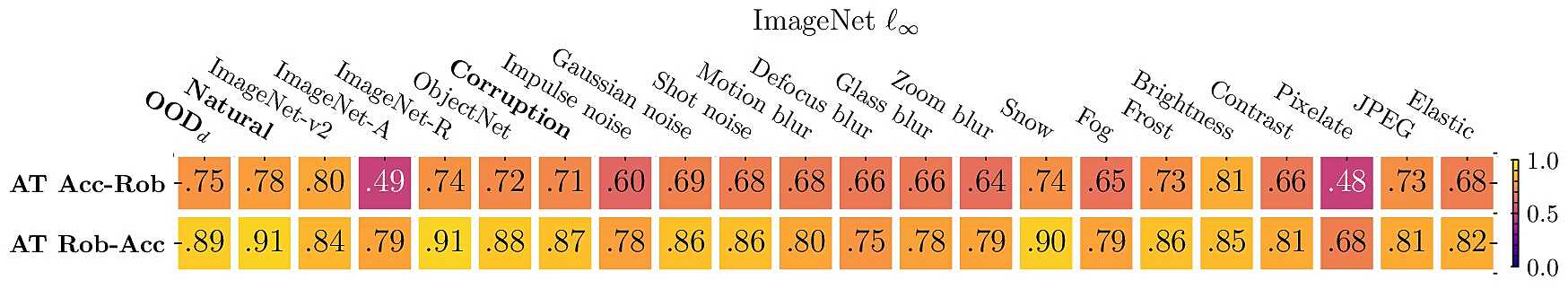}
    \end{subfigure}

    \caption{\textbf{\rsquare of regression between ID and OOD performance for Adversarially-Trained (AT) models under various dataset shifts}. "Acc-Rob" denotes the linear model between ID accuracy (x) and OOD robustness (y) and "Rob-Acc" for ID robustness (x) and OOD accuracy (y).}
    \label{fig: dataset shfit correlation r2 CAAC}
\end{figure}

\newpage
\subsection{Unsupervised OOD Robustness Prediction}

\begin{figure}[!h]
    \centering
    
    \begin{subfigure}{.33\linewidth}
        \includegraphics[width=\linewidth]{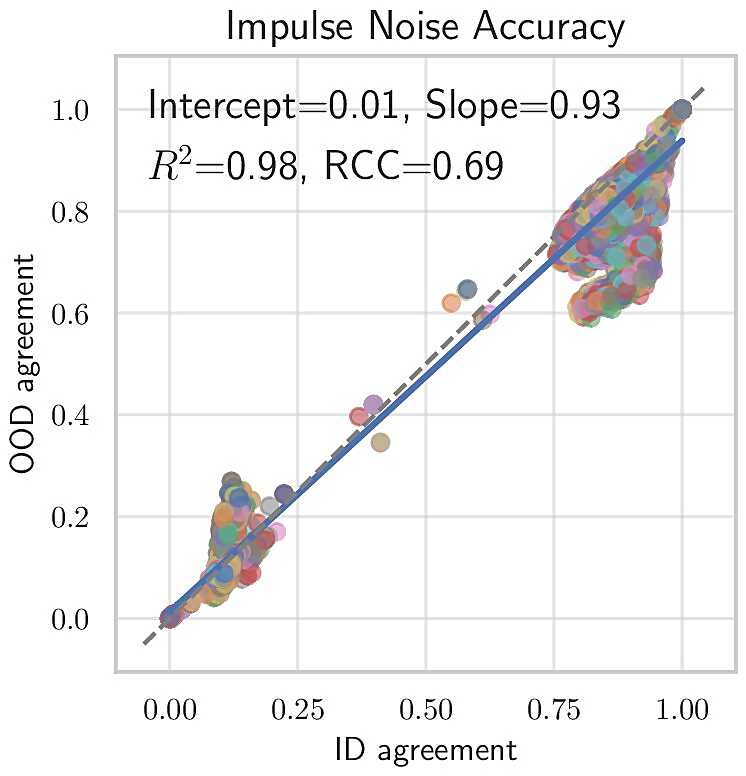}
    \end{subfigure}
    \hspace{5mm}
    \begin{subfigure}{.33\linewidth}
        \includegraphics[width=\linewidth]{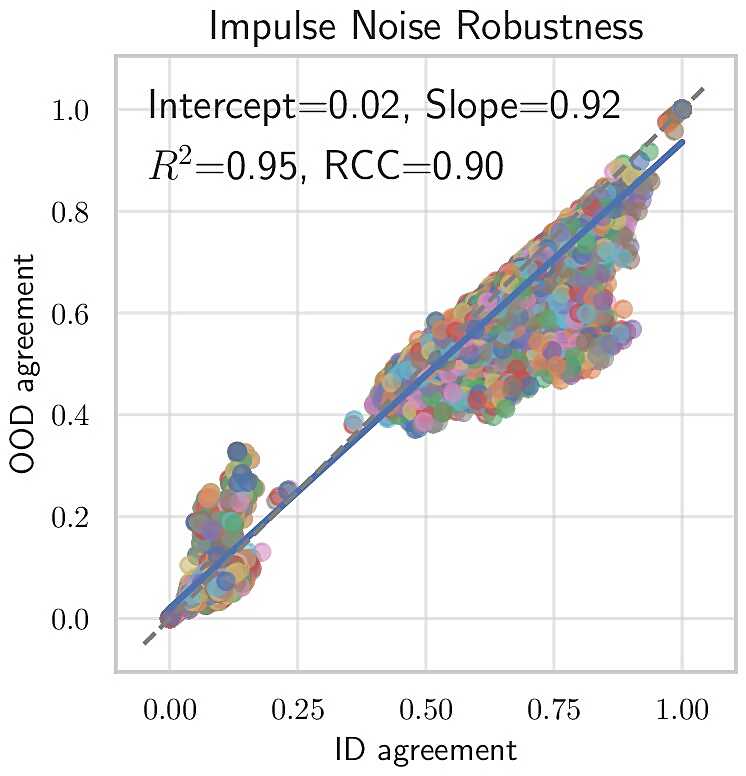}
    \end{subfigure}
    \caption{\textbf{Correlation between ID and OOD prediction agreement on adversarial examples for CIFAR10 \linf AT models}.}
    \label{fig: agreement line impulse noise}
\end{figure}




    


\newpage
\subsection{Predicted Upper Limit of OOD Accuracy and Robustness} 
\label{appendix: predicted upper limit}
    



\begin{figure}[!h]
    \centering

    \begin{subfigure}{.7\linewidth}
    \includegraphics[width=\linewidth]{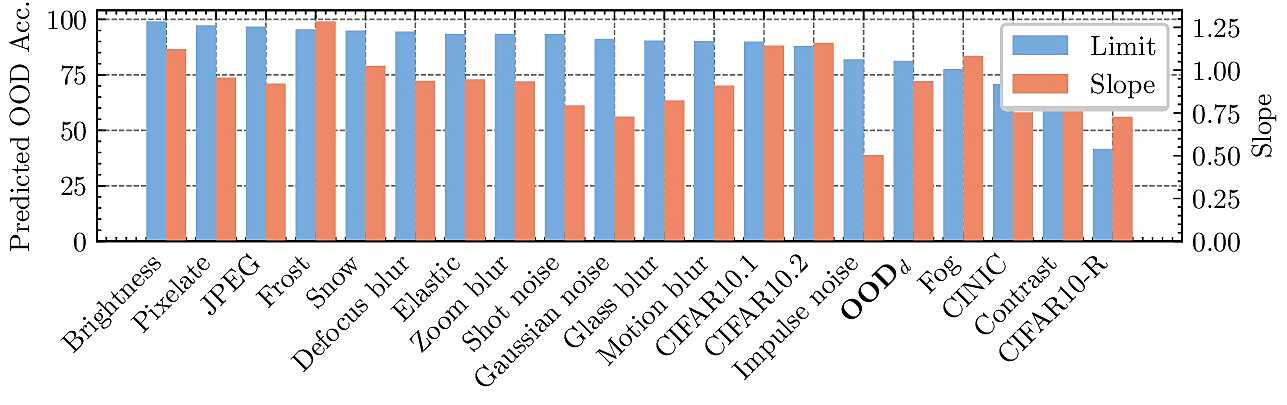}
    \vspace{-7mm}
    \caption{\textbf{CIFAR10 \linf Accuracy}.}
    \end{subfigure}
    
    \begin{subfigure}{.7\linewidth}
        \includegraphics[width=\linewidth]{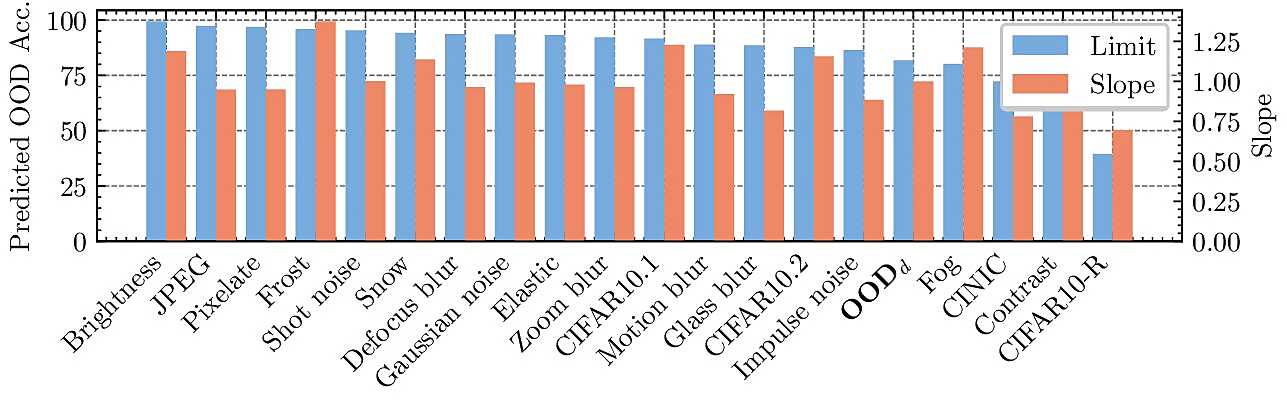}
        \vspace{-7mm}
        \caption{\textbf{CIFAR10 \lt Accuracy}.}
    \end{subfigure}
    
    \begin{subfigure}{.7\linewidth}
        \includegraphics[width=\linewidth]{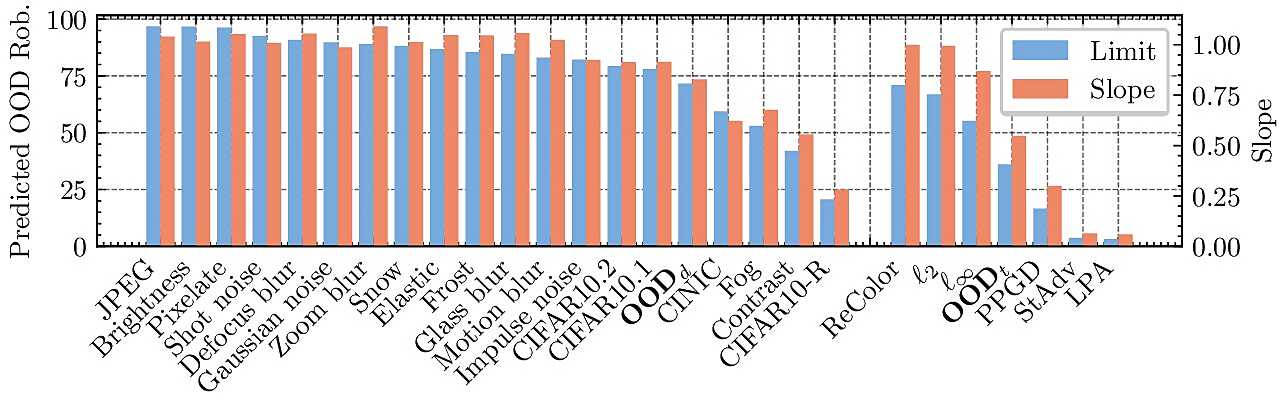}
        \vspace{-7mm}
        \caption{\textbf{CIFAR10 \lt Robustness}.}
    \end{subfigure}

    \begin{subfigure}{.7\linewidth}
        \includegraphics[width=\linewidth]{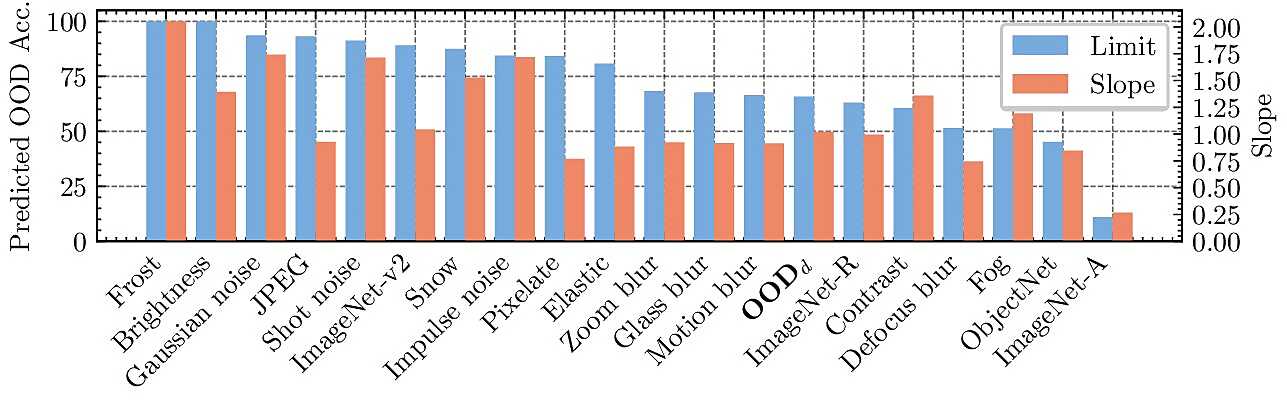}
        \vspace{-7mm}
        \caption{\textbf{ImageNet \linf Accuracy}.}
    \end{subfigure}

    \begin{subfigure}{.7\linewidth}
        \includegraphics[width=\linewidth]{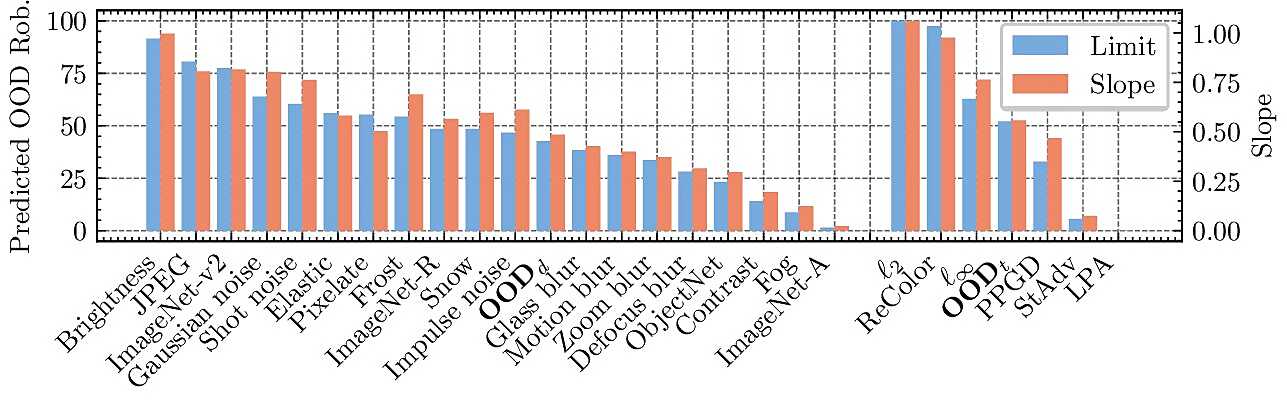}
        \vspace{-7mm}
        \caption{\textbf{ImageNet \linf Robustness}.}
    \end{subfigure}
    
    \vspace{-3mm}    
    \caption{\textbf{The estimated upper limit of OOD performance and the conversion rate, a.k.a. slope, to OOD performance from ID performance under various distribution shifts}.}
    \label{fig: predicted upper limit and slope}
\end{figure}

\newpage
\section{Catastrophic Degradation of Robustness}
\label{appendix: catastrophic degradation}
We observe this issue on only one implementation, using WideResNet28-10 with extra synthetic data (model id: \textit{Rade2021Helper\_ddpm} on RobustBench), from \citet{rade_reducing_2022} for CIFAR10 \linf.
There are three other implementations of this method on RobustBench. None of them, including the one using ResNet18 with extra synthetic data, is observed to suffer from this issue. It seems that catastrophic degradation in this case is specific to the implementation or training dynamics. 

On the other hand, catastrophic degradation consistently happens on the models trained with AutoAugment or IDBH but not other tested data augmentations. It suggests the possibility that a certain image transformation operation exclusively used by AutoAugment and IDBH cause this issue. Besides, catastrophic degradation also consistently happens on the models trained using the receipt of \cite{debenedetti_light_2023} under Gaussian and shot noise shifts. However, it employs a wide range of training techniques, so further experiments are required to identify the specific cause. 

\section{How Inferior Models Affect the Correlation Analysis} 
\label{appendix: inferior models correlation}
This section studies the influence of the construction of model zoo on the result of correlation. We use the overall performance (accuracy + robustness) to filter out inferior models. As we increasing the threshold of overall performance for filtering, the average overall performance of the model zoo increases, the number of included models decreases and the weight of the models from other published sources on the regression grows up. Our locally trained models are normally inferior to the public models regarding the performance since the latter employs better optimized and more effective training methods and settings. The training methods and settings of public models are also much more diverse. 

\begin{figure}[!h]
    \centering
    \begin{subfigure}{\linewidth}
        \includegraphics[width=\linewidth]{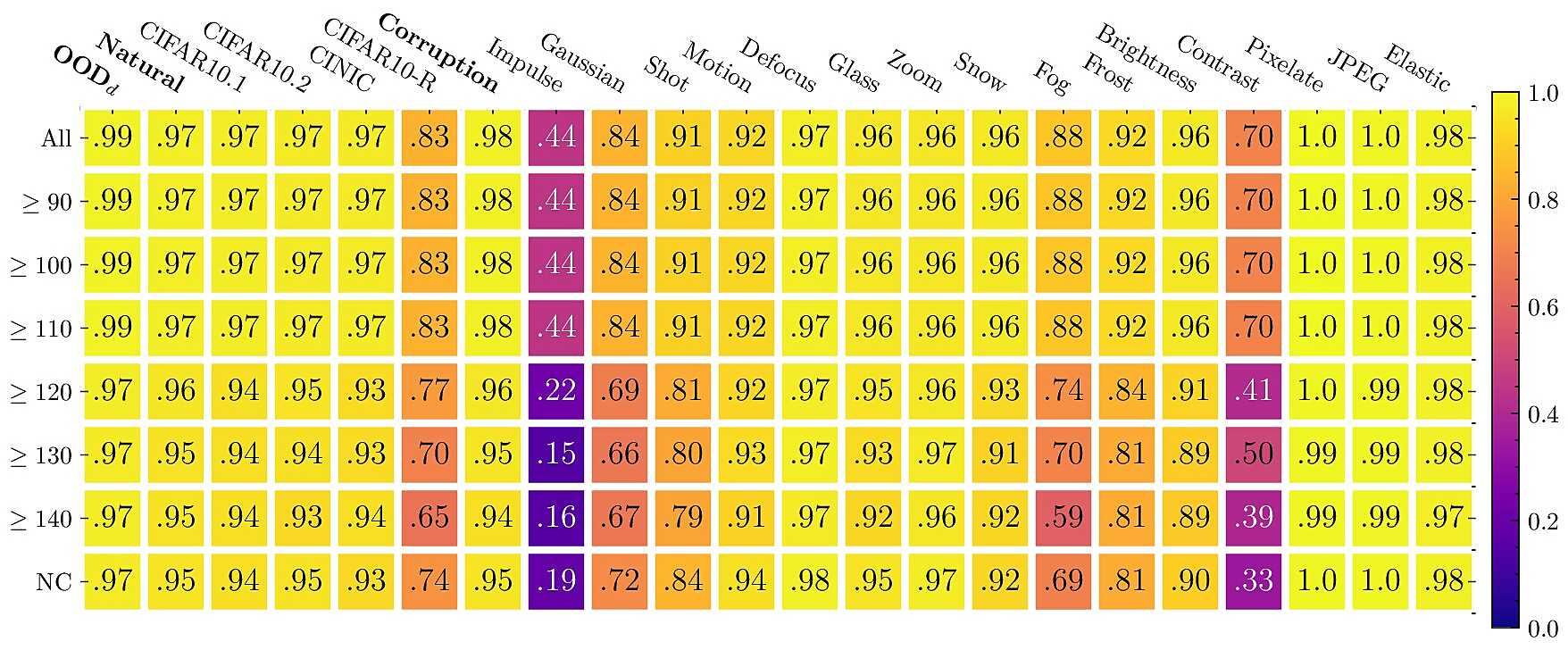}
        \caption{\textbf{\rsquare of Acc-Acc}.}
    \end{subfigure}

    \begin{subfigure}{\linewidth}
        \includegraphics[width=\linewidth]{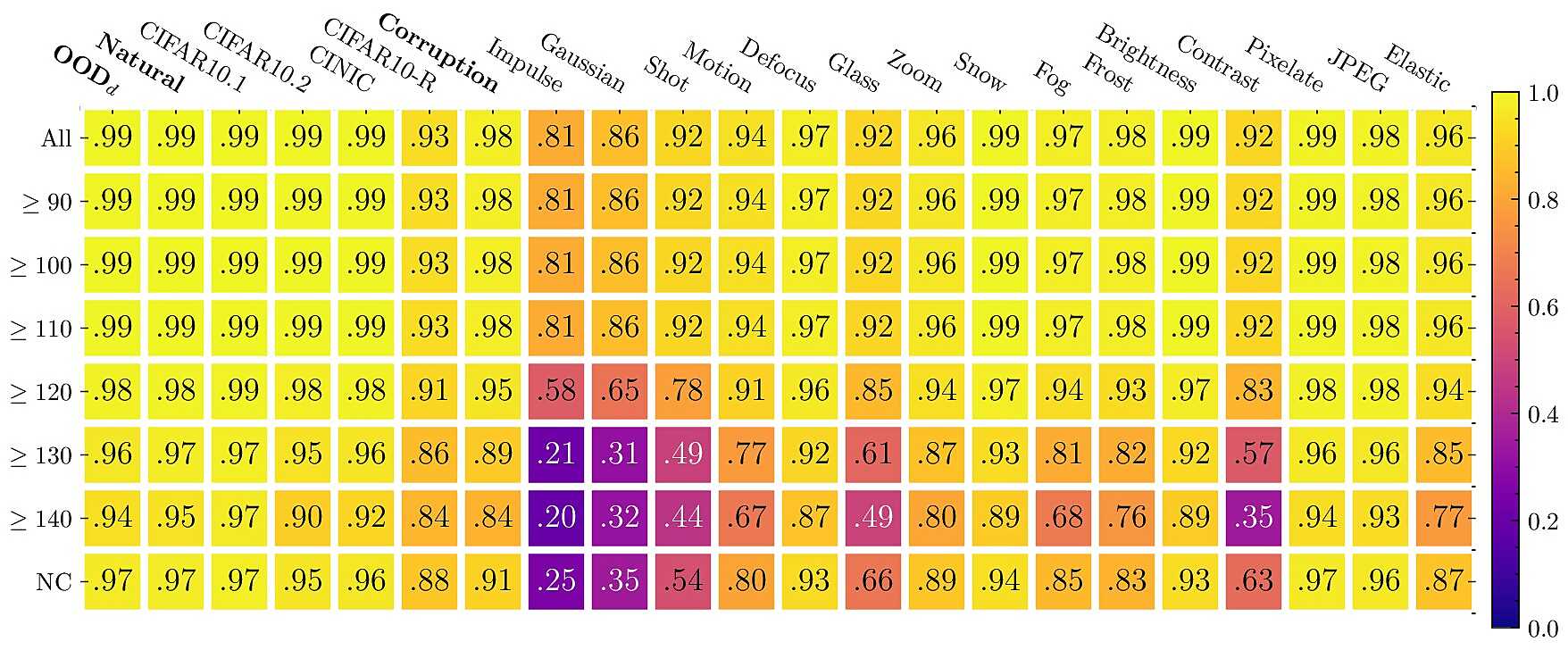}
        \caption{\textbf{\rsquare of Rob-Rob}.}
    \end{subfigure}
    
    \caption{\textbf{The change of \rsquare under various dataset shifts as the models with lower overall performance are removed from regression for CIFAR10 \linf}. Each row, with the filtering threshold labeled at the lead, corresponds to a new filtered model zoo and the regression conducted it. "NC" refers to No Custom models, so all models are retrieved from either RobustBench or other published works.}
    \label{fig: inferior dataset shifts cifar linf}
\end{figure}

\begin{figure}[!h]
    \centering
    
    \begin{subfigure}{\linewidth}
        \includegraphics[width=\linewidth]{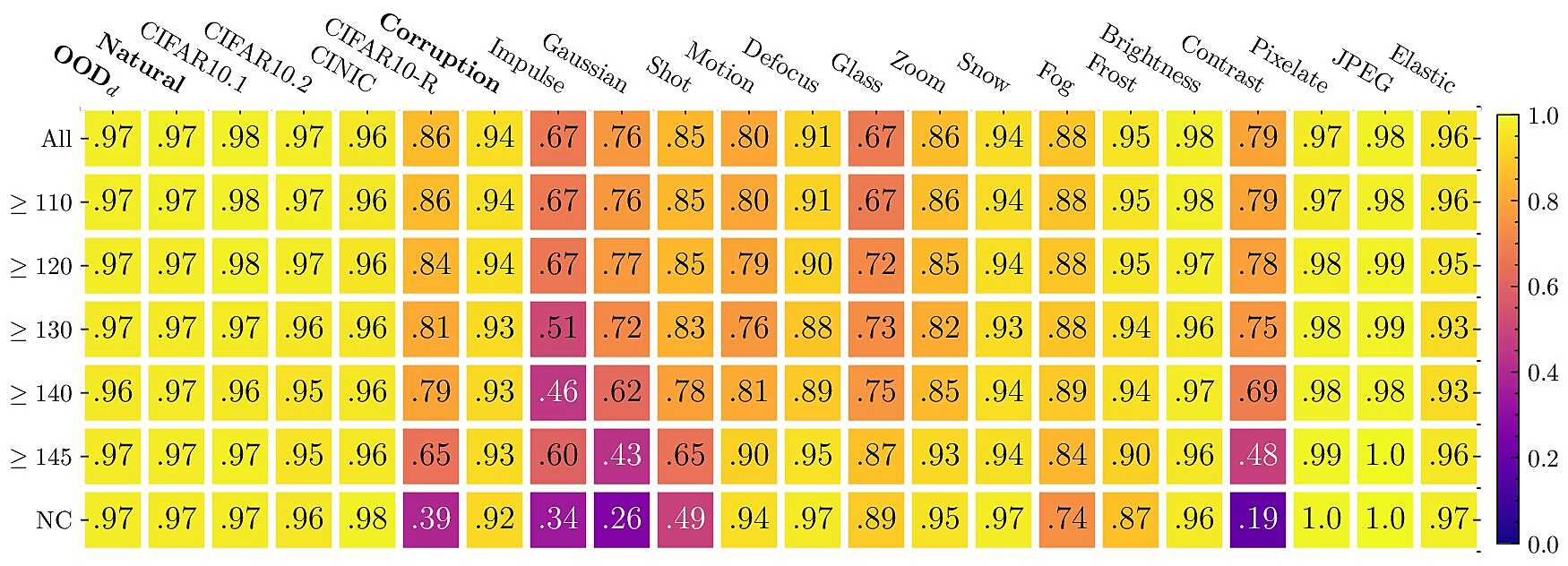}
        \caption{\textbf{\rsquare of Acc-Acc}.}
    \end{subfigure}

    \begin{subfigure}{\linewidth}
        \includegraphics[width=\linewidth]{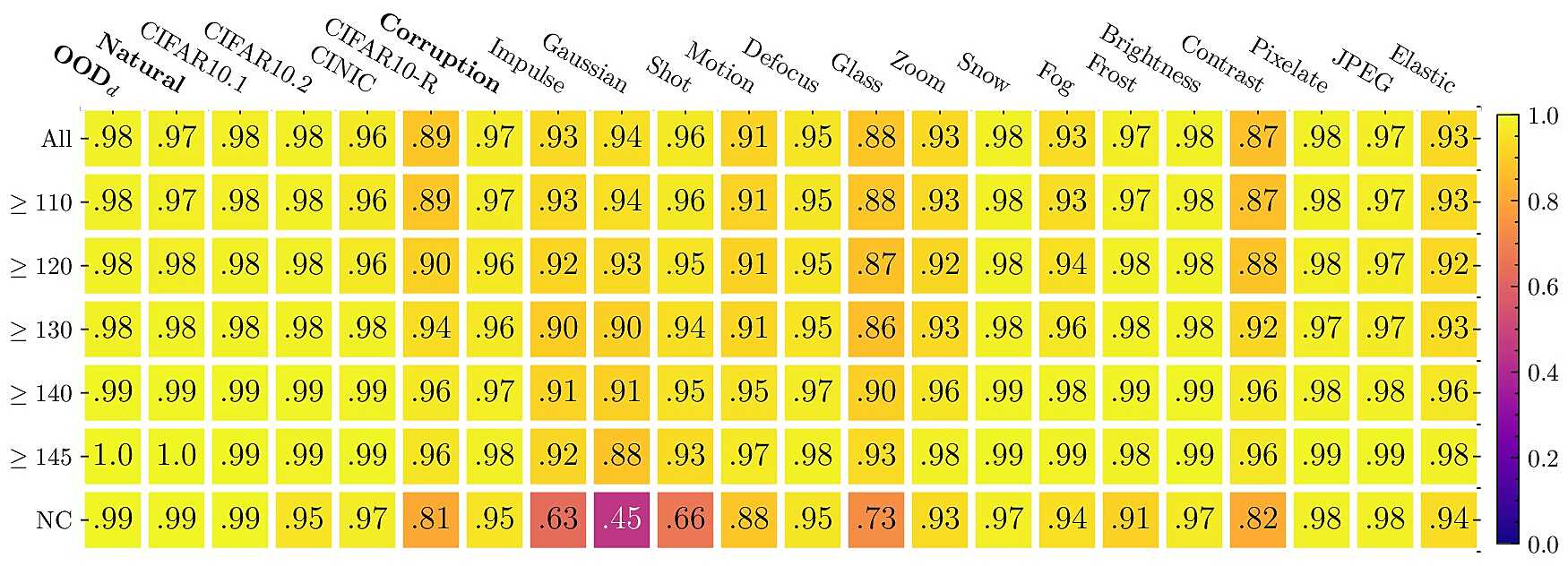}
        \caption{\textbf{\rsquare of Rob-Rob}.}
    \end{subfigure}
    
    \caption{\textbf{The change of \rsquare under various dataset shifts as the models with lower overall performance are removed from regression for CIFAR10 \lt}. Each row, with the filtering threshold labeled at the lead, corresponds to a new filtered model zoo and the regression conducted it. "NC" refers to No Custom models, so all models are retrieved from either RobustBench or other published works.}
    \label{fig: inferior dataset shifts cifar l2}
\end{figure}

\begin{figure}[!h]
    \centering    

    \begin{subfigure}{.48\linewidth}
        \includegraphics[width=\linewidth]{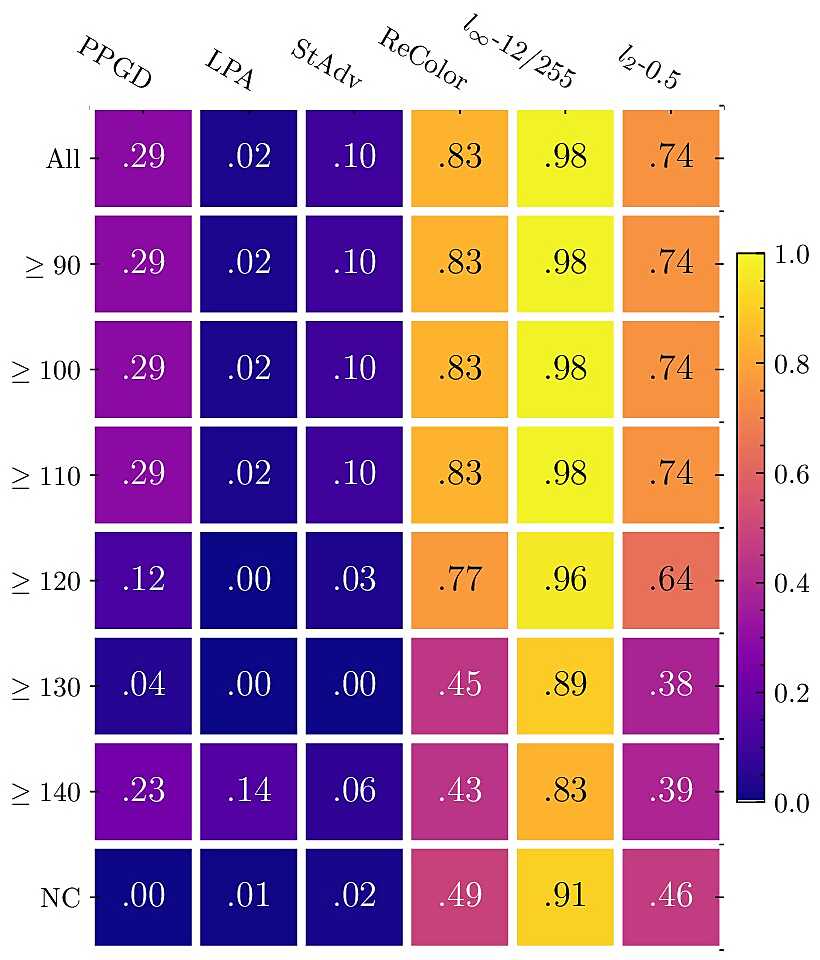}
        \caption{\textbf{\rsquare of ID seen vs. unforeseen robustness for CIFAR10 $\ell_{\infty}$}.}
    \end{subfigure}
    \hspace{2mm}
    \begin{subfigure}{.48\linewidth}
        \includegraphics[width=\linewidth]{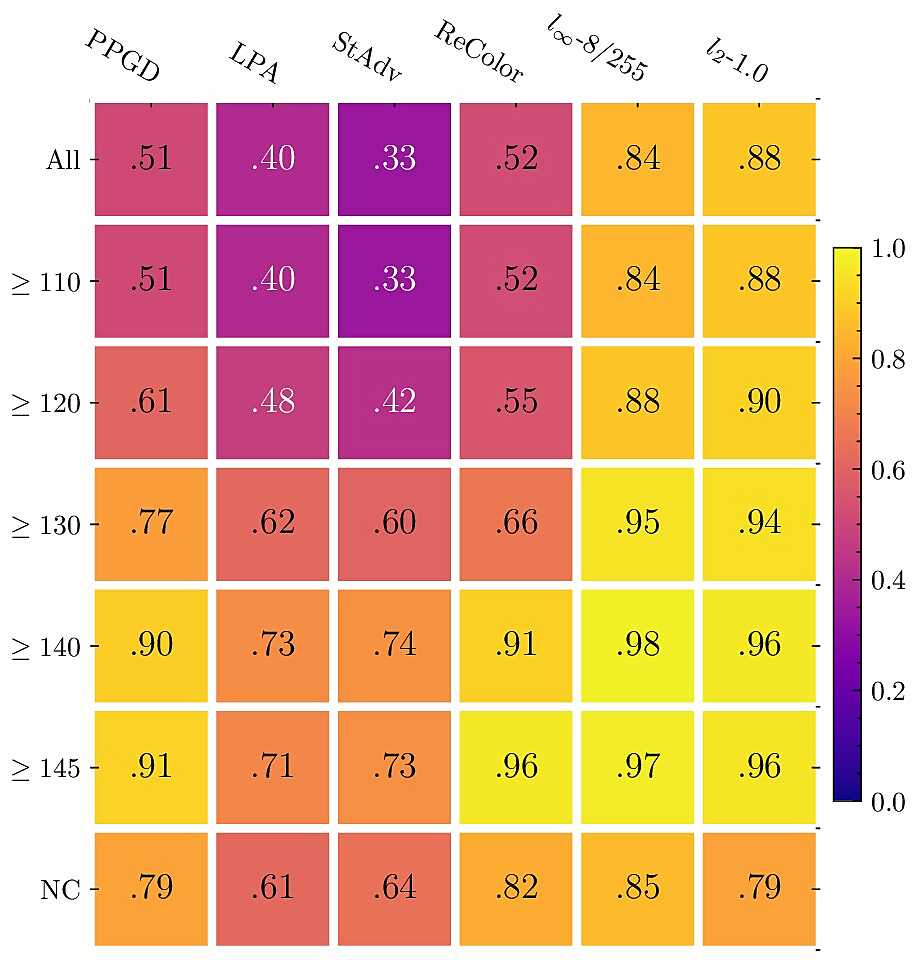}
        \caption{\textbf{\rsquare ID seen vs. unforeseen robustness for CIFAR10 $\ell_{2}$}.}
    \end{subfigure}    
    \caption{\textbf{The change of \rsquare under various threat shifts as the models with lower overall performance are removed from regression}. Each row, with the filtering threshold labeled at the lead, corresponds to a new filtered model zoo and the regression conducted it. "NC" refers to No Custom models, so all models are retrieved from either RobustBench or other published works.}
    \label{fig: inferior threat shifts}
\end{figure}

The correlation for particular shifts varies considerably as more inferior models removed. \rsquare declines considerably under CIFAR10-R, noise, fog, glass blur, frost and contrast for both Acc-Acc and Rob-Rob on CIFAR10 \linf (\cref{fig: inferior dataset shifts cifar linf}) and \lt (\cref{fig: inferior dataset shifts cifar l2}). A similar trend is also observed for threat shifts, ReColor and different \pnorm for CIFAR10 \linf as shown in \cref{fig: inferior threat shifts}. It suggests that the weak correlation under these shifts mainly results from those high-performance public models, and is likely related to the fact that these models include much diverse training methods and settings. For example, all observed catastrophic degradation under the noise shifts occur in the public models. 
Note that the locally trained models have a large diversity in model architectures particularly within the family of CNNs, but it seems that this architectural diversity does not effect the correlation as much as other factors.

\begin{figure}[!h]
    \centering
    \begin{subfigure}{\linewidth}
        \includegraphics[width=\linewidth, trim=0 25 0 25, clip]{images/legend.jpg}
    \end{subfigure}
    
    \begin{subfigure}{.325\linewidth}
        \includegraphics[width=\linewidth]{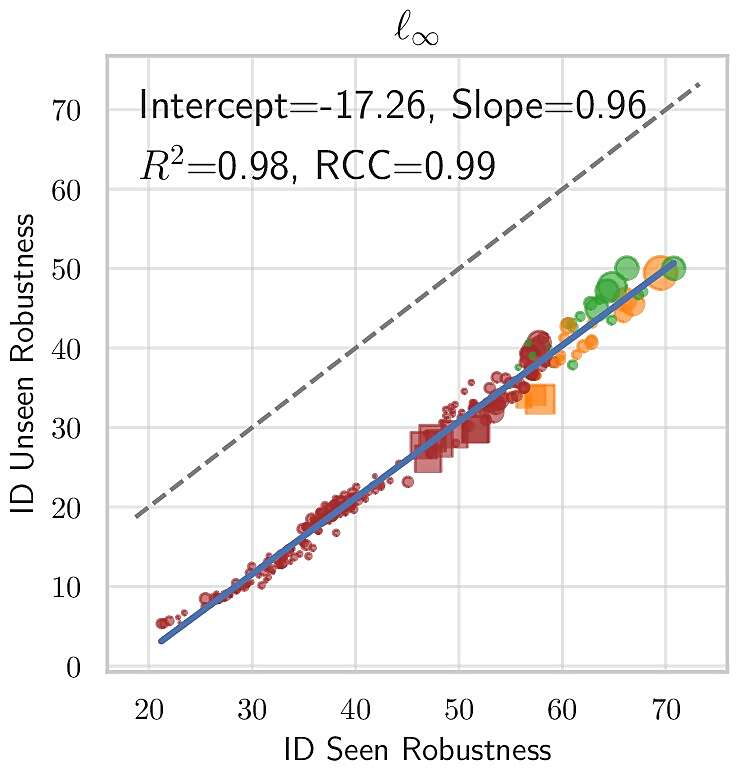}
    \end{subfigure}
    \begin{subfigure}{.325\linewidth}
        \includegraphics[width=\linewidth]{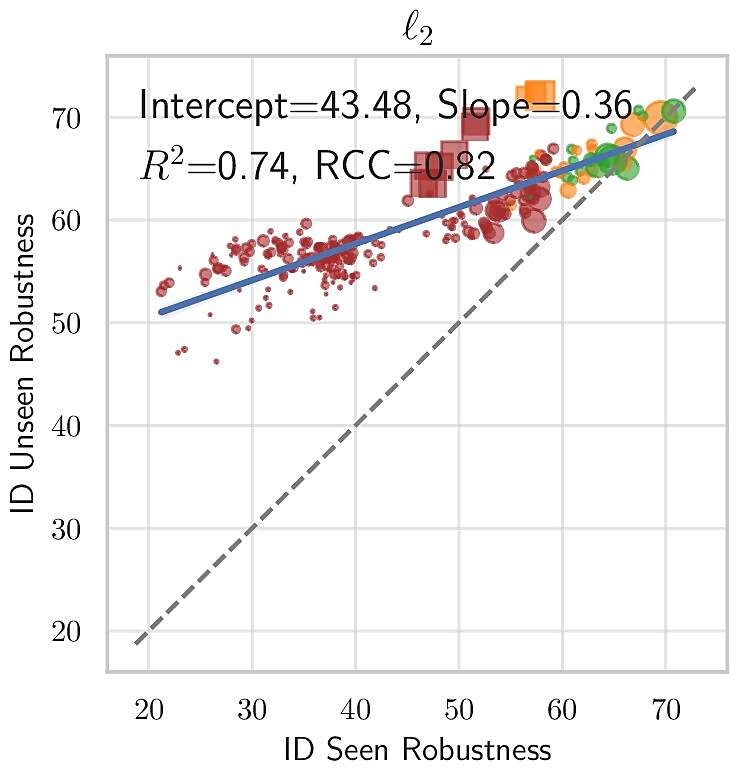}
    \end{subfigure}
    \begin{subfigure}{.325\linewidth}
        \includegraphics[width=\linewidth]{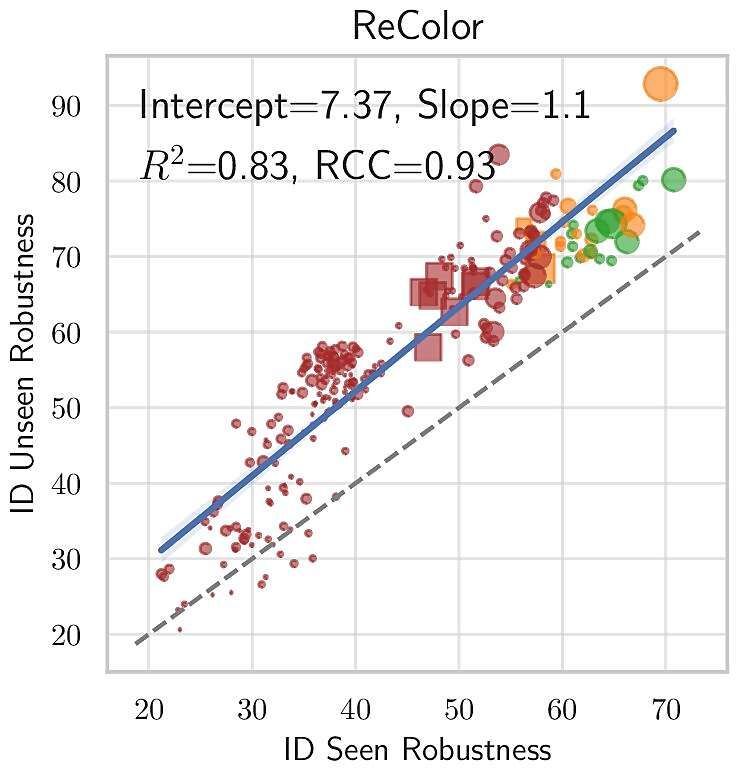}
    \end{subfigure}
    \begin{subfigure}{.325\linewidth}
        \includegraphics[width=\linewidth]{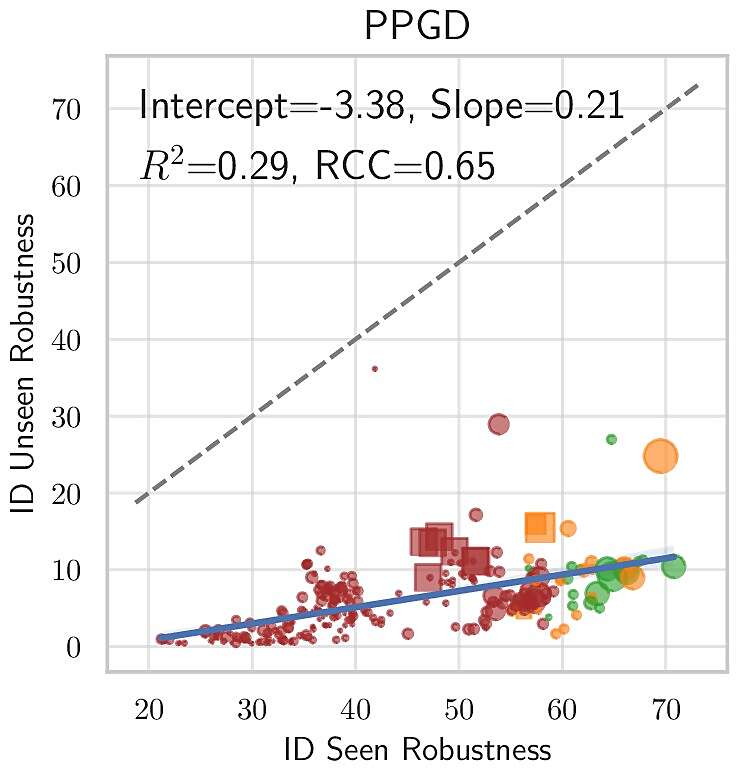}
    \end{subfigure}
    \begin{subfigure}{.325\linewidth}
        \includegraphics[width=\linewidth]{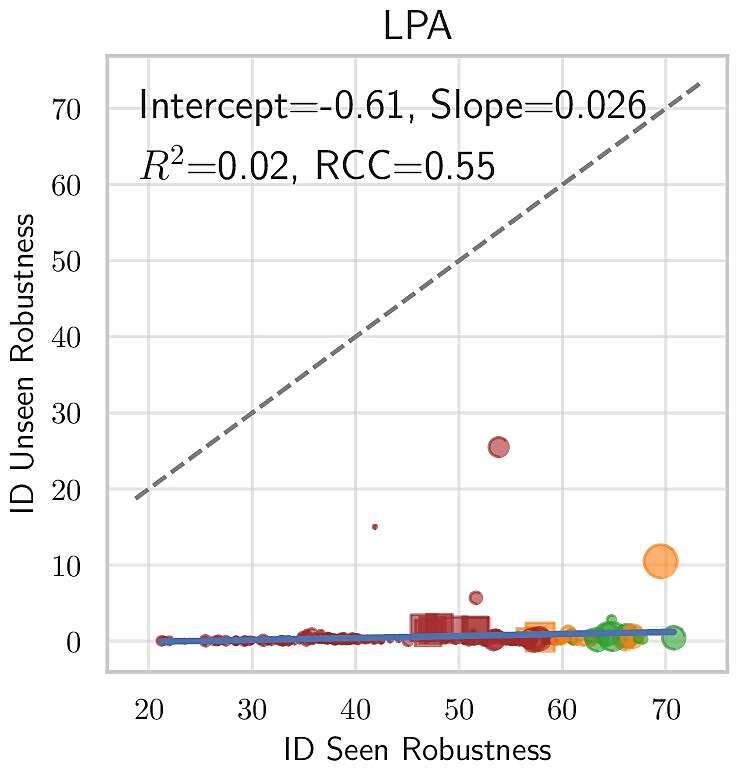}
    \end{subfigure}
    \begin{subfigure}{.325\linewidth}
        \includegraphics[width=\linewidth]{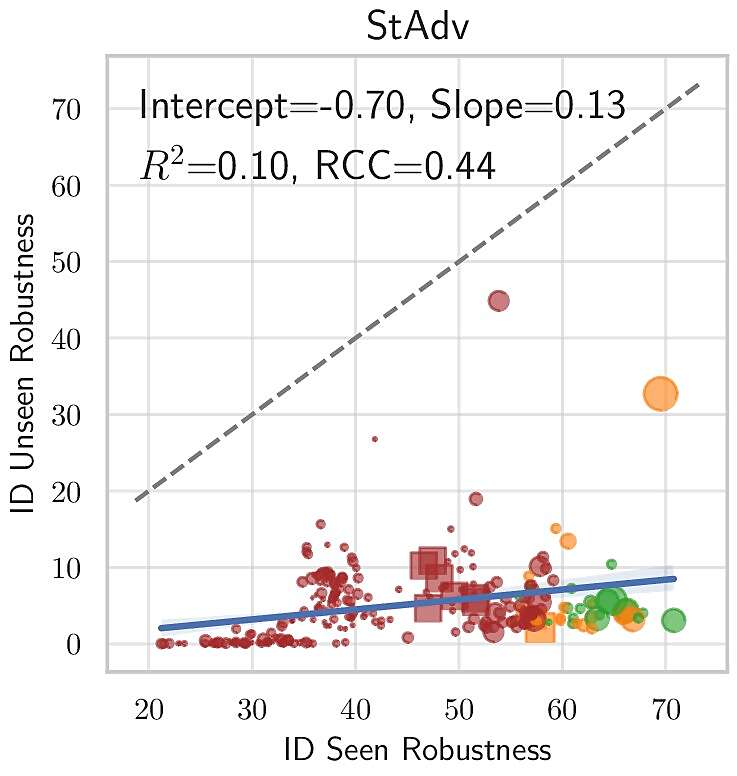}
    \end{subfigure}
    \caption{\textbf{Correlation between seen and unforeseen robustness on ID data for CIFAR10 \linf AT models}.}
    \label{fig: threat shift plots cifar10 linf}
\end{figure}

\begin{figure}[!h]
    \centering
    \begin{subfigure}{\linewidth}
        \includegraphics[width=\linewidth, trim=0 25 0 25, clip]{images/legend.jpg}
    \end{subfigure}
    \begin{subfigure}{.325\linewidth}
        \includegraphics[width=\linewidth]{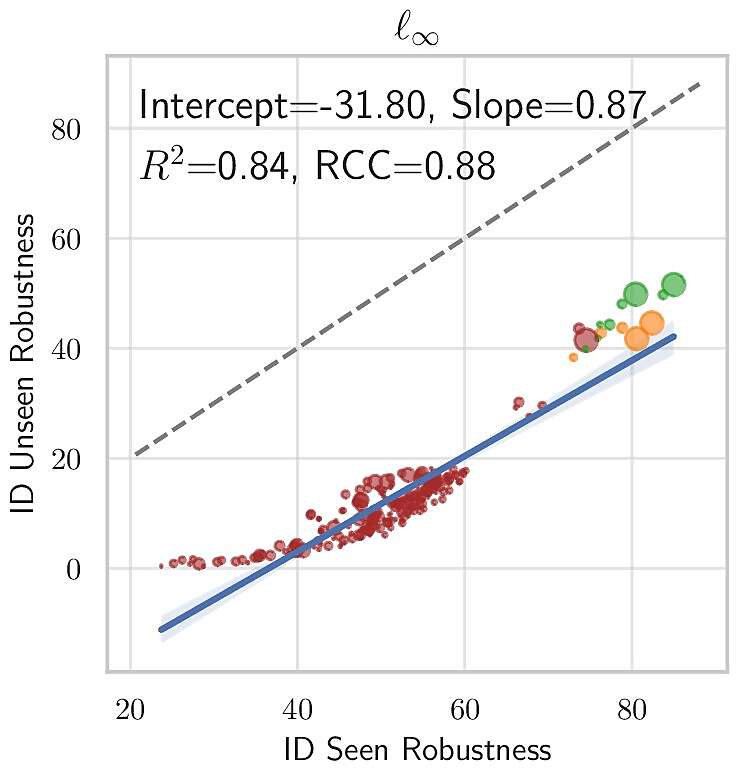}
    \end{subfigure}
    \begin{subfigure}{.325\linewidth}
        \includegraphics[width=\linewidth]{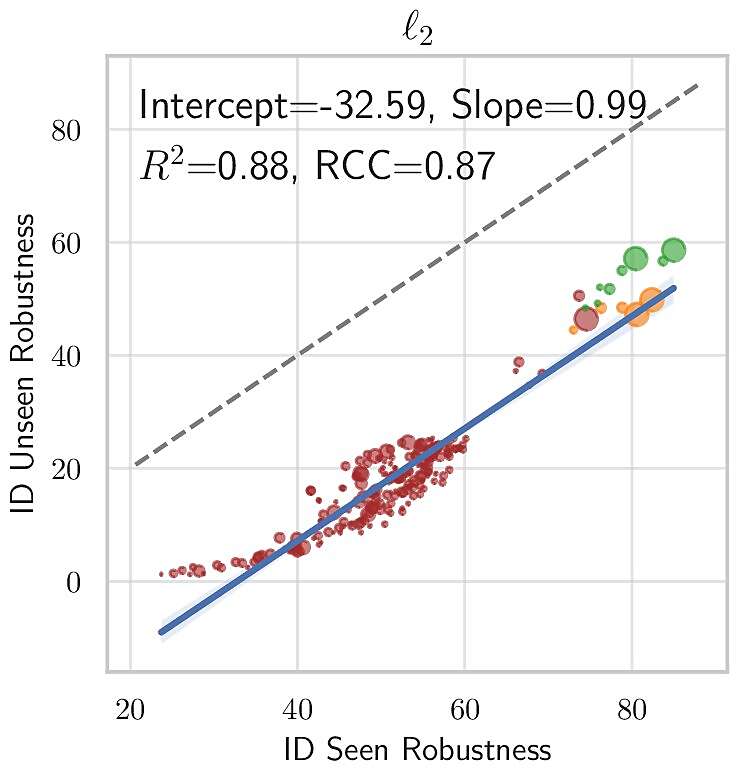}
    \end{subfigure}
    \begin{subfigure}{.325\linewidth}
        \includegraphics[width=\linewidth]{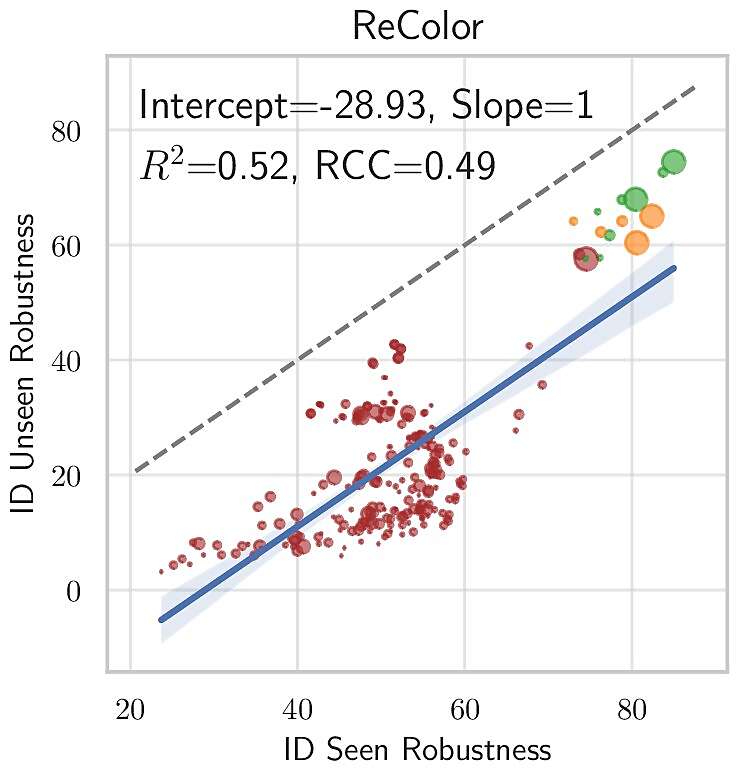}
    \end{subfigure}
    \begin{subfigure}{.325\linewidth}
        \includegraphics[width=\linewidth]{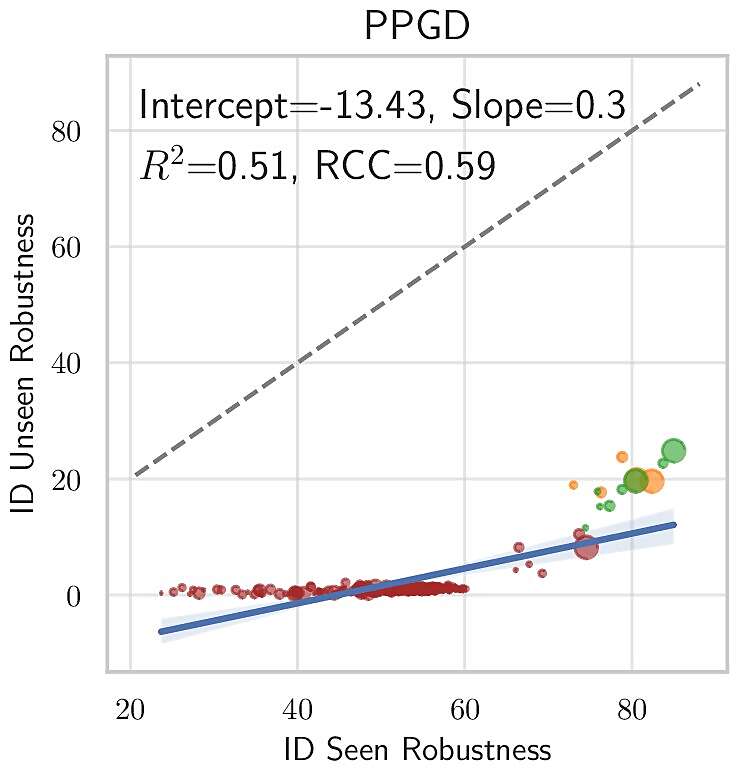}
    \end{subfigure}
    \begin{subfigure}{.325\linewidth}
        \includegraphics[width=\linewidth]{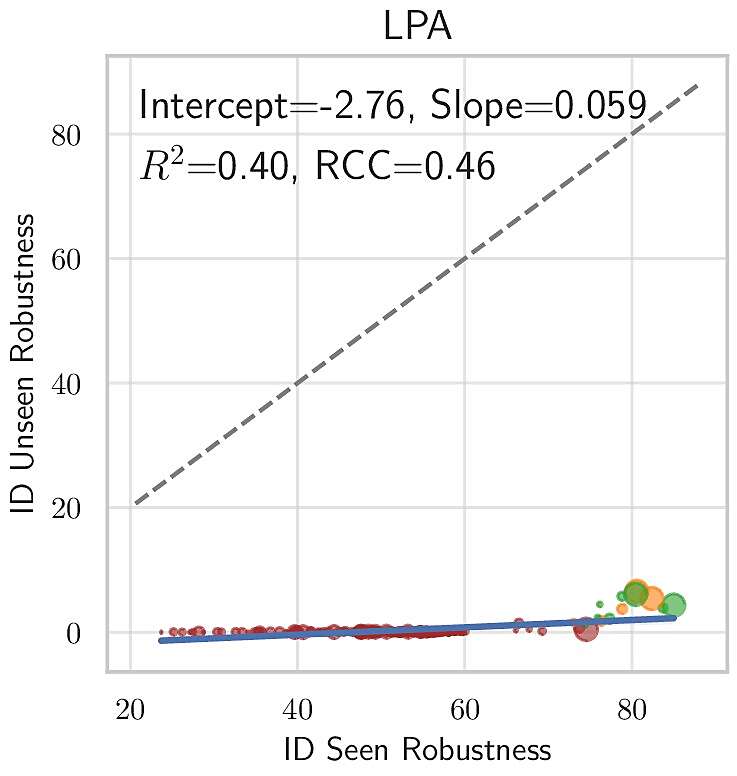}
    \end{subfigure}
    \begin{subfigure}{.325\linewidth}
        \includegraphics[width=\linewidth]{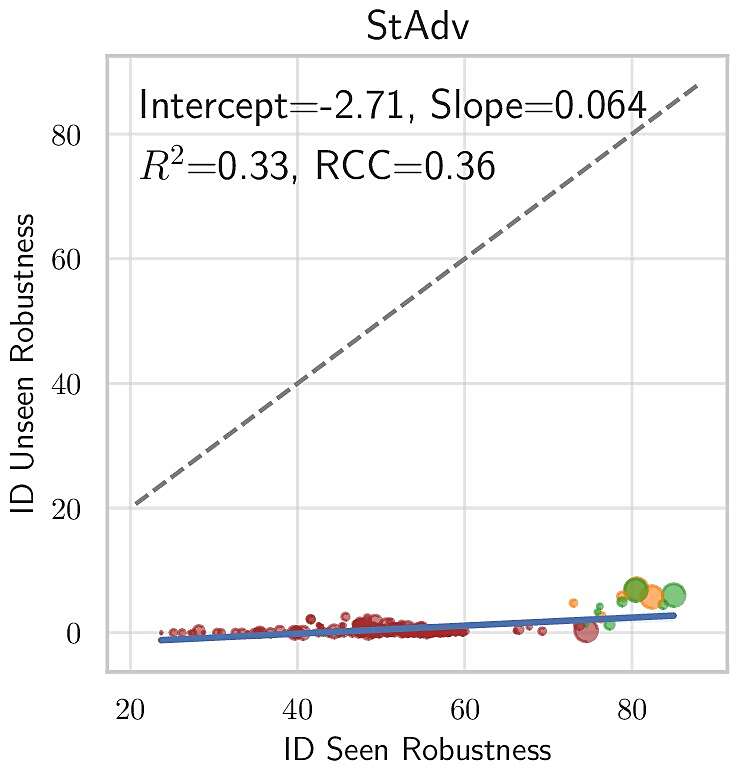}
    \end{subfigure}
    \caption{\textbf{Correlation between seen and unforeseen robustness on ID data for CIFAR10 \lt AT models}.}
    \label{fig: threat shift plots cifar10 l2}
\end{figure}

\begin{figure}[!h]
    \centering
    \begin{subfigure}{\linewidth}
        \includegraphics[width=\linewidth, trim=0 25 0 25, clip]{images/legend.jpg}
    \end{subfigure}
    \begin{subfigure}{.325\linewidth}
        \includegraphics[width=\linewidth]{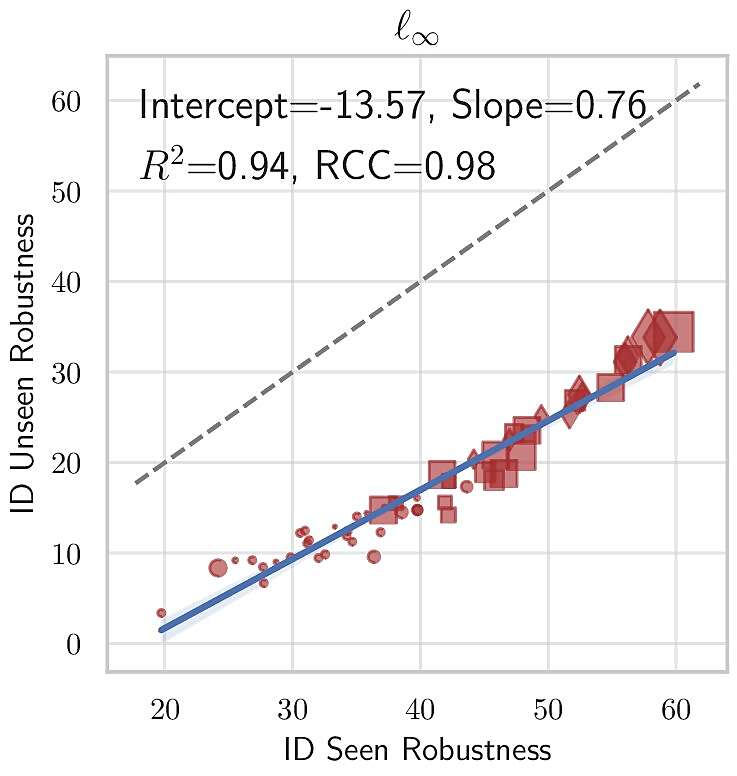}
    \end{subfigure}
    \begin{subfigure}{.325\linewidth}
        \includegraphics[width=\linewidth]{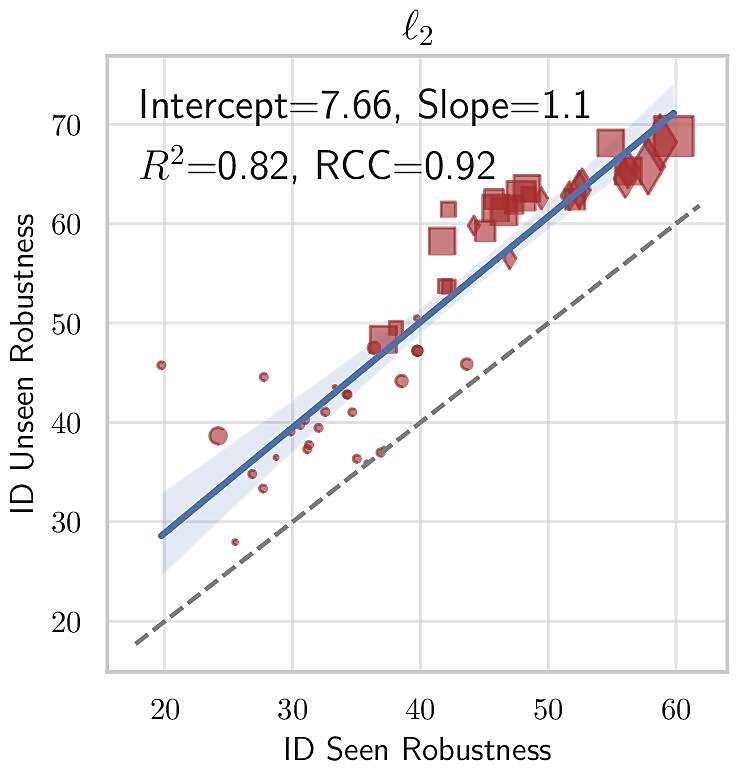}
    \end{subfigure}
    \begin{subfigure}{.325\linewidth}
        \includegraphics[width=\linewidth]{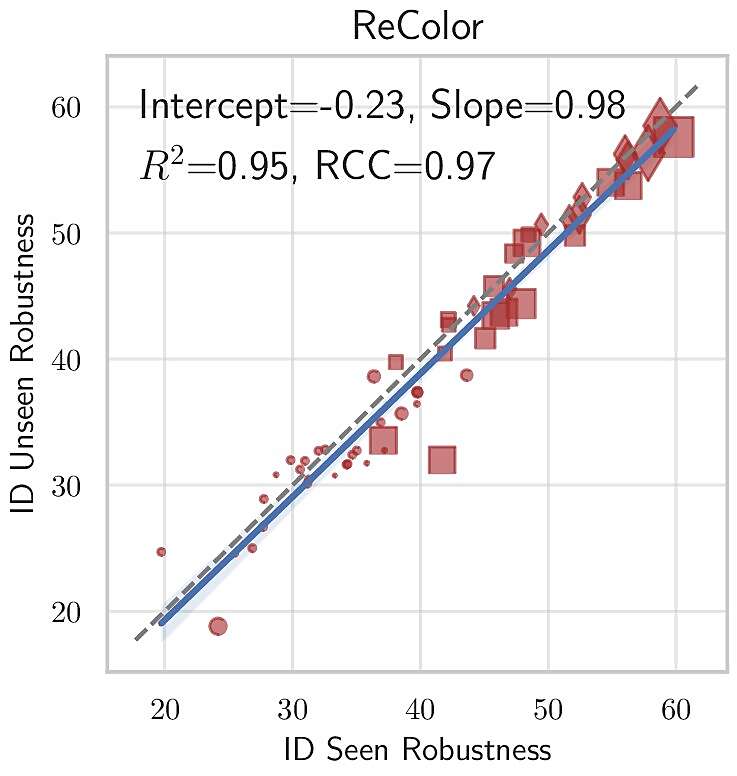}
    \end{subfigure}
    \begin{subfigure}{.325\linewidth}
        \includegraphics[width=\linewidth]{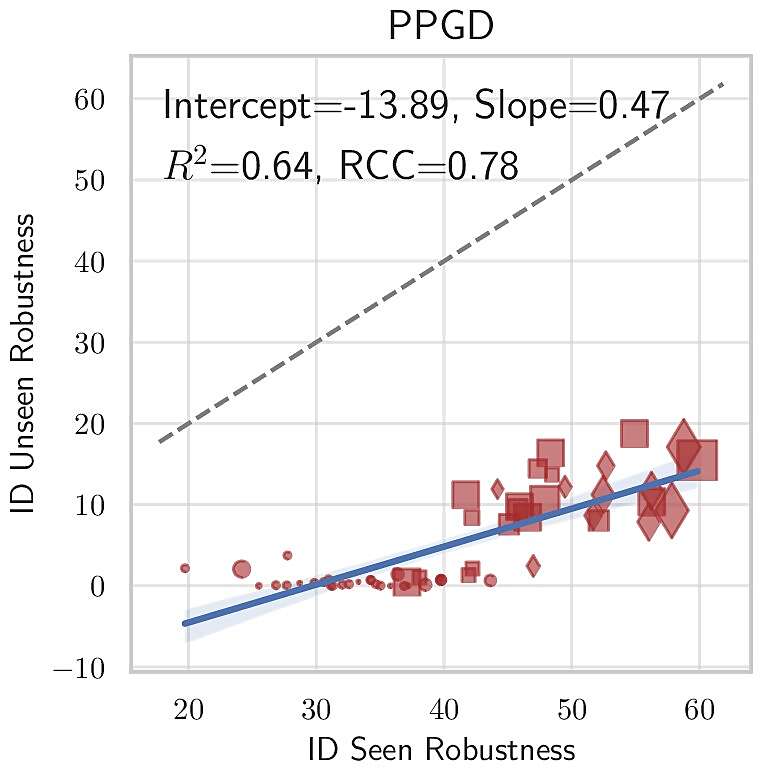}
    \end{subfigure}
    \begin{subfigure}{.325\linewidth}
        \includegraphics[width=\linewidth]{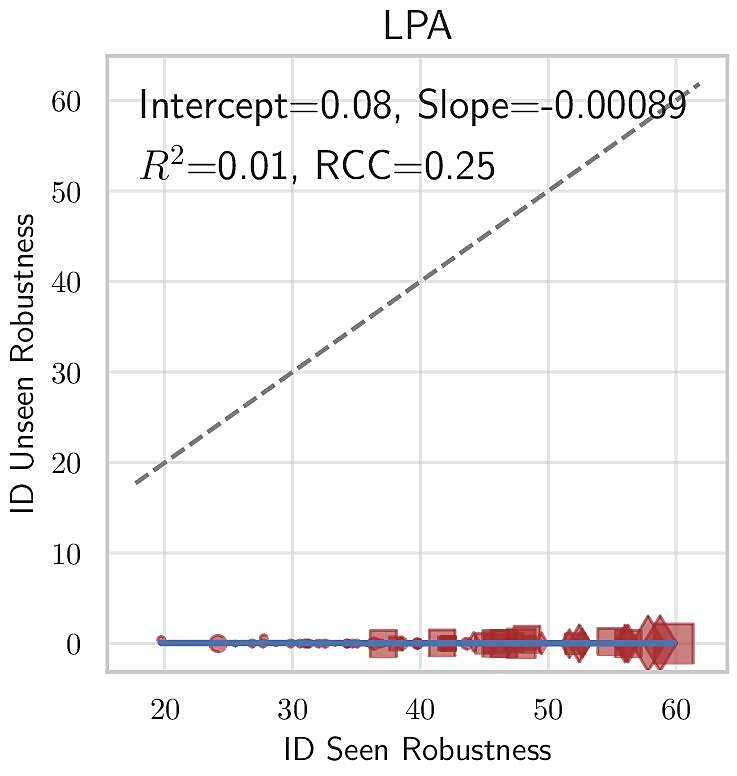}
    \end{subfigure}
    \begin{subfigure}{.325\linewidth}
        \includegraphics[width=\linewidth]{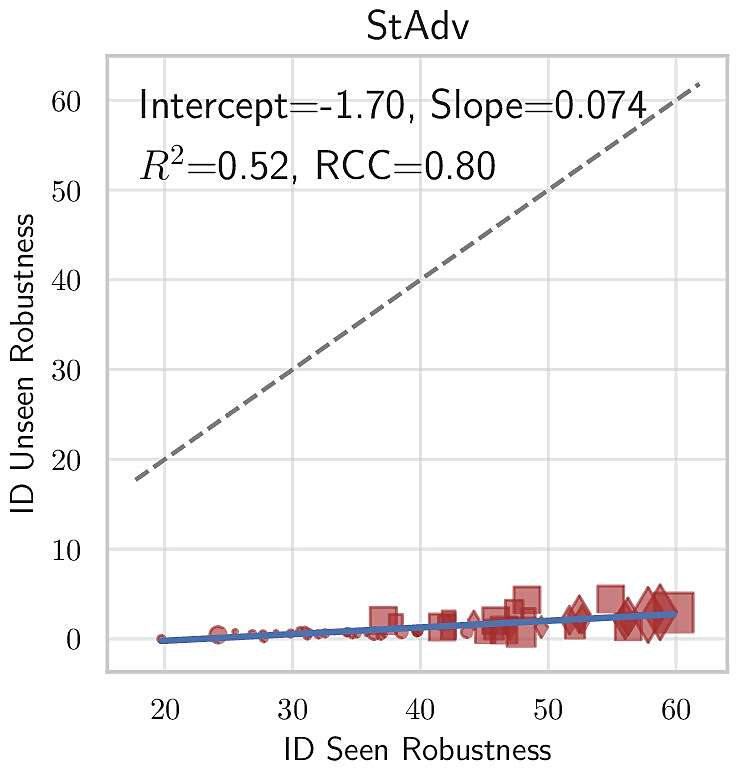}
    \end{subfigure}
    \caption{\textbf{Correlation between seen and unforeseen robustness on ID data for ImageNet \linf AT models}.}
    \label{fig: threat shift imagenet}
\end{figure}

In contrast, correlation is improved for most threat shifts for CIFAR10 \lt as shown in \cref{fig: inferior threat shifts}. As shown in \cref{fig: threat shift plots cifar10 l2}, the locally trained (inferior) models and the public (high-performance) models have divergent linear trends (most evident in the plot of PPGD). That's why removing models from either group will enhance the correlation.
Note that such divergence is not evident in the figures of CIFAR10 \linf (\cref{fig: threat shift plots cifar10 linf}) and ImageNet \linf (\cref{fig: threat shift imagenet}). 

\subsection{No Evident Correlation when ID and
OOD Metrics Misalign}
\label{appendix: why no correlation metrics misalign}
Inferior models also cause OOD robustness to not consistently increase with the ID accuracy, i.e., the poor correlation between ID accuracy (robustness) and OOD robustness (accuracy) because they have high accuracy yet poor robustness. 
These models are mainly produced by some of our custom training receipts and take a considerable proportion of our CIFAR-10 model zoo, whereas the model zoo of ImageNet is dominated by ones from public sources.

\clearpage
\section{Methods for Improving OOD Adversarial Robustness} \label{appendix: effective robust intervention}
All models used in this analysis are retrieved from RobustBench or other published works to ensure they are well-trained by the techniques to be examined. The specific experiment setting for each model can be found in its original paper.

\input{tables/robust-intervention}

    


\clearpage
\section{Plots of ID-OOD Correlation per Dataset Shift}
\label{appendix: correlation per dataset shift}

\begin{figure*}[!h]
    \centering
    \begin{subfigure}{\linewidth}
        \includegraphics[width=\linewidth, trim=0 25 0 25, clip]{images/legend.jpg}
    \end{subfigure}
    \begin{subfigure}{.245\textwidth}
        \includegraphics[width=\linewidth]{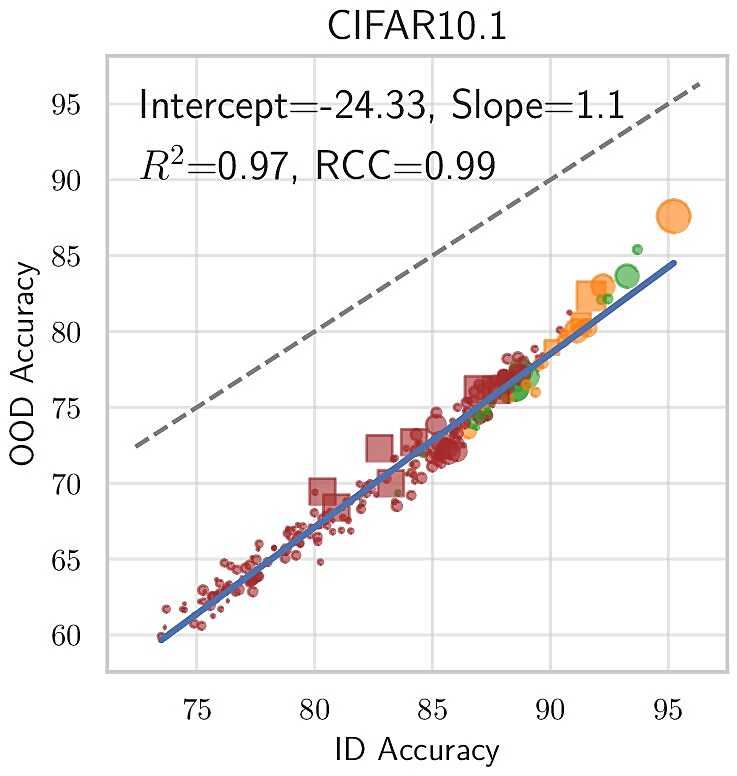}
    \end{subfigure}
    \begin{subfigure}{.245\textwidth}
        \includegraphics[width=\linewidth]{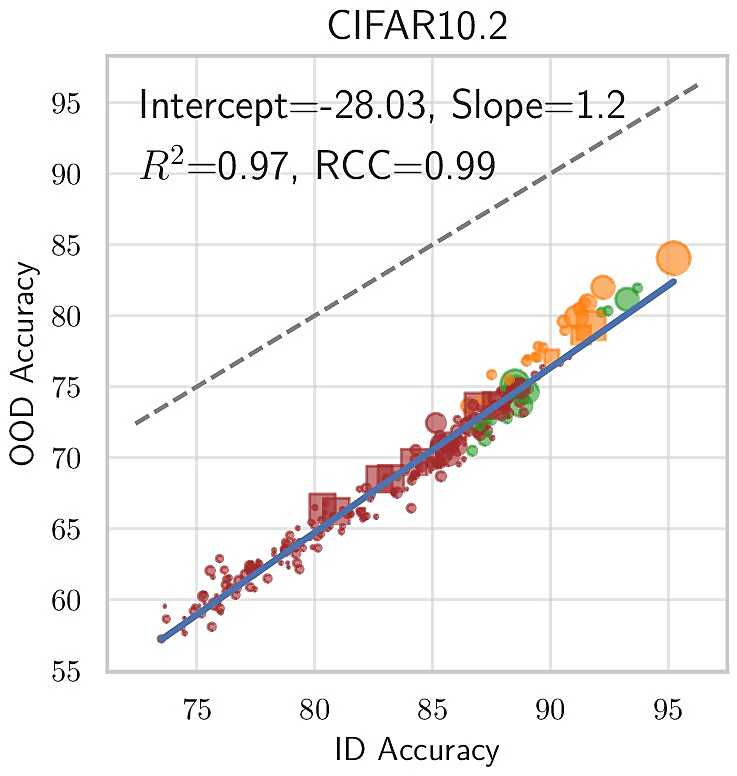}
    \end{subfigure}
    \begin{subfigure}{.245\textwidth}
        \includegraphics[width=\linewidth]{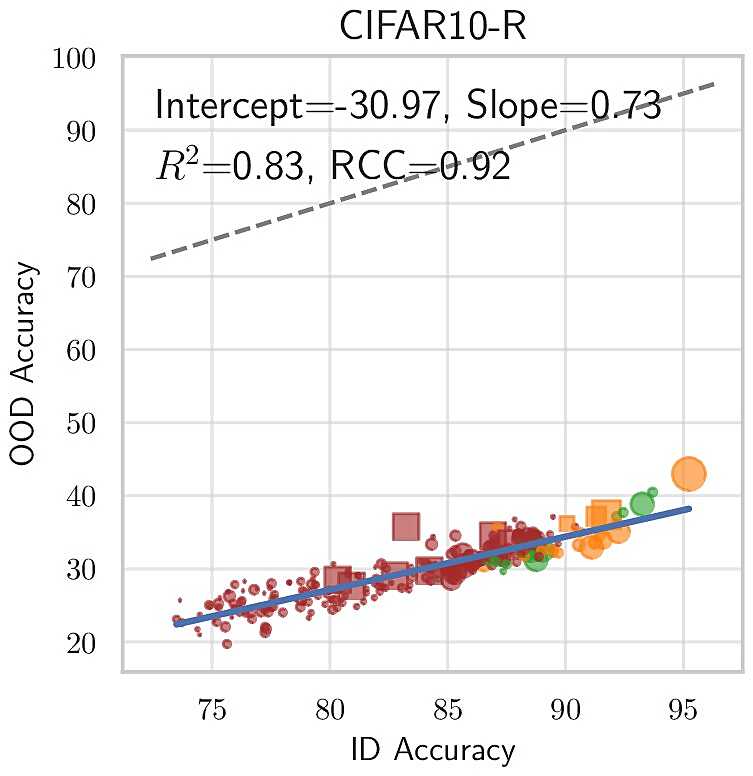}
    \end{subfigure}
    \begin{subfigure}{.245\textwidth}
        \includegraphics[width=\linewidth]{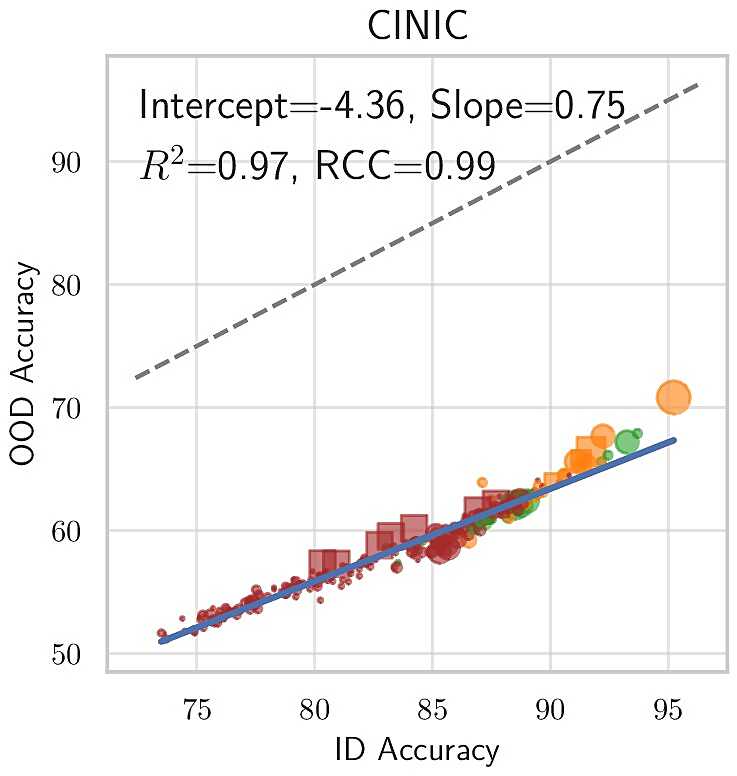}
    \end{subfigure}

    \begin{subfigure}{.245\textwidth}
        \includegraphics[width=\linewidth]{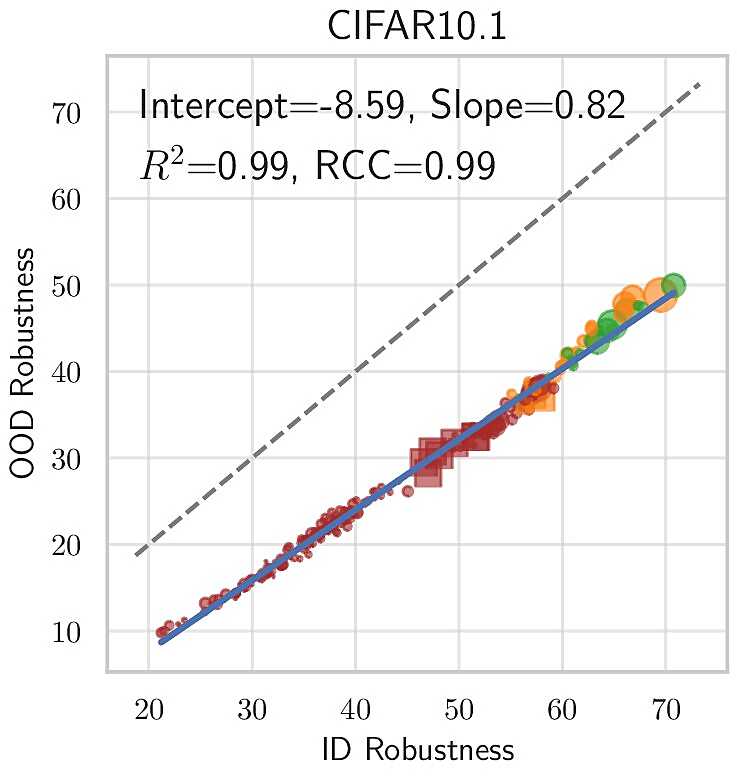}
    \end{subfigure}
    \begin{subfigure}{.245\textwidth}
        \includegraphics[width=\linewidth]{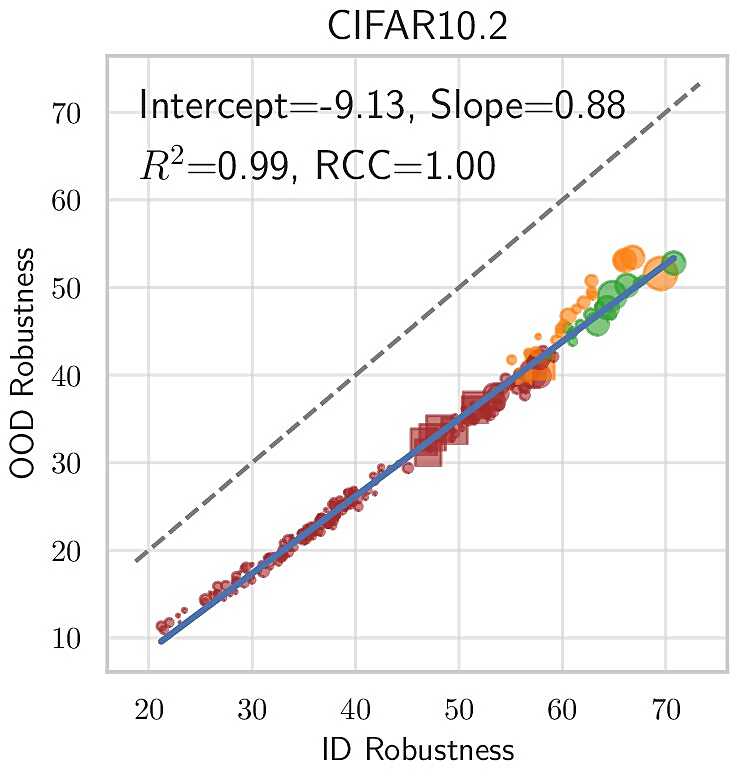}
    \end{subfigure}
    \begin{subfigure}{.245\textwidth}
        \includegraphics[width=\linewidth]{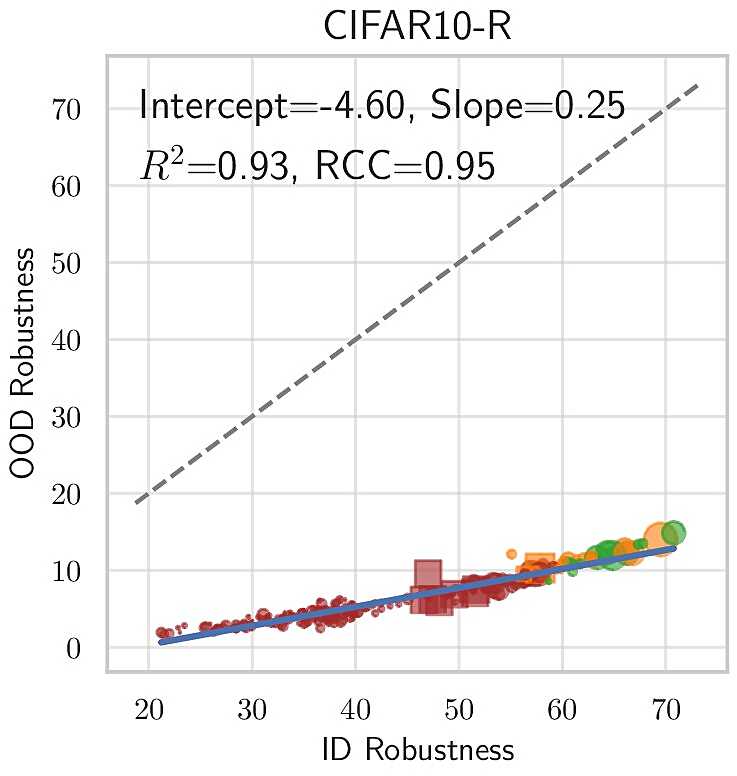}
    \end{subfigure}
    \begin{subfigure}{.245\textwidth}
        \includegraphics[width=\linewidth]{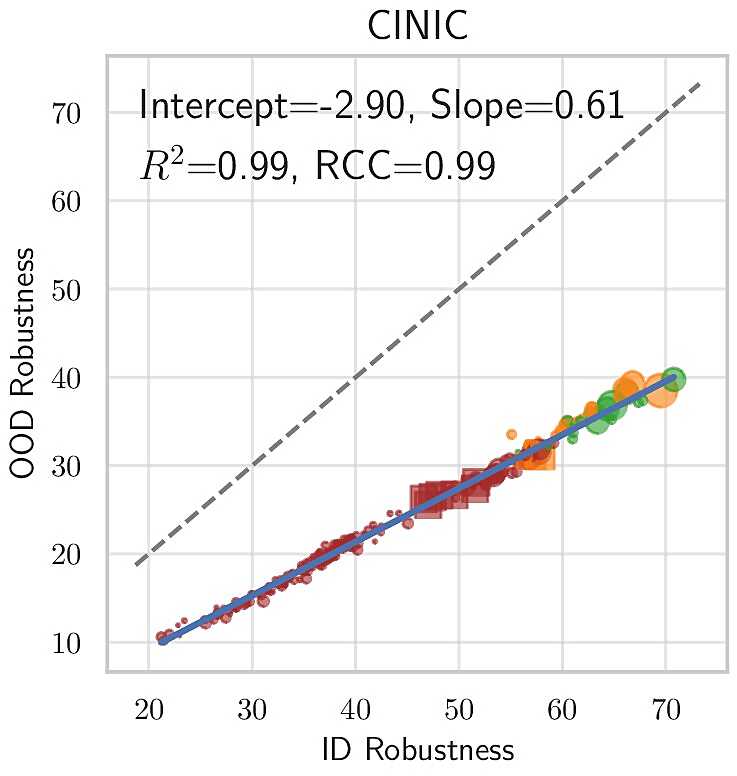}
    \end{subfigure}

    \begin{subfigure}{.245\textwidth}
        \includegraphics[width=\linewidth]{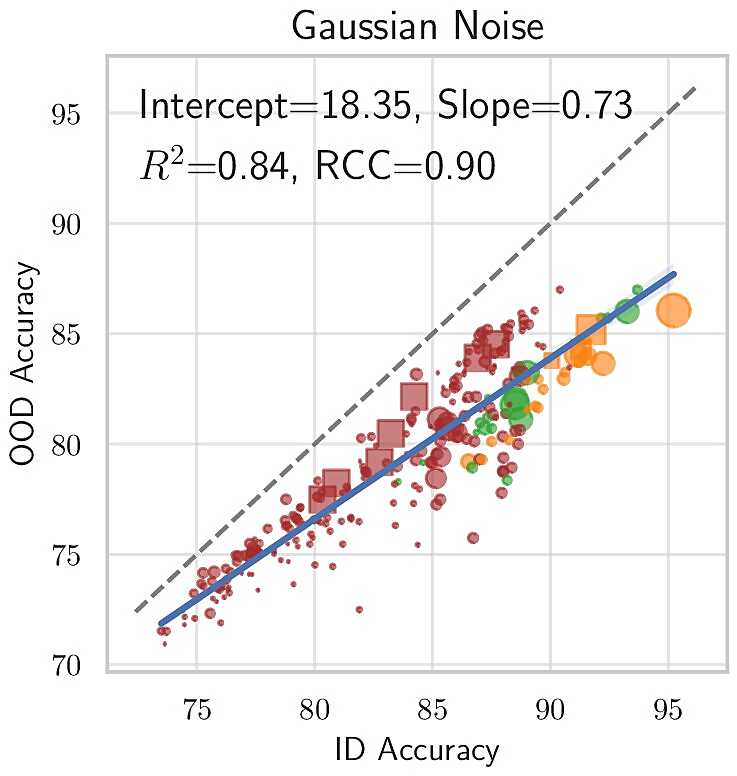}
    \end{subfigure}
    \begin{subfigure}{.245\textwidth}
        \includegraphics[width=\linewidth]{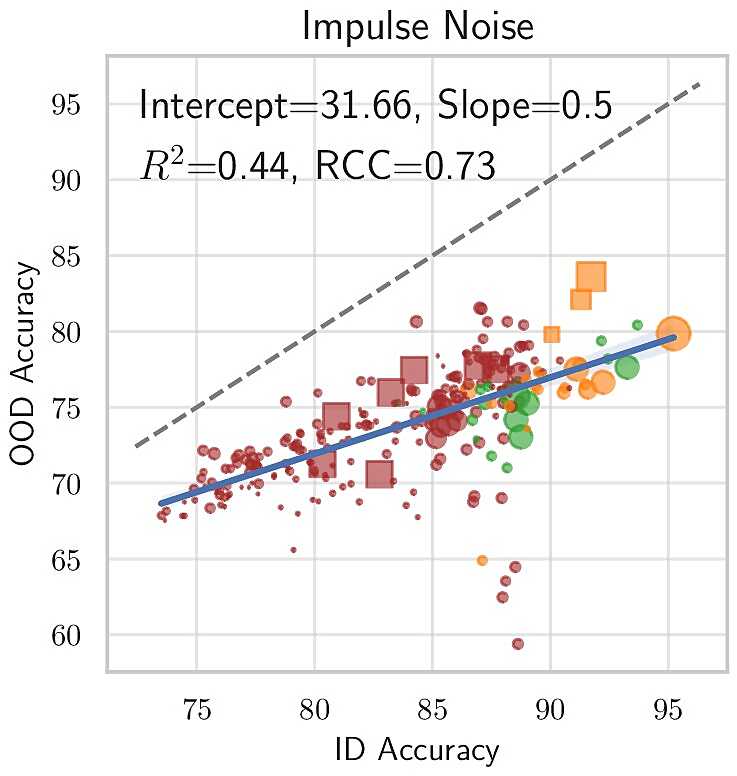}
    \end{subfigure}
    \begin{subfigure}{.245\textwidth}
        \includegraphics[width=\linewidth]{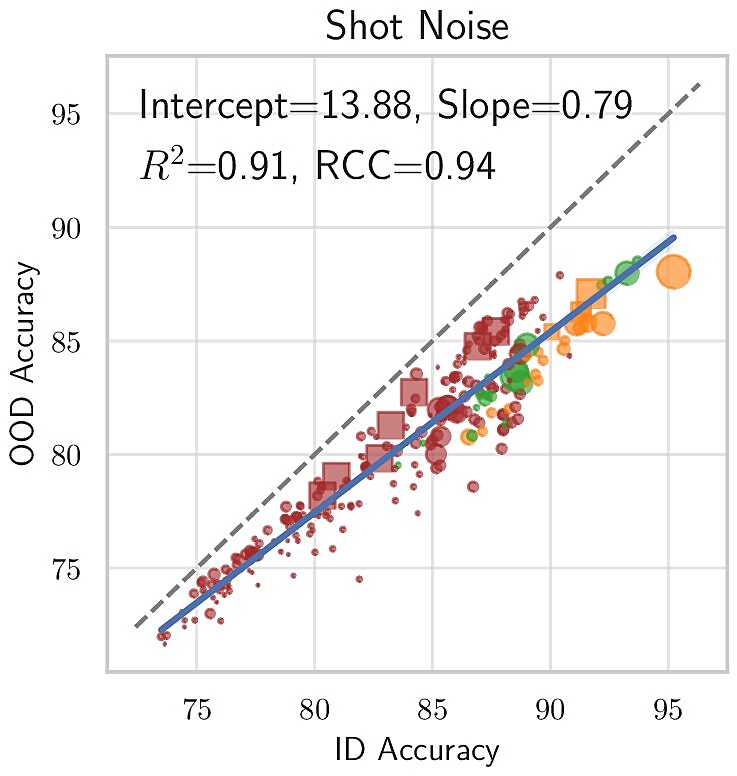}
    \end{subfigure}
    \begin{subfigure}{.245\textwidth}
        \includegraphics[width=\linewidth]{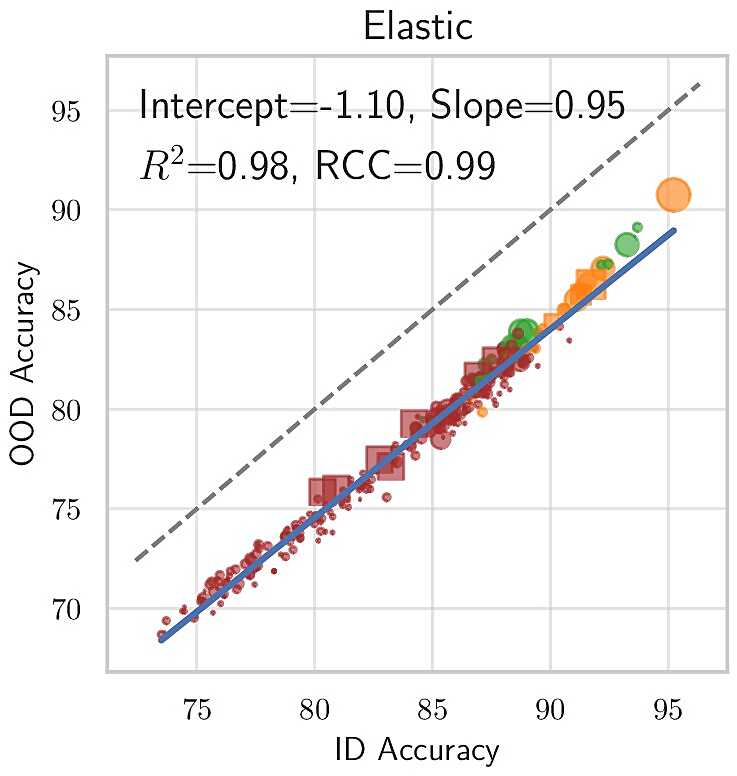}
    \end{subfigure}

    \begin{subfigure}{.245\textwidth}
        \includegraphics[width=\linewidth]{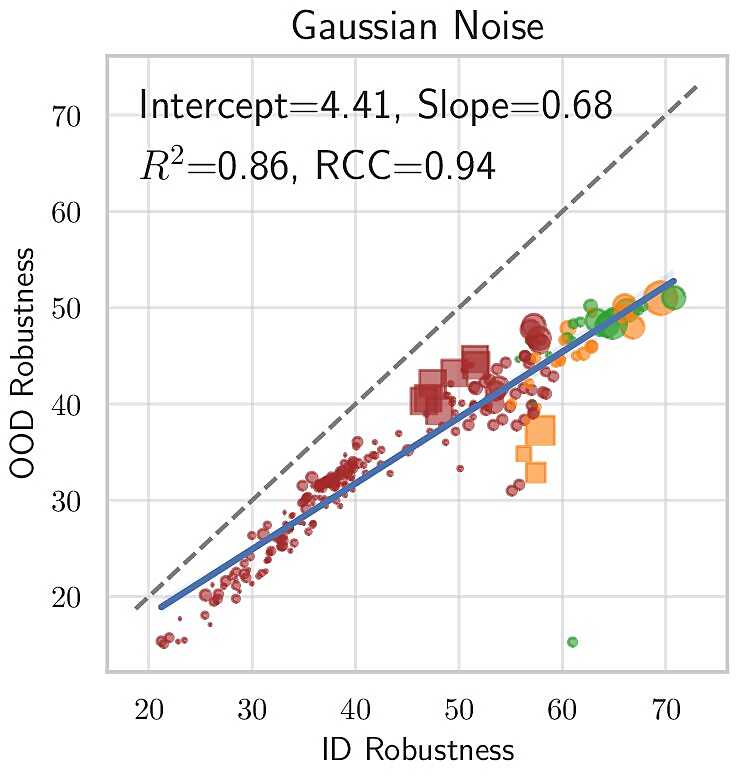}
    \end{subfigure}
    \begin{subfigure}{.245\textwidth}
        \includegraphics[width=\linewidth]{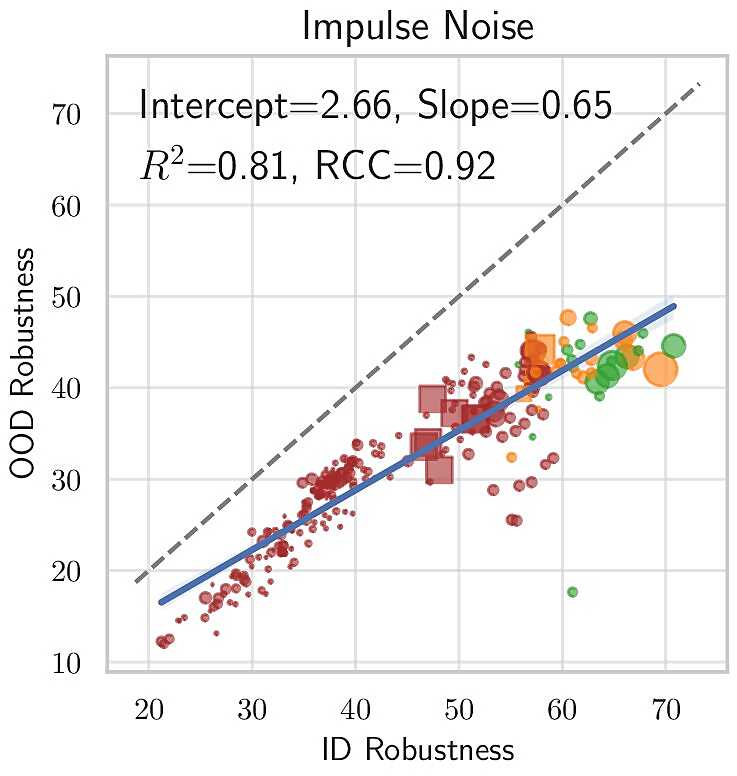}
    \end{subfigure}
    \begin{subfigure}{.245\textwidth}
        \includegraphics[width=\linewidth]{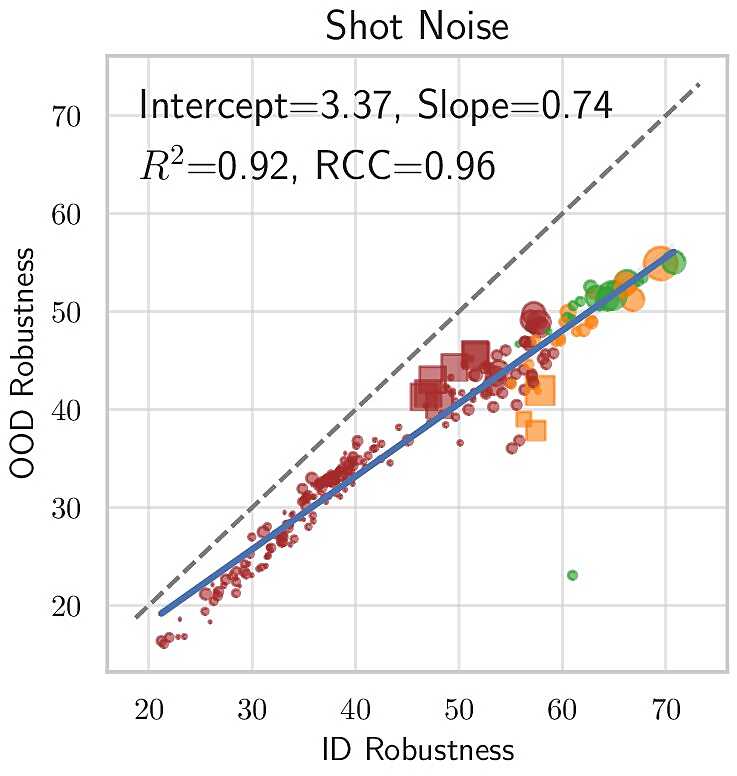}
    \end{subfigure}
    \begin{subfigure}{.245\textwidth}
        \includegraphics[width=\linewidth]{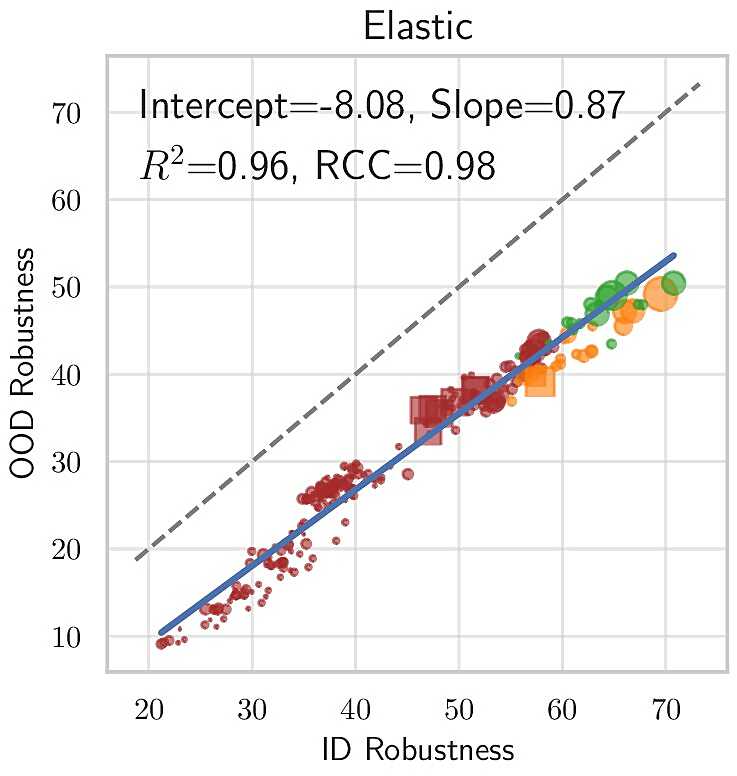}
    \end{subfigure}
    
    \caption{\textbf{Correlation between ID accuracy and OOD accuracy (odd rows); ID robustness and OOD robustness (even rows) for CIFAR10 \linf AT models}.}
    \label{fig: correlate per dataset shift cifar10 linf}
    
\end{figure*}

\begin{figure*}[!h]
    \centering
    \begin{subfigure}{\linewidth}
        \includegraphics[width=\linewidth, trim=0 25 0 25, clip]{images/legend.jpg}
    \end{subfigure}

    \begin{subfigure}{.245\textwidth}
        \includegraphics[width=\linewidth]{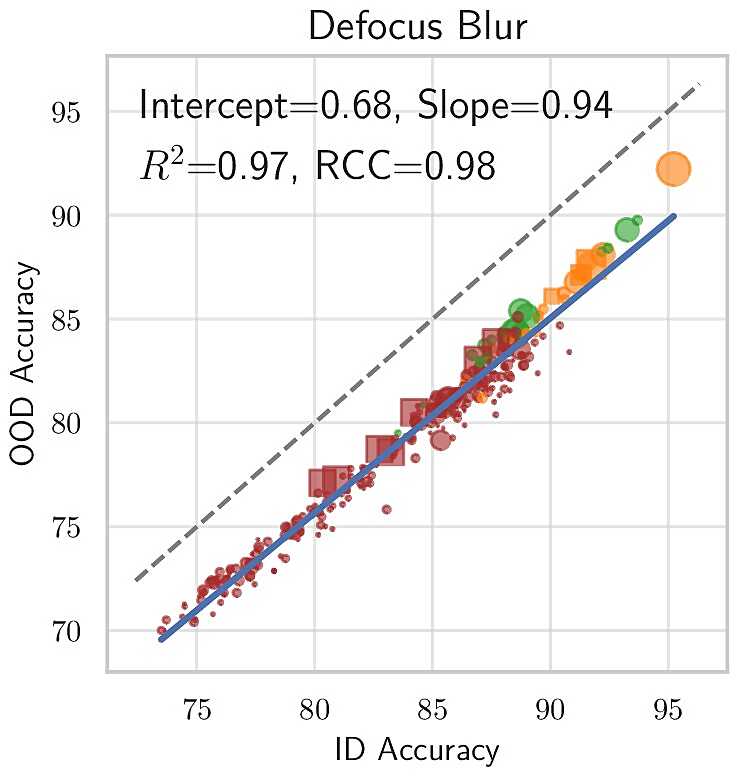}
    \end{subfigure}
    \begin{subfigure}{.245\textwidth}
        \includegraphics[width=\linewidth]{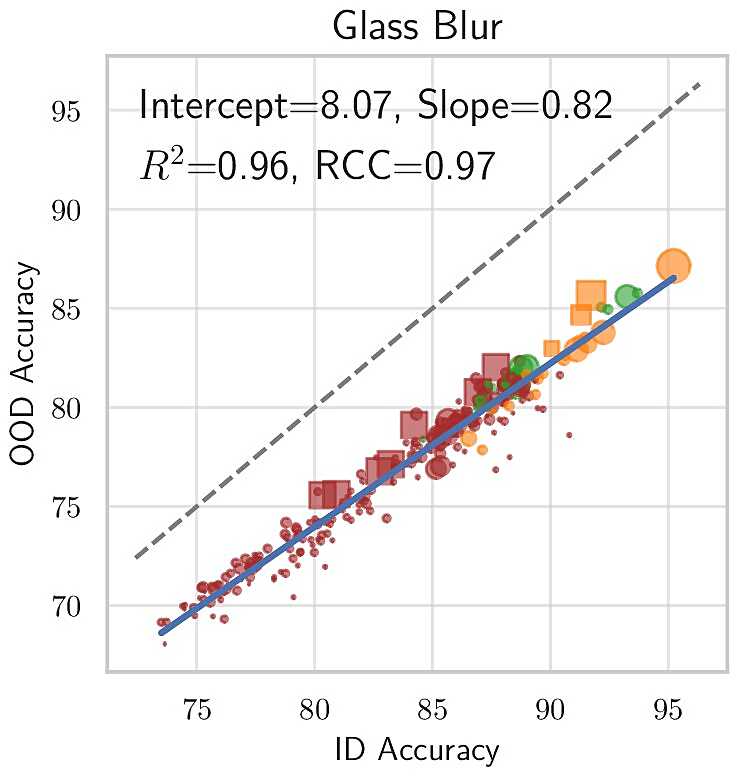}
    \end{subfigure}
    \begin{subfigure}{.245\textwidth}
        \includegraphics[width=\linewidth]{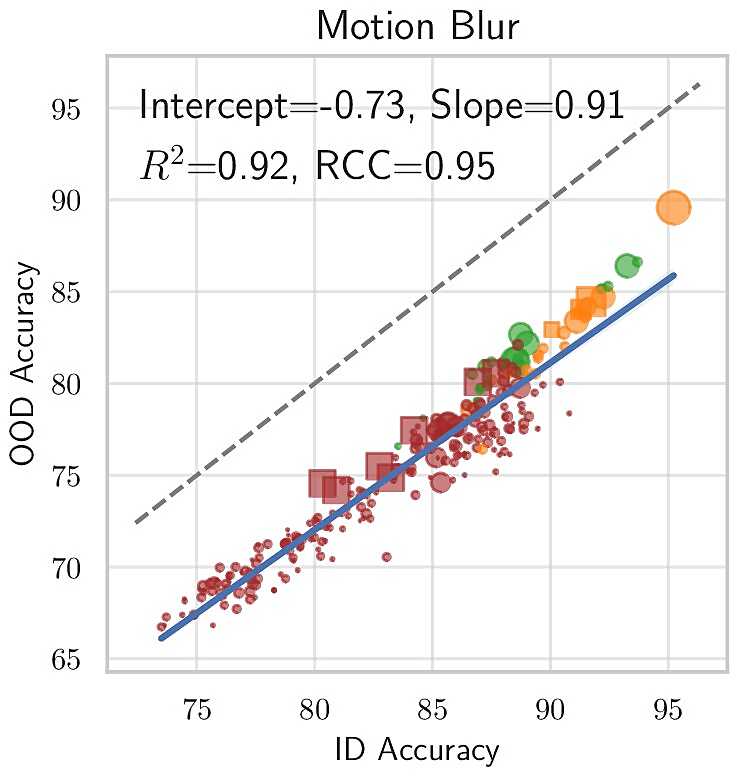}
    \end{subfigure}
    \begin{subfigure}{.245\textwidth}
        \includegraphics[width=\linewidth]{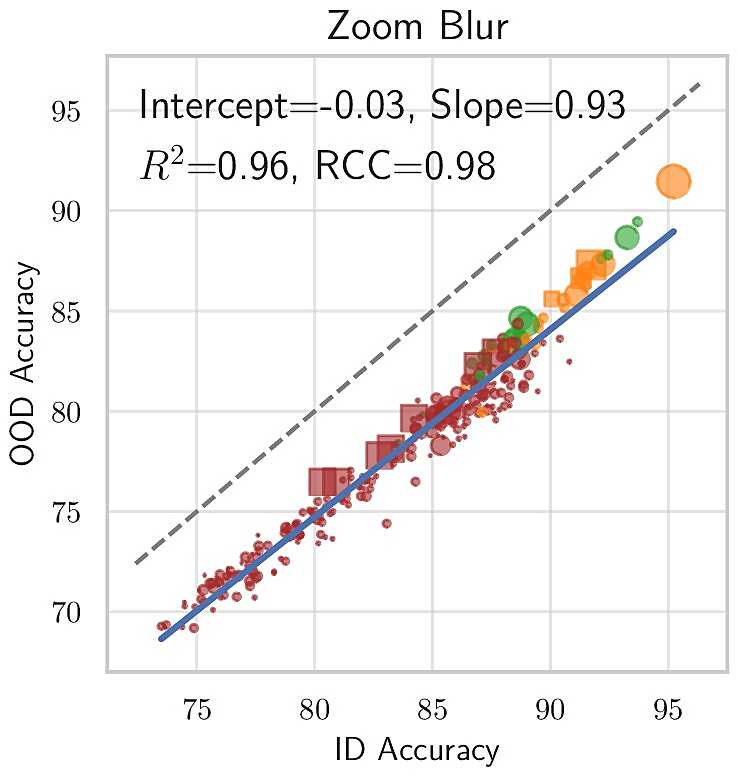}
    \end{subfigure}

    \begin{subfigure}{.245\textwidth}
        \includegraphics[width=\linewidth]{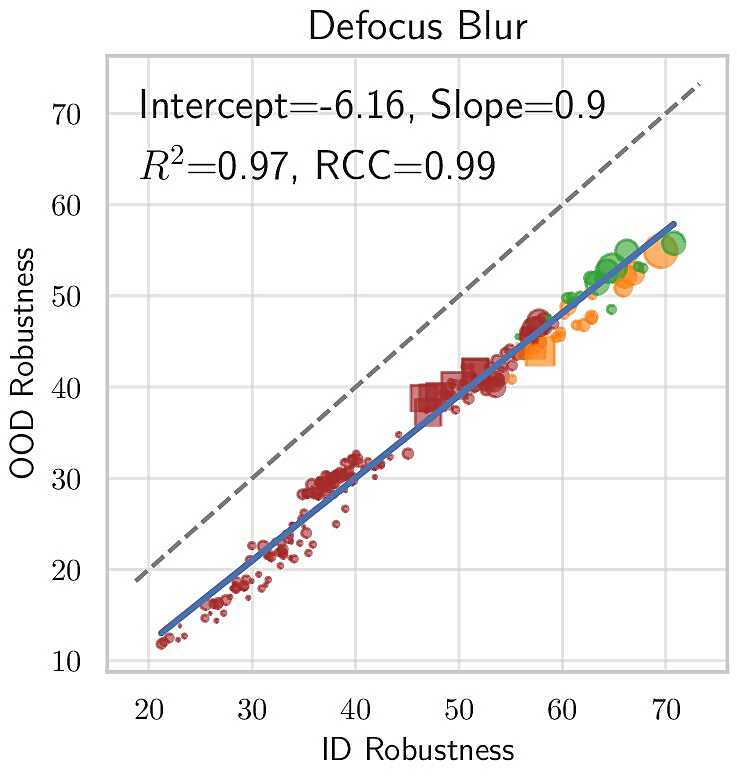}
    \end{subfigure}
    \begin{subfigure}{.245\textwidth}
        \includegraphics[width=\linewidth]{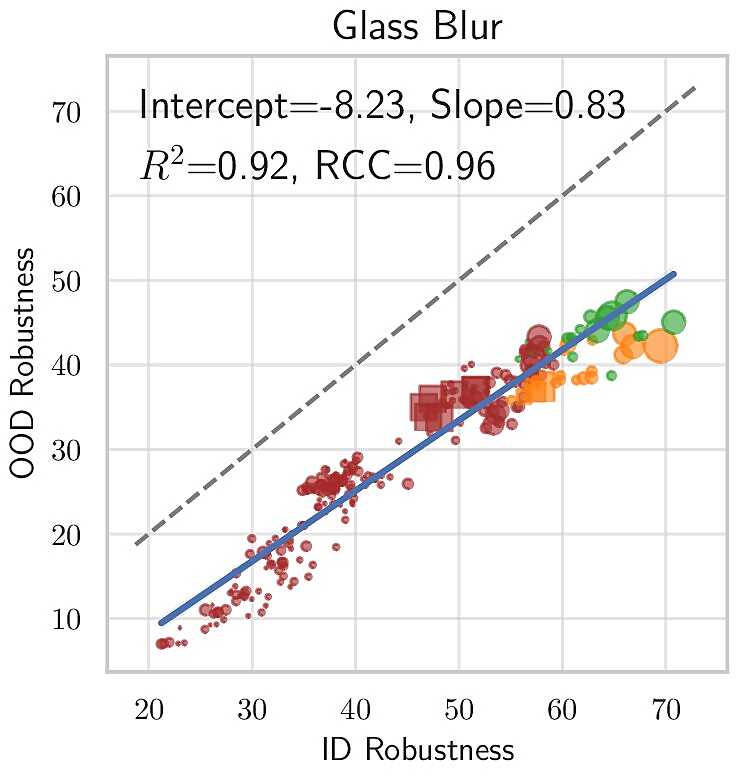}
    \end{subfigure}
    \begin{subfigure}{.245\textwidth}
        \includegraphics[width=\linewidth]{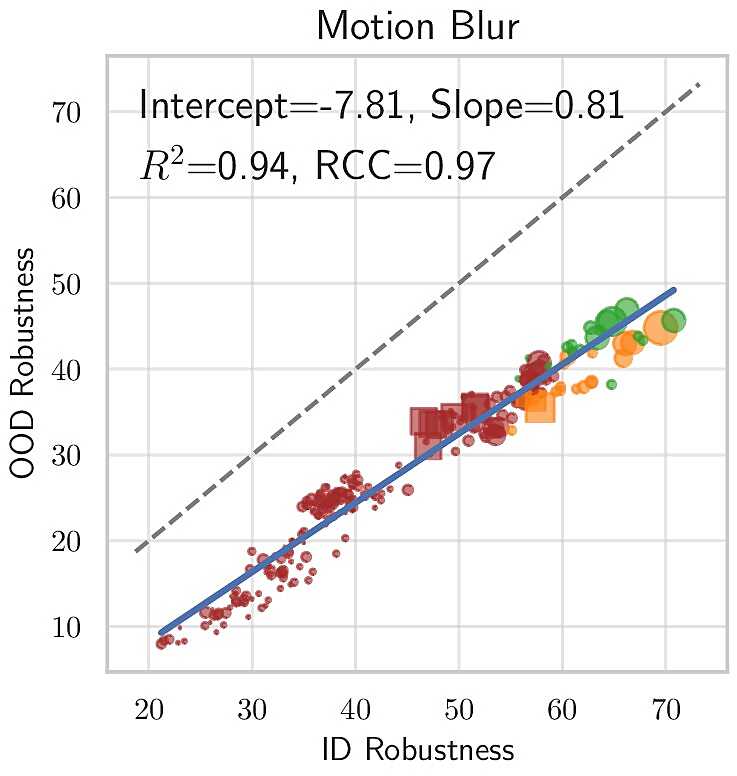}
    \end{subfigure}
    \begin{subfigure}{.245\textwidth}
        \includegraphics[width=\linewidth]{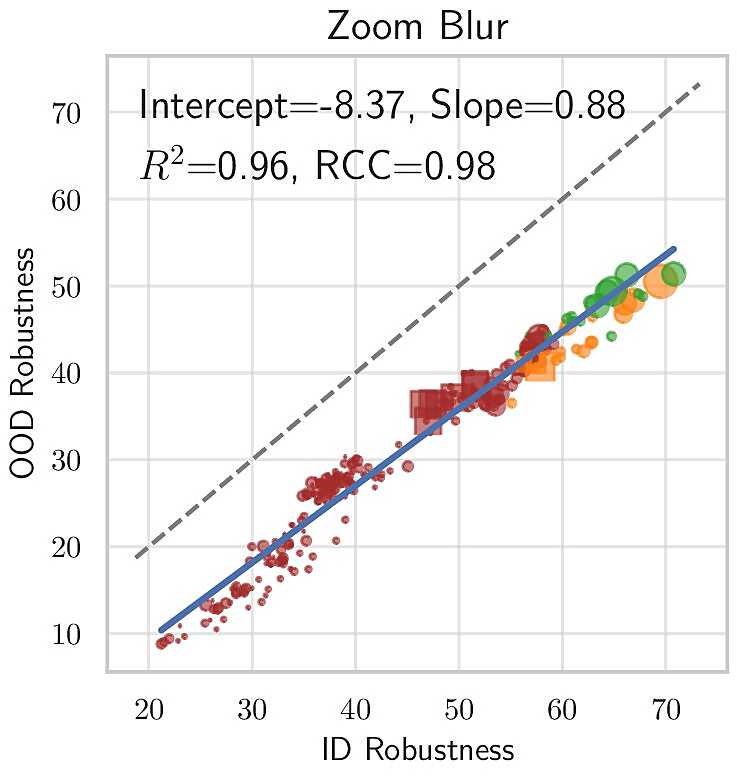}
    \end{subfigure}

    \begin{subfigure}{.245\linewidth}
        \includegraphics[width=\linewidth]{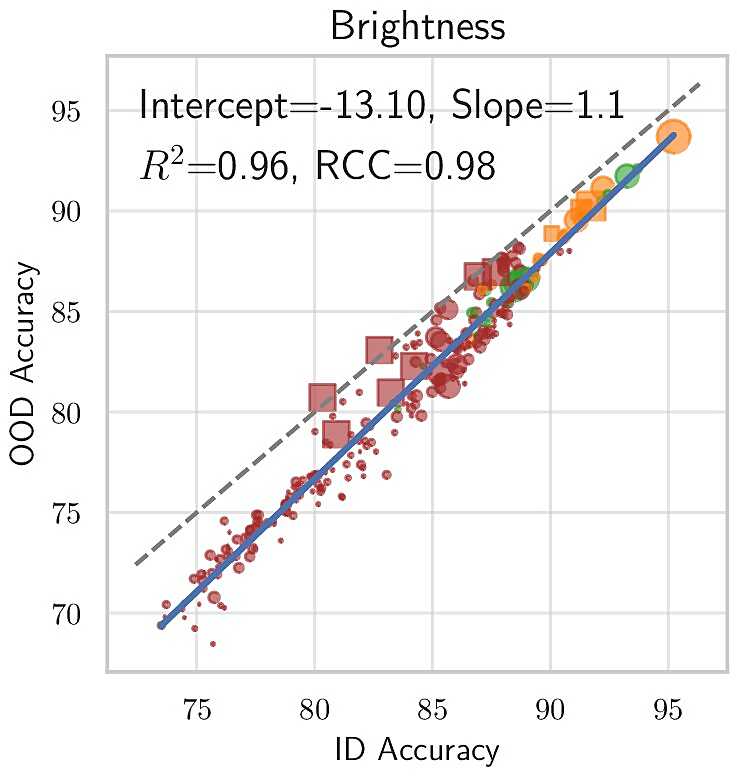}
    \end{subfigure}
    \begin{subfigure}{.245\linewidth}
        \includegraphics[width=\linewidth]{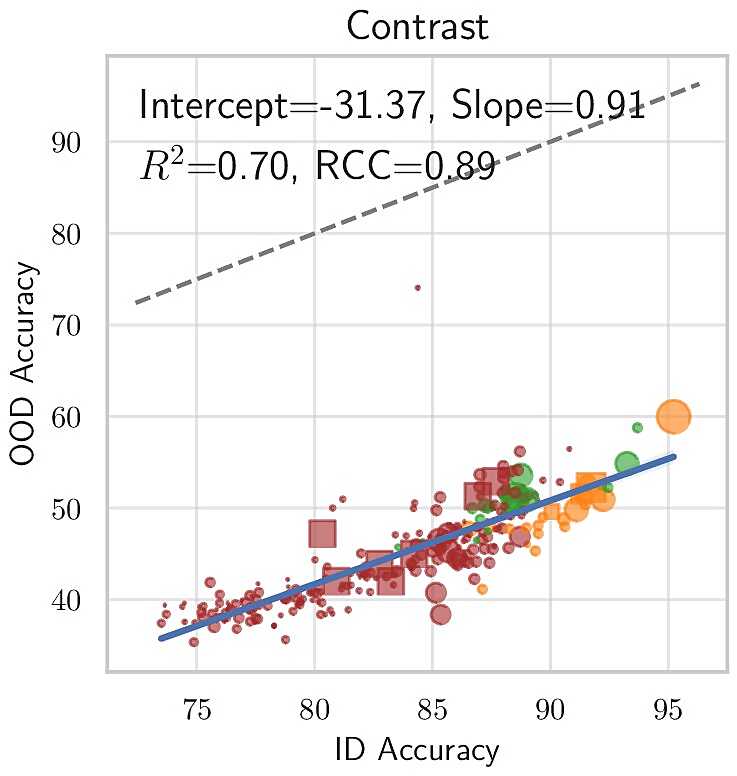}
    \end{subfigure}
    \begin{subfigure}{.245\linewidth}
        \includegraphics[width=\linewidth]{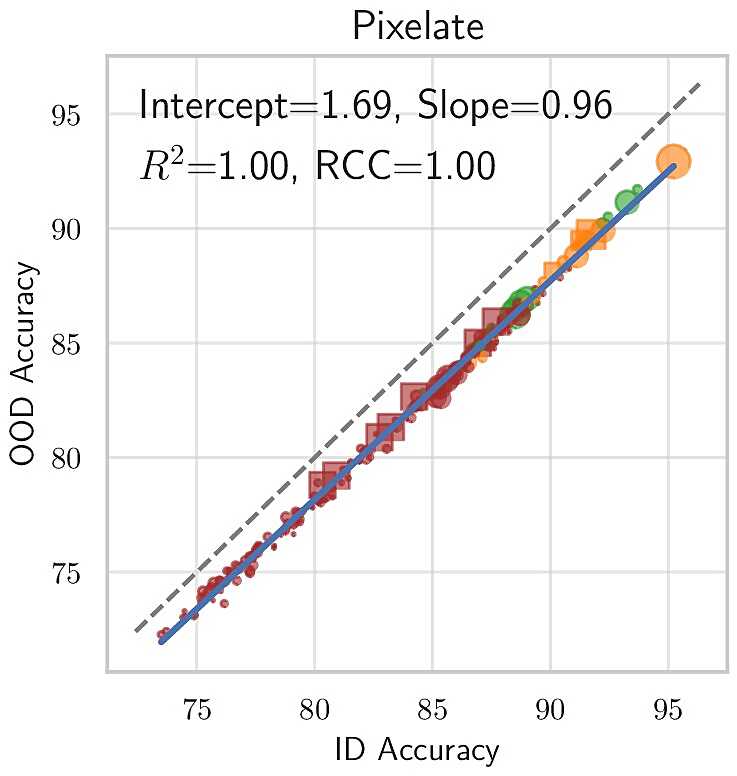}
    \end{subfigure}
    \begin{subfigure}{.245\linewidth}
        \includegraphics[width=\linewidth]{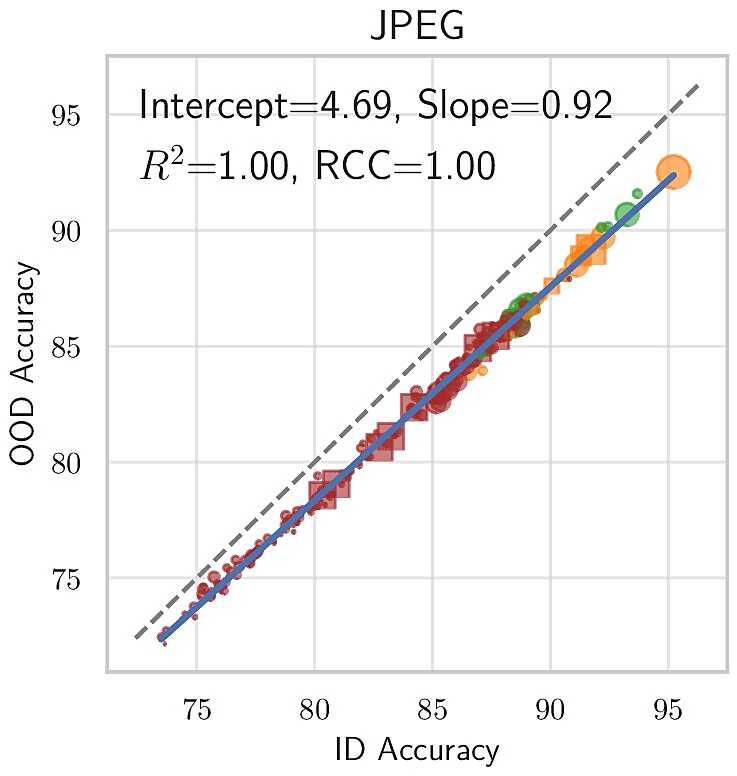}
    \end{subfigure}

    \begin{subfigure}{.245\linewidth}
        \includegraphics[width=\linewidth]{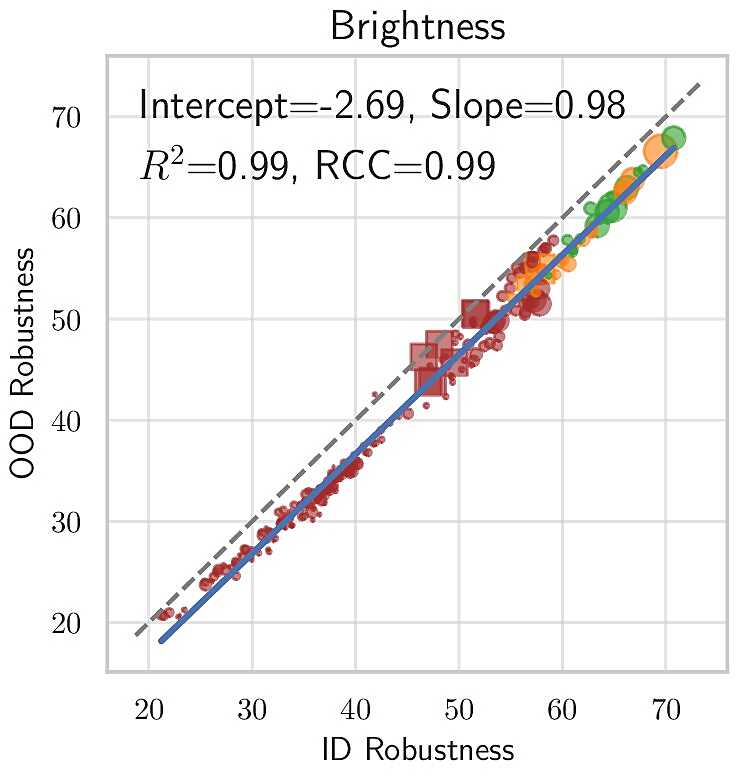}
    \end{subfigure}
    \begin{subfigure}{.245\linewidth}
        \includegraphics[width=\linewidth]{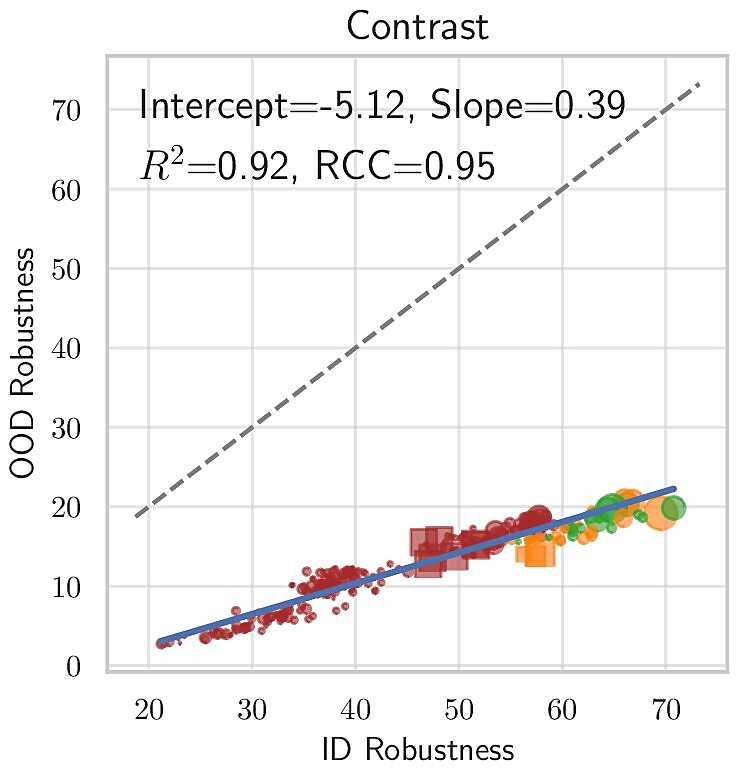}
    \end{subfigure}
    \begin{subfigure}{.245\linewidth}
        \includegraphics[width=\linewidth]{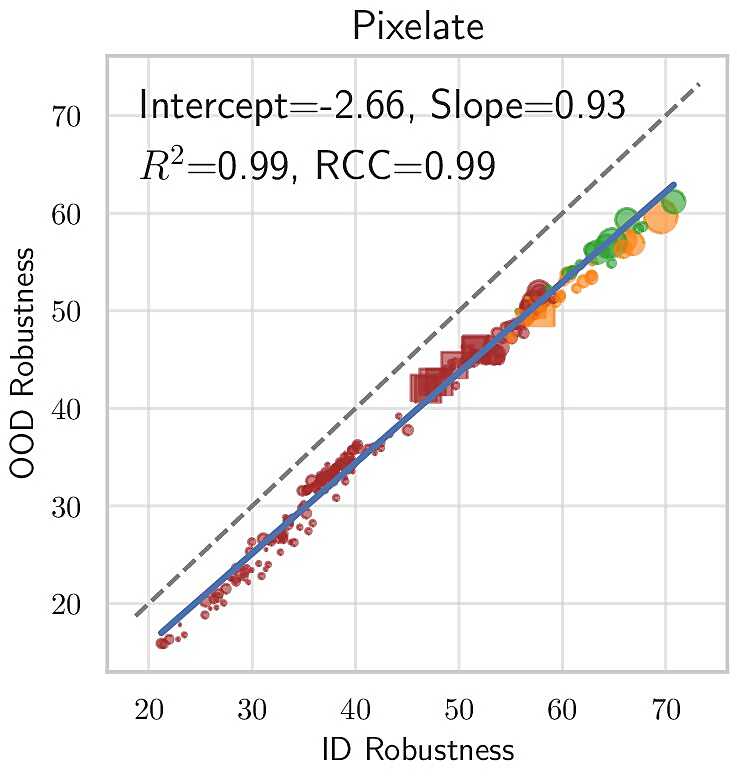}
    \end{subfigure}
    \begin{subfigure}{.245\linewidth}
        \includegraphics[width=\linewidth]{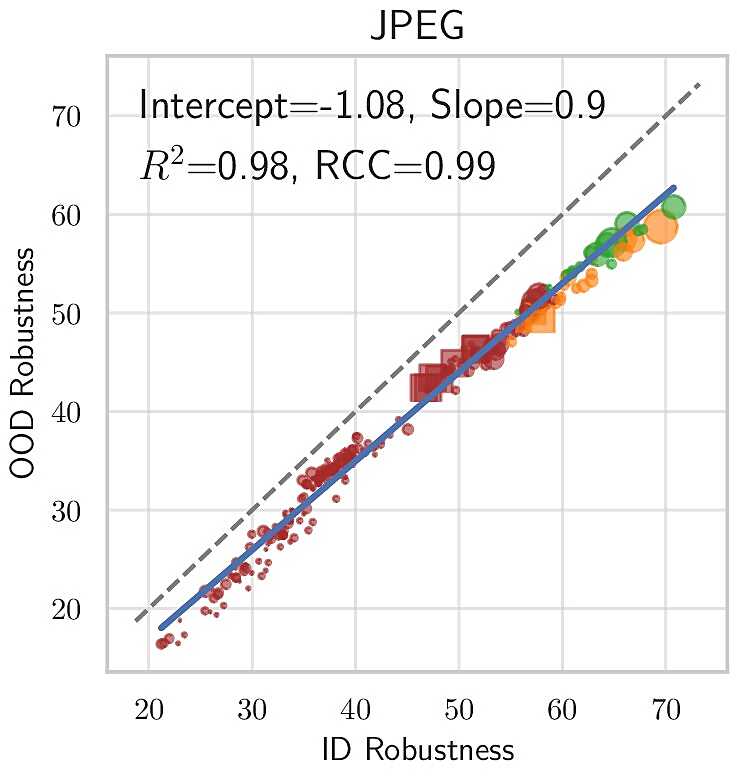}
    \end{subfigure}
    
    \caption{\textbf{Correlation between ID accuracy and OOD accuracy (odd rows); ID robustness and OOD robustness (even rows) for CIFAR10 \linf AT models}.}
    
\end{figure*}

\begin{figure}[!h]
    \centering
    \begin{subfigure}{\linewidth}
        \includegraphics[width=\linewidth, trim=0 25 0 25, clip]{images/legend.jpg}
    \end{subfigure}

    \begin{subfigure}{.245\linewidth}
        \includegraphics[width=\linewidth]{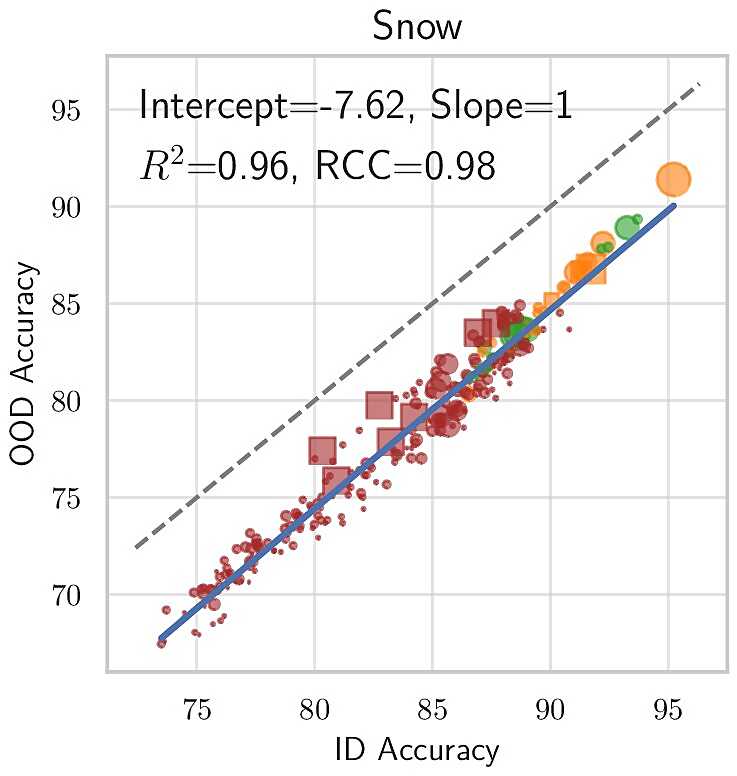}
    \end{subfigure}
    \begin{subfigure}{.245\linewidth}
        \includegraphics[width=\linewidth]{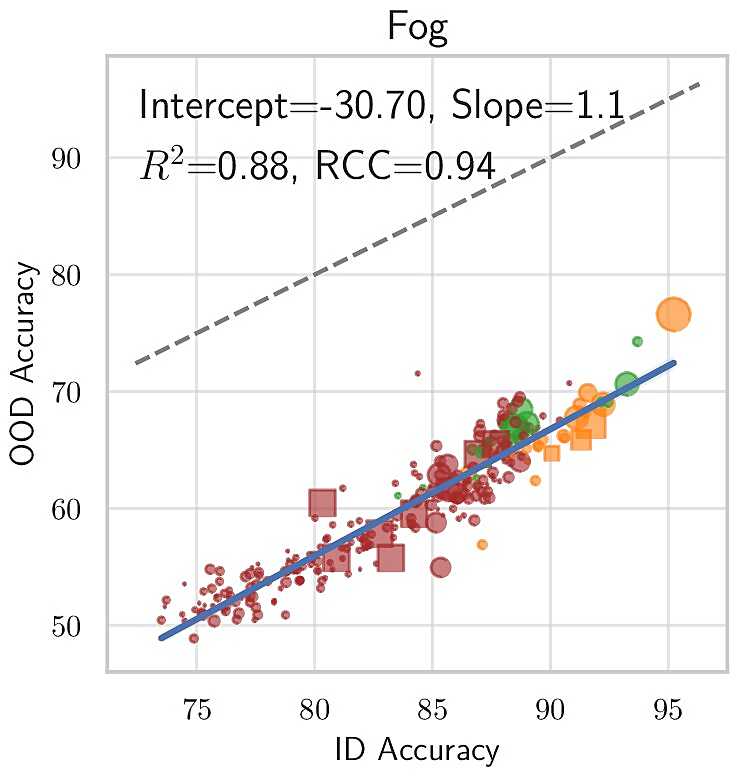}
    \end{subfigure}
    \begin{subfigure}{.245\linewidth}
        \includegraphics[width=\linewidth]{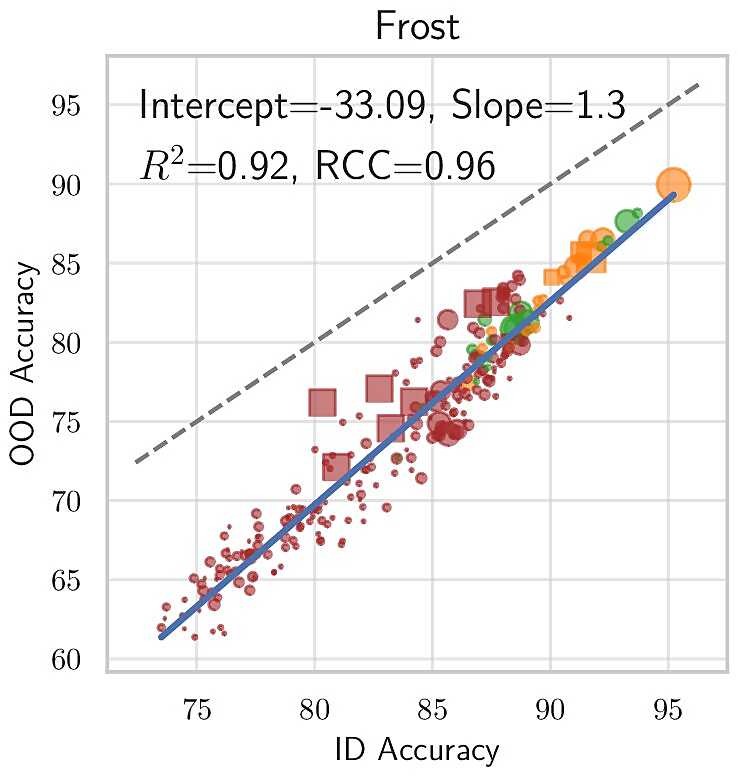}
    \end{subfigure}

    \begin{subfigure}{.245\linewidth}
        \includegraphics[width=\linewidth]{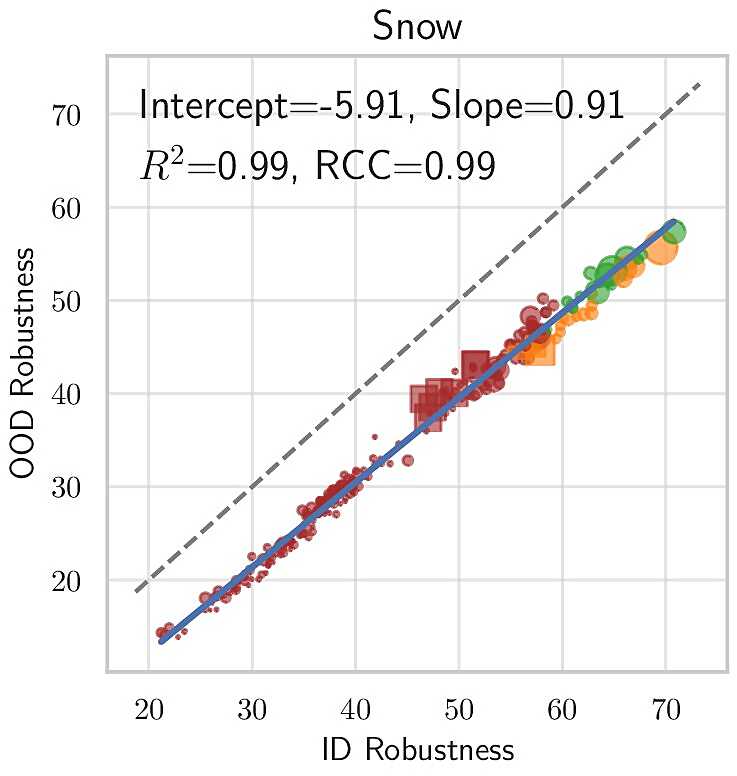}
    \end{subfigure}
    \begin{subfigure}{.245\linewidth}
        \includegraphics[width=\linewidth]{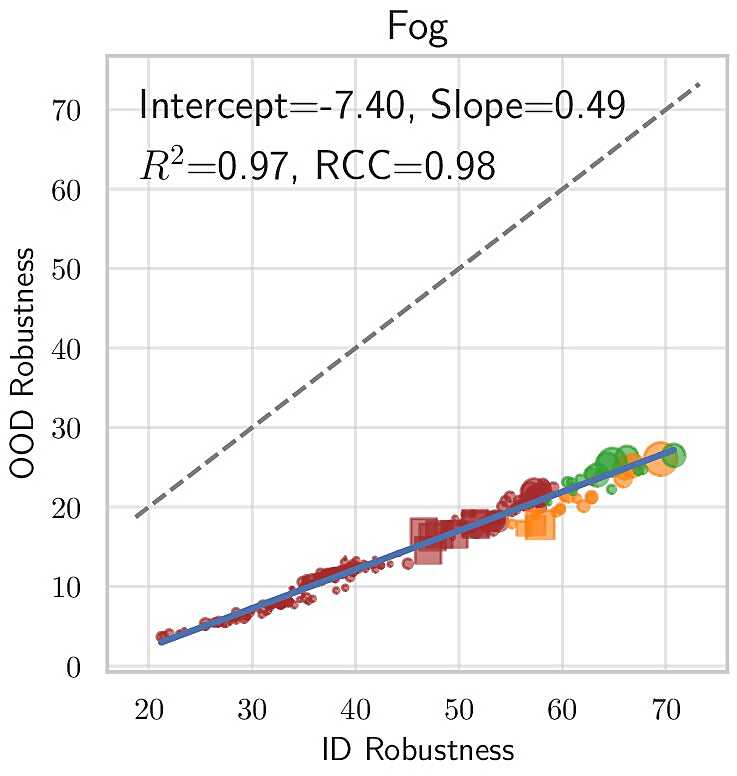}
    \end{subfigure}
    \begin{subfigure}{.245\linewidth}
        \includegraphics[width=\linewidth]{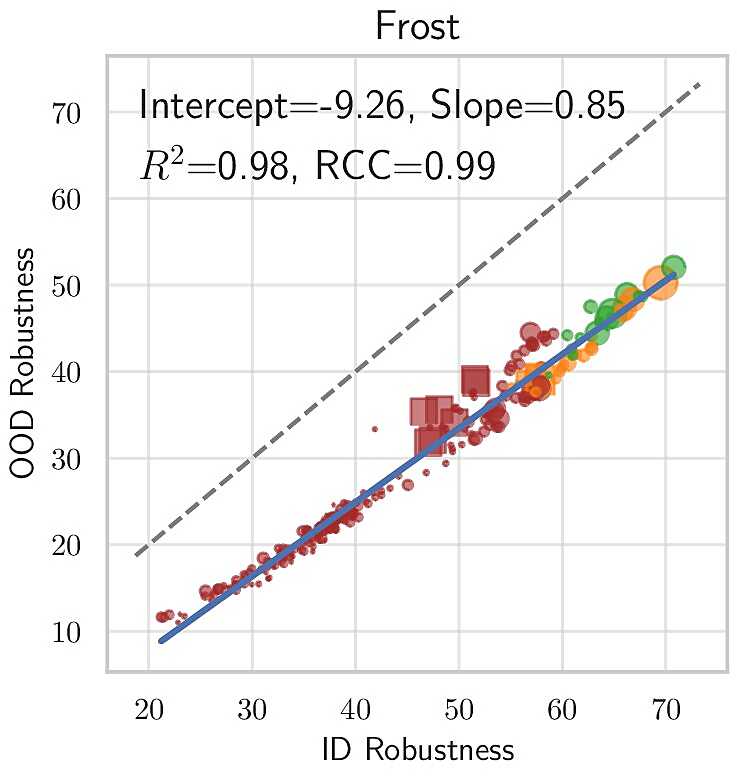}
    \end{subfigure}

    \begin{subfigure}{.245\linewidth}
        \includegraphics[width=\linewidth]{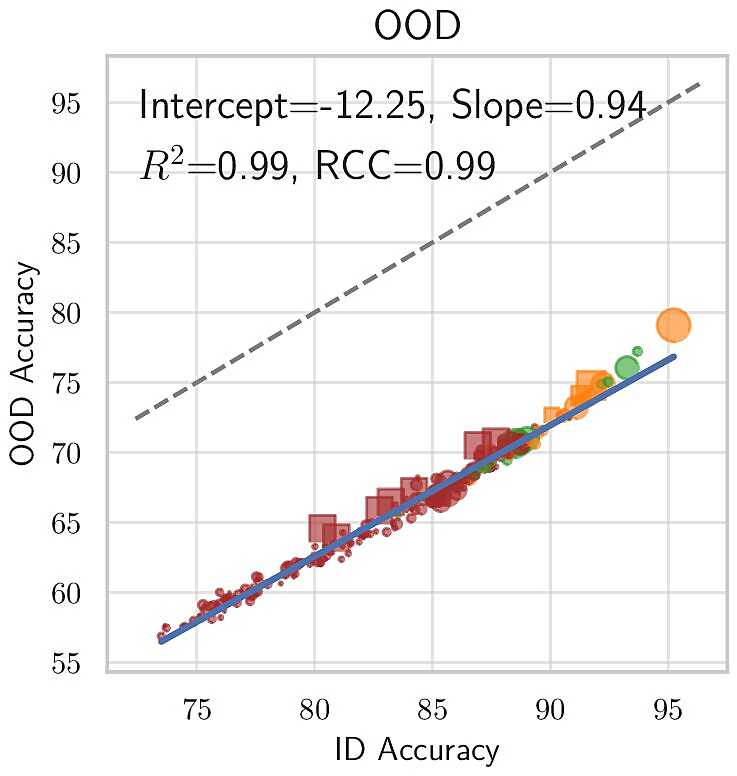}
    \end{subfigure}
    \begin{subfigure}{.245\linewidth}
        \includegraphics[width=\linewidth]{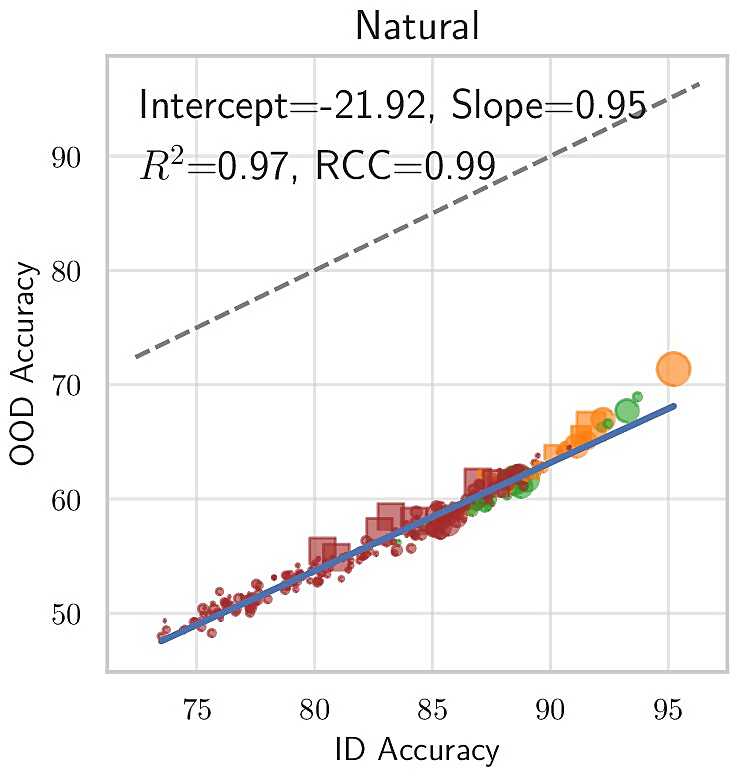}
    \end{subfigure}
    \begin{subfigure}{.245\linewidth}
        \includegraphics[width=\linewidth]{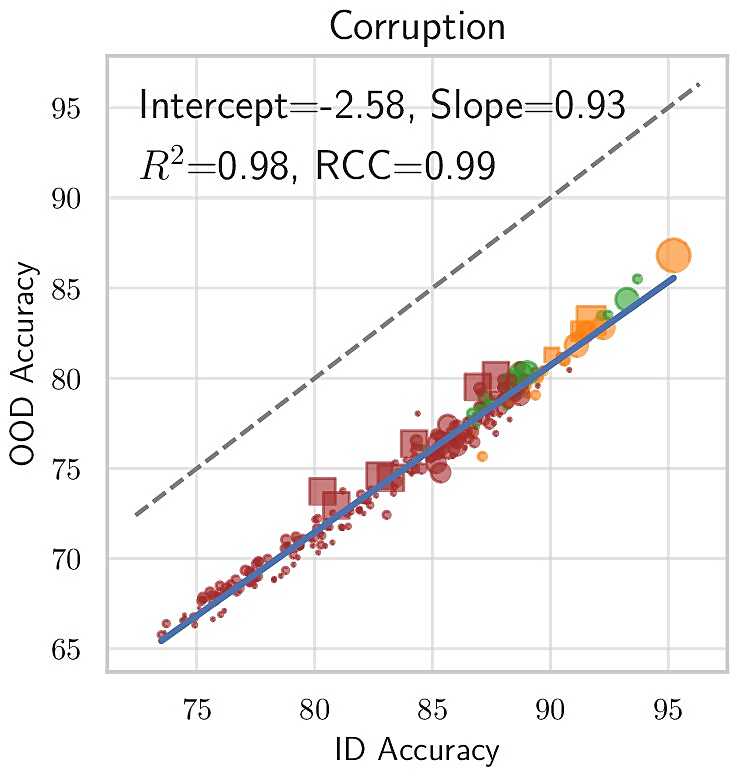}
    \end{subfigure}

    \begin{subfigure}{.245\linewidth}
        \includegraphics[width=\linewidth]{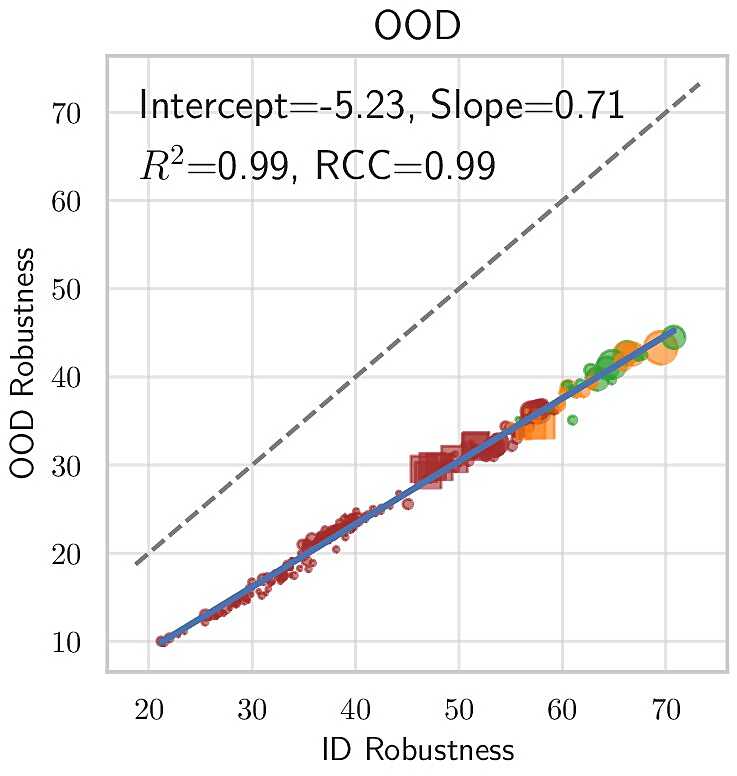}
    \end{subfigure}
    \begin{subfigure}{.245\linewidth}
        \includegraphics[width=\linewidth]{images/CIFAR10-Linf/ID-seen-OOD-seen/N.jpg}
    \end{subfigure}
    \begin{subfigure}{.245\linewidth}
        \includegraphics[width=\linewidth]{images/CIFAR10-Linf/ID-seen-OOD-seen/C.jpg}
    \end{subfigure}

    \caption{\textbf{Correlation between ID accuracy and OOD accuracy (odd rows); ID robustness and OOD robustness (even rows) for CIFAR10 \linf AT models}.}
\end{figure}


\begin{figure}[!h]
    \centering
    \begin{subfigure}{\linewidth}
        \includegraphics[width=\linewidth, trim=0 25 0 25, clip]{images/legend.jpg}
    \end{subfigure}
    \begin{subfigure}{.245\linewidth}
        \includegraphics[width=\linewidth]{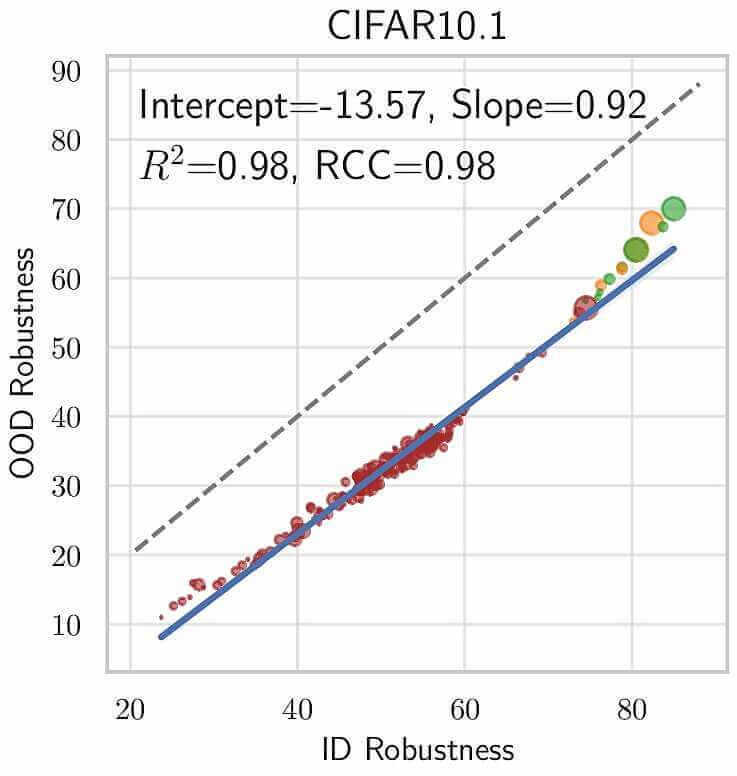}
    \end{subfigure}
    \begin{subfigure}{.245\linewidth}
        \includegraphics[width=\linewidth]{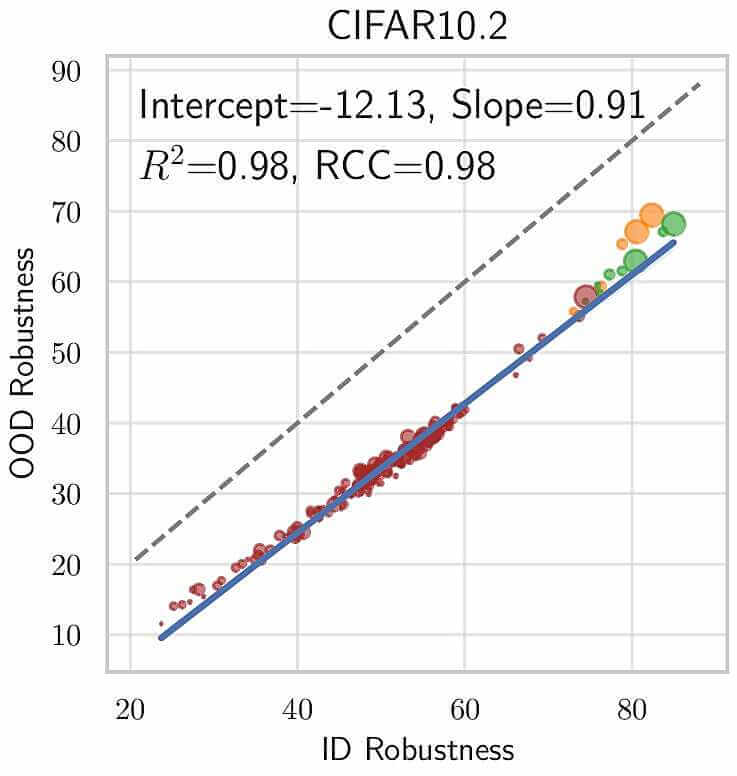}
    \end{subfigure}
    \begin{subfigure}{.245\linewidth}
        \includegraphics[width=\linewidth]{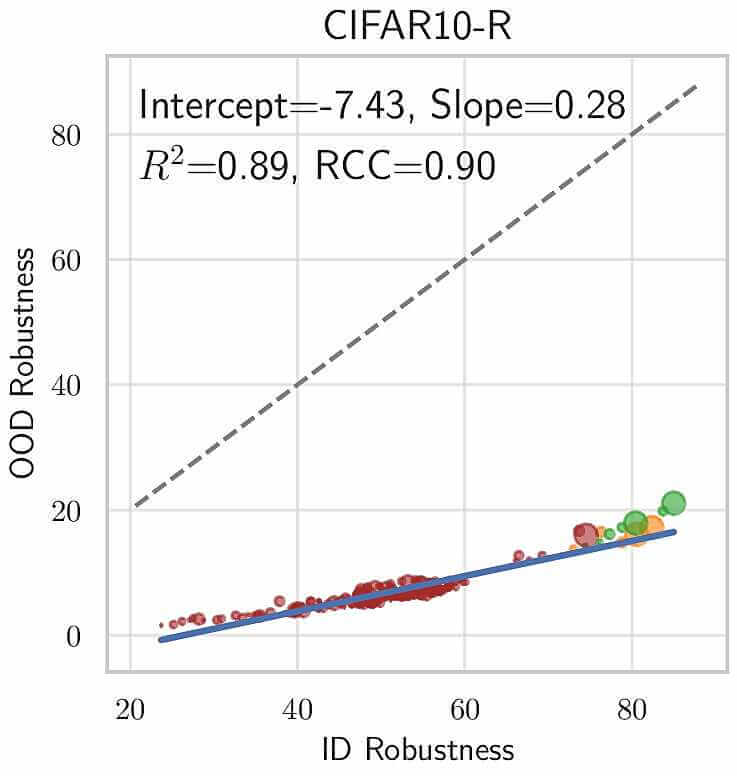}
    \end{subfigure}
    \begin{subfigure}{.245\linewidth}
        \includegraphics[width=\linewidth]{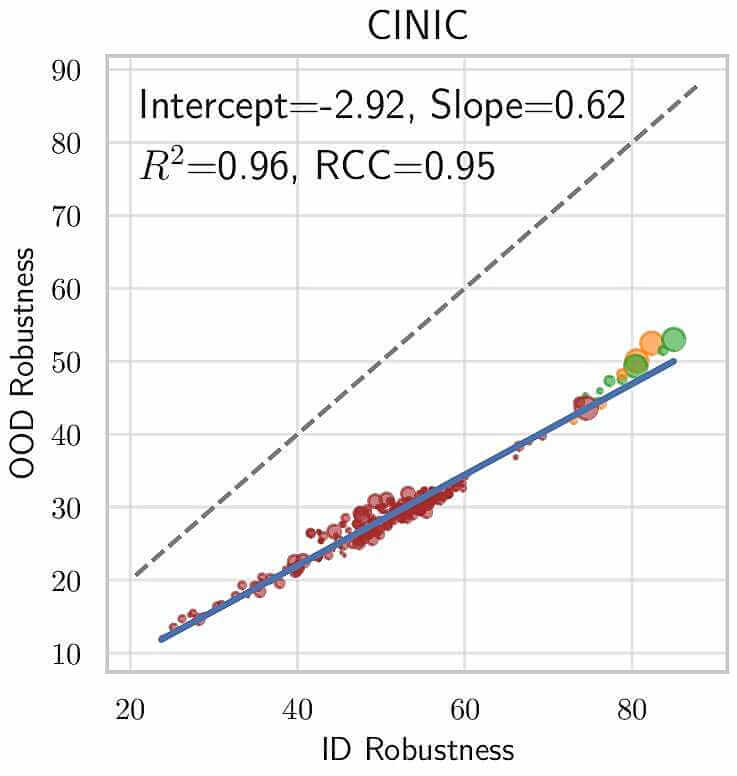}
    \end{subfigure}

    \begin{subfigure}{.245\linewidth}
        \includegraphics[width=\linewidth]{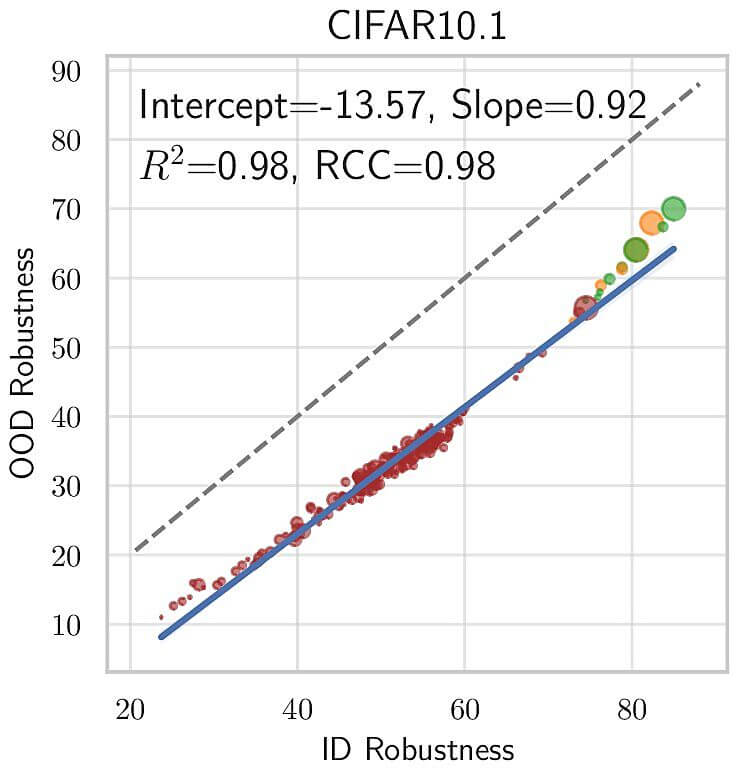}
    \end{subfigure}
    \begin{subfigure}{.245\linewidth}
        \includegraphics[width=\linewidth]{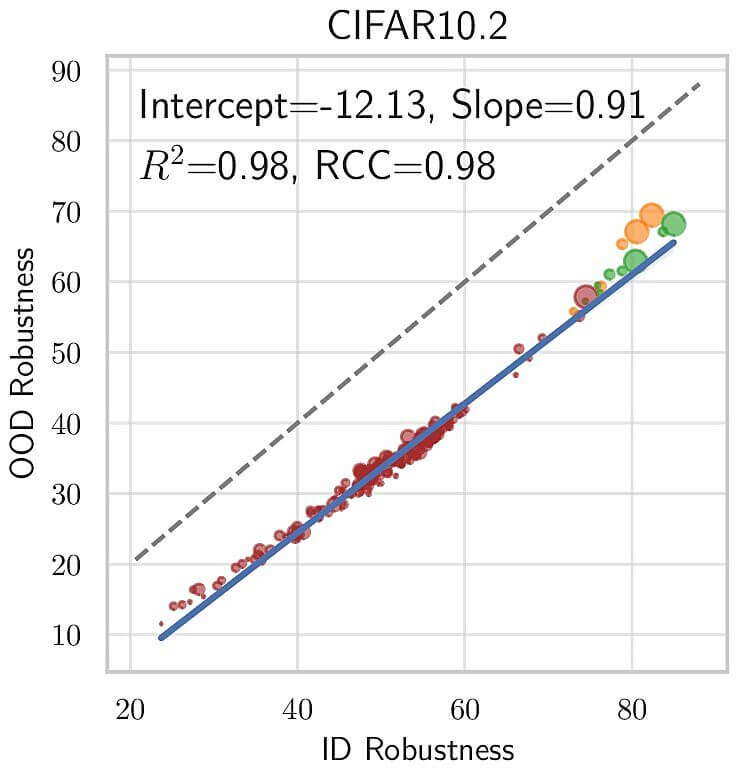}
    \end{subfigure}
    \begin{subfigure}{.245\linewidth}
        \includegraphics[width=\linewidth]{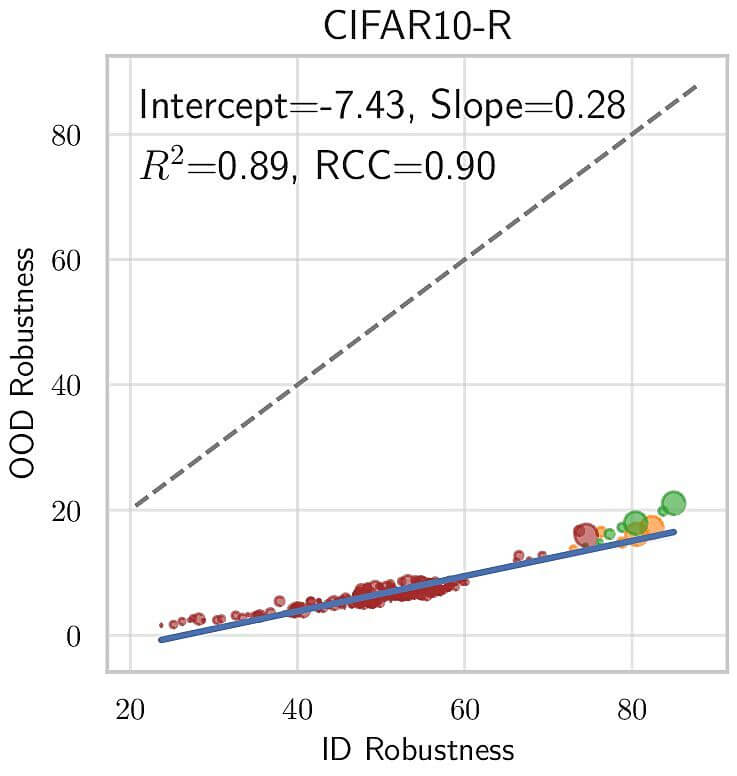}
    \end{subfigure}
    \begin{subfigure}{.245\linewidth}
        \includegraphics[width=\linewidth]{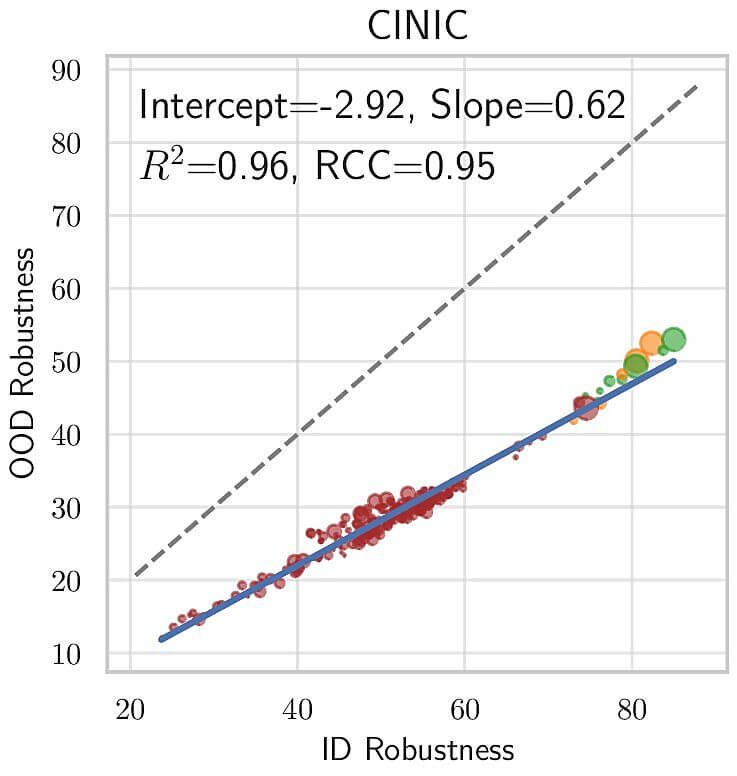}
    \end{subfigure}

    \begin{subfigure}{.245\linewidth}
        \includegraphics[width=\linewidth]{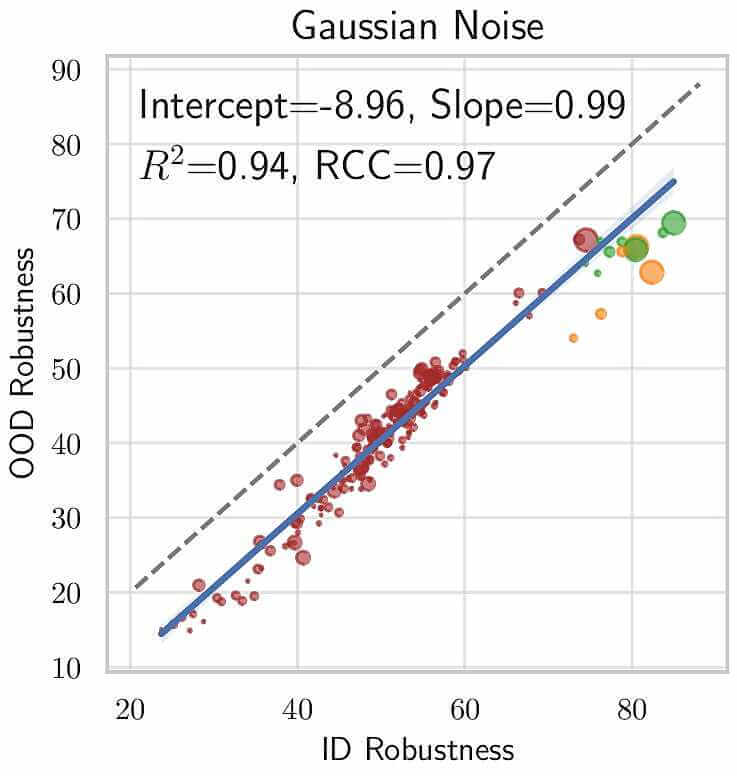}
    \end{subfigure}
    \begin{subfigure}{.245\linewidth}
        \includegraphics[width=\linewidth]{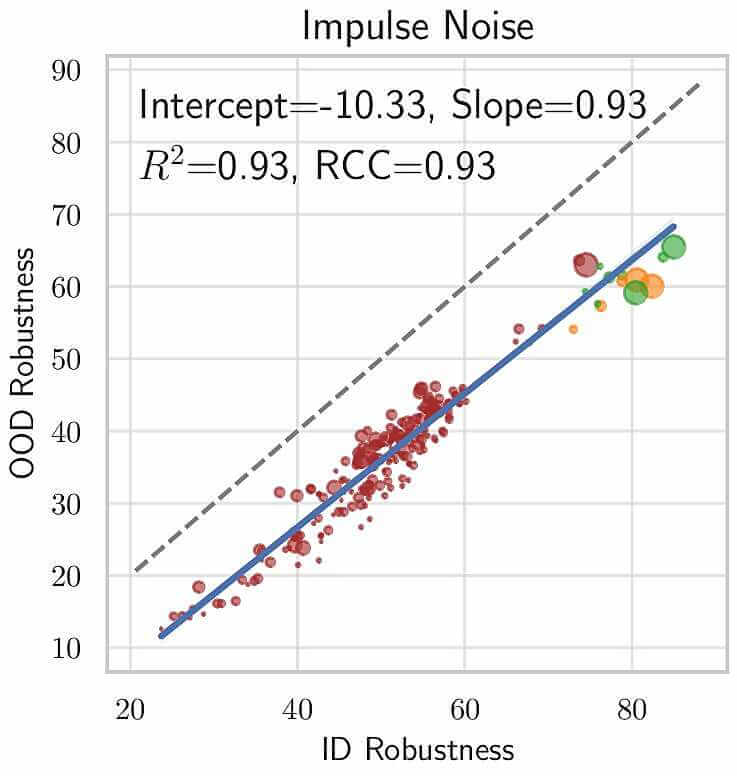}
    \end{subfigure}
    \begin{subfigure}{.245\linewidth}
        \includegraphics[width=\linewidth]{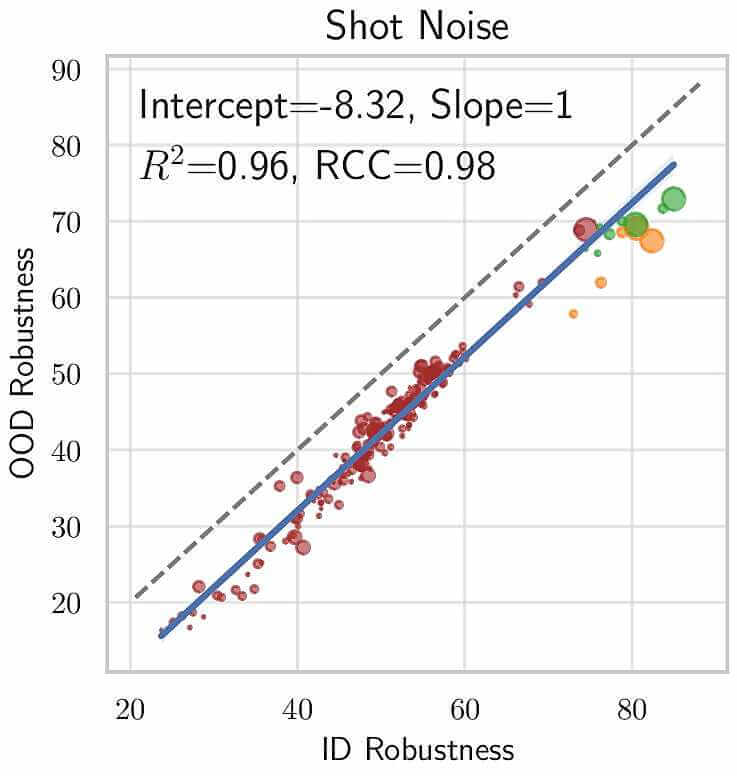}
    \end{subfigure}
    \begin{subfigure}{.245\linewidth}
        \includegraphics[width=\linewidth]{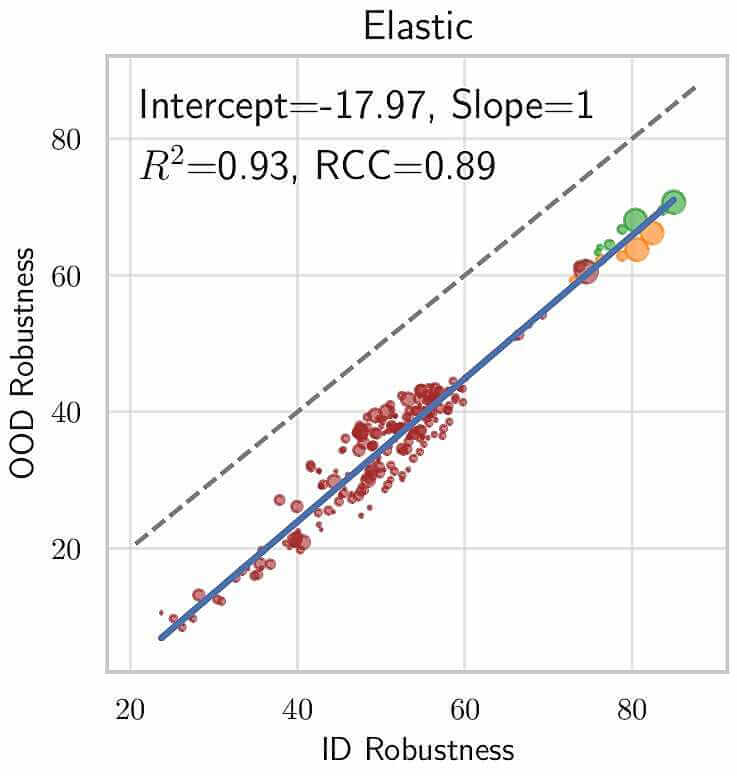}
    \end{subfigure}

    \begin{subfigure}{.245\linewidth}
        \includegraphics[width=\linewidth]{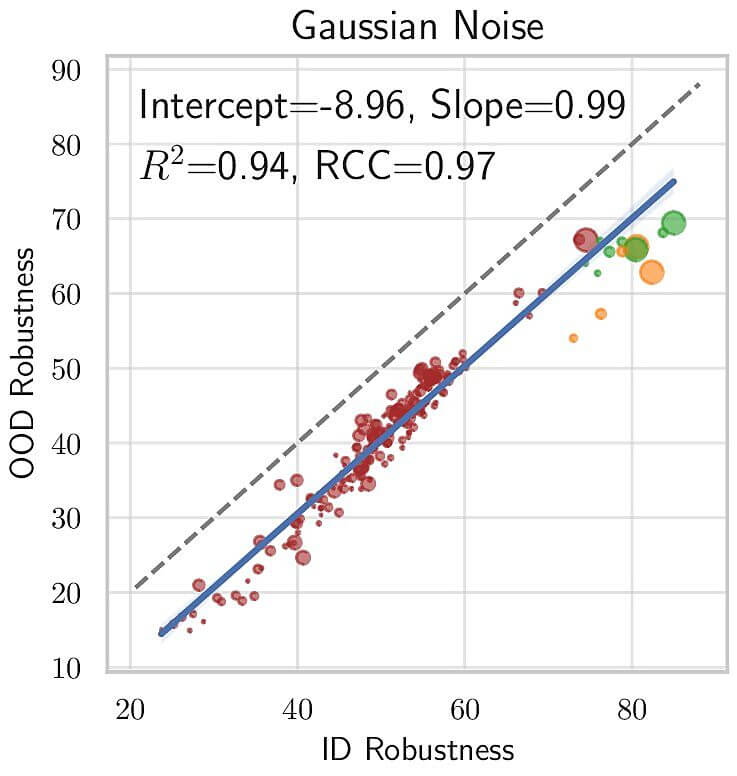}
    \end{subfigure}
    \begin{subfigure}{.245\linewidth}
        \includegraphics[width=\linewidth]{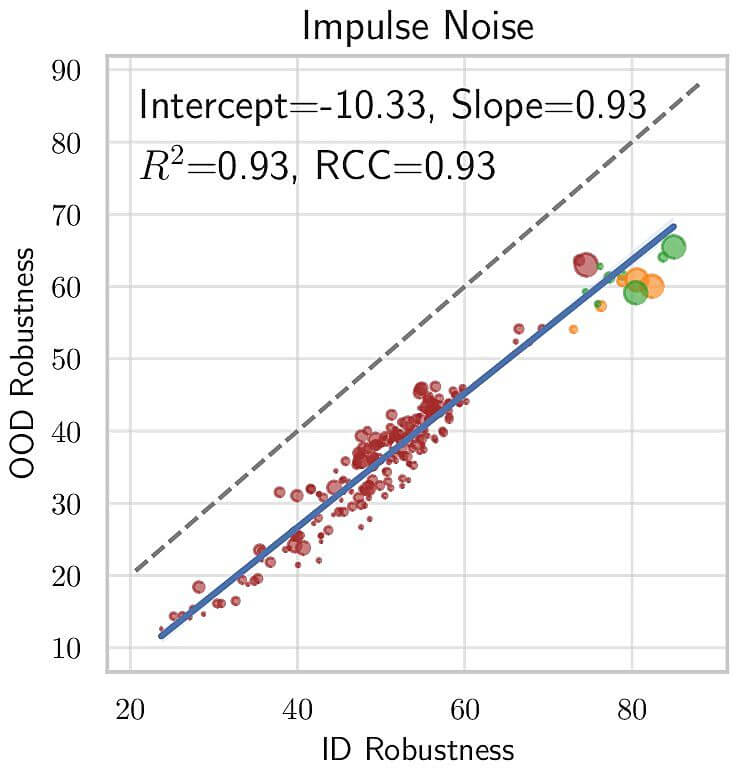}
    \end{subfigure}
    \begin{subfigure}{.245\linewidth}
        \includegraphics[width=\linewidth]{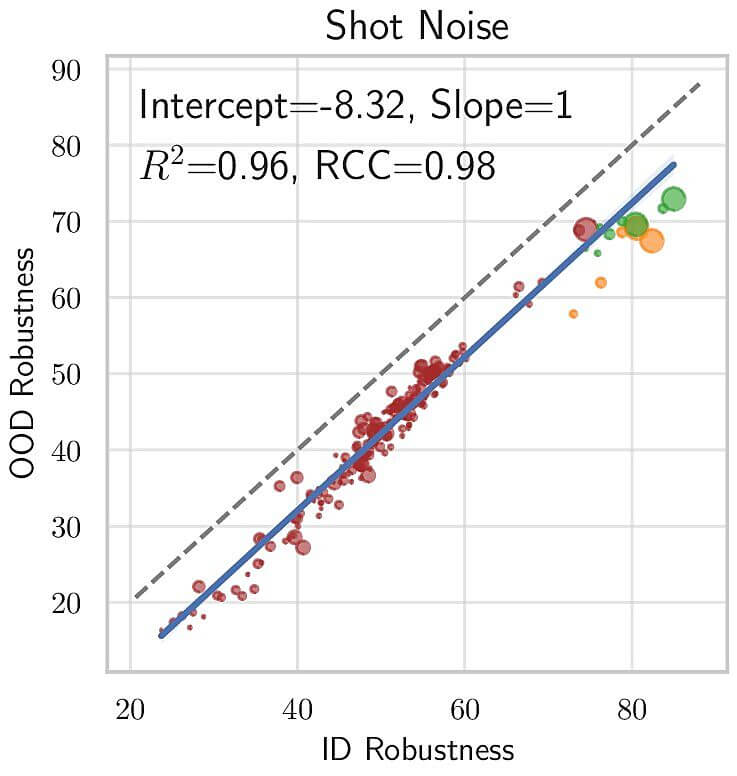}
    \end{subfigure}
    \begin{subfigure}{.245\linewidth}
        \includegraphics[width=\linewidth]{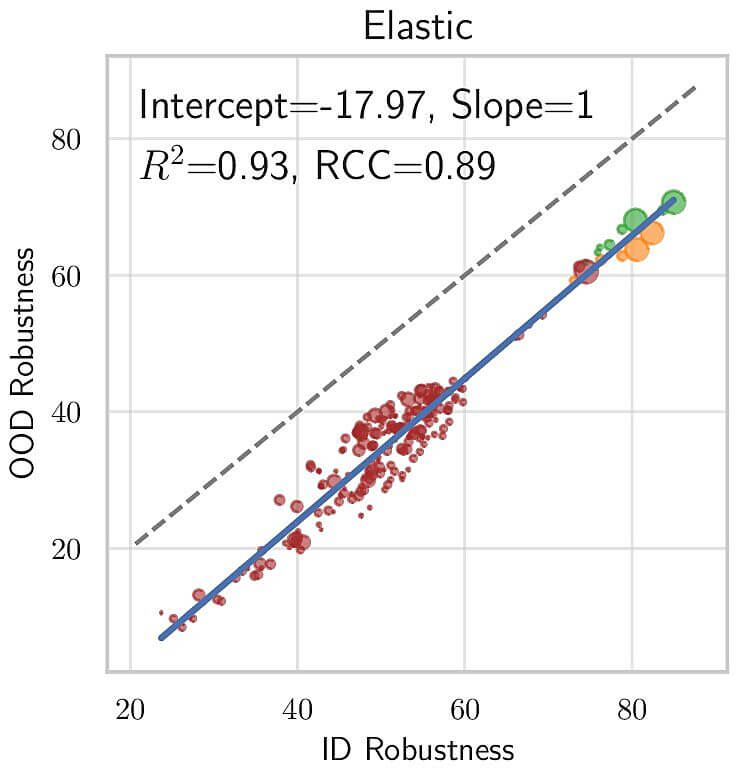}
    \end{subfigure}

    \caption{\textbf{Correlation between ID accuracy and OOD accuracy (odd rows); ID robustness and OOD robustness (even rows) for CIFAR10 \lt AT models}.}

\end{figure}

\begin{figure}[!h]
    \centering
    \begin{subfigure}{\linewidth}
        \includegraphics[width=\linewidth, trim=0 25 0 25, clip]{images/legend.jpg}
    \end{subfigure}

    \begin{subfigure}{.245\linewidth}
        \includegraphics[width=\linewidth]{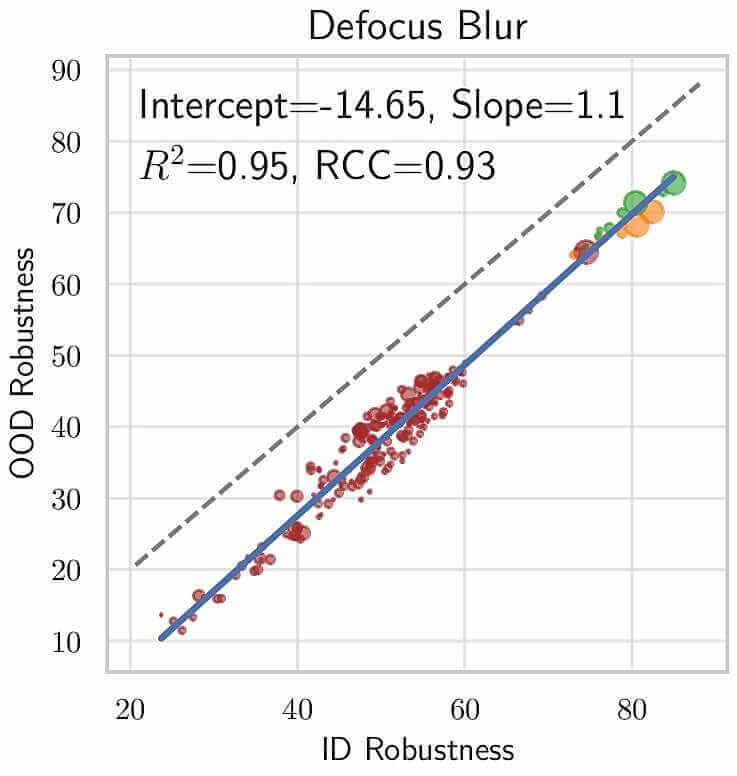}
    \end{subfigure}
    \begin{subfigure}{.245\linewidth}
        \includegraphics[width=\linewidth]{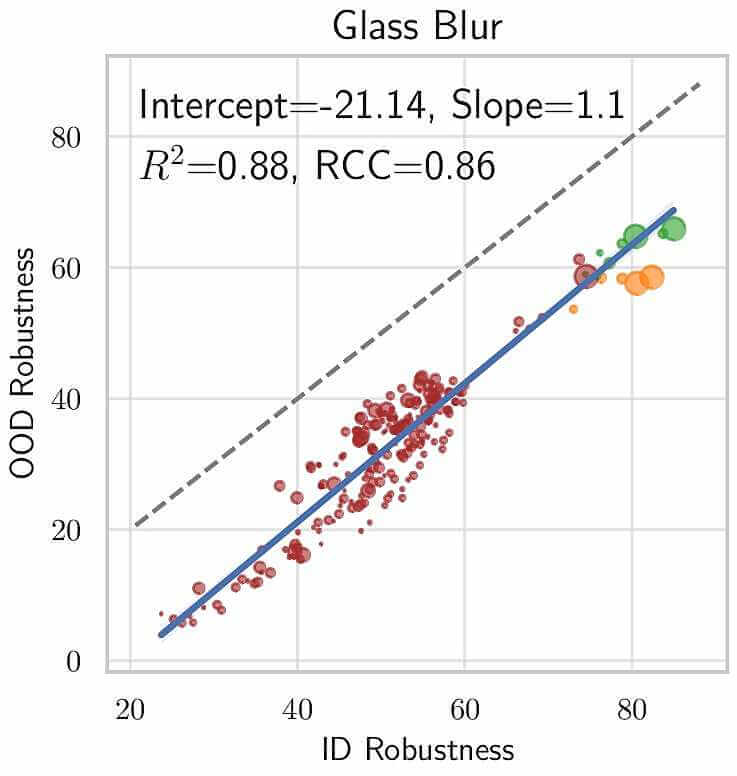}
    \end{subfigure}
    \begin{subfigure}{.245\linewidth}
        \includegraphics[width=\linewidth]{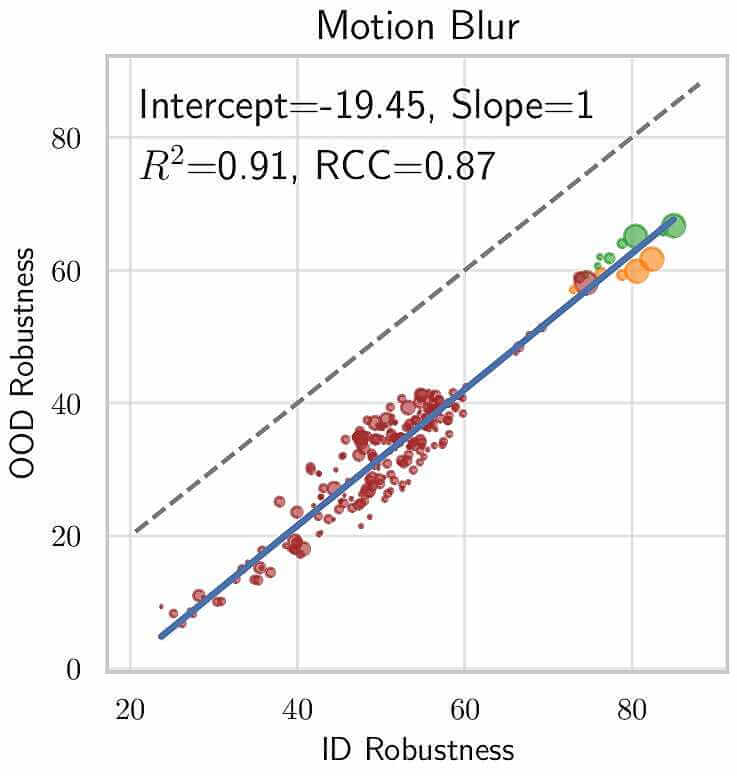}
    \end{subfigure}
    \begin{subfigure}{.245\linewidth}
        \includegraphics[width=\linewidth]{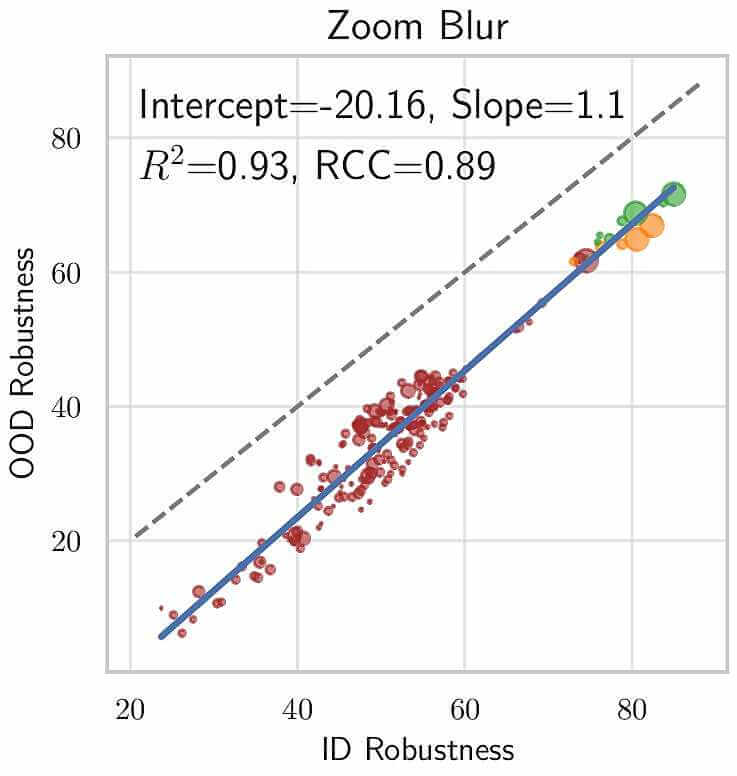}
    \end{subfigure}

    \begin{subfigure}{.245\linewidth}
        \includegraphics[width=\linewidth]{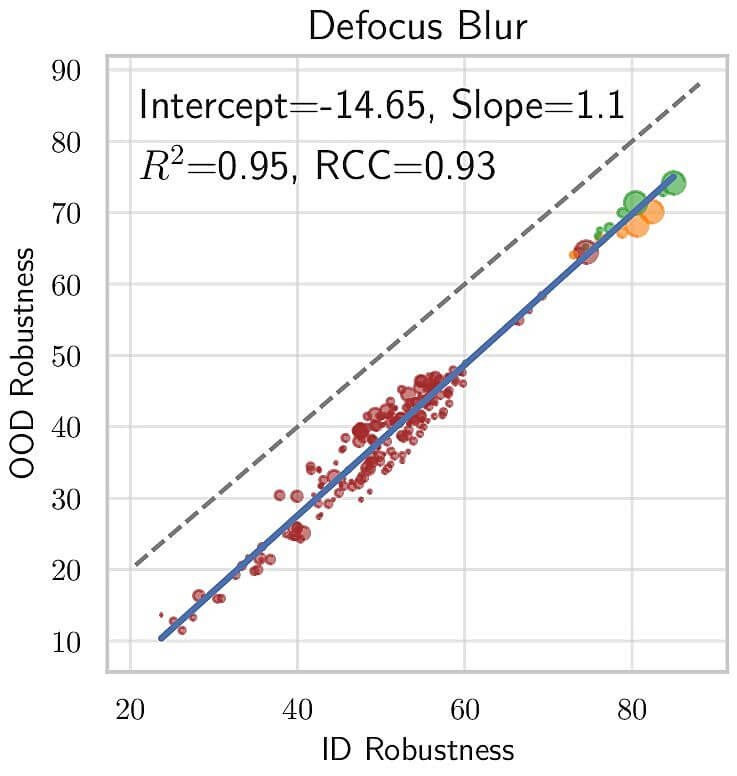}
    \end{subfigure}
    \begin{subfigure}{.245\linewidth}
        \includegraphics[width=\linewidth]{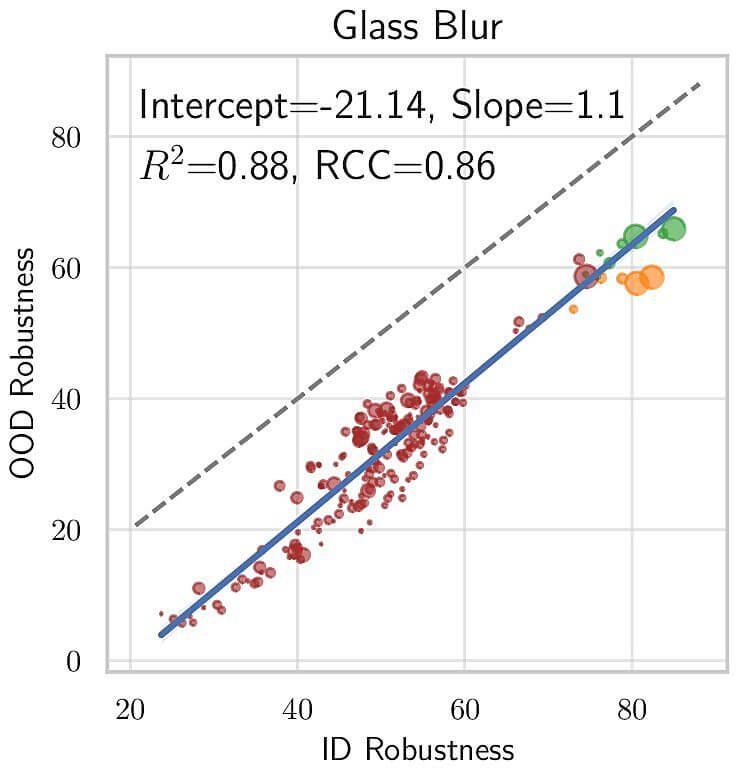}
    \end{subfigure}
    \begin{subfigure}{.245\linewidth}
        \includegraphics[width=\linewidth]{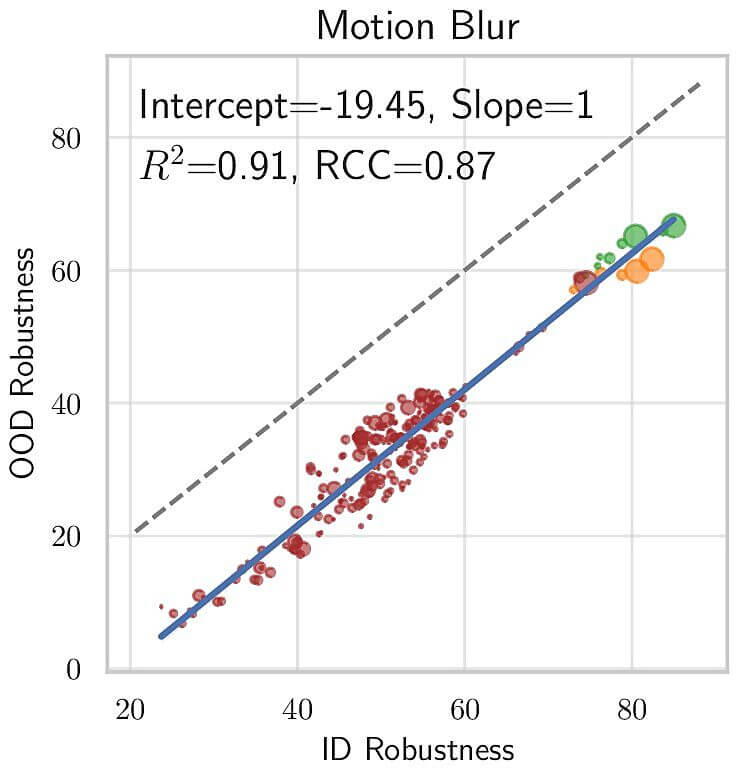}
    \end{subfigure}
    \begin{subfigure}{.245\linewidth}
        \includegraphics[width=\linewidth]{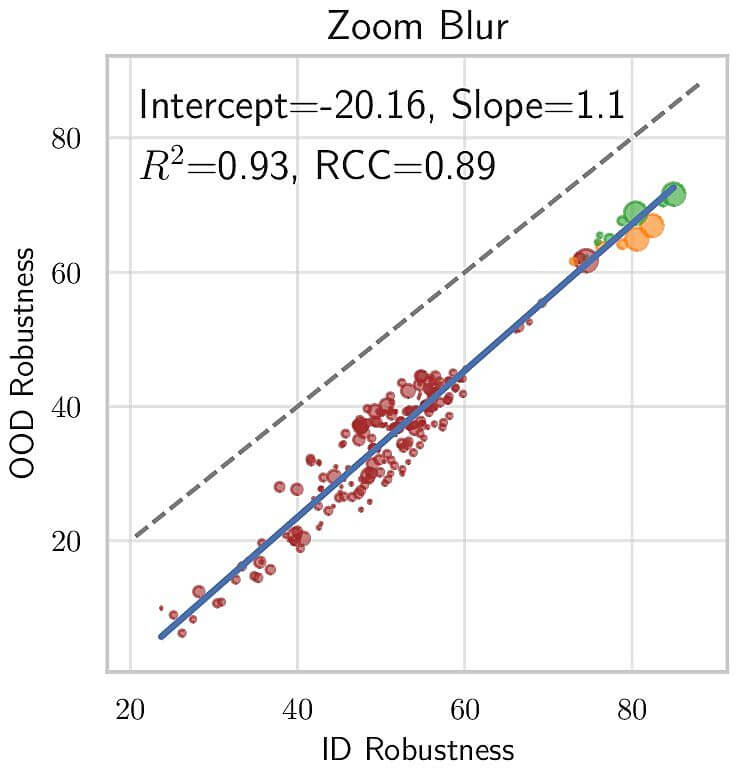}
    \end{subfigure}

    \begin{subfigure}{.245\linewidth}
        \includegraphics[width=\linewidth]{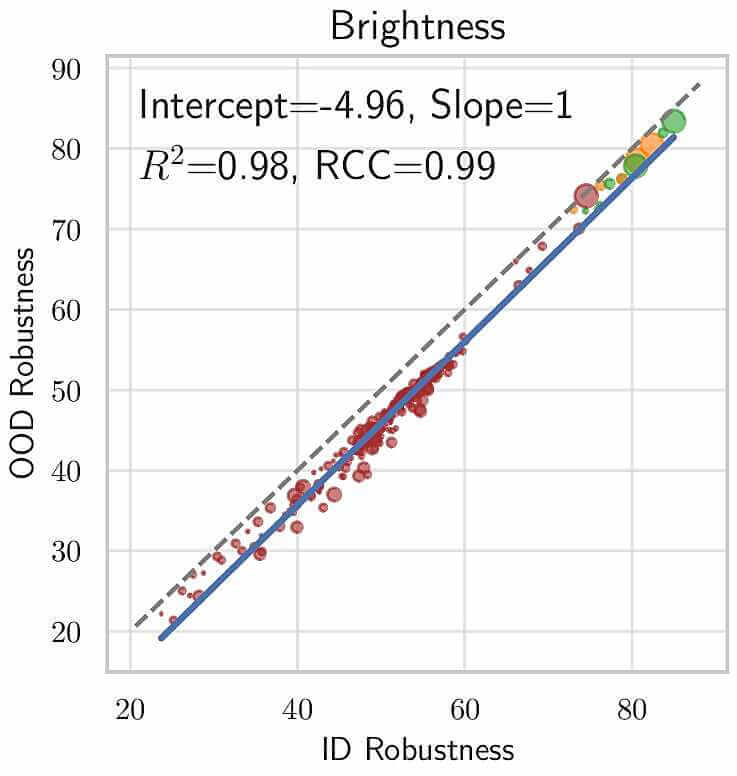}
    \end{subfigure}
    \begin{subfigure}{.245\linewidth}
        \includegraphics[width=\linewidth]{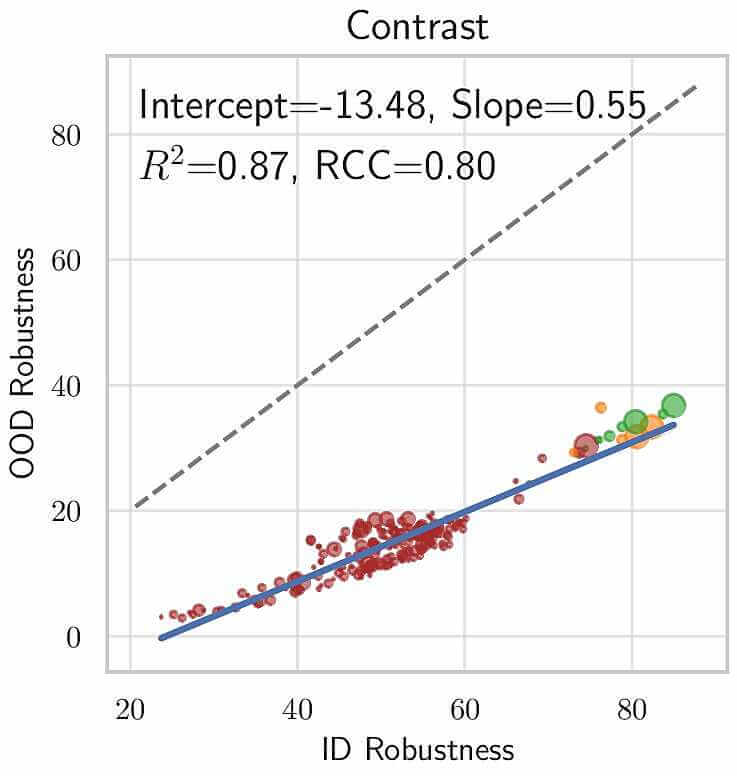}
    \end{subfigure}
    \begin{subfigure}{.245\linewidth}
        \includegraphics[width=\linewidth]{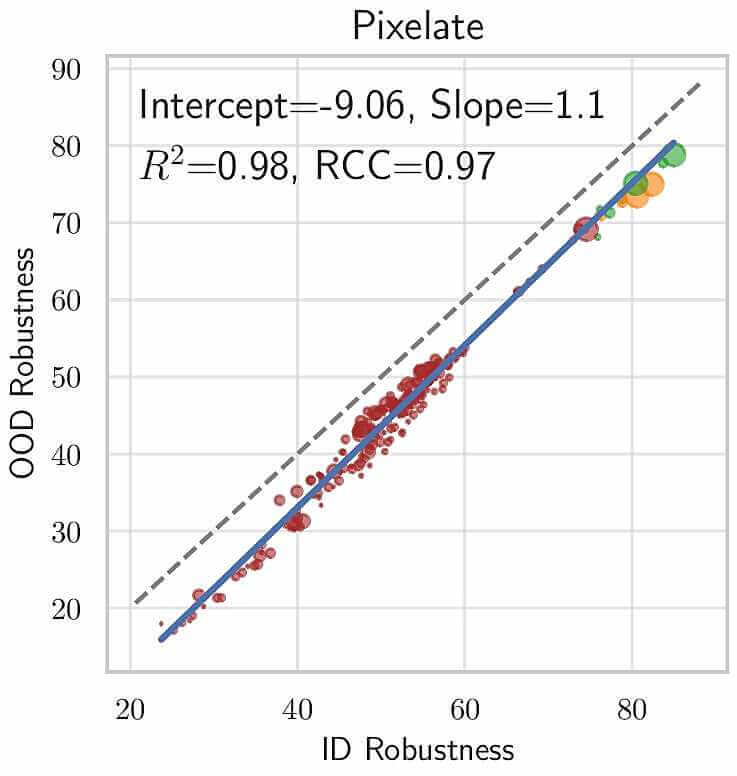}
    \end{subfigure}
    \begin{subfigure}{.245\linewidth}
        \includegraphics[width=\linewidth]{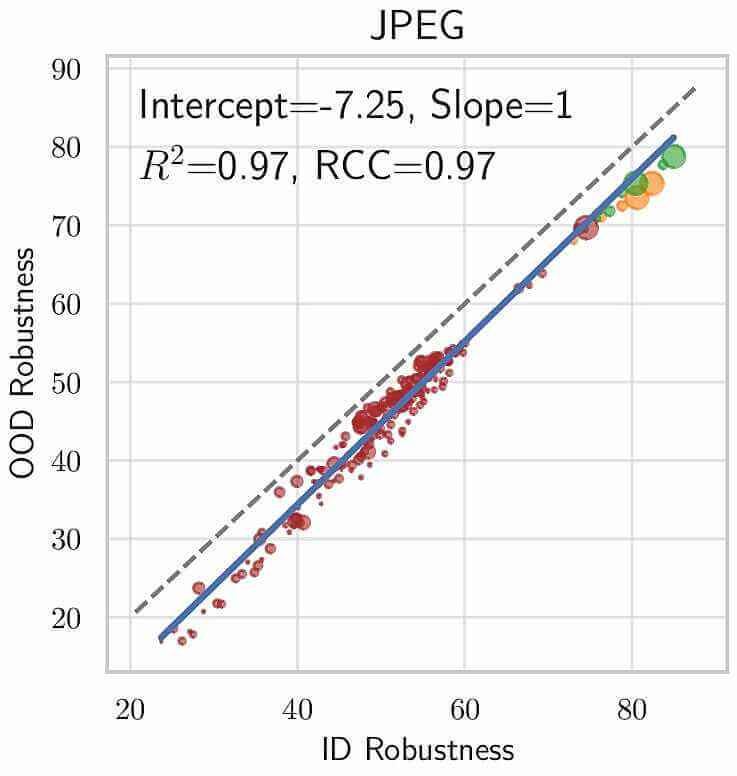}
    \end{subfigure}

    \begin{subfigure}{.245\linewidth}
        \includegraphics[width=\linewidth]{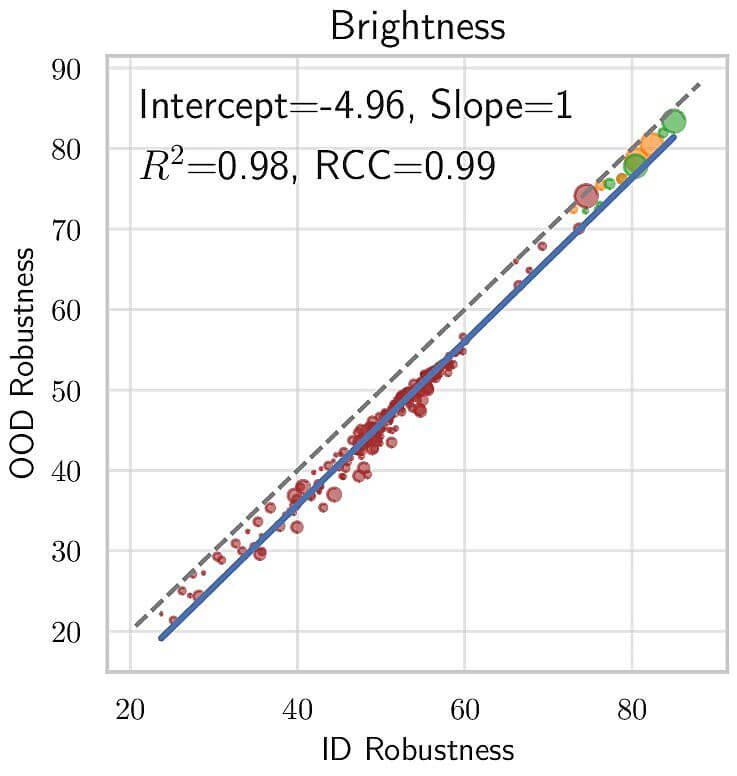}
    \end{subfigure}
    \begin{subfigure}{.245\linewidth}
        \includegraphics[width=\linewidth]{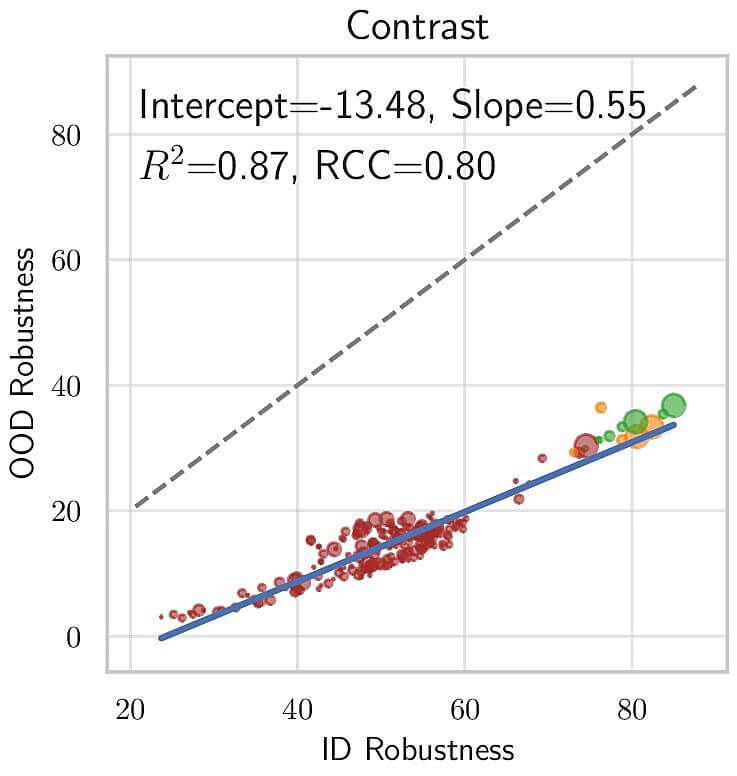}
    \end{subfigure}
    \begin{subfigure}{.245\linewidth}
        \includegraphics[width=\linewidth]{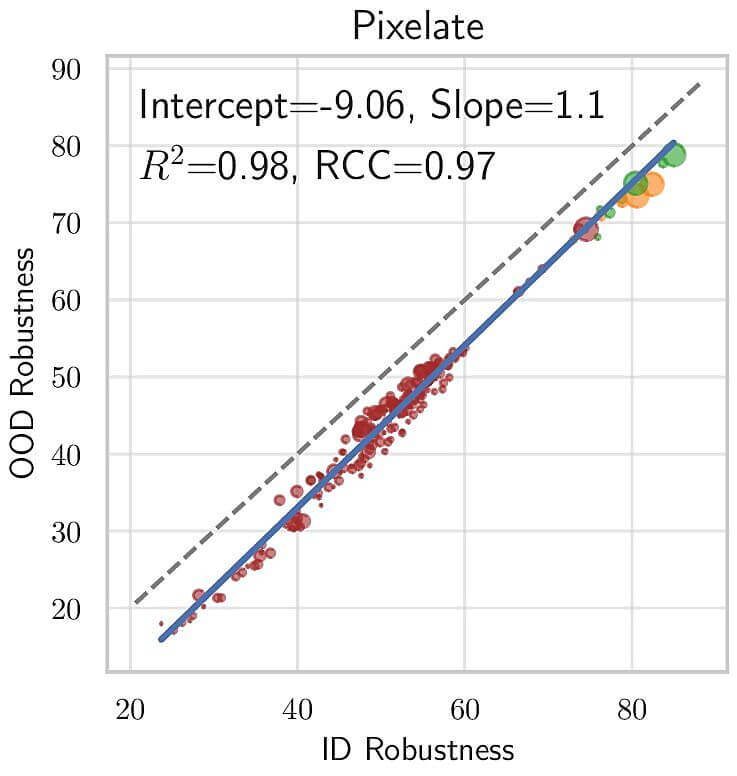}
    \end{subfigure}
    \begin{subfigure}{.245\linewidth}
        \includegraphics[width=\linewidth]{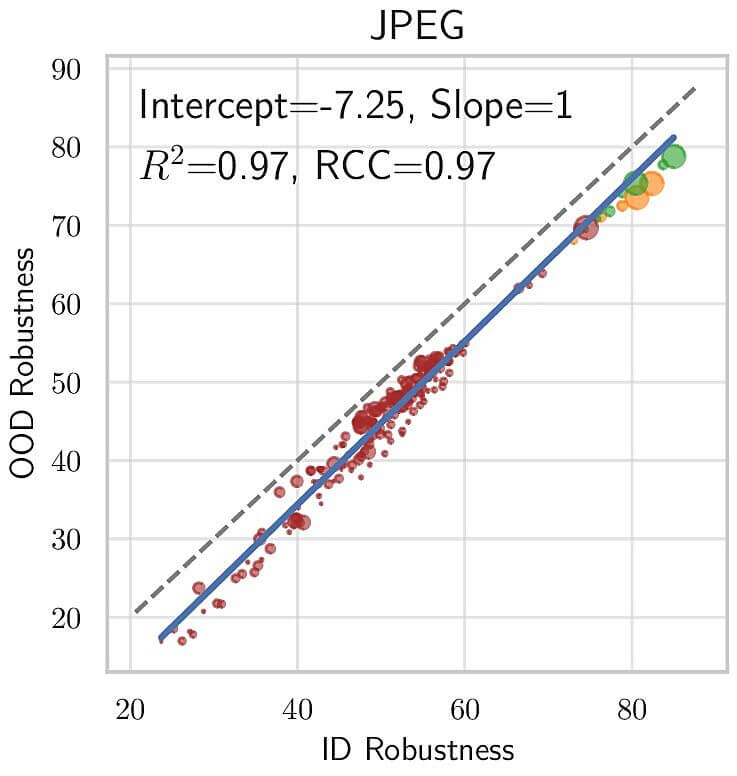}
    \end{subfigure}
    
    \caption{\textbf{Correlation between ID accuracy and OOD accuracy (odd rows); ID robustness and OOD robustness (even rows) for CIFAR10 \lt AT models}.}

\end{figure}

\begin{figure}[!h]
    \centering
    \begin{subfigure}{\linewidth}
        \includegraphics[width=\linewidth, trim=0 25 0 25, clip]{images/legend.jpg}
    \end{subfigure}
    
    \begin{subfigure}{.245\linewidth}
        \includegraphics[width=\linewidth]{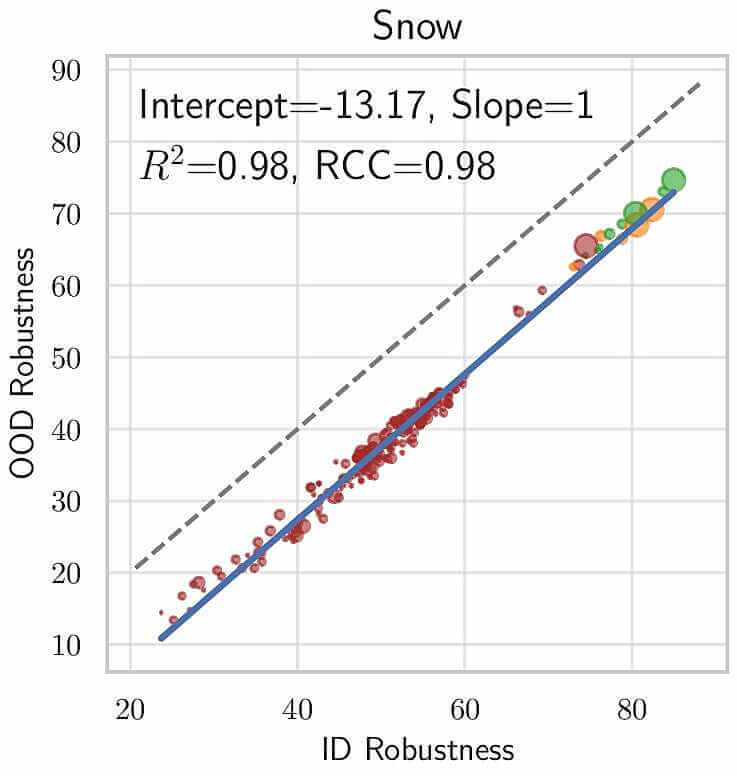}
    \end{subfigure}
    \begin{subfigure}{.245\linewidth}
        \includegraphics[width=\linewidth]{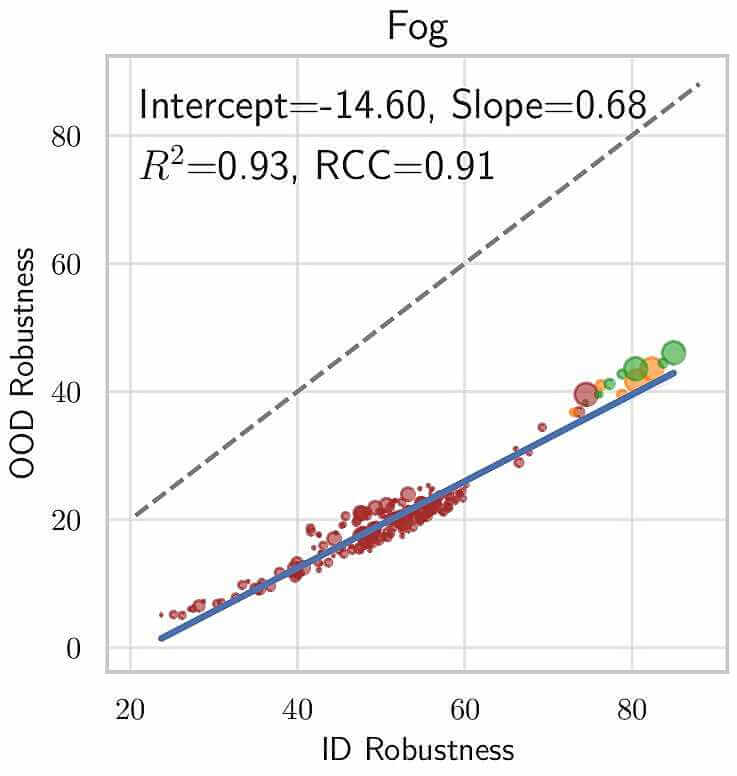}
    \end{subfigure}
    \begin{subfigure}{.245\linewidth}
        \includegraphics[width=\linewidth]{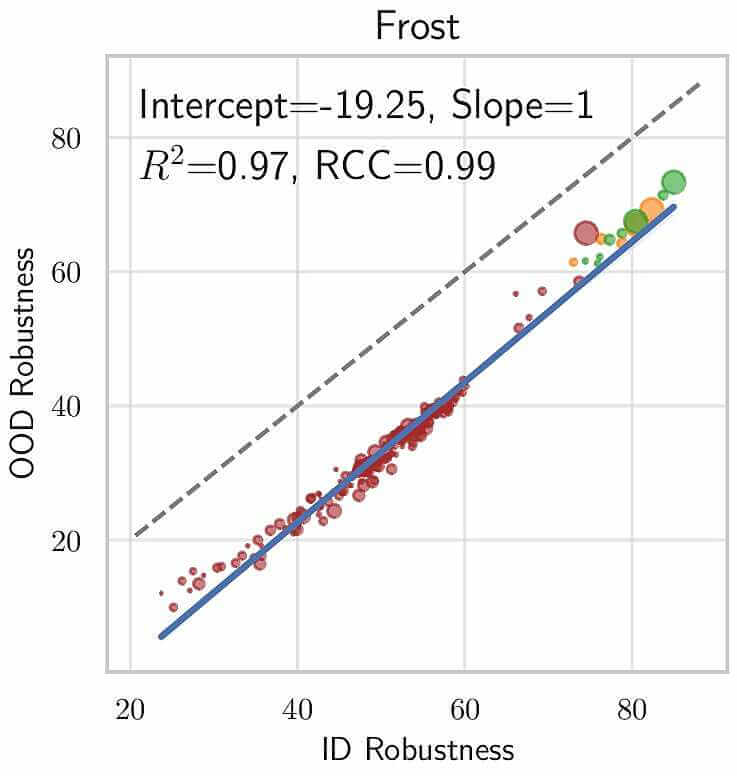}
    \end{subfigure} \hfill

    \begin{subfigure}{.245\linewidth}
        \includegraphics[width=\linewidth]{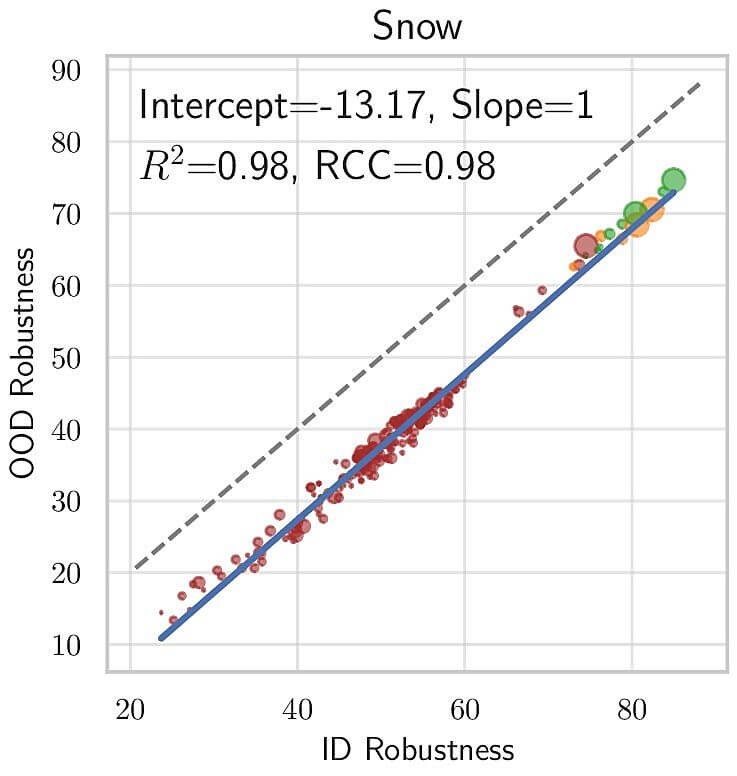}
    \end{subfigure}
    \begin{subfigure}{.245\linewidth}
        \includegraphics[width=\linewidth]{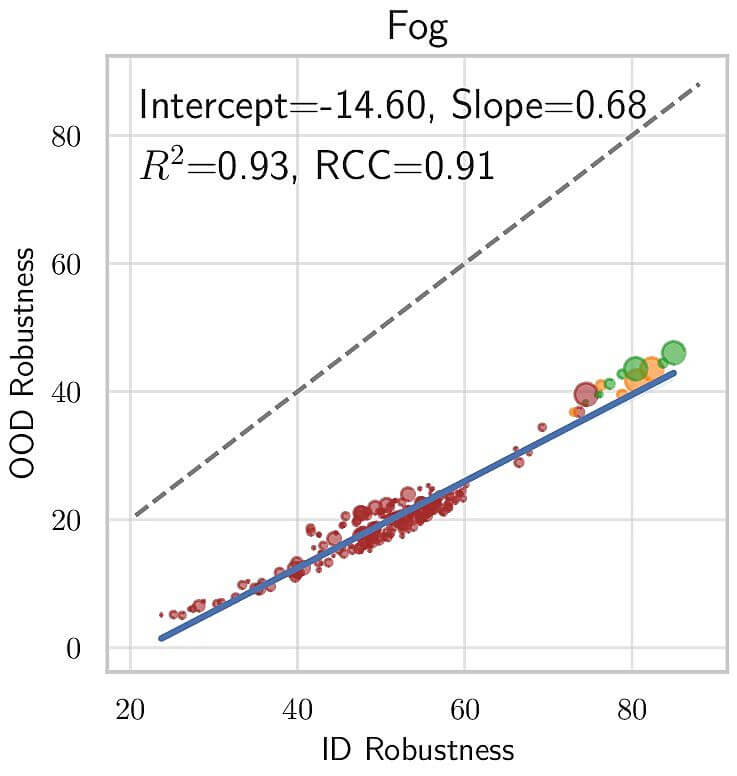}
    \end{subfigure}
    \begin{subfigure}{.245\linewidth}
        \includegraphics[width=\linewidth]{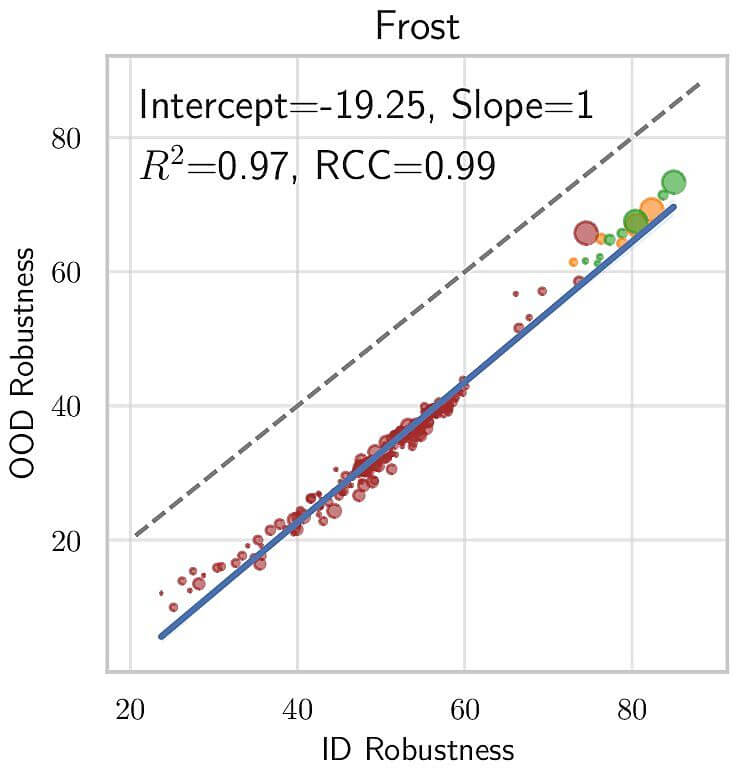}
    \end{subfigure} \hfill

        \begin{subfigure}{.245\linewidth}
        \includegraphics[width=\linewidth]{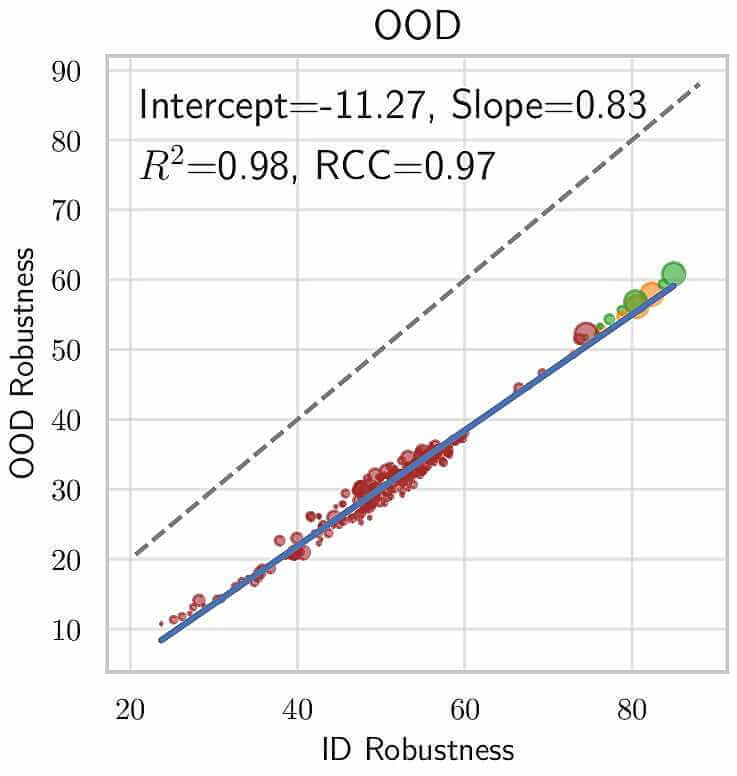}
    \end{subfigure}
    \begin{subfigure}{.245\linewidth}
        \includegraphics[width=\linewidth]{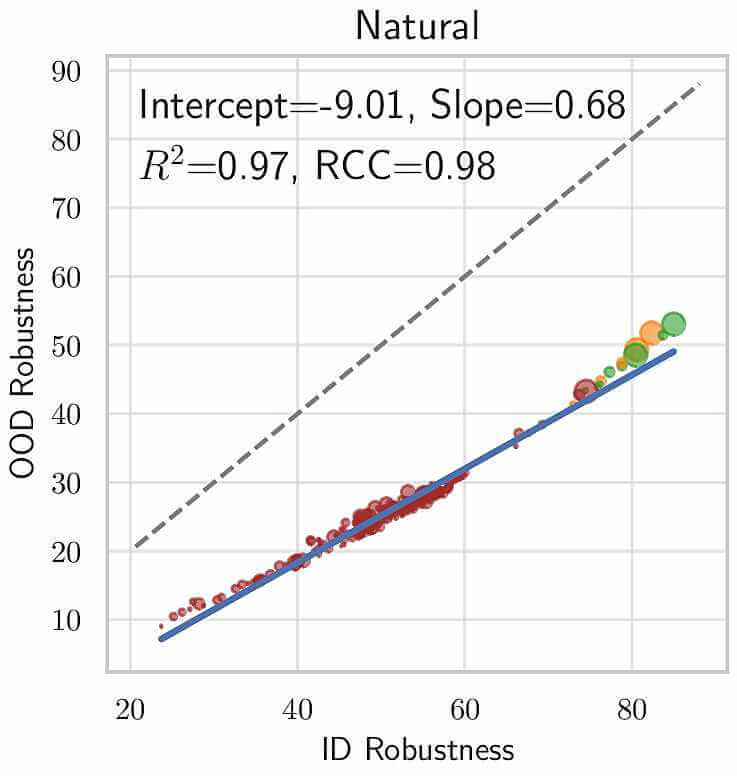}
    \end{subfigure}
    \begin{subfigure}{.245\linewidth}
        \includegraphics[width=\linewidth]{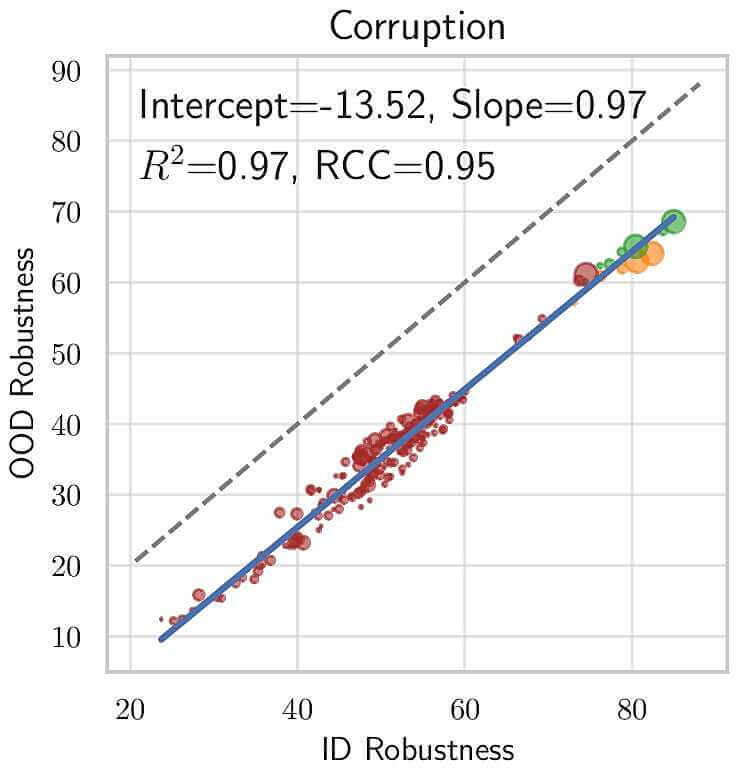}
    \end{subfigure}

    \begin{subfigure}{.245\linewidth}
        \includegraphics[width=\linewidth]{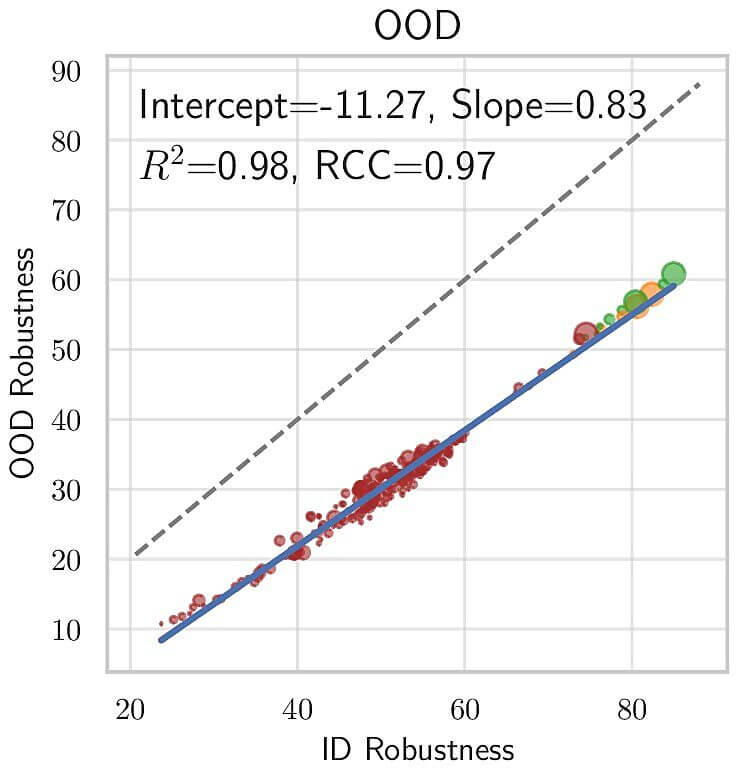}
    \end{subfigure}
    \begin{subfigure}{.245\linewidth}
        \includegraphics[width=\linewidth]{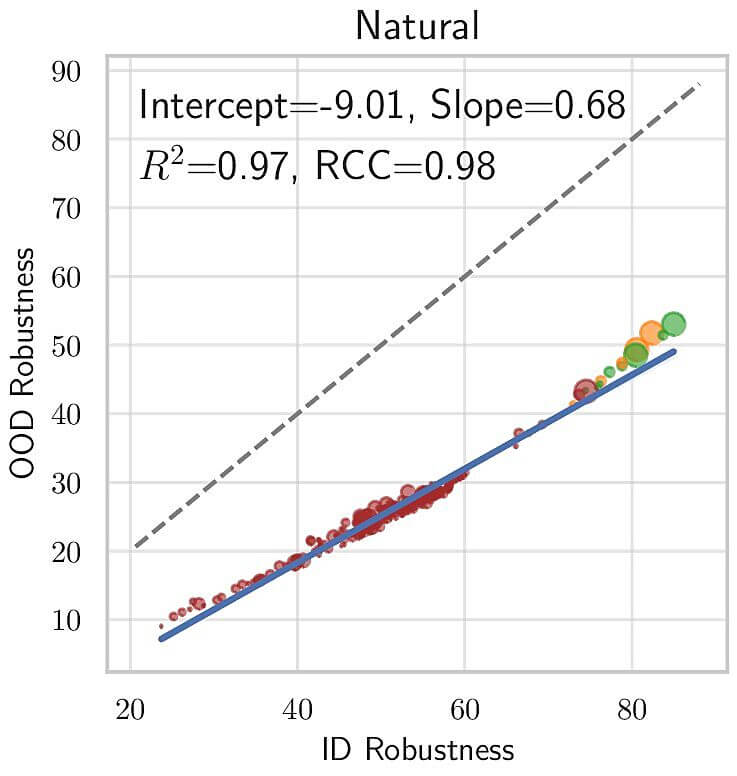}
    \end{subfigure}
    \begin{subfigure}{.245\linewidth}
        \includegraphics[width=\linewidth]{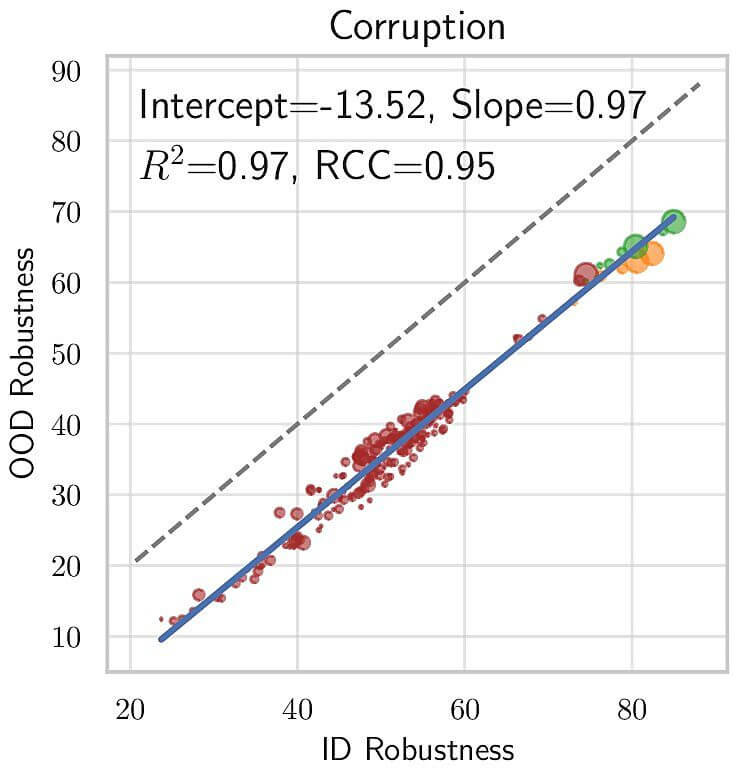}
    \end{subfigure}
    
    \caption{\textbf{Correlation between ID accuracy and OOD accuracy (odd rows); ID robustness and OOD robustness (even rows) for CIFAR10 \lt AT models}.}
\end{figure}


\begin{figure}[!h]
    \centering
    \begin{subfigure}{\linewidth}
        \includegraphics[width=\linewidth, trim=0 25 0 25, clip]{images/legend.jpg}
    \end{subfigure}
    
    \begin{subfigure}{.245\linewidth}
        \includegraphics[width=\linewidth]{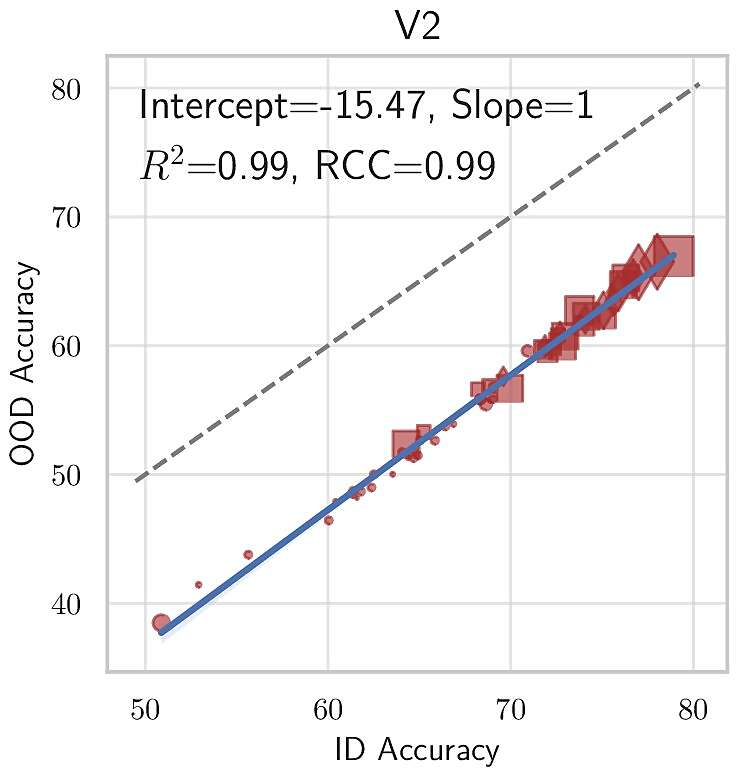}
    \end{subfigure}
    \begin{subfigure}{.245\linewidth}
        \includegraphics[width=\linewidth]{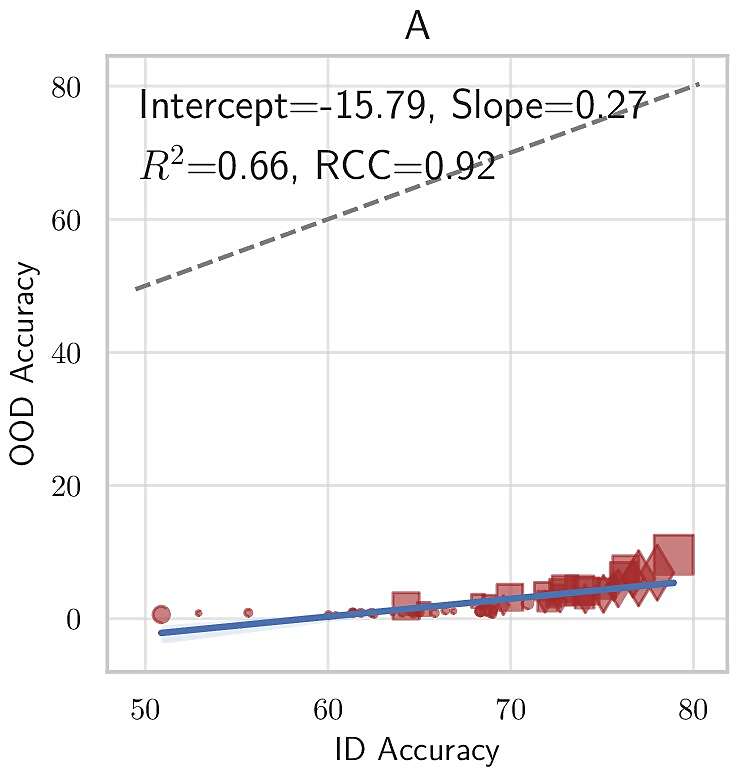}
    \end{subfigure}
    \begin{subfigure}{.245\linewidth}
        \includegraphics[width=\linewidth]{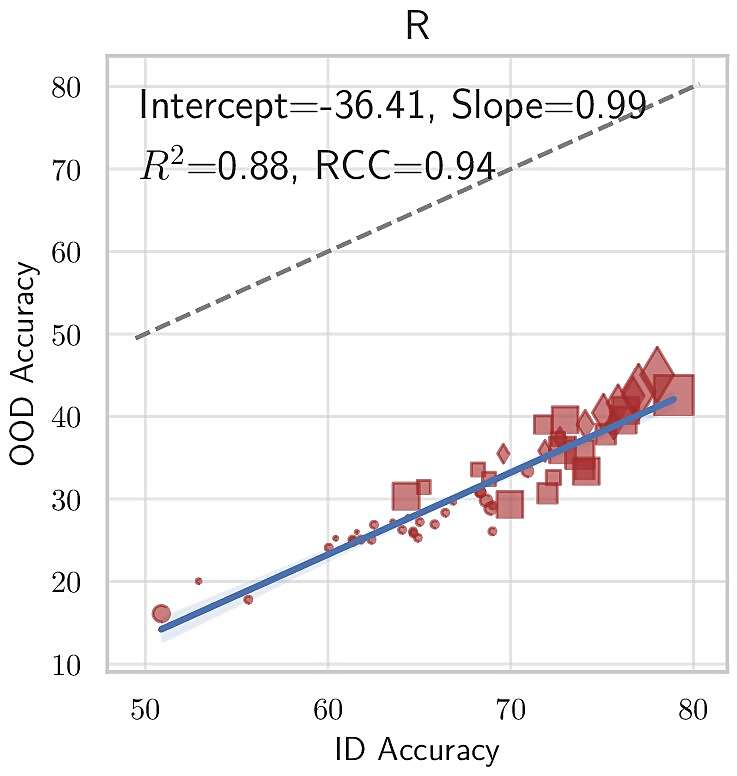}
    \end{subfigure}
    \begin{subfigure}{.245\linewidth}
        \includegraphics[width=\linewidth]{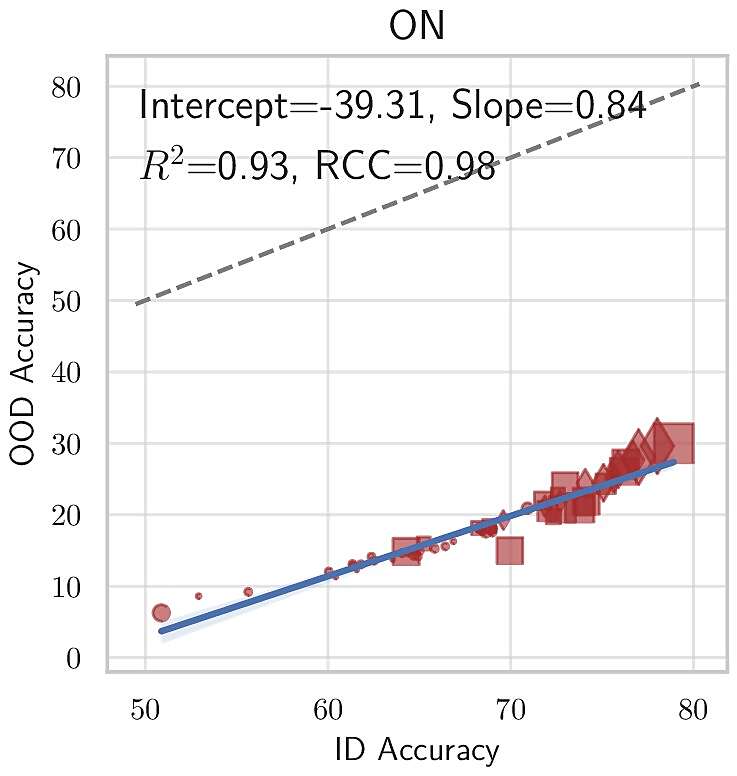}
    \end{subfigure}

    \begin{subfigure}{.245\linewidth}
        \includegraphics[width=\linewidth]{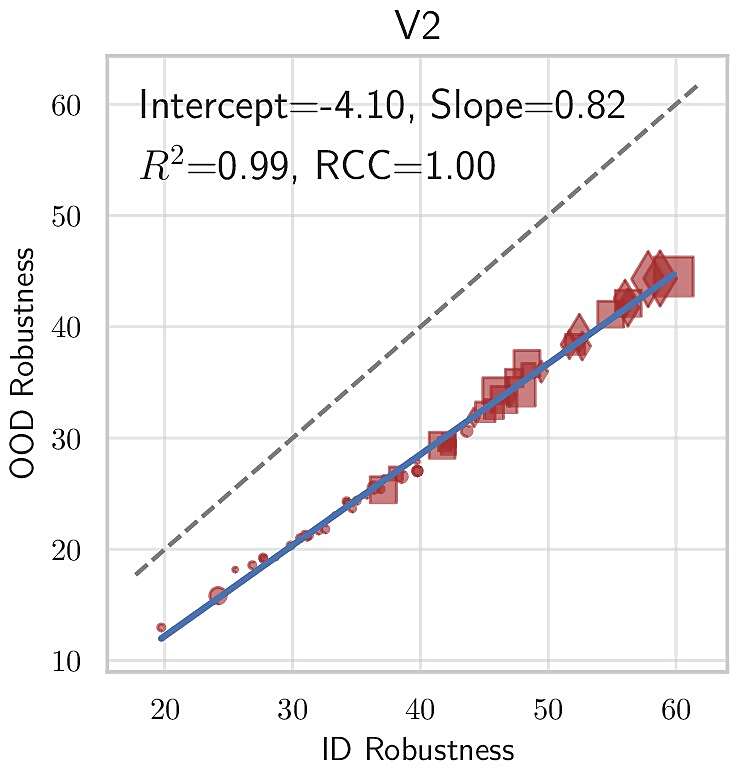}
    \end{subfigure}
    \begin{subfigure}{.245\linewidth}
        \includegraphics[width=\linewidth]{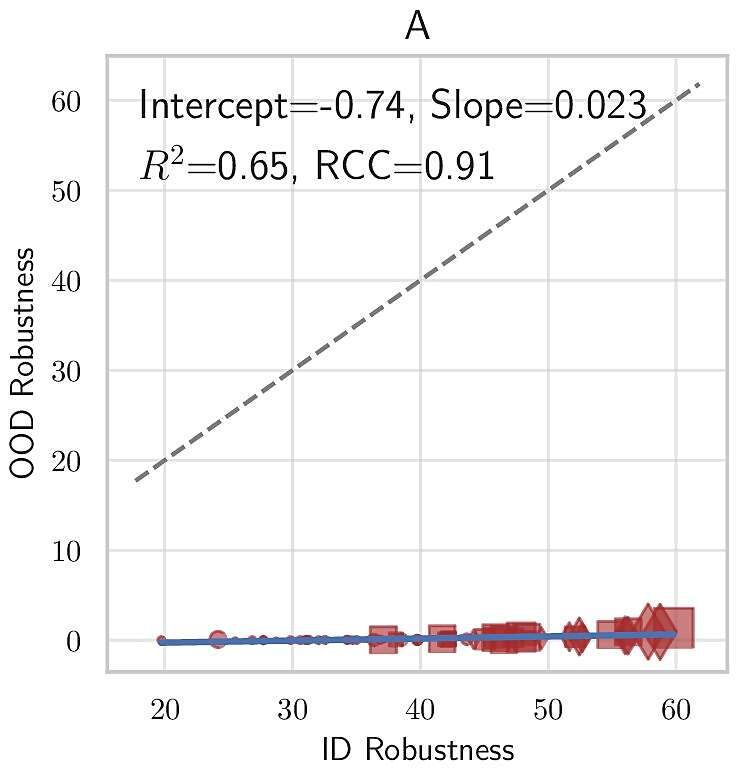}
    \end{subfigure}
    \begin{subfigure}{.245\linewidth}
        \includegraphics[width=\linewidth]{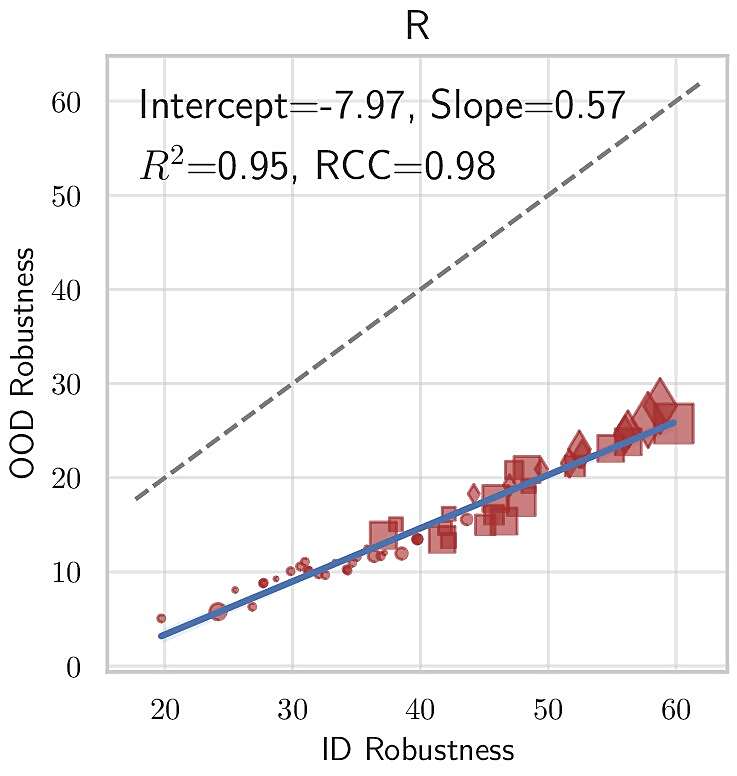}
    \end{subfigure}
    \begin{subfigure}{.245\linewidth}
        \includegraphics[width=\linewidth]{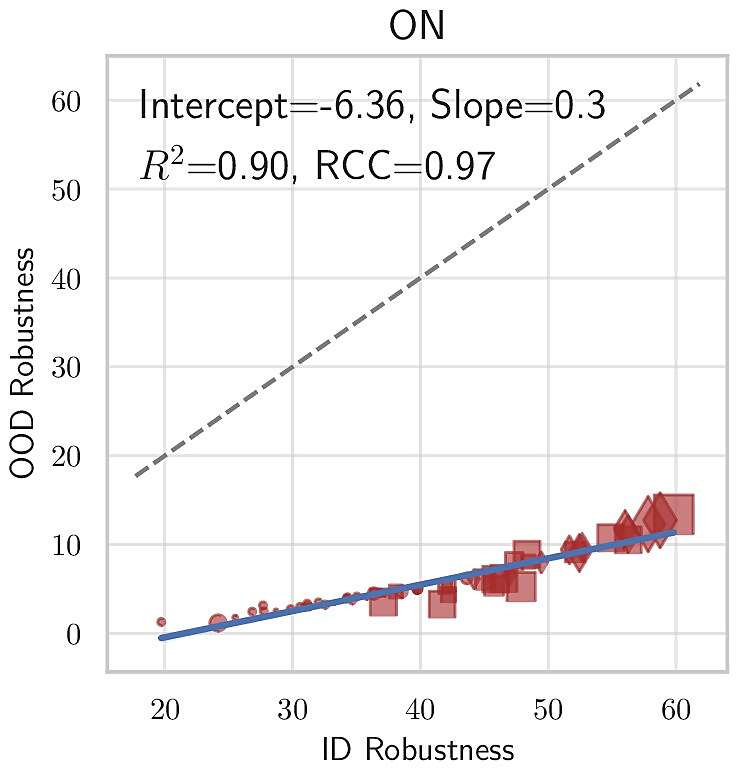}
    \end{subfigure}

    \begin{subfigure}{.245\linewidth}
        \includegraphics[width=\linewidth]{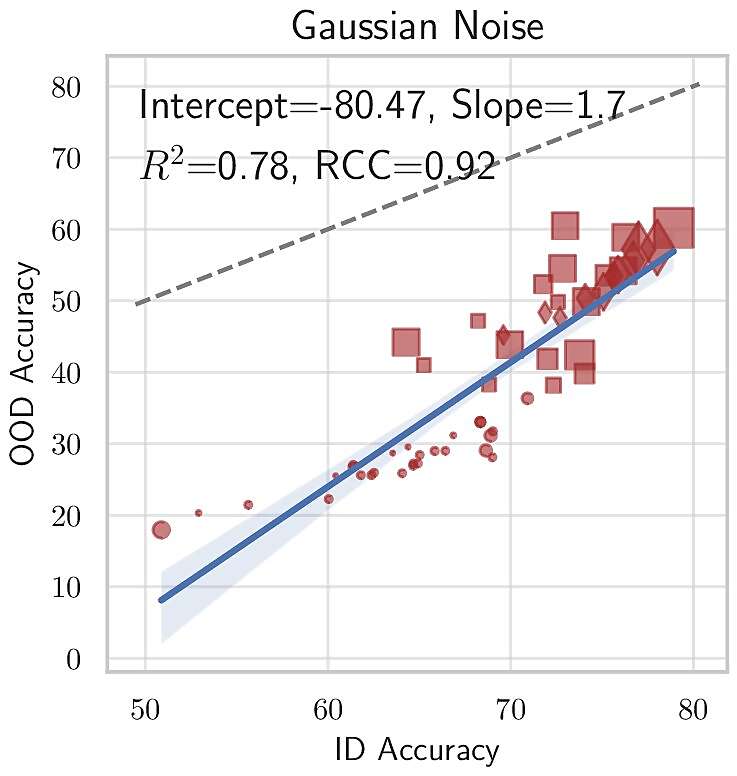}
    \end{subfigure}
    \begin{subfigure}{.245\linewidth}
        \includegraphics[width=\linewidth]{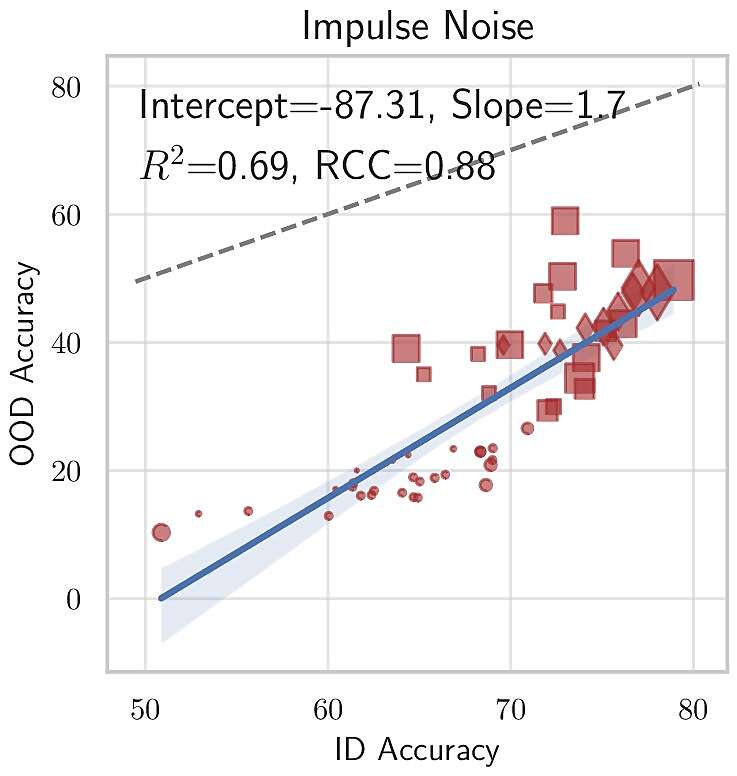}
    \end{subfigure}
    \begin{subfigure}{.245\linewidth}
        \includegraphics[width=\linewidth]{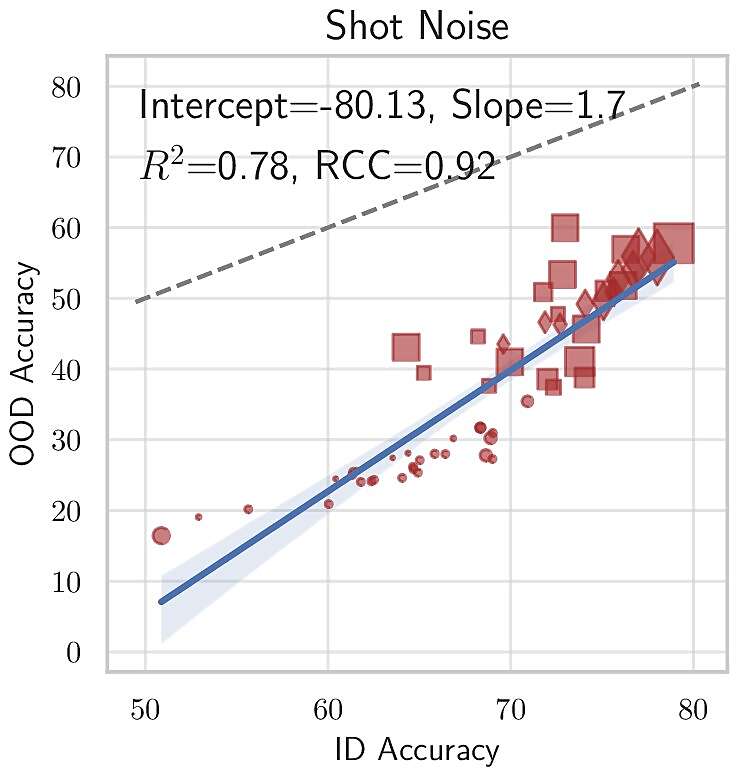}
    \end{subfigure}
    \begin{subfigure}{.245\linewidth}
        \includegraphics[width=\linewidth]{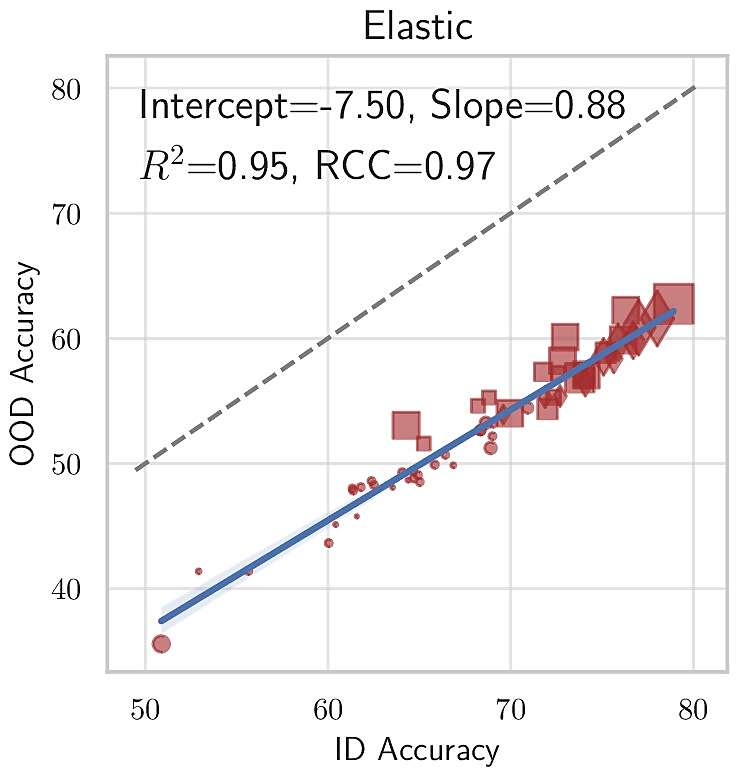}
    \end{subfigure}

    \begin{subfigure}{.245\linewidth}
        \includegraphics[width=\linewidth]{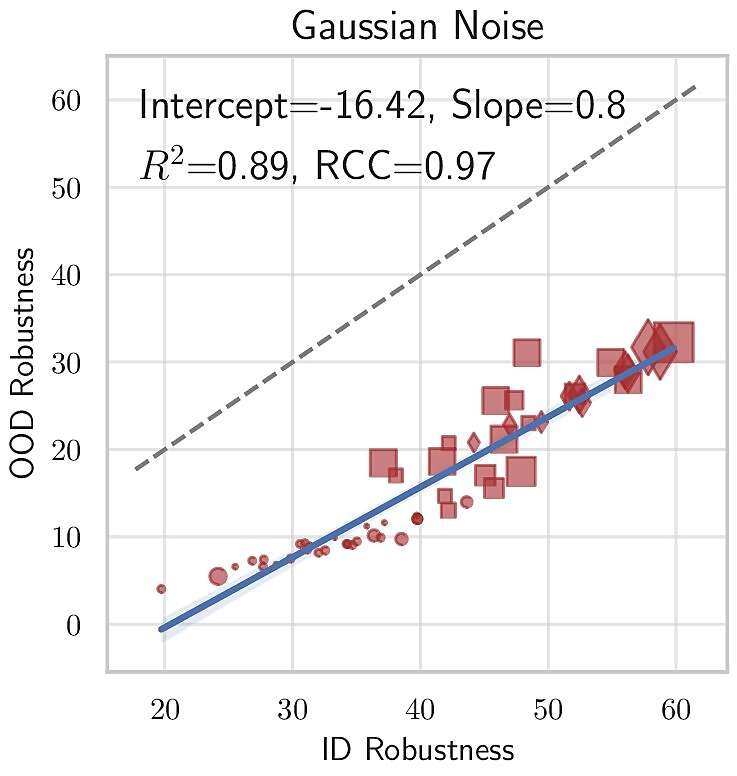}
    \end{subfigure}
    \begin{subfigure}{.245\linewidth}
        \includegraphics[width=\linewidth]{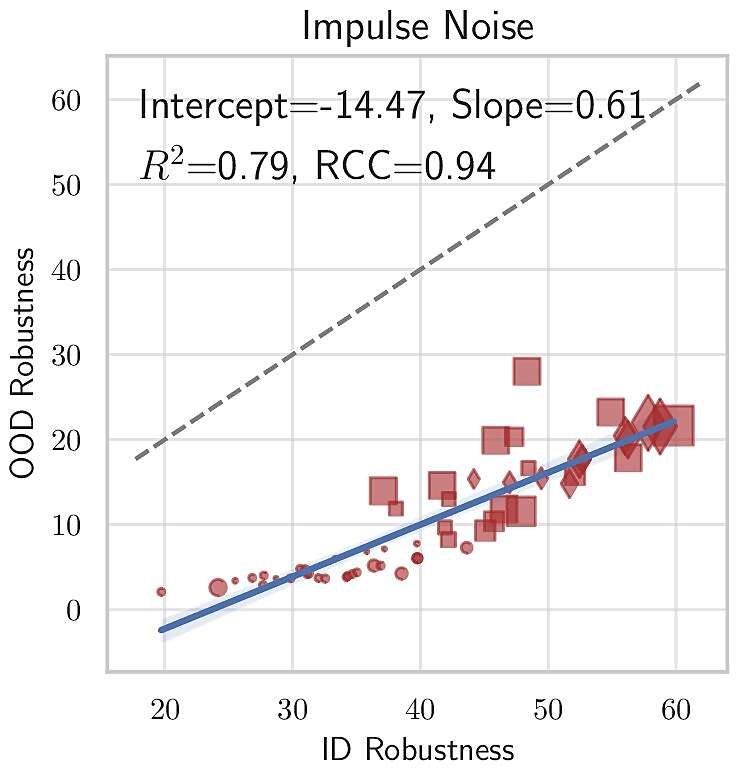}
    \end{subfigure}
    \begin{subfigure}{.245\linewidth}
        \includegraphics[width=\linewidth]{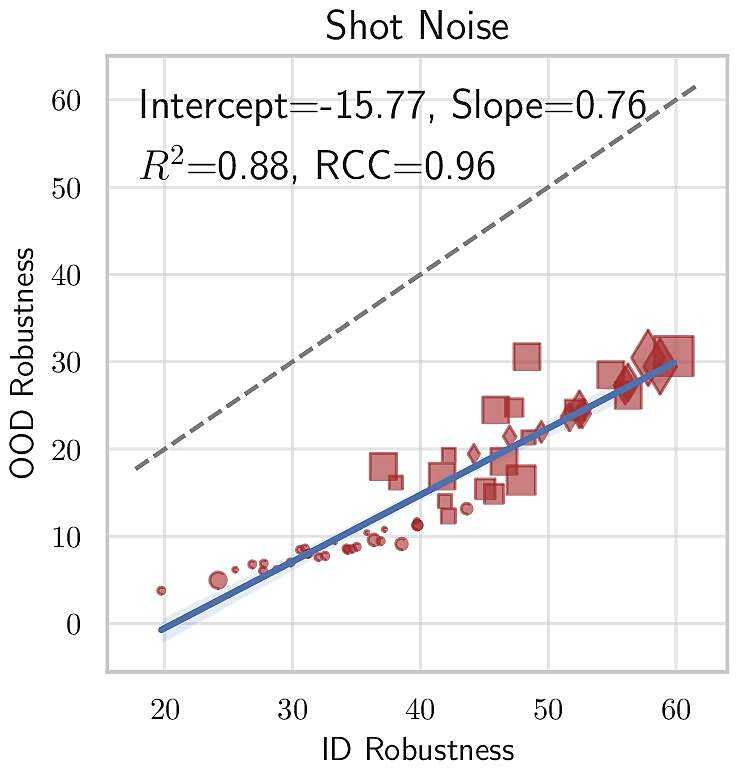}
    \end{subfigure}
    \begin{subfigure}{.245\linewidth}
        \includegraphics[width=\linewidth]{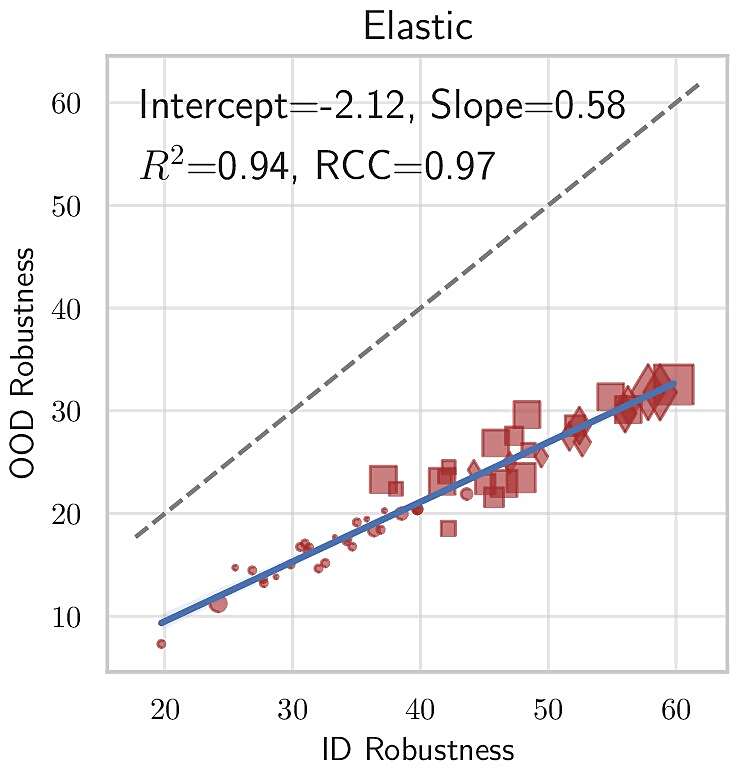}
    \end{subfigure}

    \caption{\textbf{Correlation between ID accuracy and OOD accuracy (odd rows); ID robustness and OOD robustness (even rows) for ImageNet \linf AT models}.}

\end{figure}

\begin{figure}[!h]
    \centering
    \begin{subfigure}{\linewidth}
        \includegraphics[width=\linewidth, trim=0 25 0 25, clip]{images/legend.jpg}
    \end{subfigure}

    \begin{subfigure}{.245\linewidth}
        \includegraphics[width=\linewidth]{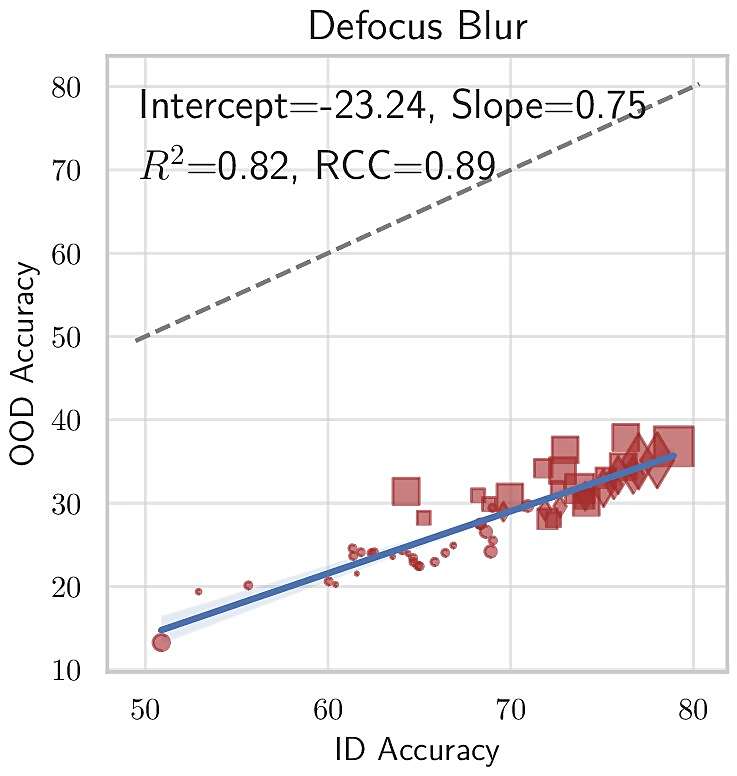}
    \end{subfigure}
    \begin{subfigure}{.245\linewidth}
        \includegraphics[width=\linewidth]{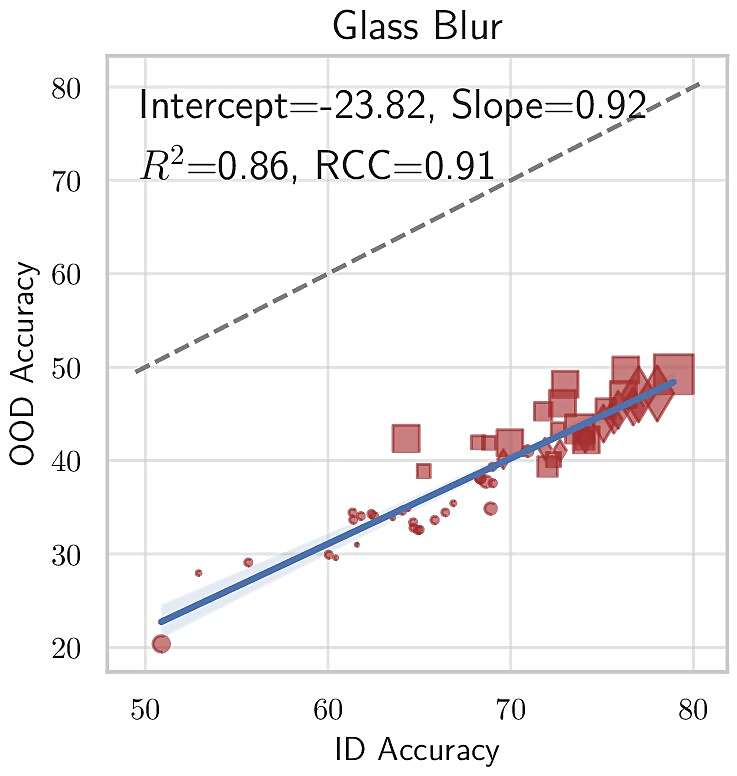}
    \end{subfigure}
    \begin{subfigure}{.245\linewidth}
        \includegraphics[width=\linewidth]{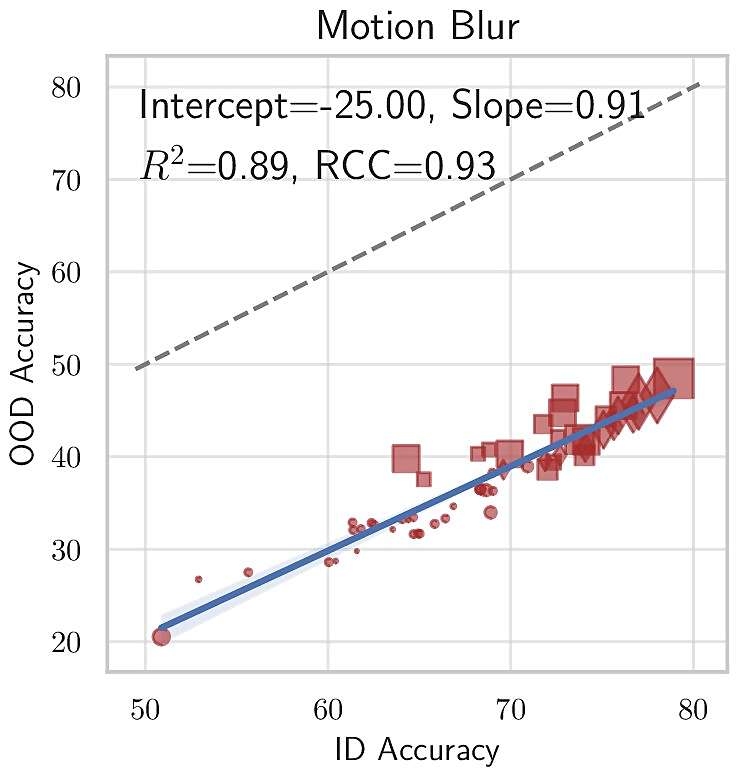}
    \end{subfigure}
    \begin{subfigure}{.245\linewidth}
        \includegraphics[width=\linewidth]{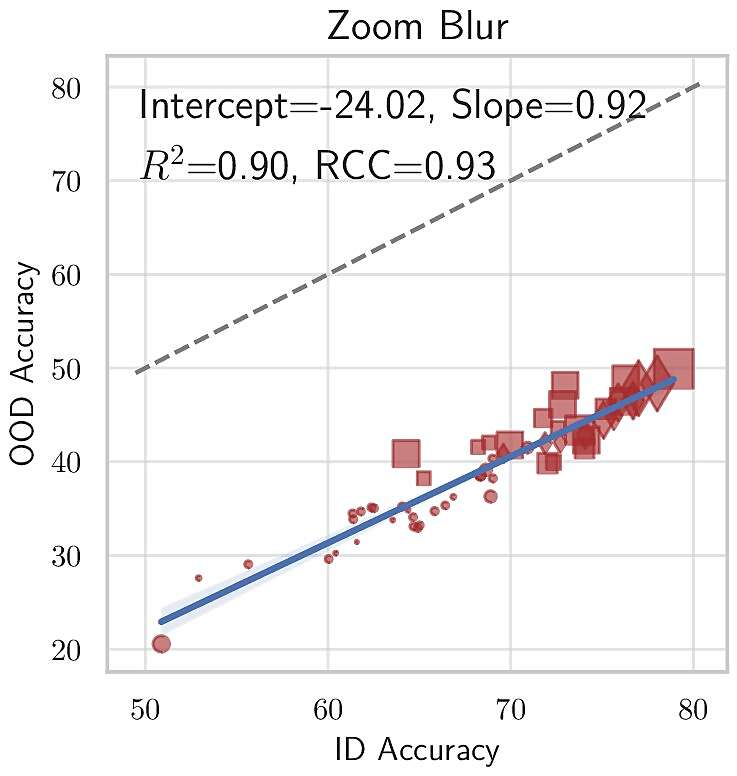}
    \end{subfigure}

    \begin{subfigure}{.245\linewidth}
        \includegraphics[width=\linewidth]{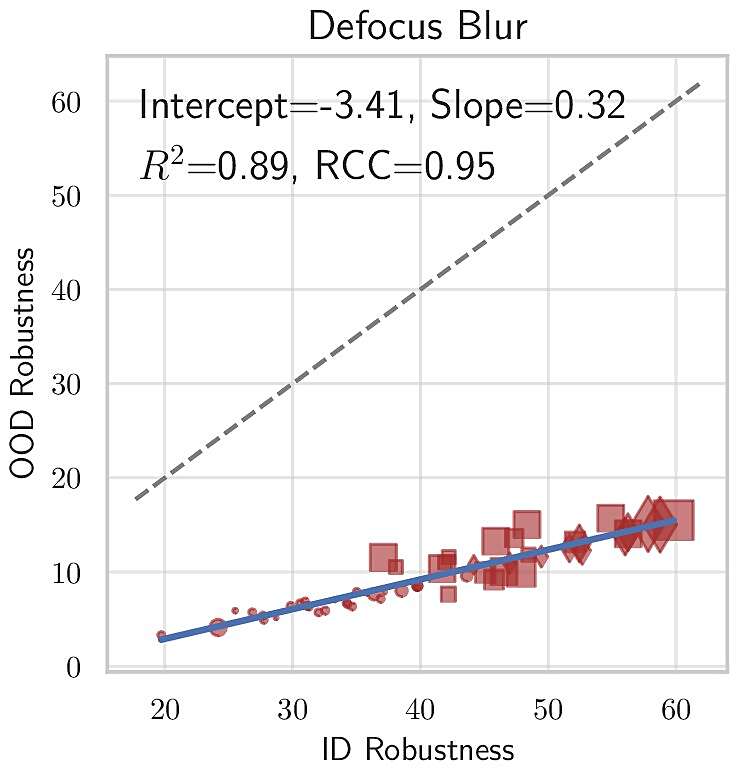}
    \end{subfigure}
    \begin{subfigure}{.245\linewidth}
        \includegraphics[width=\linewidth]{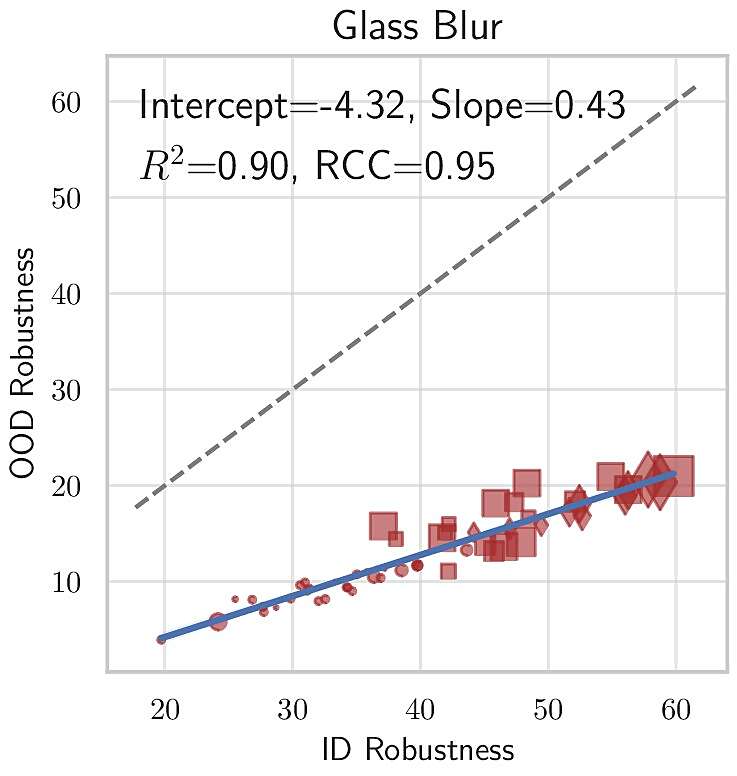}
    \end{subfigure}
    \begin{subfigure}{.245\linewidth}
        \includegraphics[width=\linewidth]{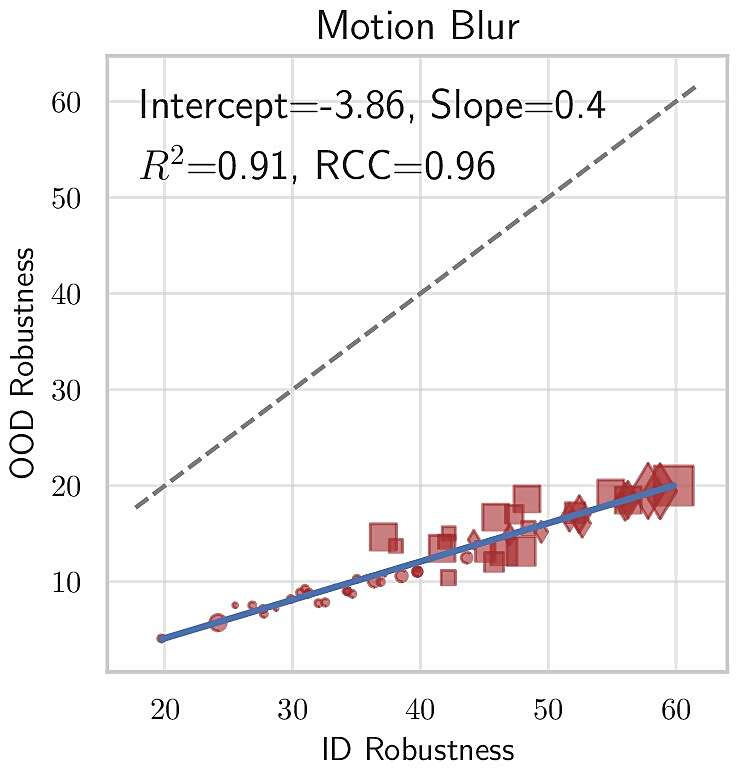}
    \end{subfigure}
    \begin{subfigure}{.245\linewidth}
        \includegraphics[width=\linewidth]{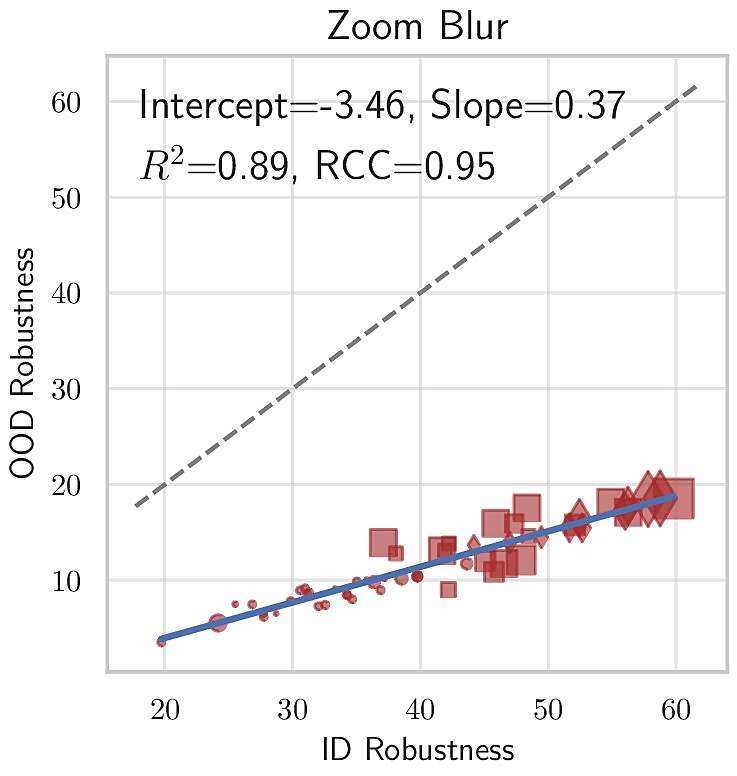}
    \end{subfigure}

    \begin{subfigure}{.245\linewidth}
        \includegraphics[width=\linewidth]{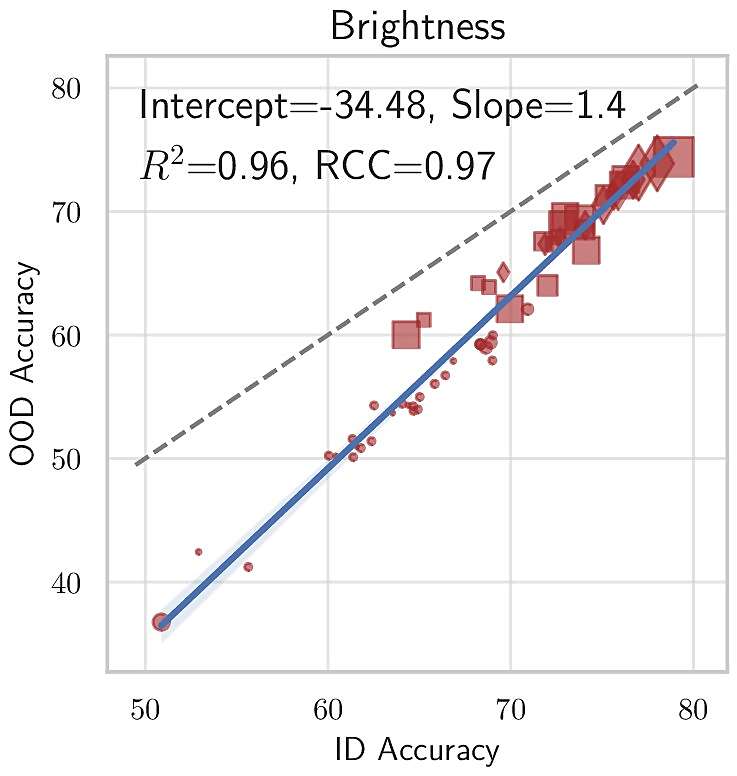}
    \end{subfigure}
    \begin{subfigure}{.245\linewidth}
        \includegraphics[width=\linewidth]{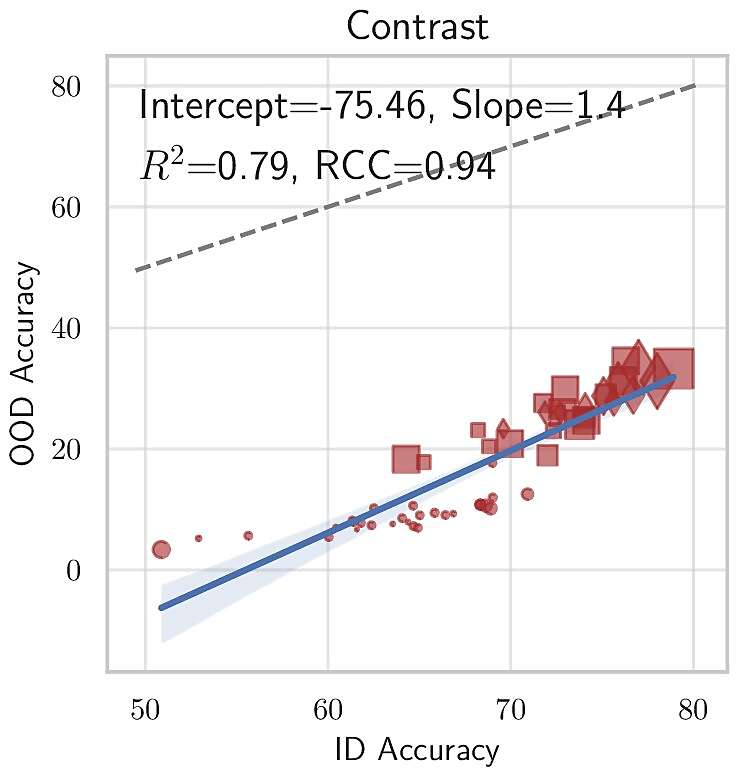}
    \end{subfigure}
    \begin{subfigure}{.245\linewidth}
        \includegraphics[width=\linewidth]{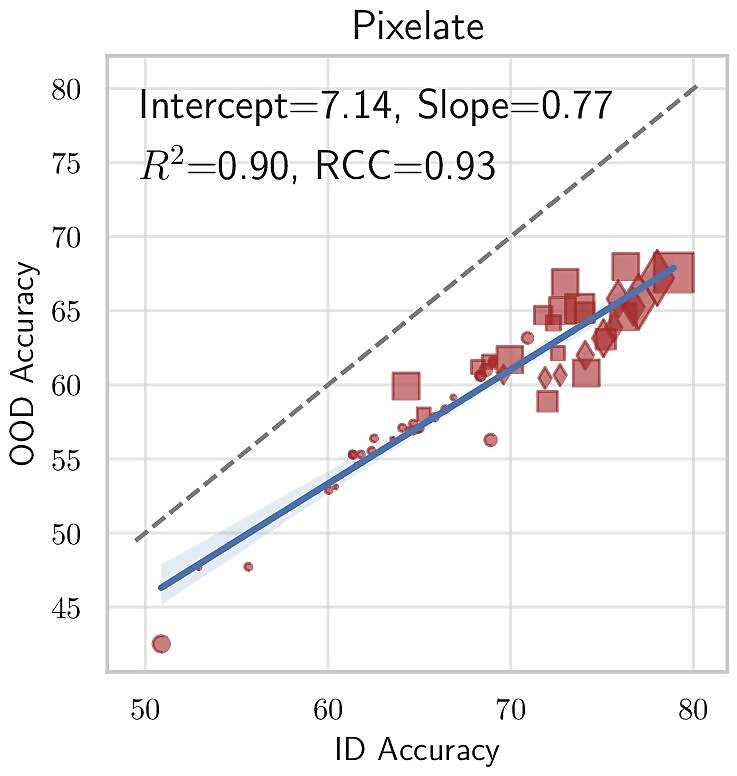}
    \end{subfigure}
    \begin{subfigure}{.245\linewidth}
        \includegraphics[width=\linewidth]{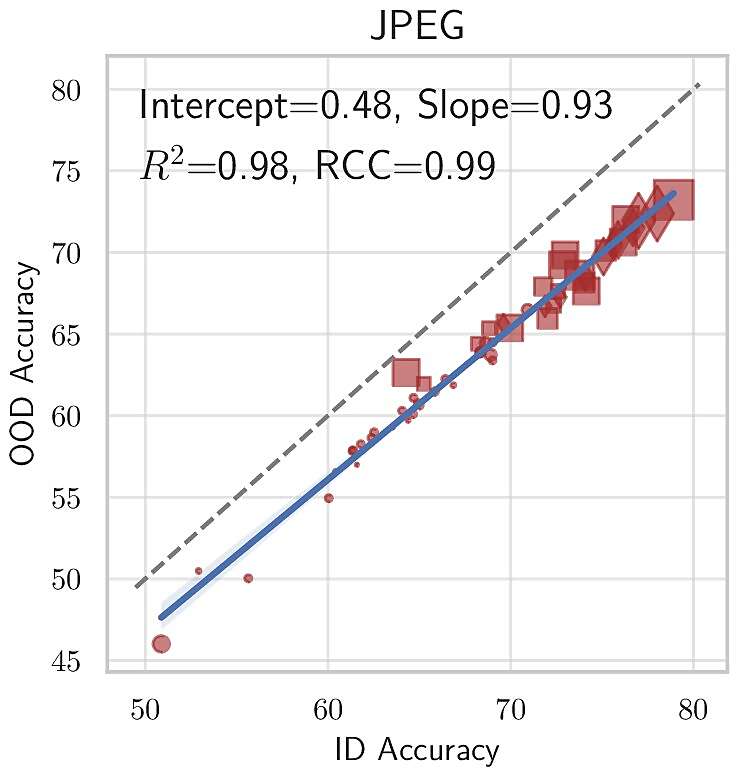}
    \end{subfigure}

    \begin{subfigure}{.245\linewidth}
        \includegraphics[width=\linewidth]{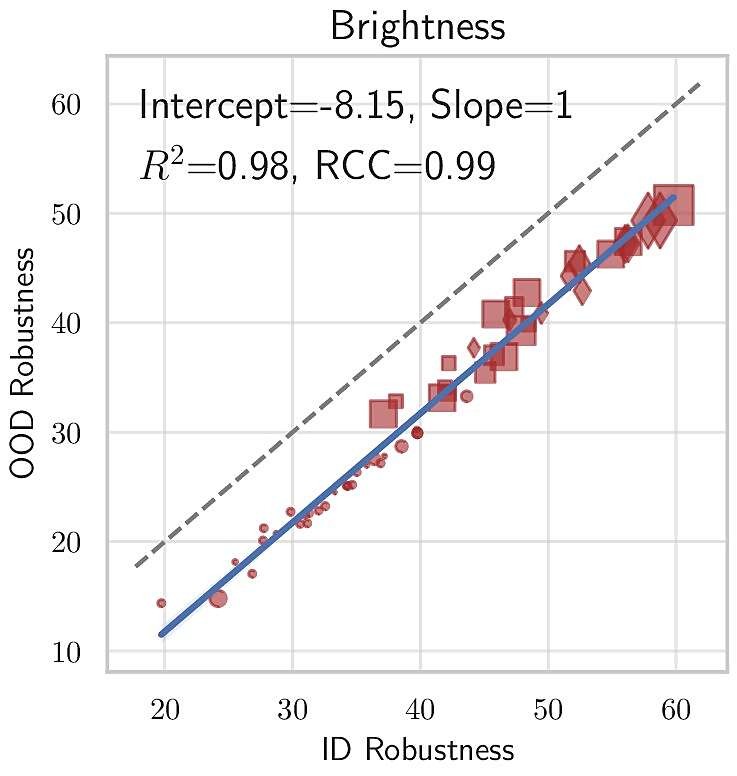}
    \end{subfigure}
    \begin{subfigure}{.245\linewidth}
        \includegraphics[width=\linewidth]{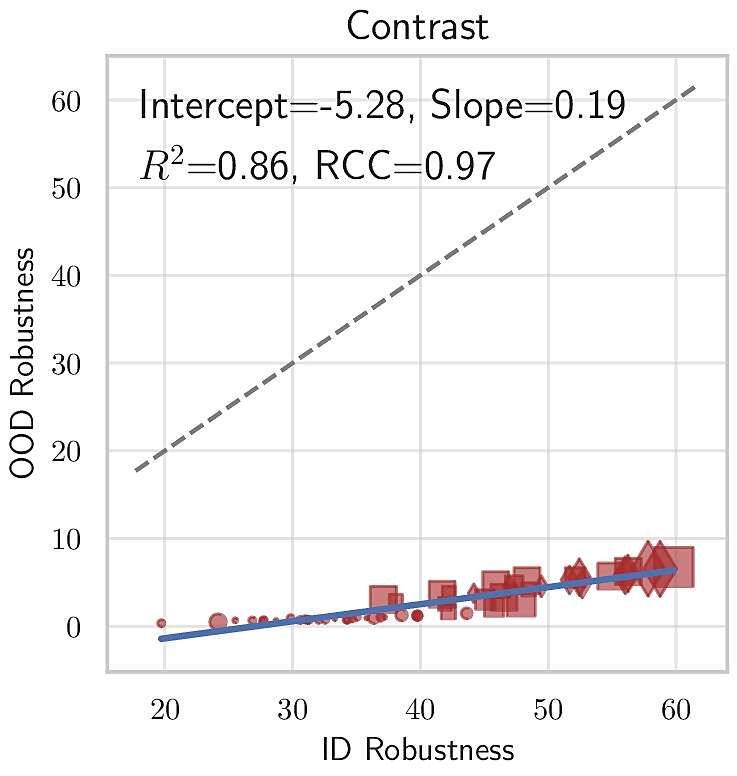}
    \end{subfigure}
    \begin{subfigure}{.245\linewidth}
        \includegraphics[width=\linewidth]{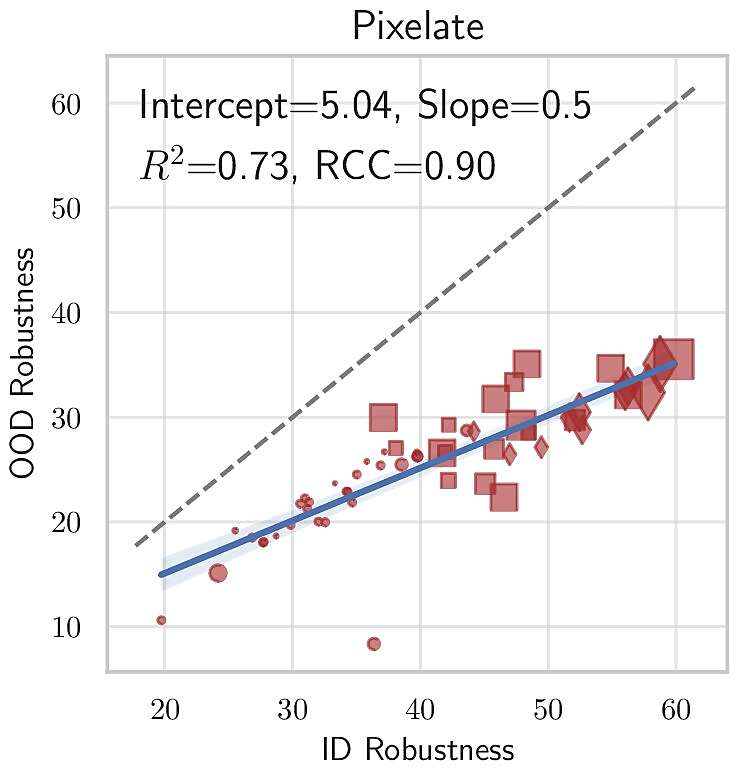}
    \end{subfigure}
    \begin{subfigure}{.245\linewidth}
        \includegraphics[width=\linewidth]{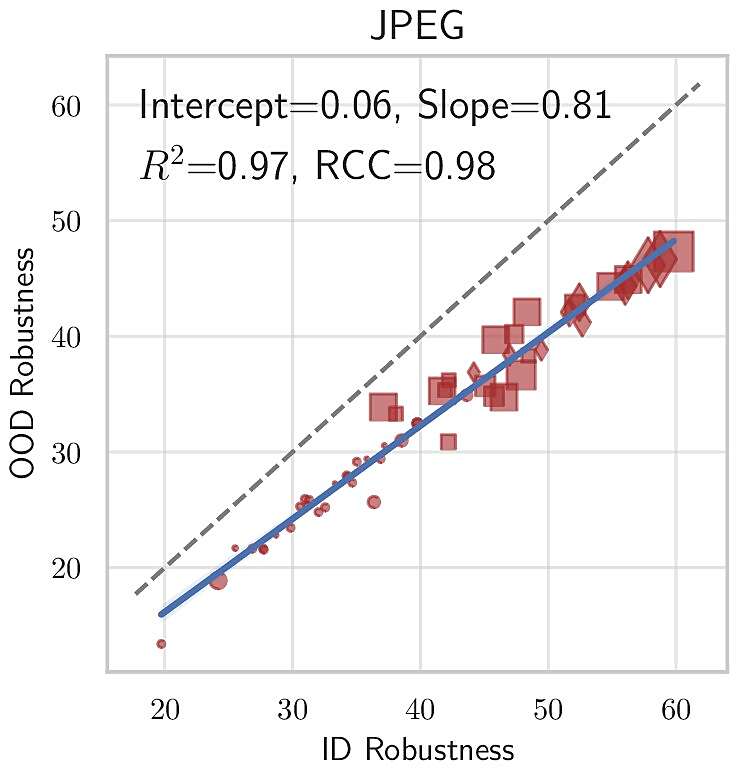}
    \end{subfigure}
    
    \caption{\textbf{Correlation between ID accuracy and OOD accuracy (odd rows); ID robustness and OOD robustness (even rows) for ImageNet \linf AT models}.}

\end{figure}

\begin{figure}[!h]
    \centering
    \begin{subfigure}{\linewidth}
        \includegraphics[width=\linewidth, trim=0 25 0 25, clip]{images/legend.jpg}
    \end{subfigure}

    \begin{subfigure}{.245\linewidth}
        \includegraphics[width=\linewidth]{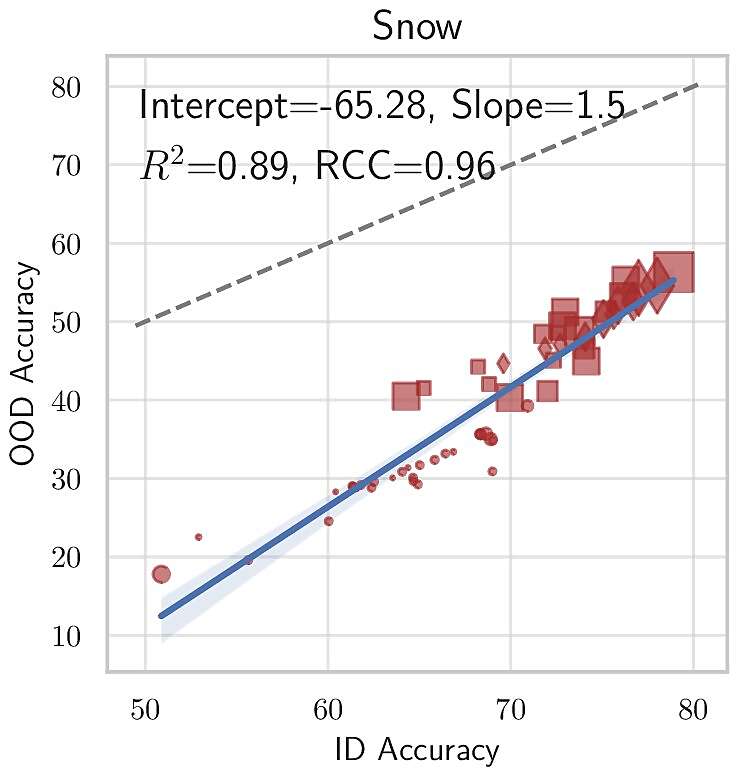}
    \end{subfigure}
    \begin{subfigure}{.245\linewidth}
        \includegraphics[width=\linewidth]{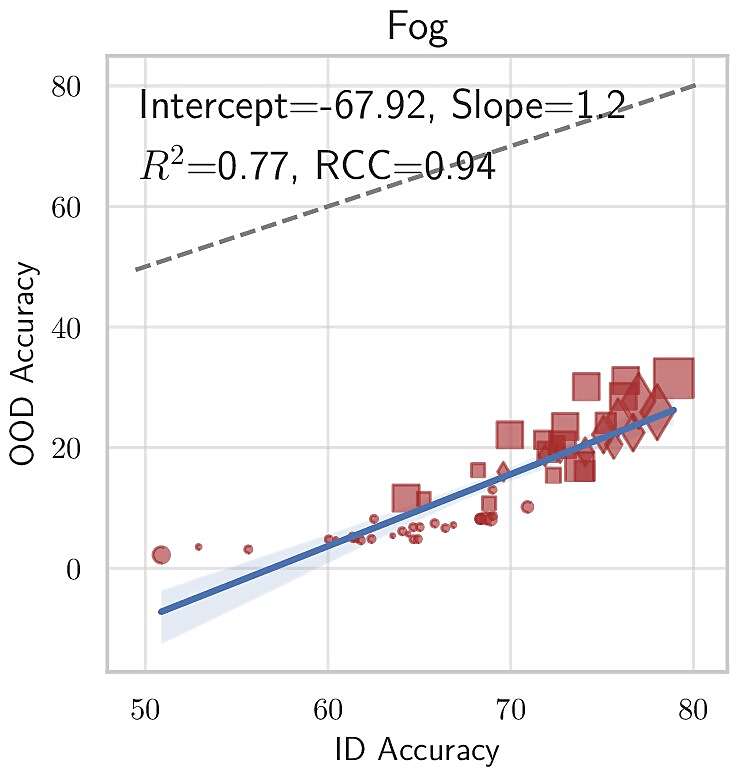}
    \end{subfigure}
    \begin{subfigure}{.245\linewidth}
        \includegraphics[width=\linewidth]{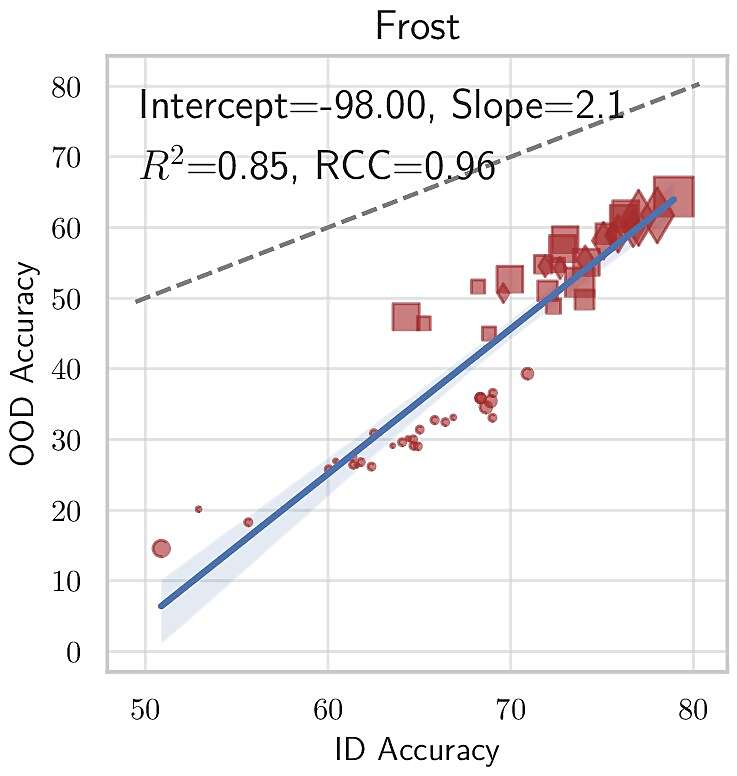}
    \end{subfigure} \hfill

    \begin{subfigure}{.245\linewidth}
        \includegraphics[width=\linewidth]{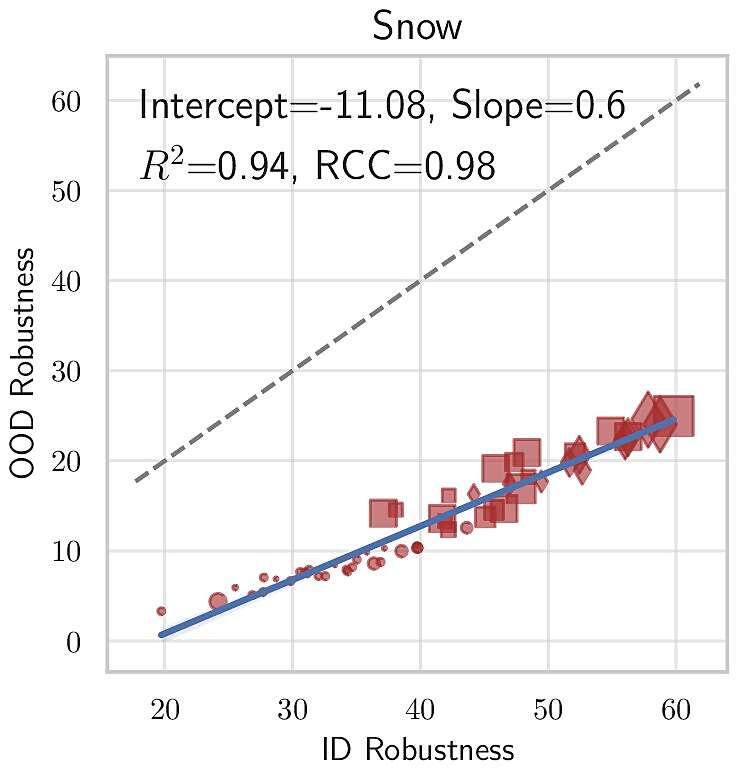}
    \end{subfigure}
    \begin{subfigure}{.245\linewidth}
        \includegraphics[width=\linewidth]{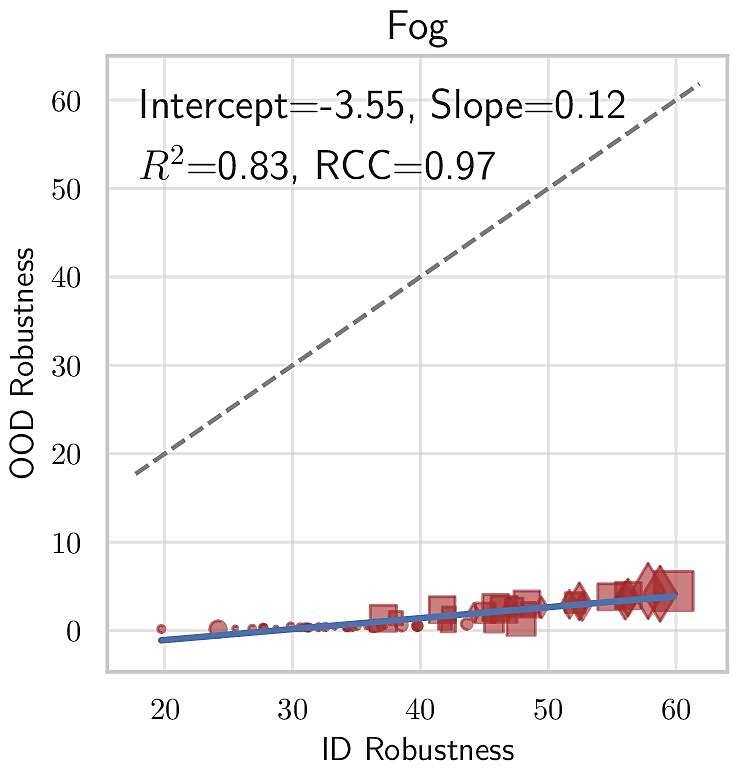}
    \end{subfigure}
    \begin{subfigure}{.245\linewidth}
        \includegraphics[width=\linewidth]{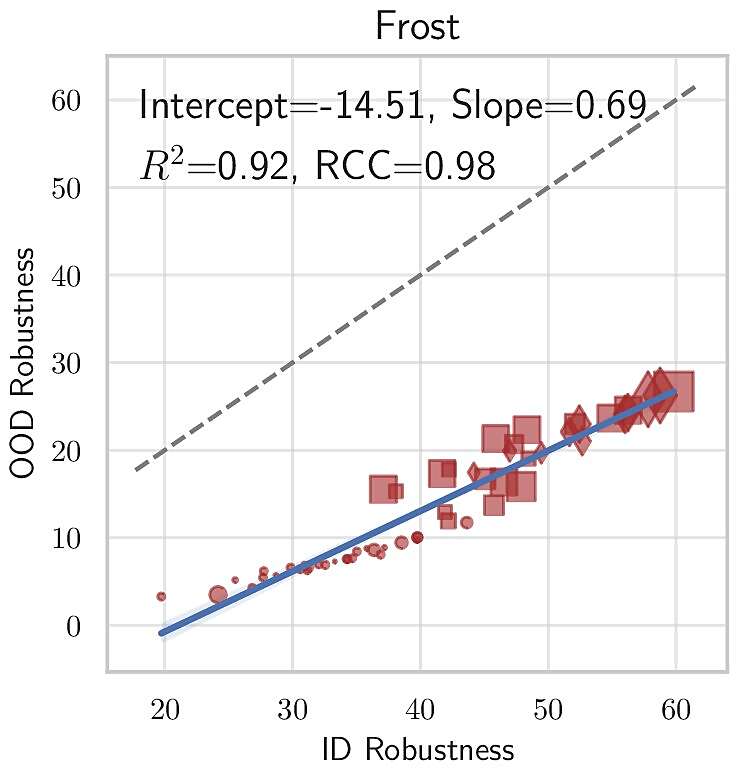}
    \end{subfigure} \hfill

    \begin{subfigure}{.245\linewidth}
        \includegraphics[width=\linewidth]{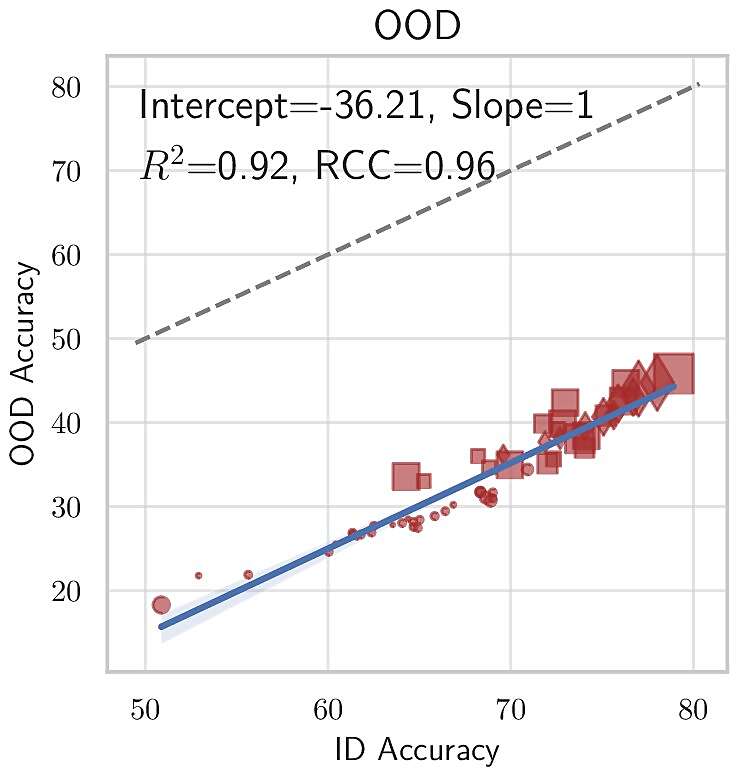}
    \end{subfigure}
    \begin{subfigure}{.245\linewidth}
        \includegraphics[width=\linewidth]{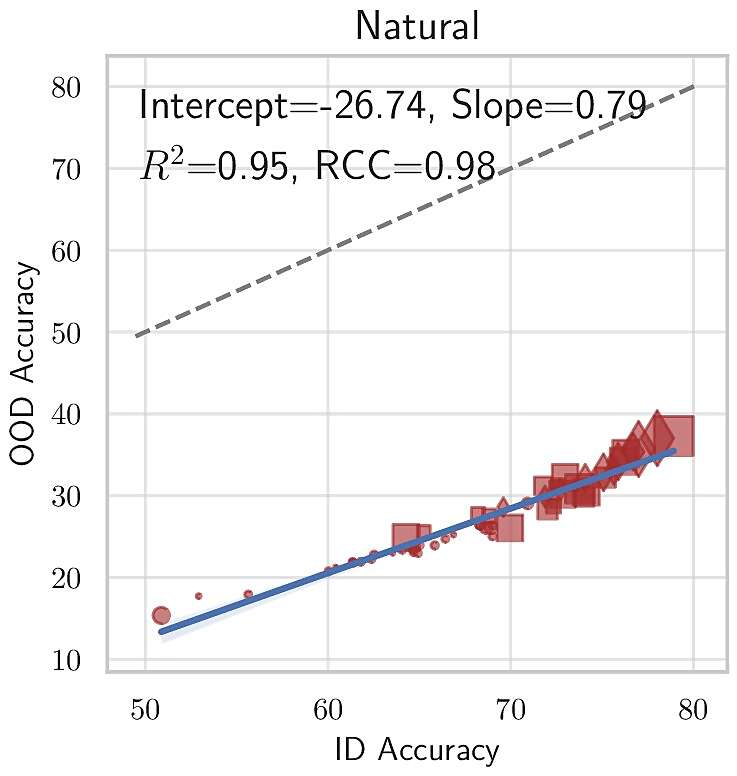}
    \end{subfigure}
    \begin{subfigure}{.245\linewidth}
        \includegraphics[width=\linewidth]{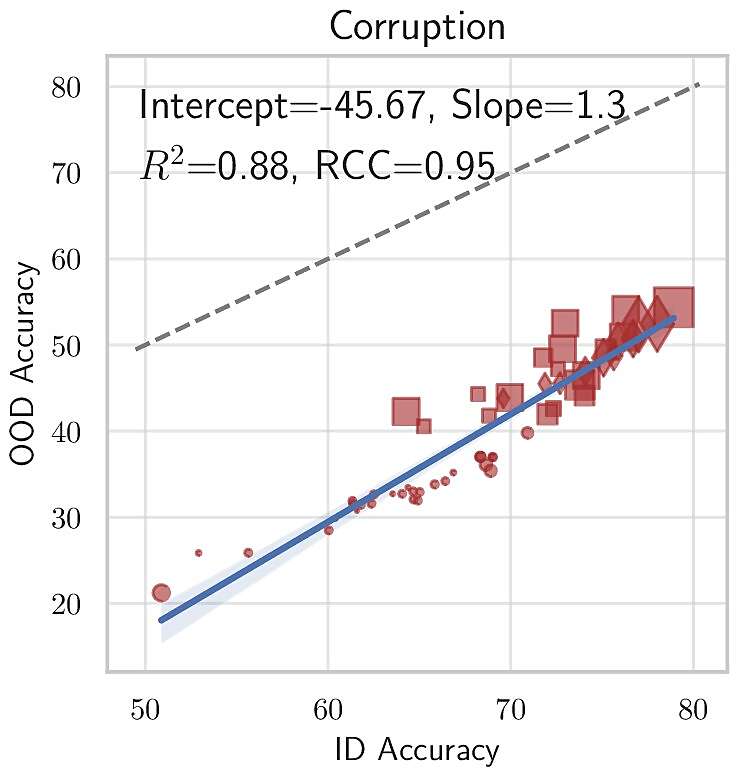}
    \end{subfigure}

    \begin{subfigure}{.245\linewidth}
        \includegraphics[width=\linewidth]{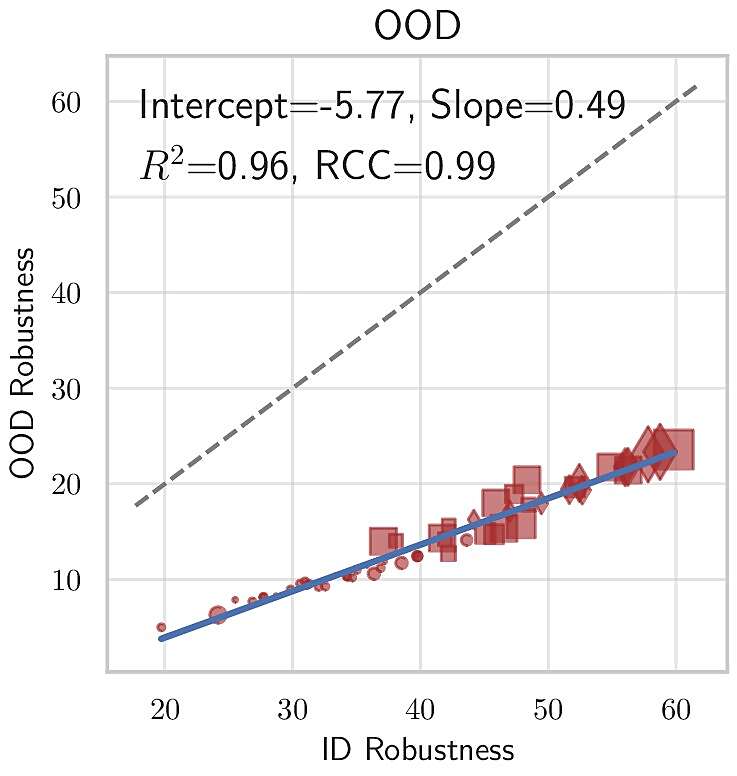}
    \end{subfigure}
    \begin{subfigure}{.245\linewidth}
        \includegraphics[width=\linewidth]{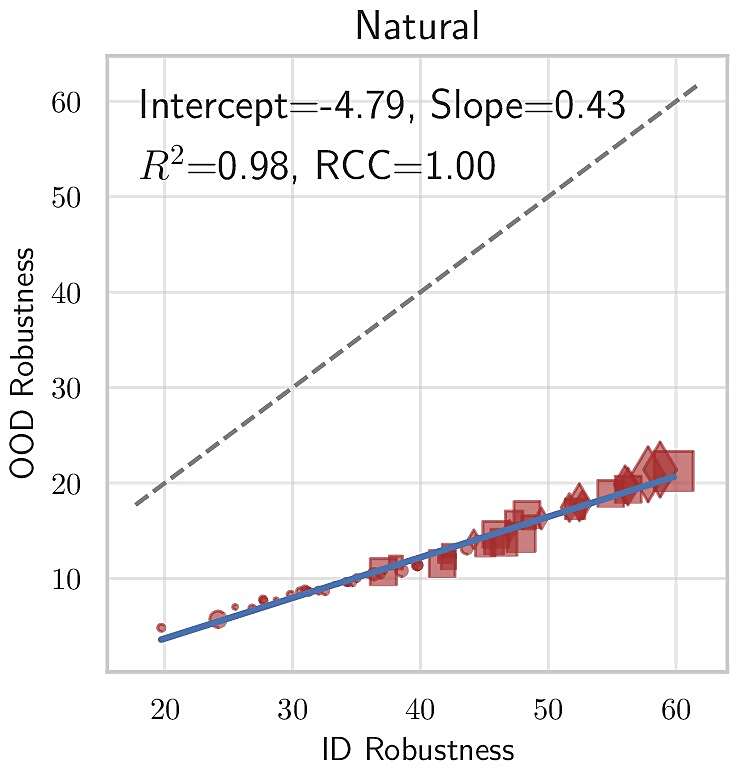}
    \end{subfigure}
    \begin{subfigure}{.245\linewidth}
        \includegraphics[width=\linewidth]{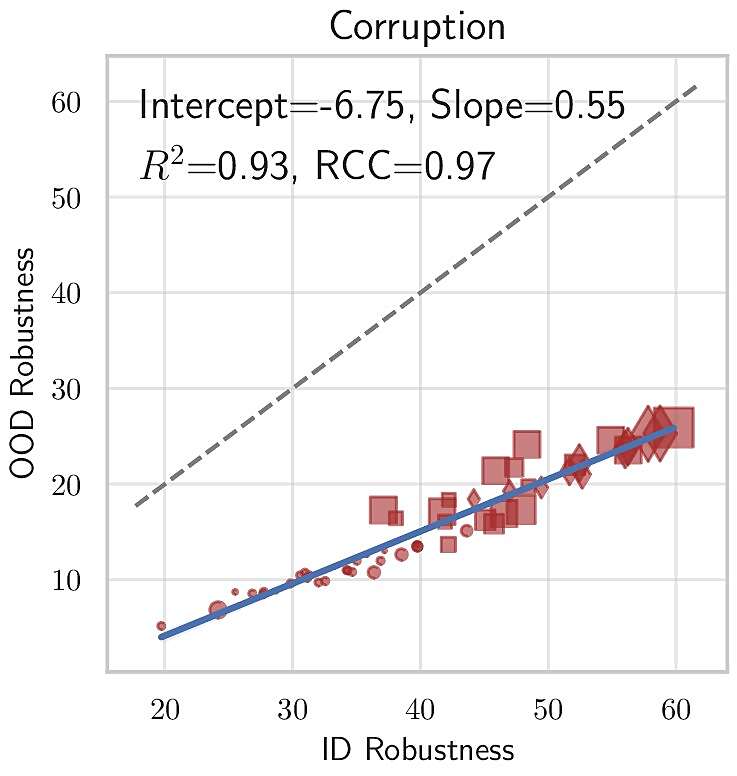}
    \end{subfigure}
    
    \caption{\textbf{Correlation between ID accuracy and OOD accuracy (odd rows); ID robustness and OOD robustness (even rows) for ImageNet \linf AT models}.}
\end{figure}

\clearpage
\section{Plots of ID-OOD Correlation per Threat Shift}
\label{appendix: correlation per threat shift}

\end{document}

%% file: tables/benchmark-cifar10-linf.tex
\begin{table*}[!ht]
\centering
\caption{\textbf{Performance, evaluated with OODRobustBench, of state-of-the-art models trained on CIFAR10 to be robust to \linf attacks}. The top 3 results for each metric are highlighted in \textbf{bold} and/or \uline{underscore}. Significant ranking discrepancies are indicated in {\color[HTML]{FF0000}red}. The "OOD" column presents the average of the robustness to \oodd and \oodt. The complete leaderboard, featuring a total of 396 models, is available at \url{https://oodrobustbench.github.io/}.}
\label{tab: benchmark result}
\resizebox{\linewidth}{!}{%
\begin{tabular}{@{}llcccccccc@{}}
\toprule
  \multicolumn{1}{c}{\multirow{2}{*}{Method}} &
  \multicolumn{1}{c}{\multirow{2}{*}{Model}} &
  \multicolumn{2}{c}{Accuracy (\%)} &
  \multicolumn{4}{c}{Robustness (\%)} &
  \multicolumn{2}{c}{Ranking} \\ \cmidrule(l){3-4} \cmidrule(l){5-8} \cmidrule(l){9-10}
  \multicolumn{1}{c}{} &
  \multicolumn{1}{c}{} &
  ID &
  \oodd &
  ID &
  \oodd &
  \oodt &
  OOD &
  ID &
  OOD \\ 
  \midrule
  BDM \citep{wang_better_2023} & WRN-70-16 &
  {\ul 93.2} &
  {\ul 76.0} &
  {\ul \textbf{70.7}} &
  {\ul \textbf{44.4}} &
  35.8 &
  \textbf{40.1} &
  1 &
  2 \\
  AS \citep{bai_improving_2023} & WRN-70-16 + ResNet-152 &
  {\ul \textbf{95.2}} &
  {\ul \textbf{79.0}} &
  \textbf{69.5} &
  \textbf{43.3} &
  {\ul \textbf{46.7}} &
  {\ul \textbf{45.0}} &
  2 &
  1 \\
  DKL \citep{cui_decoupled_2023} &  WRN-28-10 &
  92.1 &
  74.8 &
  {\ul 67.7} &
  42.4 &
  35.4 &
  38.9 &
  3 &
  4 \\

  BDM \citep{wang_better_2023} & WRN-28-10&
  92.4 &
  75.0 &
  67.3 &
  42.3 &
  35.2 &
  38.8 &
  4 &
  5 \\
 
  FDA \citep{rebuffi_fixing_2021} & WRN-70-16 &
  92.2 &
  74.8 &
  66.7 &
  {\ul 42.6} &
  33.6 &
  38.1 &
  5 &
  9 \\
 
  DDPM \citep{gowal_improving_2021} & WRN-70-16 &
  88.7 &
  70.6 &
  66.2 &
  42.7 &
  33.6 &
  38.2 &
  6 &
  8 \\
 
  Uncovering \citep{gowal_uncovering_2021} & WRN-70-16 &
  91.1 &
  73.2 &
  66.0 &
  42.5 &
  34.0 &
  38.2 &
  7 &
  7 \\
 
  RobustResNet \citep{huang_revisiting_2023} & WRN-A4 &
  91.5 &
  73.8 &
  65.8 &
  41.7 &
  33.3 &
  37.5 &
  8 &
  12 \\
 
  FDA \citep{rebuffi_fixing_2021} & WRN-106-16 &
  88.5 &
  70.6 &
  64.8 &
  41.4 &
  33.9 &
  37.6 &
  9 &
  10 \\
 
      DyART \citep{xu_exploring_2023} & WRN-28-10 &
  \textbf{93.6} &
  \textbf{77.2} &
  64.7 &
  39.6 &
  {\ul 37.0} &
  38.3 &
  10 &
  6 \\
 
  PORT \citep{sehwag_robust_2022} & ResNet-152 &
  87.2 &
  69.2 &
  62.7 &
  40.7 &
  32.3 &
  36.5 &
  17 &
  15 \\
 
  HAT \citep{rade_reducing_2022} & WRN-28-10 &
  88.1 &
  69.4 &
  60.9 &
  35.1 &
  30.2 &
  32.6 &
  {\color[HTML]{FF0000}22} &
  {\color[HTML]{FF0000}57} \\
 
  AWP \citep{wu_adversarial_2020} & WRN-28-10 &
  88.2 &
  69.8 &
  60.1 &
  38.2 &
  31.3 &
  34.8 &
  26 &
  27 \\
 
  RST \citep{carmon_unlabeled_2019} & WRN-28-10 &
  89.6 &
  71.5 &
  59.8 &
  36.7 &
  31.1 &
  33.9 &
  28 &
  38 \\
 
  MART \citep{wang_improving_2020} & WRN-28-10 &
  87.5 &
  70.2 &
  56.7 &
  35.5 &
  32.6 &
  34.0 &
  52 &
  35 \\
 
  HE \citep{pang_boosting_2020} & WRN-34-20 &
  85.1 &
  66.9 &
  53.8 &
  32.4 &
  \textbf{46.2} &
  {\ul 39.3} &
  {\color[HTML]{FF0000}70} &
  {\color[HTML]{FF0000}3} \\
 
  FAT \citep{zhang_attacks_2020} & WRN-34-10 &
  84.5 &
  65.9 &
  53.6 &
  32.9 &
  31.8 &
  32.4 &
  71 &
  59 \\
 
  Overfitting \citep{rice_overfitting_2020} & WRN-34-20 &
  85.3 &
  66.4 &
  53.5 &
  32.0 &
  27.8 &
  29.9 &
  72 &
  89 \\
 
  TRADES \citep{zhang_theoretically_2019} & WRN-34-10 &
  84.9 &
  66.5 &
  52.6 &
  31.6 &
  26.5 &
  29.1 &
  76 &
  99 \\
 
  FBF \citep{wong_fast_2020} & PreActResNet-18 &
  83.3 &
  64.9 &
  43.3 &
  25.3 &
  24.8 &
  25.0 &
  111 &
  112 \\
  \bottomrule
\end{tabular}%
}
\end{table*}

%% file: tables/benchmark-cifar10-l2.tex
\begin{table}[!h]
\centering
\caption{\textbf{Performance, evaluated with OODRobustBench, of state-of-the-art models trained on CIFAR10 to be robust to \lt attacks}. Top 3 results under each metric are highlighted by \textbf{bold} and/or \uline{underscore}. The column ``OOD" gives the overall OOD robustness which is the mean of the robustness to \oodd and \oodt.}
\label{tab: benchmark result cifar10 l2}
\resizebox{\textwidth}{!}{%
\begin{tabular}{@{}llcccccccc@{}}
\toprule
\multicolumn{1}{c}{\multirow{2}{*}{Method}} &
\multicolumn{1}{c}{\multirow{2}{*}{Model}} &
  \multicolumn{2}{c}{Accuracy} &
  \multicolumn{4}{c}{Robustness} &
  \multicolumn{2}{c}{Ranking (Rob.)} \\ \cmidrule(l){3-4} \cmidrule(l){5-8} \cmidrule(l){9-10} 
\multicolumn{1}{c}{} & \multicolumn{1}{c}{}                & ID                   & \oodd          & ID             & \oodd          & \oodt          & OOD            & ID & OOD \\ \midrule
BDM \citep{wang_better_2023} & WRN-70-16 &
  \textbf{95.54} &
  {\ul \textbf{80.04}} &
  {\ul \textbf{84.97}} &
  {\ul \textbf{60.83}} &
  {\ul \textbf{36.65}} &
  {\ul \textbf{48.74}} &
  1 &
  1 \\
BDM \citep{wang_better_2023} & WRN-28-10             & {\ul 95.16}          & {\ul 79.28}    & \textbf{83.69} & \textbf{59.39} & \textbf{35.04} & \textbf{47.21} & 2  & 2   \\
FDA \citep{rebuffi_fixing_2021} & WRN-70-16 (extra) & {\ul \textbf{95.74}} & \textbf{79.90} & {\ul 82.36}    & {\ul 57.94}    & 31.71          & 44.82          & 3  & 4   \\
Uncovering \citep{gowal_uncovering_2021} & WRN-70-16 (extra)            & 94.74                & 78.78          & 80.56          & 56.18          & 30.48          & 43.33          & 4  & 6   \\
FDA \citep{rebuffi_fixing_2021} & WRN-70-16 (DDPM)  & 92.41                & 75.95          & 80.42          & 56.82          & {\ul 34.58}    & {\ul 45.70}    & 5  & 3   \\
RATIO \citep{vedaldi_adversarial_2020} & WRN-34-10 (extra)  & 93.97                & 77.40          & 78.81          & 54.71          & 31.62          & 43.16          & 6  & 7   \\
FDA \citep{rebuffi_fixing_2021} & WRN-28-10  & 91.79                & 75.26          & 78.79          & 55.63          & 33.32          & 44.48          & 7  & 5   \\
PORT \citep{sehwag_robust_2022} &   WRN-34-10                 & 90.93                & 74.00          & 77.29          & 54.33          & 29.44          & 41.88          & 8  & 8   \\
RATIO \citep{vedaldi_adversarial_2020} & WRN-34-10        & 92.23                & 76.43          & 76.27          & 52.83          & 29.25          & 41.04          & 9  & 11  \\
HATE \citep{rade_reducing_2022} & PreActResNet-18            & 90.57                & 73.55          & 76.14          & 53.35          & 29.69          & 41.52          & 10 & 9   \\
FDA \citep{rebuffi_fixing_2021} & RreActResNet-18    & 90.33                & 72.96          & 75.87          & 52.21          & 30.06          & 41.14          & 11 & 10  \\
Uncovering \citep{gowal_uncovering_2021} & WRN-70-16                & 90.89                & 74.71          & 74.51          & 52.20          & 25.76          & 38.98          & 12 & 15  \\
PORT \citep{sehwag_robust_2022} & ResNet-18                  & 89.76                & 72.31          & 74.42          & 51.76          & 26.68          & 39.22          & 13 & 13  \\
AWP \citep{wu_adversarial_2020} & WRN-34-10                   & 88.51                & 71.23          & 73.66          & 51.53          & 27.50          & 39.52          & 14 & 12  \\
RATIO \citep{vedaldi_adversarial_2020}   & ResNet-50       & 91.07                & 74.24          & 72.99          & 49.32          & 28.72          & 39.02          & 15 & 14  \\
PGD10 \citep{robustness} & ResNet-50              & 90.83                & 73.85          & 69.25          & 46.65          & 17.71          & 32.18          & 16 & 16  \\
Overfitting \citep{rice_overfitting_2020}    & PreActResNet-18             & 88.67                & 71.27          & 67.69          & 44.76          & 18.58          & 31.67          & 17 & 17  \\
DDN \citep{rony_decoupling_2019} & WRN-38-10                  & 89.04                & 71.77          & 66.46          & 44.54          & 18.31          & 31.42          & 18 & 18  \\
MMA \citep{ding_mma_2020}     & WRN-28-4                    & 88.00                & 72.32          & 66.09          & 43.79          & 16.52          & 30.15          & 19 & 20  \\ \bottomrule
\end{tabular}%
}
\end{table}

%% file: tables/benchmark-imagenet.tex
\begin{table}[!h]
\centering
\caption{\textbf{Performance, evaluated with OODRobustBench, of state-of-the-art models trained on ImageNet to be robust to \linf attacks}. Top 3 results under each metric are highlighted by \textbf{bold} and/or \uline{underscore}. The column ``OOD" gives the overall OOD robustness which is the mean of the robustness to \oodd and \oodt.}
\label{tab: benchmark result imagenet linf}
\resizebox{\textwidth}{!}{%
\begin{tabular}{@{}llcccccccc@{}}
\toprule
\multicolumn{1}{c}{\multirow{2}{*}{Method}} &
\multicolumn{1}{c}{\multirow{2}{*}{Model}} &
  \multicolumn{2}{c}{Accuracy} &
  \multicolumn{4}{c}{Robustness} &
  \multicolumn{2}{c}{Ranking (Rob.)} \\ \cmidrule(l){3-4} \cmidrule(l){5-8} \cmidrule(l){9-10}
\multicolumn{1}{c}{}  &   \multicolumn{1}{c}{}                          & ID    & \oodd       & ID    & \oodd & \oodt       & OOD   & ID & OOD \\ \midrule
Comprehensive \citep{liu_comprehensive_2023} & Swin-L &
  {\ul \textbf{78.92}} &
  {\ul \textbf{45.84}} &
  {\ul \textbf{59.82}} &
  {\ul \textbf{23.59}} &
  \textbf{29.88} &
  {\ul \textbf{26.74}} &
  1 &
  1 \\
Comprehensive \citep{liu_comprehensive_2023} & ConvNeXt-L &
  \textbf{78.02} &
  \textbf{44.74} &
  \textbf{58.76} &
  \textbf{23.35} &
  {\ul \textbf{30.10}} &
  \textbf{26.72} &
  2 &
  2 \\
Revisiting \citep{singh_revisiting_2023} & ConvNeXt-L-ConvStem &
  {\ul 77.00} &
  44.05 &
  {\ul 57.82} &
  {\ul 23.09} &
  27.98 &
  {\ul 25.53} &
  3 &
  3 \\
Comprehensive \citep{liu_comprehensive_2023} & Swin-B             & 76.16 & 42.58       & 56.26 & 21.45 & 27.02       & 24.24 & 4  & 7   \\
Revisiting \citep{singh_revisiting_2023} & ConvNeXt-B-ConvStem & 75.88 & 42.29       & 56.24 & 21.77 & 27.89       & 24.83 & 5  & 5   \\
Comprehensive \citep{liu_comprehensive_2023} & ConvNeXt-B         & 76.70 & 43.06       & 56.02 & 21.74 & 26.97       & 24.36 & 6  & 6   \\
Revisiting \citep{singh_revisiting_2023} & ViT-B-ConvStem      & 76.30 & {\ul 44.67} & 54.90 & 21.76 & {\ul 28.98} & 25.37 & 7  & 4   \\
Revisiting \citep{singh_revisiting_2023} & ConvNeXt-S-ConvStem & 74.08 & 39.55       & 52.66 & 19.35 & 26.87       & 23.11 & 8  & 9   \\
Revisiting \citep{singh_revisiting_2023} & ConvNeXt-B          & 75.08 & 40.68       & 52.44 & 20.09 & 26.06       & 23.07 & 9  & 10  \\
Comprehensive \citep{liu_comprehensive_2023} & Swin-S             & 75.20 & 40.84       & 52.10 & 19.67 & 24.73       & 22.20 & 10 & 12  \\
Comprehensive \citep{liu_comprehensive_2023} & ConvNeXt-S         & 75.64 & 40.91       & 51.66 & 19.40 & 25.00       & 22.20 & 11 & 11  \\
Revisiting \citep{singh_revisiting_2023} & ConvNeXt-T-ConvStem & 72.70 & 38.15       & 49.46 & 17.97 & 25.32       & 21.65 & 12 & 14  \\
Revisiting \citep{singh_revisiting_2023} & ViT-S-ConvStem      & 72.58 & 39.24       & 48.46 & 17.83 & 25.43       & 21.63 & 13 & 15  \\
Revisiting \citep{singh_revisiting_2023} & ViT-B               & 72.98 & 42.38       & 48.34 & 20.43 & 26.26       & 23.34 & 14 & 8   \\
Light \citep{debenedetti_light_2023} & XCiT-L12           & 73.78 & 38.10       & 47.88 & 15.84 & 23.22       & 19.53 & 15 & 18  \\
Revisiting \citep{singh_revisiting_2023} & ViT-M               & 71.78 & 39.88       & 47.34 & 18.95 & 25.25       & 22.10 & 16 & 13  \\
Revisiting \citep{singh_revisiting_2023} & ConvNeXt-T          & 71.88 & 37.70       & 46.98 & 17.13 & 21.36       & 19.25 & 17 & 19  \\
Easy \citep{mao2022easyrobust} & Swin-B                  & 74.14 & 38.45       & 46.54 & 15.36 & 22.19       & 18.78 & 18 & 20  \\
Comprehensive \citep{liu_comprehensive_2023} & ViT-B               & 72.84 & 39.88       & 45.90 & 18.01 & 22.95       & 20.48 & 19 & 16  \\
Light \citep{debenedetti_light_2023} & XCiT-M12           & 74.04 & 37.00       & 45.76 & 14.73 & 22.82       & 18.77 & 20 & 21  \\ \bottomrule
\end{tabular}%
}
\end{table}

%% file: tables/robust-intervention.tex
\newlength{\authcolwidth}
\settowidth{\authcolwidth}{Training}

\begin{table}[!h]
\centering
\caption{\textbf{The effect of training with extra data on the OOD generalization of accuracy and robustness}.}
\label{tab: robust intervention extra data}
\resizebox{\columnwidth}{!}{%
\tabcolsep3pt
\begin{tabular}{@{}lcp{\authcolwidth}lccccccccc@{}}
\toprule
\multicolumn{1}{c}{\multirow{2}{*}{Dataset}} &
  Threat &
  \multicolumn{1}{c}{\multirow{2}{*}{Training}} &
  \multicolumn{1}{c}{Model} &
  Extra &
  \multicolumn{2}{c}{ID} &
  \multicolumn{4}{c}{\oodd} &
  \multicolumn{2}{c}{\oodt} \\ \cmidrule(l){6-7} \cmidrule(l){8-11} \cmidrule(l){12-13} 
\multicolumn{1}{c}{} &
  Model &
  \multicolumn{1}{c}{} &
  \multicolumn{1}{c}{Architecture} &
  Data &
  Acc. &
  Rob. &
  Acc. &
  Rob. &
  EAcc. &
  ERob. &
  Rob. &
  ERob. \\ \midrule
\multirow{3}{*}{CIFAR10} &
  \multirow{3}{*}{Linf} &
  \multirow{3}{*}{\parbox{\authcolwidth}{\cite{gowal_uncovering_2021}}} &
  \multirow{3}{*}{WideResNet70-16} &
  - &
  85.29 &
  57.24 &
  66.98 &
  35.90 &
  -0.56 &
  0.30 &
  29.39 &
  -2.18 \\
 &
   &
   &
   &
  Synthetic &
  88.74 &
  {\ul \textbf{66.24}} &
  70.68 &
  {\ul \textbf{42.76}} &
  -0.08 &
  {\ul \textbf{0.74}} &
  33.65 &
  -2.13 \\
 &
   &
   &
  &
  Real &
  {\ul \textbf{91.10}} &
  66.03 &
  {\ul \textbf{73.24}} &
  42.58 &
  {\ul \textbf{0.26}} &
  0.71 &
  {\ul \textbf{34.00}} &
  {\ul \textbf{-1.67}} \\ \bottomrule
\end{tabular}%
}
\end{table}

\begin{table}[!h]
\centering
\caption{\textbf{The effect of data augmentation on the OOD generalization of accuracy and robustness}. The results reported in \cref{fig: intervention data aug} are the mean of the results on ViT and WideResNets.}
\label{tab: robust intervention data aug}
\resizebox{\columnwidth}{!}{%
\tabcolsep3pt
\begin{tabular}{@{}lcp{\authcolwidth}llllllllll@{}}
\toprule
\multicolumn{1}{c}{\multirow{2}{*}{Dataset}} &
  Threat &
  \multicolumn{1}{c}{\multirow{2}{*}{Training}} &
  \multicolumn{1}{c}{Model} &
  \multicolumn{1}{c}{Data} &
  \multicolumn{2}{c}{ID} &
  \multicolumn{4}{c}{\oodd} &
  \multicolumn{2}{c}{\oodt} \\ \cmidrule(l){6-7} \cmidrule(l){8-11} \cmidrule(l){12-13} 
\multicolumn{1}{c}{} &
  Model &
  \multicolumn{1}{c}{} &
  \multicolumn{1}{c}{Architecture} &
  \multicolumn{1}{c}{Augmentation} &
  \multicolumn{1}{c}{Acc.} &
  \multicolumn{1}{c}{Rob.} &
  \multicolumn{1}{c}{Acc.} &
  \multicolumn{1}{c}{Rob.} &
  \multicolumn{1}{c}{EAcc.} &
  \multicolumn{1}{c}{ERob.} &
  \multicolumn{1}{c}{Rob.} &
  \multicolumn{1}{c}{ERob.} \\ \midrule
\multirow{12}{*}{CIFAR10} &
  \multirow{12}{*}{Linf} &
  \multirow{12}{*}{\parbox{\authcolwidth}{\cite{li_data_2023}}} &
  \multirow{6}{*}{ViT-B} &
  RandomCrop &
  {\ul 83.23} &
  47.02 &
  {\ul 66.48} &
  28.85 &
  {\ul 0.86} &
  0.54 &
  27.36 &
  0.57 \\
 &
   &
   &
   &
  Cutout &
  \textbf{84.22} &
  \textbf{49.57} &
  \textbf{67.23} &
  \textbf{30.68} &
  0.69 &
  0.56 &
  29.74 &
  1.75 \\
 &
   &
   &
   &
  CutMix &
  80.92 &
  47.45 &
  63.93 &
  {\ul 29.89} &
  0.48 &
  \textbf{1.27} &
  {\ul 30.48} &
  {\ul 3.49} \\
 &
   &
   &
   &
  TrivialAugment &
  80.33 &
  46.61 &
  64.59 &
  29.56 &
  {\ul \textbf{1.69}} &
  {\ul \textbf{1.54}} &
  30.40 &
  {\ul \textbf{3.80}} \\
 &
   &
   &
   &
  AutoAugment &
  82.75 &
  {\ul 48.11} &
  65.89 &
  29.78 &
  0.73 &
  {\ul 0.69} &
  {\ul \textbf{30.90}} &
  \textbf{3.60} \\
 &
   &
   &
   &
  IDBH &
  {\ul \textbf{86.92}} &
  {\ul \textbf{51.55}} &
  {\ul \textbf{70.51}} &
  {\ul \textbf{32.08}} &
  \textbf{1.45} &
  0.54 &
  \textbf{30.59} &
  1.68 \\ \cmidrule(l){4-13} 
 &
   &
   &
  \multirow{6}{*}{WideResNet34-10} &
  RandomCrop &
  86.52 &
  52.42 &
  68.11 &
  31.55 &
  -0.58 &
  \textbf{-0.61} &
  26.47 &
  -2.84 \\
 &
   &
   &
   &
  Cutout &
  86.77 &
  53.31 &
  68.40 &
  31.03 &
  -0.53 &
  -1.76 &
  27.00 &
  -2.74 \\
 &
   &
   &
   &
  CutMix &
  87.41 &
  53.89 &
  68.97 &
  31.71 &
  -0.55 &
  -1.50 &
  28.50 &
  {\ul \textbf{-1.50}} \\
 &
   &
   &
   &
  TrivialAugment &
  {\ul 86.98} &
  {\ul 54.18} &
  {\ul 69.85} &
  \textbf{32.94} &
  {\ul \textbf{0.73}} &
  {\ul \textbf{-0.47}} &
  \textbf{28.62} &
  {\ul -1.52} \\
 &
   &
   &
   &
  AutoAugment &
  \textbf{87.93} &
  \textbf{55.10} &
  \textbf{70.05} &
  {\ul 32.17} &
  {\ul 0.04} &
  -1.90 &
  {\ul \textbf{29.06}} &
  \textbf{-1.51} \\
 &
   &
   &
   &
  IDBH &
  {\ul \textbf{88.62}} &
  {\ul \textbf{55.56}} &
  {\ul \textbf{70.96}} &
  {\ul \textbf{32.99}} &
  \textbf{0.30} &
  {\ul -1.41} &
  {\ul 28.58} &
  -2.21 \\ \bottomrule
\end{tabular}%
}
\end{table}

\begin{table}[!h]
\centering
\caption{\textbf{The effect of model architecture on the OOD generalization of accuracy and robustness}.}
\label{tab: robust intervention model arch}
\resizebox{\columnwidth}{!}{%
\tabcolsep3pt
\begin{tabular}{@{}lcllccccccccc@{}}
\toprule
\multicolumn{1}{c}{\multirow{2}{*}{Dataset}} &
  Threat &
  \multicolumn{1}{c}{\multirow{2}{*}{Training}} &
  \multicolumn{1}{c}{Model} &
  Model &
  \multicolumn{2}{c}{ID} &
  \multicolumn{4}{c}{\oodd} &
  \multicolumn{2}{c}{\oodt} \\ \cmidrule(l){6-7} \cmidrule(l){8-11} \cmidrule(l){12-13} 
\multicolumn{1}{c}{} &
  Model &
  \multicolumn{1}{c}{} &
  \multicolumn{1}{c}{Architecture} &
  Size (M) &
  Acc. &
  Rob. &
  Acc. &
  Rob. &
  EAcc. &
  ERob. &
  Rob. &
  ERob. \\ \midrule
\multirow{4}{*}{ImageNet} &
  \multirow{4}{*}{\linf} &
  \multirow{4}{*}{\cite{liu_comprehensive_2023}} &
  ResNet152 &
  60.19 &
  70.92 &
  43.62 &
  34.43 &
  14.13 &
  -1.71 &
  -1.26 &
  17.23 &
  -3.47 \\
 &
   &
   &
  ConvNeXt-B &
  {\ul \textbf{88.59}} &
  {\ul \textbf{76.70}} &
  56.02 &
  {\ul \textbf{43.06}} &
  {\ul \textbf{21.74}} &
  1.03 &
  0.33 &
  26.97 &
  -0.63 \\
 &
   &
   &
  ViT-B &
  86.57 &
  72.84 &
  45.90 &
  39.88 &
  18.01 &
  {\ul \textbf{1.78}} &
  {\ul \textbf{1.51}} &
  22.95 &
  {\ul \textbf{0.98}} \\
 &
   &
   &
  Swin-B &
  87.77 &
  76.16 &
  {\ul \textbf{56.26}} &
  42.58 &
  21.45 &
  1.10 &
  -0.07 &
  {\ul \textbf{27.02}} &
  -0.72 \\ \bottomrule
\end{tabular}%
}
\end{table}

\begin{table}[!h]
\centering
\caption{\textbf{The effect of model size on the OOD generalization of accuracy and robustness}. The results reported in \cref{fig: intervention model size} are averaged over three architectures at the corresponding relatively model size. For example, the result of "small" is averaged over WideResNet28-10, ResNet50 and ConvNeXt-S-ConvStem.}
\label{tab: robust intervention model size}
\resizebox{\columnwidth}{!}{%
\tabcolsep3pt
\begin{tabular}{@{}lcp{\authcolwidth}lccccccccc@{}}
\toprule
\multicolumn{1}{c}{\multirow{2}{*}{Dataset}} &
  \multicolumn{1}{c}{Threat} &
  \multicolumn{1}{c}{\multirow{2}{*}{Training}} &
  \multicolumn{1}{c}{Model} &
  Model &
  \multicolumn{2}{c}{ID} &
  \multicolumn{4}{c}{\oodd} &
  \multicolumn{2}{c}{\oodt} \\ \cmidrule(l){6-7} \cmidrule(l){8-11} \cmidrule(l){12-13}
\multicolumn{1}{c}{} &
  \multicolumn{1}{c}{Model} &
  \multicolumn{1}{c}{} &
  \multicolumn{1}{c}{Architecture} &
  Size &
  Acc. &
  Rob. &
  Acc. &
  Rob. &
  EAcc. &
  ERob. &
  Rob. &
  ERob. \\ \midrule
\multirow{3}{*}{CIFAR10} &
  \multirow{3}{*}{\linf} &
  \multirow{3}{*}{\parbox{\authcolwidth}{\cite{rebuffi_fixing_2021}}} &
  WideResNet28-10 &
  36.48 &
  87.33 &
  60.88 &
  69.35 &
  38.54 &
  -0.10 &
  0.35 &
  33.63 &
  {\ul \textbf{0.36}} \\
 &
   &
   &
  WideResNet70-16 &
  266.80 &
  88.54 &
  64.33 &
  70.62 &
  41.01 &
  0.04 &
  0.35 &
  {\ul \textbf{34.12}} &
  -0.76 \\
 &
   &
   &
  WideResNet106-16 &
  {\ul \textbf{415.48}} &
  {\ul \textbf{88.50}} &
  {\ul \textbf{64.82}} &
  {\ul \textbf{70.65}} &
  {\ul \textbf{41.43}} &
  {\ul \textbf{0.11}} &
  {\ul \textbf{0.42}} &
  33.90 &
  -1.22 \\ \midrule
\multirow{3}{*}{ImageNet} &
  \multirow{3}{*}{\linf} &
  \multirow{3}{*}{\parbox{\authcolwidth}{\cite{liu_comprehensive_2023}}} &
  ResNet50 &
  25.56 &
  65.02 &
  32.02 &
  28.43 &
  9.23 &
  {\ul \textbf{-1.68}} &
  {\ul \textbf{-0.53}} &
  13.71 &
  {\ul \textbf{-0.52}} \\
 &
   &
   &
  ResNet101 &
  44.55 &
  68.34 &
  39.76 &
  31.74 &
  12.44 &
  -1.76 &
  -1.08 &
  16.82 &
  -1.72 \\
 &
   &
   &
  ResNet152 &
  {\ul \textbf{60.19}} &
  {\ul \textbf{70.92}} &
  {\ul \textbf{43.62}} &
  {\ul \textbf{34.43}} &
  {\ul \textbf{14.13}} &
  -1.71 &
  -1.26 &
  {\ul \textbf{17.23}} &
  -3.47 \\ \midrule
\multirow{3}{*}{ImageNet} &
  \multirow{3}{*}{\linf} &
  \multirow{3}{*}{\parbox{\authcolwidth}{\cite{singh_revisiting_2023}}} &
  ConvNeXt-S-ConvStem &
  50.26 &
  74.08 &
  52.66 &
  39.55 &
  19.35 &
  0.19 &
  -0.42 &
  26.87 &
  1.14 \\
 &
   &
   &
  ConvNeXt-B-ConvStem &
  88.75 &
  75.88 &
  56.24 &
  42.29 &
  21.77 &
  1.10 &
  0.26 &
  27.89 &
  0.16 \\
 &
   &
   &
  ConvNeXt-L-ConvStem &
  {\ul \textbf{198.13}} &
  {\ul \textbf{77.00}} &
  {\ul \textbf{57.82}} &
  {\ul \textbf{44.05}} &
  {\ul \textbf{23.09}} &
  {\ul \textbf{1.71}} &
  {\ul \textbf{0.80}} &
  {\ul \textbf{27.98}} &
  -0.63 \\ \bottomrule
\end{tabular}%
}
\end{table}


\begin{table}[!h]
\centering
\caption{\textbf{The effect of different adversarial training methods on the OOD generalization of accuracy and robustness}.}
\label{tab: robust intervention training}
\resizebox{\columnwidth}{!}{%
\tabcolsep3pt
\begin{tabular}{@{}lclcccccccc@{}}
\toprule
\multicolumn{1}{c}{\multirow{2}{*}{Dataset}} &
  \multicolumn{1}{c}{\multirow{2}{*}{Threat}} &
  \multicolumn{1}{c}{\multirow{2}{*}{Training}} &
  \multicolumn{2}{c}{ID} &
  \multicolumn{4}{c}{\oodd} &
  \multicolumn{2}{c}{\oodt} \\ \cmidrule(l){4-5} \cmidrule(l){6-9} \cmidrule(l){10-11} 
\multicolumn{1}{c}{} &
  \multicolumn{1}{c}{} &
  \multicolumn{1}{c}{} &
  Acc. &
  Rob. &
  Acc. &
  Rob. &
  EAcc. &
  ERob. &
  Rob. &
  ERob. \\ \midrule
\multirow{8}{*}{CIFAR10} &
  \multirow{8}{*}{\linf} &
  PGD \citep{li_data_2023} &
  {\ul \textbf{86.52}} &
  {\ul \textbf{52.42}} &
  {\ul \textbf{68.11}} &
  31.55 &
  -0.58 &
  -0.61 &
  26.47 &
  -2.84 \\
 &
   &
  VR-\linf \citep{dai_formulating_2022} &
  72.72 &
  49.92 &
  56.12 &
  {\ul \textbf{31.84}} &
  0.34 &
  {\ul \textbf{1.47}} &
  {\ul \textbf{34.70}} &
  {\ul \textbf{6.55}} \\ \cmidrule(l){3-11}

&
   &
  PGD \citep{rice_overfitting_2020} &
  {\ul \textbf{85.34}} &
  53.52 &
  66.46 &
  32.07 &
  -1.12 &
  -0.88 &
  27.89 &
  -1.94 \\
 &
   &
  HE \citep{pang_boosting_2020} &
  85.14 &
  {\ul \textbf{53.84}} &
  {\ul \textbf{66.96}} &
  {\ul \textbf{32.45}} &
  {\ul \textbf{-0.43}} &
  {\ul \textbf{-0.72}} &
  {\ul \textbf{46.20}} &
  {\ul \textbf{16.22}} \\ \cmidrule(l){3-11}
 &
   &
  PGD (locally-trained) &
  80.44 &
  38.98 &
  62.40 &
  22.18 &
  -0.60 &
  -0.39 &
  21.77 &
  -1.27 \\
 &
   &
  MMA \citep{ding_mma_2020} &
  {\ul \textbf{84.37}} &
  {\ul \textbf{41.86}} &
  {\ul \textbf{68.22}} &
  {\ul \textbf{24.65}} &
  {\ul \textbf{1.54}} &
  {\ul \textbf{0.02}} &
  {\ul \textbf{35.12}} &
  {\ul \textbf{10.74}} \\ \cmidrule(l){3-11}
 &
   &
  PGD \citet{gowal_uncovering_2021} &
  91.10 &
  66.03 &
  73.24 &
  42.58 &
  0.26 &
  {\ul \textbf{0.71}} &
  34.00 &
  -1.67 \\
 &
   &
  AS \cite{bai_improving_2023} &
  {\ul \textbf{95.23}} &
  {\ul \textbf{69.50}} &
  {\ul \textbf{79.09}} &
  {\ul \textbf{43.32}} &
  {\ul \textbf{2.25}} &
  -1.03 &
  {\ul \textbf{46.71}} &
  {\ul \textbf{9.41}} \\
  \bottomrule
\end{tabular}%
}
\end{table}